\documentclass[10pt,twocolumn,letterpaper]{article}

\usepackage{cvpr}              % To produce the CAMERA-READY version
\usepackage{graphicx}
\usepackage{amsmath}
\usepackage{amssymb}
\usepackage{amsfonts}
\usepackage{booktabs}
\usepackage{gensymb}

\usepackage[pagebackref,breaklinks,colorlinks]{hyperref}

\usepackage[capitalize]{cleveref}
\crefname{section}{Sec.}{Secs.}
\Crefname{section}{Section}{Sections}
\Crefname{table}{Table}{Tables}
\crefname{table}{Tab.}{Tabs.}

\def\cvprPaperID{9672} % *** Enter the CVPR Paper ID here
\def\confName{CVPR}
\def\confYear{2022}

\begin{document}

%%%%%%%%% TITLE - PLEASE UPDATE
\title{Multi-instance Point Cloud Registration by Efficient Correspondence Clustering}

\author{
Weixuan Tang and Danping Zou\thanks{Corresponding author. This work was supported by National Science Foundation of China (62073214).}\\
Shanghai Key Laboratory of Navigation and Location Based Services\\
Shanghai Jiao Tong University, Shanghai,China\\
{\tt\small \{weixuantang,dpzou\}@sjtu.edu.cn}}

\maketitle

%%%%%%%%% ABSTRACT
\begin{abstract}
We address the problem of estimating the poses of multiple instances of the source point cloud within a target point cloud.
Existing solutions require sampling a lot of hypotheses to detect possible instances and reject the outliers, whose robustness and efficiency degrade notably when the number of instances and outliers increase.
We propose to directly group the set of noisy correspondences into different clusters based on a distance invariance matrix. The instances and outliers are automatically identified through clustering. Our method is robust and fast. We evaluated our method on both synthetic and real-world datasets. The results show that our approach can correctly register up to $20$ instances with an F1 score of $90.46\%$ in the presence of $70\%$ outliers, which performs significantly better and at least $10\times$ faster than existing methods. (Source code : \href{https://github.com/SJTU-ViSYS/multi-instant-reg}{https://github.com/SJTU-ViSYS/multi-instant-reg})
\end{abstract}

%%%%%%%%% BODY TEXT
\section{Introduction}
\label{sec:intro}
\begin{figure}[ht]
\centering
    \includegraphics[width=0.32\textwidth]{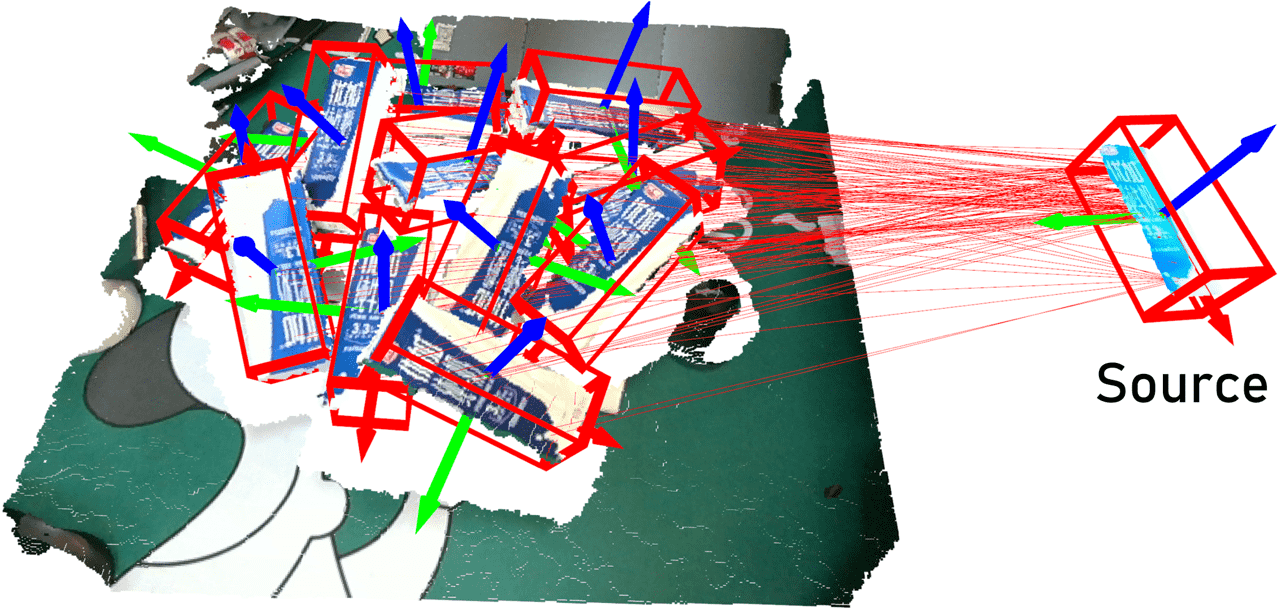}
    \caption{Multi-instance point cloud registration: Given a source point cloud of an object, multi-instance registration needs to estimate the pose of each object within the target point cloud.}
\label{fig:problem}
    \end{figure}

Three-dimensional point cloud registration \cite{paulj1992method}\cite{TEASER}\cite{DCP} mainly focuses on estimating one single transformation between the source point cloud and the target point cloud. However, we may sometimes want to estimate multiple transformations between point clouds. For instance, we have a 3D scan of an object and may want to find the poses of the same objects on the table within the target point cloud as shown in Figure \ref{fig:problem}. This problem, named multi-instance point cloud registration here, has been less investigated in the literature. It is non-trivial to extend existing point cloud registration methods to solve this problem.

The major challenge is to identify different clusters of corresponding points belonging to different instances within the set of noisy correspondences. One solution is to adopt a 3D object detector or apply instance segmentation to the target point cloud. After that, the pose of each instance can be estimated by a conventional point cloud registration method. However, this approach needs to train a detector or a segmentation network\cite{votenet}\cite{Occuseg} for specific objects or classes, which does not apply to unknown objects or arbitrary 3D scans.
Another solution is via multi-model fitting \cite{Tlinkage}\cite{RPA}\cite{Coverage} or \cite{MCT}\cite{CONSAC}\cite{ProgressiveX2}. Existing multi-model fitting methods rely on sampling valid hypotheses, which involves a large number of sampling steps when the number of models or the outlier ratio becomes high, making the efficiency and robustness of those algorithms drop drastically.

%\begin{figure}[ht]
%\includegraphics[width=0.48\textwidth]{figure/multi-instance-problem2.png}
%\caption{Multi-instance point cloud registration : Given a source point cloud of an object, multi-instance point cloud registration needs to estimate the relative pose of each object within the target point cloud. }
%\end{figure}

In this paper, we propose a robust and efficient solution to the multi-instance 3D registration problem. The key idea is to directly group the corresponding points into different clusters according to a distance invariance matrix. Specifically, the matrix is constructed by checking the distance consistency between each pair of correspondences after the point correspondences have been obtained by feature matching using descriptors like D3Feat \cite{D3Feat}, PREDATOR \cite{PREDATOR}, or SIFT \cite{lowe2004distinctive}. We find that the row or column vector of this matrix has a powerful representation capability that can be used for identifying the set of correspondences from a particular instance. We hence apply a simple and efficient clustering algorithm to divide those correspondences into cliques. The clustering is further refined by a few recursive steps involving merging similar clusters and re-assigning cluster ids to each correspondence. Finally, both the outliers and the inliers of each instance are automatically identified by a simple ranking strategy.

Our method is highly efficient since no time-consuming hypothesis sampling is required. We have conducted extensive experiments on both synthetic and real-world datasets. The results show that our method is at least ten times faster than existing methods while performing significantly better in terms of accuracy and robustness. To summary, our contributions include:
\begin{itemize}
\item We propose an efficient and robust solution to the multi-instance point cloud registration problem, which achieves superior performance in terms of accuracy, robustness, and speed.
\item We propose to use three metrics (Mean Hit Recall, Mean Hit Precision, and Mean Hit F1) to fully evaluate the performance of multi-instance point cloud registration. %To our best knowledge, no metric has been proposed for such kind of evaluation.
\item Our solution can be potentially used for zero-shot detection of 3D objects as our real-world tests demonstrate.
\end{itemize}

\begin{figure*}[ht]
    \centering
    \includegraphics[width=1\textwidth]{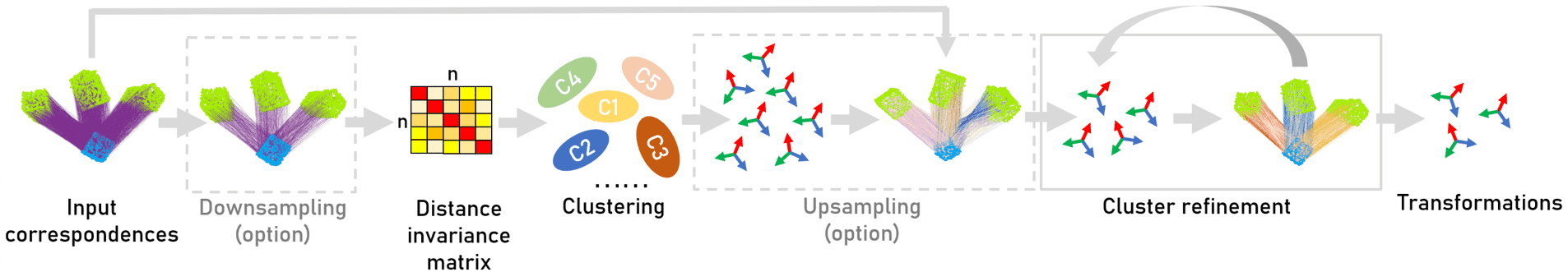} % Reduce the figure size so that it is slightly narrower than the column. Don't use precise values for figure width.This setup will avoid overfull boxes.
    \caption{The pipeline of the proposed method for multi-instance point cloud registration. A distance invariance matrix is constructed from the input correspondences, which is used to cluster the correspondences into different clusters (\textbf{Clustering}) and being refined (\textbf{Cluster refinement}). Finally, the rigid transformation (\textbf{Transformations}) related to each instance is estimated from each cluster of correspondences. To handle a large number of correspondences, two addition processes (\textbf{Downsampling} and \textbf{Upsampling}) are adopted.}
    \label{fig:pipeline}
    \end{figure*}
%-------------------------------------------------------------------------
\section{Related Work}
\textbf{Point cloud registration} can be divided into three stages: point matching, outlier rejection, and pose estimation. Most works focus on the first two stages since acquiring correct point correspondences is the key to successful registration. Point matching usually relies on features, either hand-crafted features \cite{rusu2009fast}\cite{drost2010model}
 or learning-based features \cite{3DMatch}\cite{PPFNet}\cite{3DSmoothNet}\cite{FCGF}\cite{D3Feat}. Though recent results show that the latter is superior to the hand-crafted ones in some benchmarks, those features are still far from producing perfect matching and a powerful outlier rejection mechanism is still required.
RANSAC\cite{RANSAC} and its variants(\cite{GCRansac}\cite{NGRansac}\cite{LORansac}) follow the hypothesis-and-verification process to reject outliers. This kind of method requires a lot of sampling steps when many outliers exist, becoming highly time-consuming, while could still fail to obtain the correct model. GORE\cite{bustos2017guaranteed} and PMC\cite{parra2019practical} seek to reduce the outliers by geometric consistency checks. Other methods such as FGR\cite{FGR} and TEASER\cite{TEASER} adopt robust estimators to solve the transformation directly from the noisy correspondences. By carefully tackling each subproblem, TEASER\cite{TEASER} achieved impressive performance in terms of robustness and efficiency.
There are also learning-based outlier rejection methods. DGR\cite{DGR} and 3DRegNet\cite{3DRegNet} treat outlier rejection as binary classification and predict the inlier probability for each correspondence. PointDSC\cite{PointDSC} takes a step further to embed the spatial consistency into feature learning for better training the inlier classifier.
Recently, a stream of work (e.g.PointNetLK\cite{PointNetLK}, FMR\cite{FMR}, DCP\cite{DCP}, PRNet\cite{PRNet}, RPMNet\cite{RPMNet}) tries to apply end-to-end learning to solve the registration problem. They also exhibit impressive performance, especially in low-overlap cases\cite{PREDATOR}.

Existing point cloud registration methods mostly focus on the one-to-one registration problem which estimates a single transformation between two point clouds. The multi-instance registration that aligns a source point cloud to its multiple instances in the target point cloud is however less investigated. This task is different from the multi-way registration \cite{choi2015robust} whose goal is to produce a globally consistent reconstruction from multiple fragments via pair-wise registration \cite{FGR}\cite{DGR}. The multi-instance registration requires not only rejecting outliers from the noisy correspondences but also identifying the set of inliers for individual instances, making it even more challenging than the classic registration problem.

\textbf{3D object detection and instance segmentation} are closely related to multi-instance 3D registration. Given a single point cloud, 3D object detection \cite{votenet} is to obtain the bounding box of each object of interest, while 3D instance segmentation \cite{SGPN}\cite{Occuseg} produces the instance labels for each point. Though they produce results \cite{avetisyan2019end}\cite{SceneCAD} similar to that of multi-instance registration, they need to train the prior of specific objects or categories into the network. By contrast, multi-instance registration processes two point clouds by directly aligning the source one to multiple instances in the target one, without using any priors about the contents of input 3D scans.

\textbf{Multi-model fitting}
Multi-instance registration can be approached by multi-model fitting, which aims to estimate the model parameters from the data points generated from multiple models.
% For example, given a set of 2D points sampled from different lines or circles, multi-model fitting is to estimate the parameters of each line or circle.
Existing multi-model fitting methods can be categorized into clustering-based methods and RANSAC-based ones. The clustering-based methods(e.g.\cite{Tlinkage}\cite{RPA}\cite{Coverage}) initialize a huge hypothesis set by sampling points and then calculate the preference vector about those hypotheses for each point. Those data points are clustered according to their preference vectors. Finally, the model parameters are computed from different clusters. RANSAC-based methods(e.g.\cite{PEARL}\cite{MultiX}\cite{ProgressiveX}\cite{MCT}\cite{CONSAC}\cite{ProgressiveX2}) run revised RANSAC sequentially to obtain multiple model parameters. They change the sampling weight of each point in each iteration to get different model parameters. CONSAC\cite{CONSAC} is a learning-based method that learns to weigh each point for sampling. Both clustering-based and RANSAC-based methods rely on sampling valid hypotheses. When the number of models or the outlier ratio increases, a lot of hypotheses are required to be sampled, making those algorithms highly inefficient.

\textbf{3D spatial consistency}, defined between every pair of points by a rigid transformation, is an important property for outlier rejection in 3D registration. Spectral matching \cite{leordeanu2005spectral} constructs a graph using the length consistency between each pair of correspondences and extracts the maximum clique from the graph to reject outliers. Existing methods such as TEASER \cite{TEASER}, GORE\cite{bustos2017guaranteed}, and PMC\cite{parra2019practical} also incorporate the spatial consistency in their algorithms. Recently, ROBIN\cite{shi2021robin} generalizes the concept of spatial consistency to high orders. PointDSC \cite{PointDSC} integrates the spatial consistency into an end-to-end learning pipeline to better regress the inlier probability.

Motivated by those works, we also adopt spatial consistency in our solution. Different from existing methods that apply spectral clustering\cite{leordeanu2005spectral} or approximated solutions\cite{shi2021robin} in the spatial consistency graph which is slow and has trouble dealing with multiple instances, we employ an efficient algorithm to find multiple instances within the correspondences. Specifically, we take the row vector or the column vector of the distance invariance matrix as the 'feature vector' of correspondence and run bottom-up clustering to get the inlier correspondences from different instances. Our method avoids hypothesis sampling which is the key weakness of existing multi-model fitting methods. It also does not rely on any particular features to obtain point correspondences, hence the performance can be further improved if better features (either 3D or image features) are adopted.

%%%%%%%%%%%%%%%%%%%%%%% Related Work End %%%%%%%%%%%%%%%%%%%%%%%%%%%%%%%%%%%%%%%

\section{Problem Statement}
In multi-instance point registration problem, the source point cloud $\mathbf{X}$ provides an instance of a 3D model and the target point cloud $\mathbf{Y}$ contains $K$ instances of this model, where those instances are the sets of points that may sample only a part of the 3D model. If we write the $k^{th}$ instance as $\mathbf{Y}_k$, the target point cloud $\mathbf{Y}$ can be decomposed as $
%\begin{equation}
\mathbf{Y} = \mathbf{Y}_0 \cup \mathbf{Y}_1 \cup \ldots \mathbf{Y}_k \ldots \cup \mathbf{Y}_K$.
%\end{equation}
Here we use $\mathbf{Y}_0$ to represent the part of the point cloud that does not belong to any instances.
The goal of multi-instance 3D registration is to find the rigid transformation $(\mathbf{R}_k, \mathbf{t}_k)$ that aligns the source instance $\mathbf{X}$ to each target instance $\mathbf{Y}_k$.
If we manage to obtain the correspondences between the source instance and each target instance $\mathbf{X} \leftrightarrow \mathbf{Y}_k$, the pose of the $k^{th}$ instance in the target point cloud, $(\mathbf{R}_k, \mathbf{t}_k)$, can be solved from the set of correspondences $\mathbf{X}\leftrightarrow \mathbf{Y}_k$ by minimizing the sum of alignment errors (\ref{eq:solve_rigid_transform}) \cite{SVD}:
\begin{equation}
  \underset{\mathbf{R}_k,\mathbf{t}_k}{\min}\sum_i{\parallel}\mathbf{y}_{ki}-(\mathbf{R}_k\mathbf{x}_i+\mathbf{t}_k)\parallel ^2.
\label{eq:solve_rigid_transform}
\end{equation}
Consider we have obtained a set of correspondences $\mathcal{C}$ between the source and target point clouds. The key of multi-instance registration task is to classify those correspondences into separate sets related to different instances, namely,
\begin{equation}
\mathcal{C} = \mathcal{C}_0 \cup \mathcal{C}_1\cdots \cup \mathcal{C}_K.
\end{equation}
Here $\mathcal{C}_0$ is used to represent the set of outliers. As we can see, multi-instance registration needs to not only reject outlier correspondences but also resolve the ambiguity of correspondences from different instances. This task is not easy because all instances look the same and a lot of outlier correspondences usually exist.

\section{Method}

The overview of the proposed method is shown in Fig. \ref{fig:pipeline}. Our method takes the point correspondences as the input. An invariance consistency matrix is then constructed by checking the distance consistency between correspondences. Next, those correspondences are quickly clustered into different groups by treating the column or row vectors as 'features' of those correspondences.  The clustering is done efficiently via agglomerative clustering,  which is further refined by alternatively merging similar transformations and re-assigning the cluster labels for several iterations. Optionally, we apply downsampling and upsampling processes to handle the case when the number of correspondences is large. The details are presented in the next sections.

\subsection{Invariance matrix $\&$ compatibility vector}
\label{subsec:Distance-Consistency-Graph}
The distance invariance property has been already explored in 3D registration for many years \cite{TEASER}\cite{shi2021robin}\cite{leordeanu2005spectral}, which describes that the distance between two points keeps unchanged after a rigid transformation. Namely, if $c_i :\mathbf{x}_i \leftrightarrow \mathbf{y}_i$ and $c_j : \mathbf{x}_j \leftrightarrow \mathbf{y}_j$ are two real correspondences, they should have 
%
%Let $\mathbf{x}_1, \mathbf{x}_2$ be the two points in the source point cloud and $\mathbf{y}_1, \mathbf{y}_2$ be their corresponding points found in the target point cloud. We have 
\begin{equation}
G_{ij}=|d_{ij} - d'_{ij} | < \delta
\label{eq:abs_diff}
\end{equation}
where $d_{ij} = \|\mathbf{x}_i-\mathbf{x}_j\|, d'_{ij}=\|\mathbf{y}_i -\mathbf{y}_j\|$ and $\delta $ is a threshold accounting for the noise.
Hence the difference between $d_{ij}$ and $d'_{ij}$ can be used as a metric to test whether an outlier exists, or whether the two correspondences are  from different rigid transformations. Instead of using the absolute difference defined in (\ref{eq:abs_diff}), we follow \cite{matrix} to assign a score measuring the relative difference
%\cite{SpectralTechnique16} 
between $c_i$ and $c_j$ by defining
%Ignoring the noises, we consider a simple score $G_{ij} $ of which $G_{ij}=1$ if the distances are the same or $G_{ij} = 0 $ otherwise.
\begin{equation}
G_{ij} = s_{ij}^2, s_{ij} = \min( \frac{d_{ij}}{d'_{ij}}, \frac{d'_{ij}}{d_{ij}}) \in (0, 1).
\end{equation}
A \emph{distance invariance matrix} $G$ (where we let $G_{ii} = 1)$ can be obtained by computing the scores between all the correspondence pairs. The distance invariance matrix is symmetric, where each column or row is a vector describing the compatibility between a given correspondence and other correspondences \cite{reviewof3dourlierremovingjiaqiYang}. 

\begin{figure}[ht]
\centering
\includegraphics[width=0.38\textwidth]{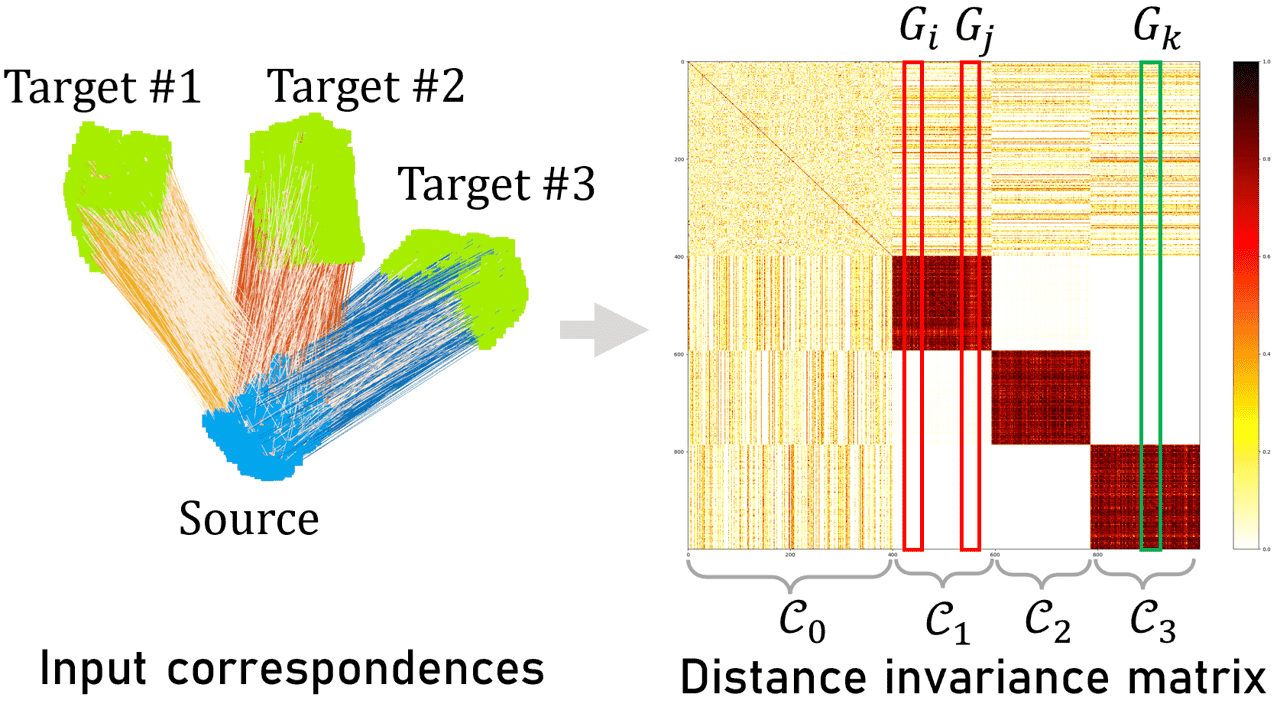} % Reduce the figure size so that it is slightly narrower than the column. Don't use precise values for figure width.This setup will avoid overfull boxes.
\caption{The column vectors (\emph{compatibility vectors}) in the distance invariance matrix contain rich information related to the instances. Here $G_i,G_j$ represent the compatibility vectors of $i^{th}$ and $j^{th}$ correspondences, which are both in the instance $\mathcal{C}_1$. We observe that $G_i$ is similar to $G_j$.  By constrast, $G_i$ differs significantly from $G_k$ since the $k^{th}$ correspondence is within the different instance $\mathcal{C}_3$. Here $\mathcal{C}_0$ represents the set of outliers. Please refer to Section \ref{subsec:Distance-Consistency-Graph} for details.}
\label{fig:matrix}
\end{figure}

We name a column vector $G_i = (G_{i1}, \ldots , G_{ij}, \ldots)^T$ as a \emph{compatibility vector} of the correspondence $c_i$.
We observe that if two correspondences belong to the same instance, their \emph{compatibility vectors} have similar patterns.
Consider two correspondences $c_i, c_j \in \mathcal{C}_s$. For any correspondence $c_k \in \mathcal{C}_s$, we have $G_{ik} \rightarrow 1, G_{jk} \rightarrow 1$ because of distance invariance. For other correspondences $c_k \in \mathcal{C}/\mathcal{C}_s$, we are likely to have $G_{ik} \rightarrow 0, G_{jk} \rightarrow 0$. In other words, $G_i,G_j$ have similar $0-1$ patterns.
 By contrast, if the two correspondences belong to different instances, their compatibility vectors are very different.
To better understand this observation, we illustrate a simple example in Figure \ref{fig:matrix}.

The compatibility vector of a correspondence can be regarded as a characteristic representation or 'feature' of this correspondence. Correspondences belonging to the same rigid transformation have similar features. Therefore, based on these compatibility vectors, we can cluster the correspondences into different groups related to inliers from different instances. %The large groups are related to different instances, while those small groups are likely outliers.

\subsection{Fast correspondence clustering}
We cluster the correspondences in a bottom-up manner which is much faster than spectral clustering adopted by existing methods \cite{parra2019practical}\cite{shi2021robin}. In the beginning, each correspondence is treated as an individual group. We then repeatedly merge the two groups with the smallest distance until the smallest distance between two groups is larger than a given value ($min\_dist\_thresh$). The way the distance between groups being defined yields different flavors of algorithms. We follow \cite{Tlinkage} to define the distance. Let $\mathbf{p}_i, \mathbf{p}_j$ be the representation vectors of two groups $i$ and $j$, the group distance is defined as
\begin{equation}
d(\mathbf{p}_i, \mathbf{p}_j)= 1-\frac{\langle \mathbf{p}_i,\mathbf{p}_j\rangle}{\parallel \mathbf{p}_i\parallel ^2+\parallel \mathbf{p}_j\parallel ^2-\langle \mathbf{p}_i,\mathbf{p}_j\rangle}.
\end{equation}
If the two groups are merged, the representation vector of the new group is updated by
$\mathbf{p}_i \leftarrow \min (\mathbf{p}_i, \mathbf{p}_j),$ where $\min(\cdot)$ denotes taking the minimum value for each dimension of the two vectors.
At the beginning of clustering, the representation vector of a group (containing only one correspondence) is set as the compatibility vector of that correspondence.

\subsection{Recursive cluster refinement} \label{sec:cluster_refinement}
After agglomerative clustering, we further refine the result by repeating the following steps until no change happens.

Step 1. Estimate the rigid transformations from the clusters where the number of correspondences is larger than a threshold $\alpha$.

Step 2. Merge similar transformations. This step will be explained in the next section.
 
Step 3. Re-assign the cluster label to each correspondence. Each correspondence is assigned to the transformation where its alignment error is the smallest. If the smallest alignment error over all the transformations is larger than $inlier\_thresh$, the correspondence is marked as an outlier.

During iteration, the correspondences become more and more gathered, so we can adjust $\alpha$ in Step 1 to increase the strength of outlier rejection. We use the following strategy to update the $\alpha$ in each iteration:
\begin{equation} 
\alpha \leftarrow \min(\alpha _0\times \theta ^{n-1},\left[N/100 \right] ),
\label{eq:alpha}
\end{equation}
where $n$ denotes the $n^{th}$ iteration, $N$ is the number of correspondences, and $\left[ \cdot \right] $ is a rounding operation. We set $\alpha_0 = 3$ and $\theta = 3$ in our experiments. The refinement process usually converges within three iterations in our experiments, hence it is also highly efficient.

\subsection{Merge duplicated transformations}
Sometimes similar transformations are generated from different clusters, which means they probably belong to the same instance. We need to merge them in this case.
Given two estimated transformations $(\mathbf{R}_1, \mathbf{t}_1)$ and $(\mathbf{R}_2, \mathbf{t}_2)$, we compute the alignment error for each correspondence, namely, $e_{ki} = \|\mathbf{y}_{i}-(\mathbf{R}_k \mathbf{x}_{i} + \mathbf{t}_k)\|^2, (k = 1,2)$. Next, we set $p_{ki} = 1$ if $e_{ki} < inlier\_thresh$, $p_{ki}=0$ otherwise. Thus, we obtain two binary sets $P_1, P_2$ for the two transformations. The criterion for merging the two transformations is
\begin{equation}
IOU = |P_1 \cap P_2|/|P_1 \cup P_2| \geq 80\%.
\label{eq:iou}
\end{equation}
If this criterion is satisfied, we drop one of the two transformations with more outliers ($p_{ki} = 0$). Then we re-assign the cluster label to each correspondence according to the one with the smallest alignment error among all the transformations.

\subsection{Extract transformations from clusters}

%After clustering, those small groups containing less than three correspondences are treated as outliers. We use the correspondences within each remaining group to compute a set of rigid transformations and evaluate the alignment errors,which is defined in (\ref{eq:solve_rigid_transform}), between those transformations and all of the input correspondences. The correspondence with the smallest alignment error over all the transformation larger than a threshold ($inlier\_thres$) are also treated as outliers. %All the outliers are saved in the $\mathcal{C}_0$ for future processing.

After clustering, we need to extract the rigid transformations from those correspondence clusters. Since we do not know about the true number of instances in the target point clouds, we need to choose those inlier clusters automatically. We first select the inlier clusters whose element number is larger than a threshold ($10$ in our experiments) and estimate transformations from those clusters. Next, we sort the transformations by their inlier numbers in descending order. The more inliers a transformation has, the higher chance it is associated with a true instance. Finally, we check the dropping ratio of the inlier number between the transformations and the first transformation (with the most inliers) by
\begin{equation}
\gamma_k = \#I_{k}/\#I_{0},\,\, k = 1,2,\ldots
\end{equation}
where $\#I_k$ denotes the number of inliers of $k^{th}$ transformation. We neglect all the transforms after $k$ if $\gamma_k <= \gamma\_thresh$. $\gamma\_thresh$ can be changed for the trade-off between recall and precision.

\subsection{Handle a large number of correspondences}
% TODO down_sample and cluster
% 1.unsample
% 为什么要downsample，downsample为什么可行
When the number of input correspondences is large, both calculating the distance invariance matrix and clustering the correspondences may become expensive. We add downsampling and upsampling processes to address this issue. The downsampling process is run before constructing the distance invariance matrix, which is done by randomly sampling a fixed number of correspondences ($1024$ in our implementation) for further processing. The upsampling process is run after clustering on the selected correspondences, which assigns all the correspondences to existing clusters. The assignment is done by selecting the transformation with the smallest alignment error as described in Section \ref{sec:cluster_refinement} (Step 3).

\section{Experiment}
We conduct experiments on both synthetic and real-world datasets by comparing our method with three state-of-the-art multi-model fitting methods: T-linkage(2014)\cite{Tlinkage}, Progressive-X(2019)\cite{ProgressiveX}, and CONSAC(2020)\cite{CONSAC}. Other multi-model fitting methods: RPA\cite{RPA} and RansaCov\cite{Coverage} are extremely slow (need months) to run our experiments, hence we do not include them. We also present the results of the state-of-the-art one-to-one registration method TEASER(2020)\cite{TEASER} for comparison. We carefully tune all the methods to achieve the best performance on the evaluation datasets within a reasonable time and memory consumption. For a fair comparison, all the methods take the same set of point correspondences as the input.

We implement our algorithm in Pytorch\cite{Pytorch}. T-linkage and Progressive-X are pure-CPU algorithms, while CONSAC is a learned-based method that runs on GPU. We run our algorithm on the same CPU (Intel Core i7-8700K) with T-linkage and Progressive-X, and the same GPU (GTX 1080Ti) with CONSAC. Our method has three parameters, among which are set as $min\_dist\_thresh=0.2$, $inlier\_thresh=0.3$ and $\gamma\_thresh=0.5$ for our experiments. All the point clouds were downsampled in $0.05m$ voxel size. Our method is not sensitive to the parameter change as the ablation study shown in the supplementary material.

As the metrics used one-to-one registration can not be used
for the multi-instance setting, we adopt three metrics from the retrieval task for evaluation: MHR (Mean Hit Recall), MHP (Mean Hit Precision), MHF1 (Mean Hit F1). Their definitions are described in the supplementary material.

\begin{figure*}[ht]
  \centering
  \begin{subfigure}{0.4\textwidth}
      \centering
      \includegraphics[height=3cm]{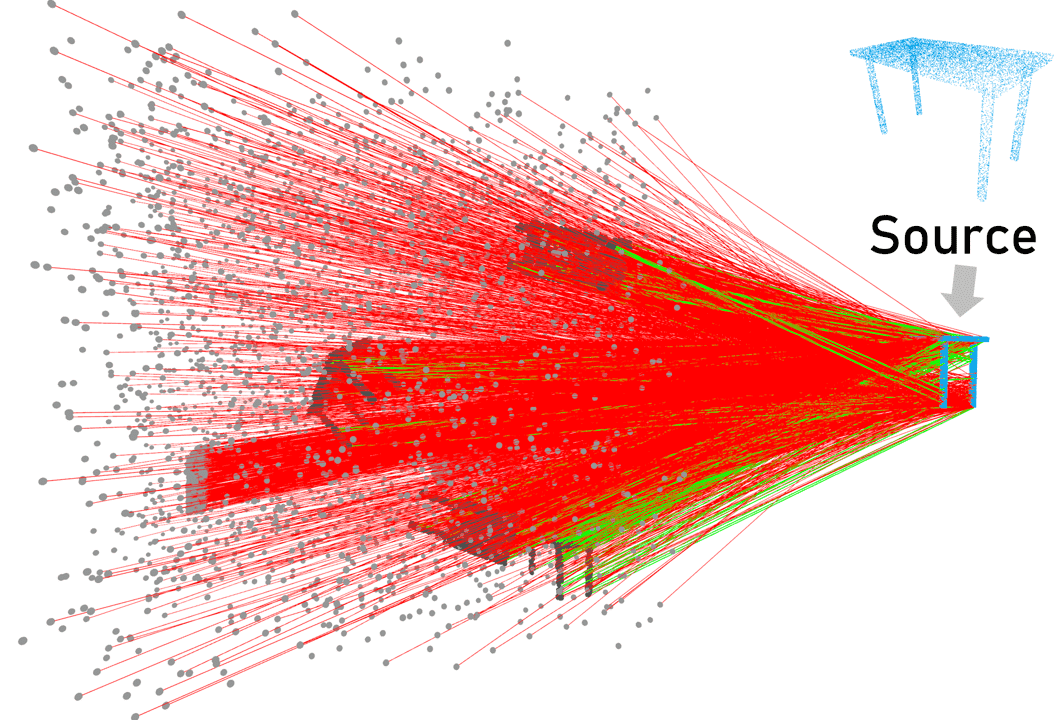}
        \caption{Input correspondences (outlier ratio : $95.5\%$) }
        \label{fig:multi-corrs}
    \end{subfigure}
    \begin{subfigure}{0.45\textwidth}
      \centering
      \includegraphics[height=3cm]{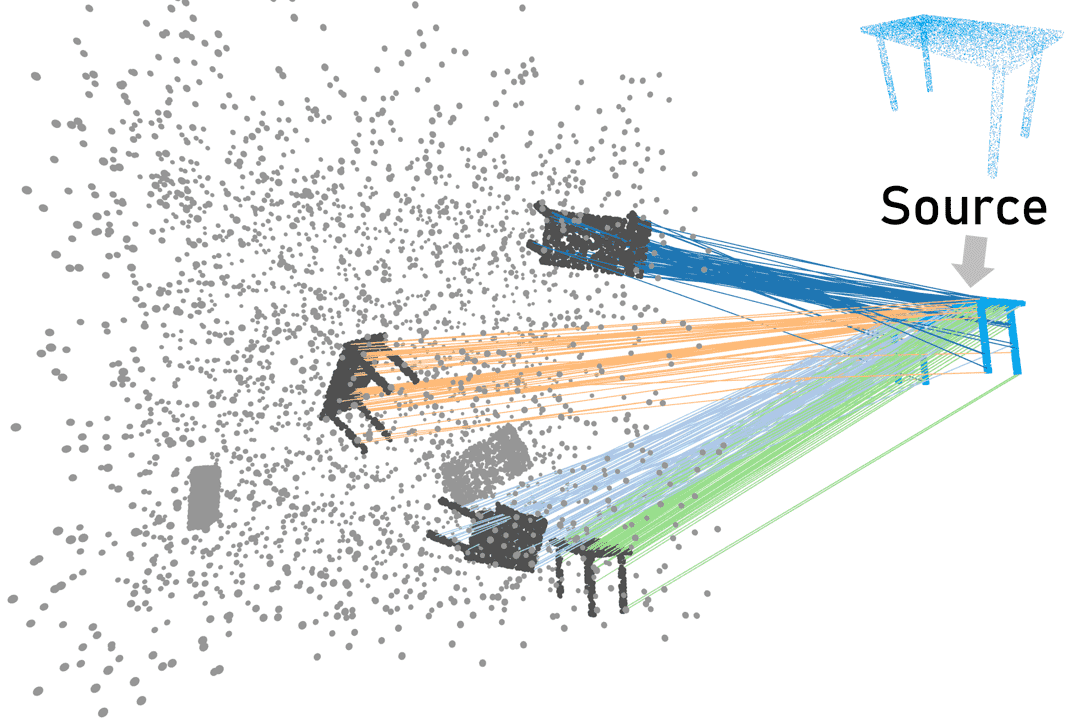}
        \caption{Our clustering result}
        \label{fig:multi-cluster-corrs}
    \end{subfigure}

    \begin{subfigure}{0.18\textwidth}
      \centering
      \includegraphics[height=2.8cm]{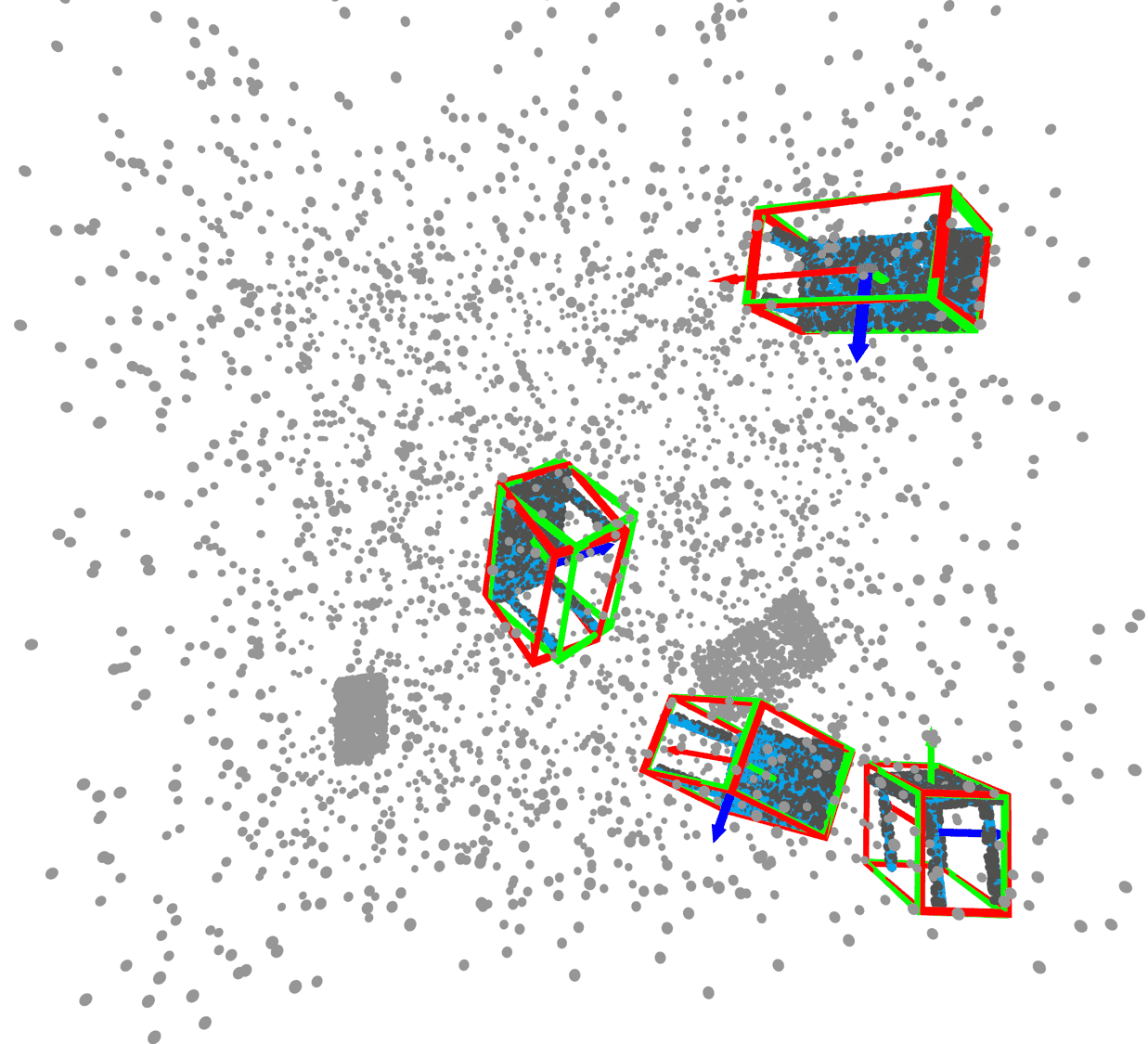}
        \caption{Ours}
        \label{fig:multi-result}
    \end{subfigure}
    \begin{subfigure}{0.18\textwidth}
      \centering
      \includegraphics[height=2.8cm]{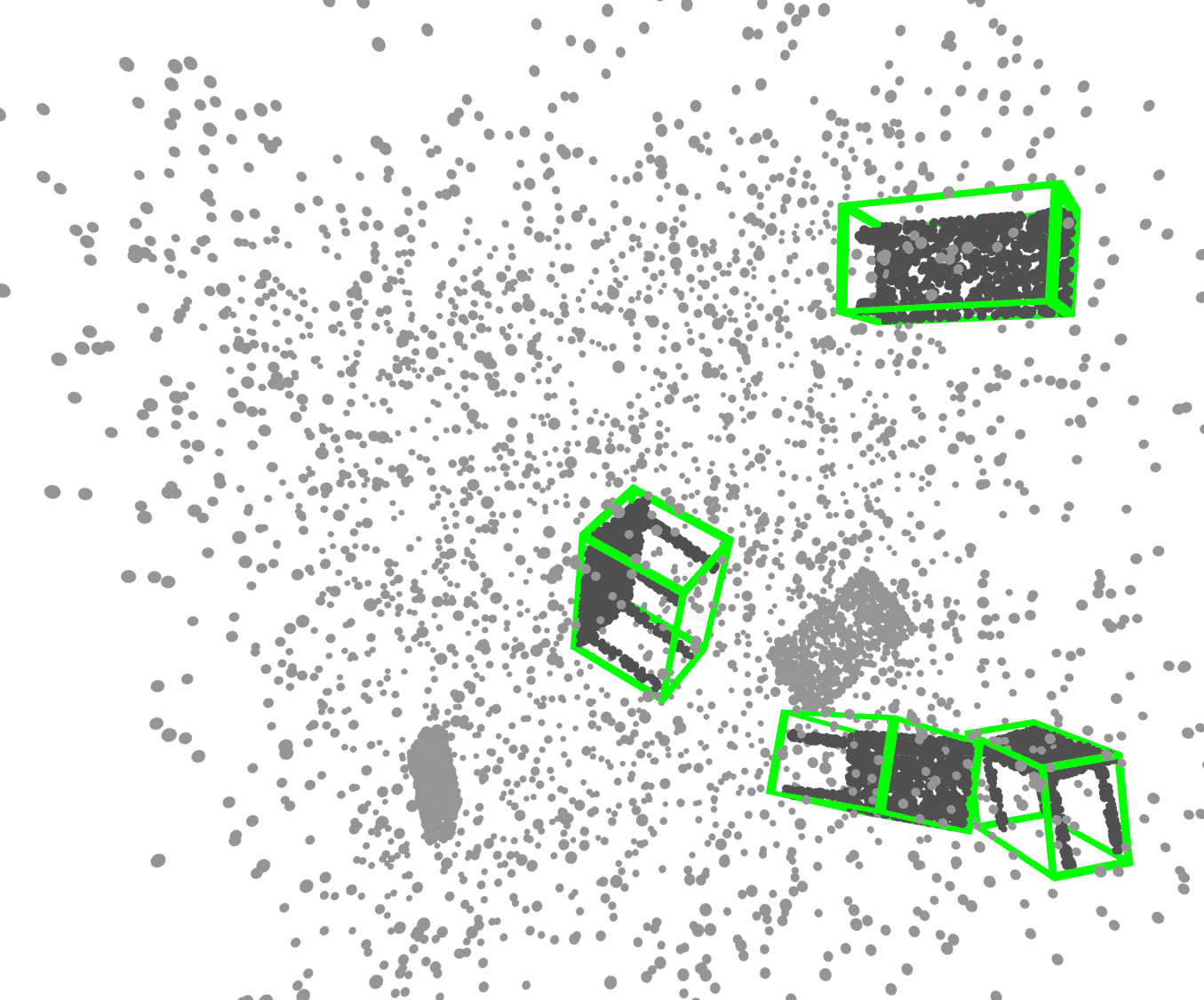}
        \caption{T-Linkage(2014)\cite{Tlinkage}}
        \label{fig:multi-tlinkage1}
    \end{subfigure}
    \begin{subfigure}{0.2\textwidth}
      \centering
      \includegraphics[height=2.8cm]{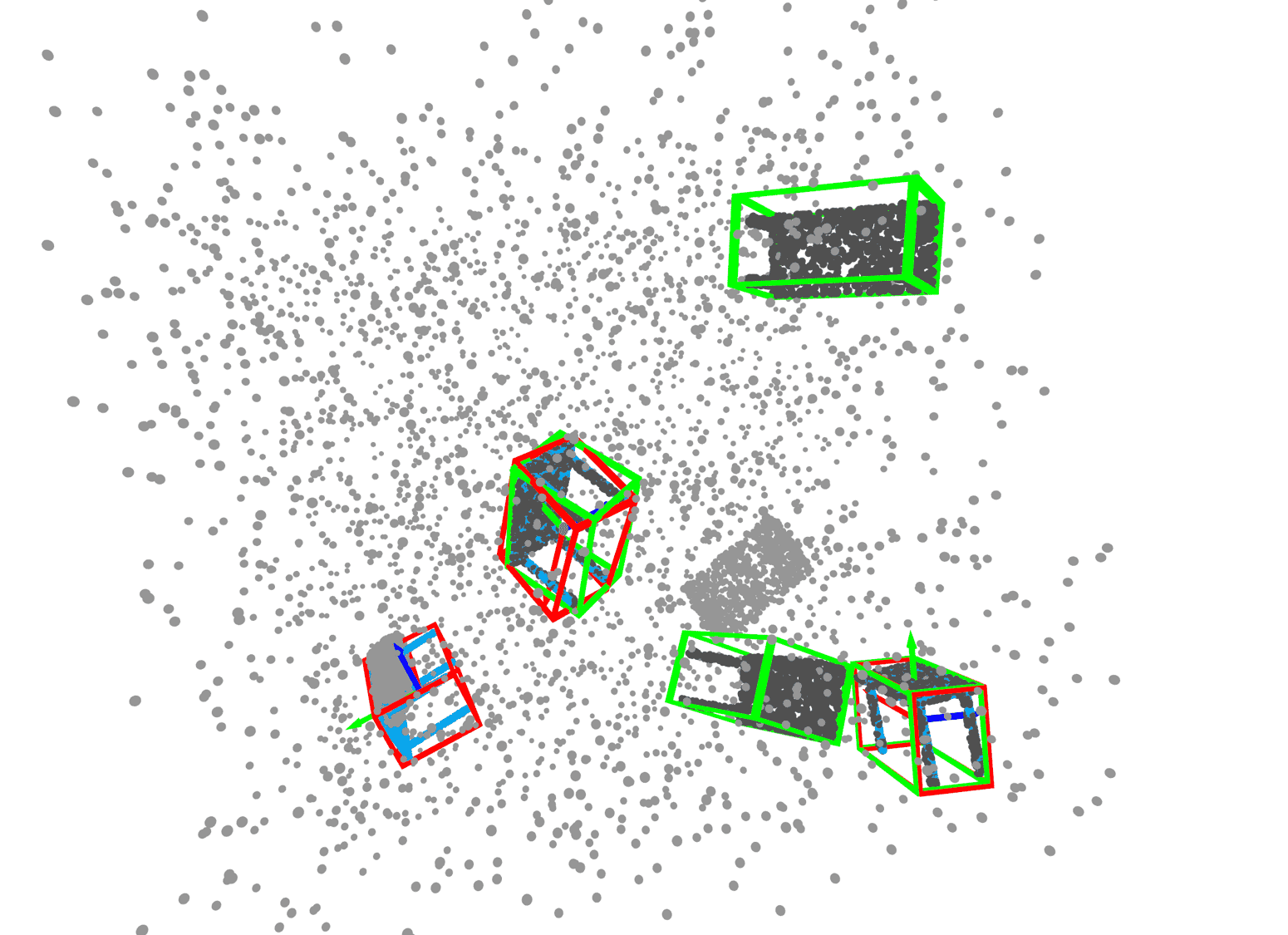}
        \caption{Progressive-X(2019)\cite{ProgressiveX}}
        \label{fig:multi-prox}
    \end{subfigure}
    \begin{subfigure}{0.2\textwidth}
      \centering
      \includegraphics[height=2.8cm]{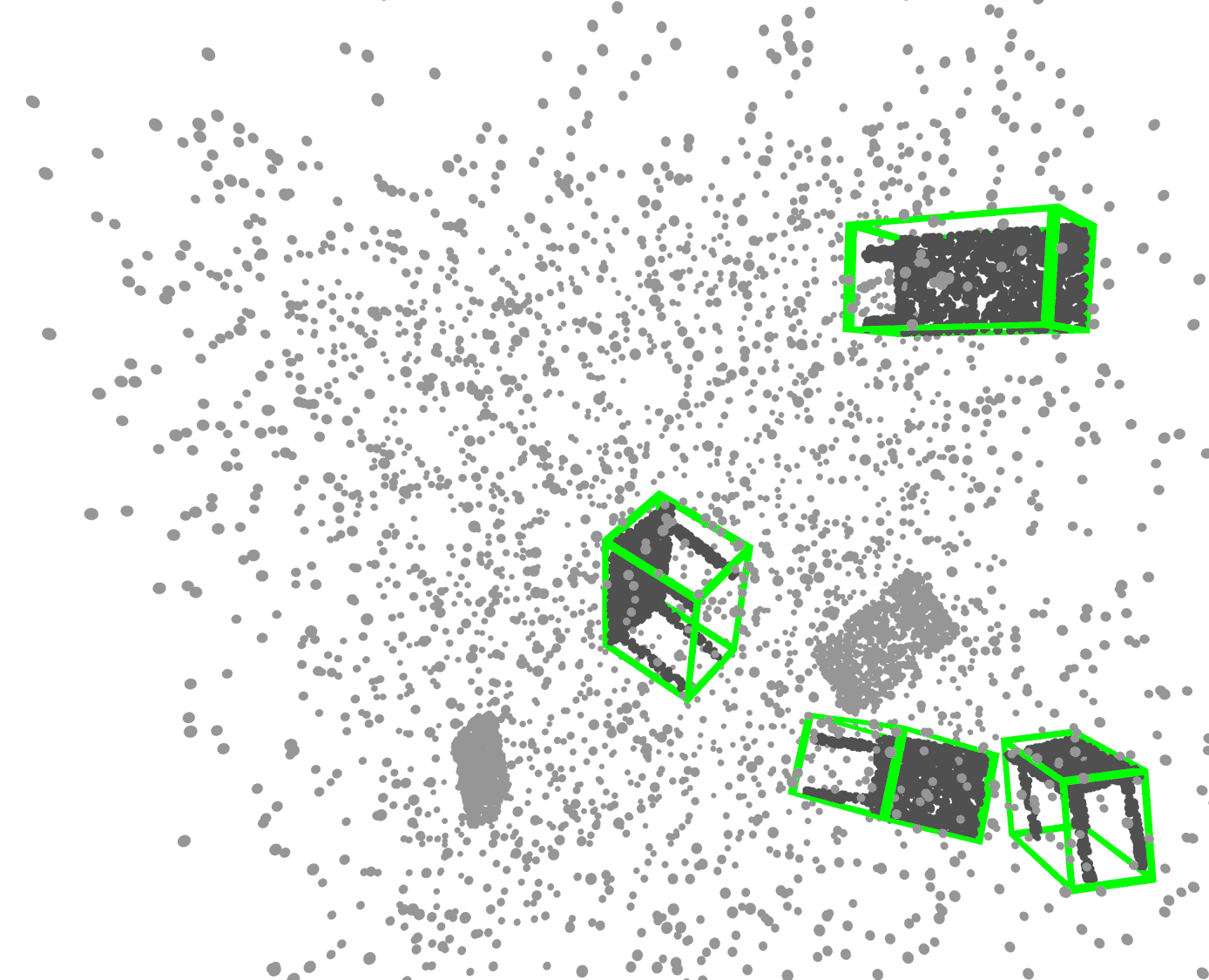}
        \caption{CONSAC(2020)\cite{CONSAC}}
        \label{fig:multi-consac}
    \end{subfigure}
    \begin{subfigure}{0.18\textwidth}
      \centering
      \includegraphics[height=2.8cm]{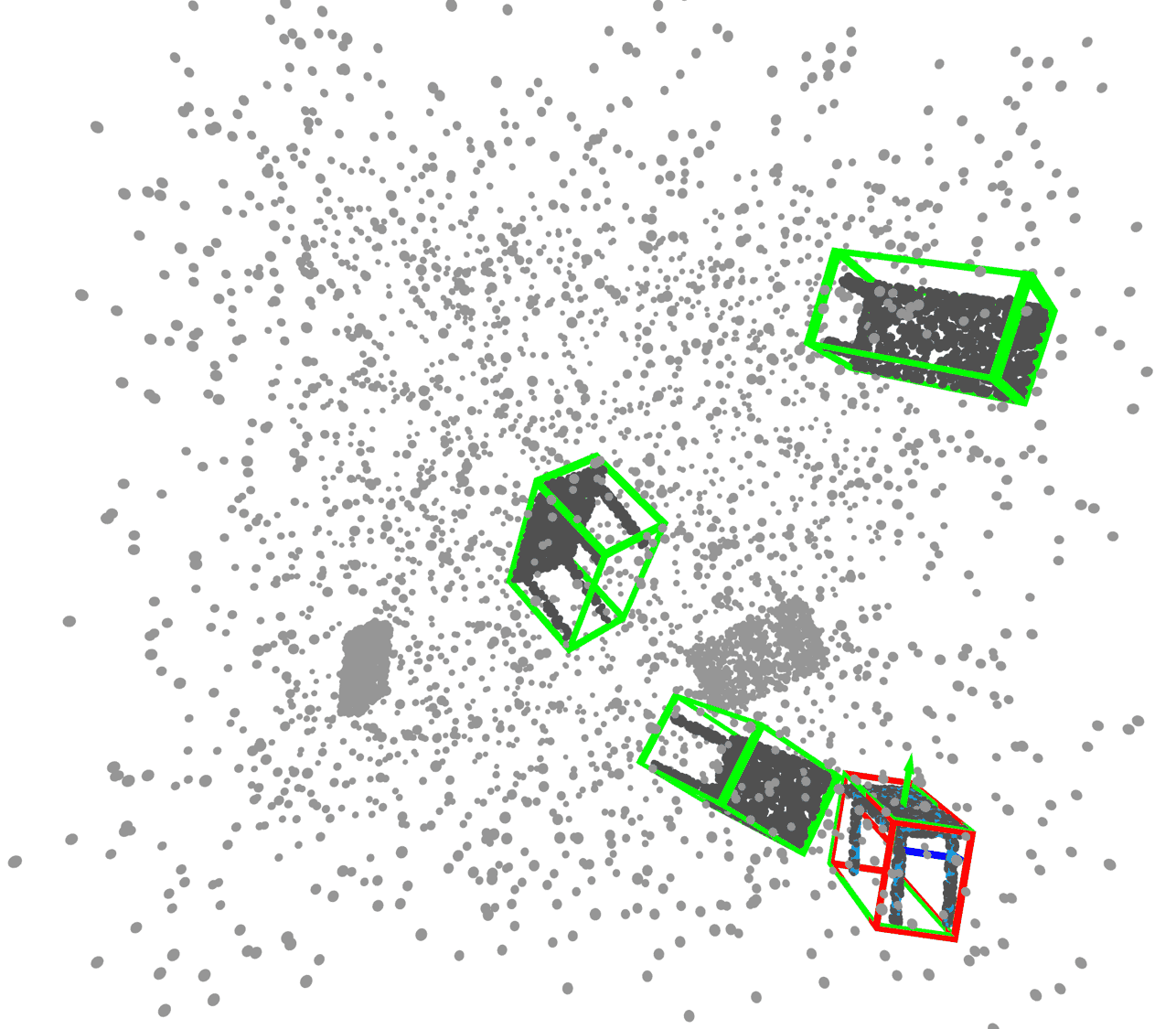}
        \caption{TEASER(2020)\cite{TEASER}}
        \label{fig:multi-teaser}
    \end{subfigure}
    % \begin{subfigure}{0.18\textwidth}
    %   \centering
    %   \includegraphics[height=2.5cm]{scan2cad-cad-ransac1.png}
    %     \caption{RANSAC}
    %     \label{fig:scan2cad_cad-ransac1}
    % \end{subfigure}

%\includegraphics[width=0.9\columnwidth]{figure/} % Reduce the figure size so that it is slightly narrower than the column. Don't use precise values for figure width.This setup will avoid overfull boxes.
\caption{\textbf{Results on the synthetic dataset.} (a) Input correspondences by matching PREDATOR\cite{PREDATOR} features. The inlier and outliers are visualized in green and red respectively. (b) Our clustering result is visualized by different colors (only inliers are shown). In (c-g), we visualize estimated poses in red boxes and ground truth poses in green boxes. Our method (c) registers all instances. T-linkage (d) and CONSAC (f) fail to register any instances. Progressive-X (e) registers 2 instances but produces a wrong registration. TEASER (g) registers one instance.}
\label{fig:predatormm}
\end{figure*}

\subsection{Synthetic datasets}
We generate a synthetic dataset from a pre-sampled Modelnet40 dataset\cite{ModelNet40} from PointNet++\cite{pointnet2}.
We downsample each point cloud to $256$ points and randomly generate $K$ (up to $20$ in our tests) transformations to form a target point cloud. The target point cloud is also mixed with other objects and random points to better mimic real-world cases.

%based on the Modelnet40\cite{ModelNet40} for evaluation. We randomly select a source point cloud $X$ from Modelnet40 and generate $K$ transformations to form one target point cloud $Y$. In our experiment, we set $K=20$. We randomly sample up to $K$ point clouds which is different from $X$. We randomly transform them and add them to $Y$ as noise. In addtion, we generate $3000$ noise points in a fix cube that is enough to cover all the transformed objects. We use a pre-sampled Modelnet40 dataset from PointNet++\cite{pointnet2}, in which each point cloud has $10000$ points, we randomly down-sample them to $256$ points to generate our data. We split the dataset into a train set with $1843$ samples, a validation set with $611$ samples, and a test set with $623$ samples.

\begin{figure}[ht]
    \centering
    \begin{subfigure}{0.21\textwidth}
        \centering   
        \includegraphics[width=\linewidth]{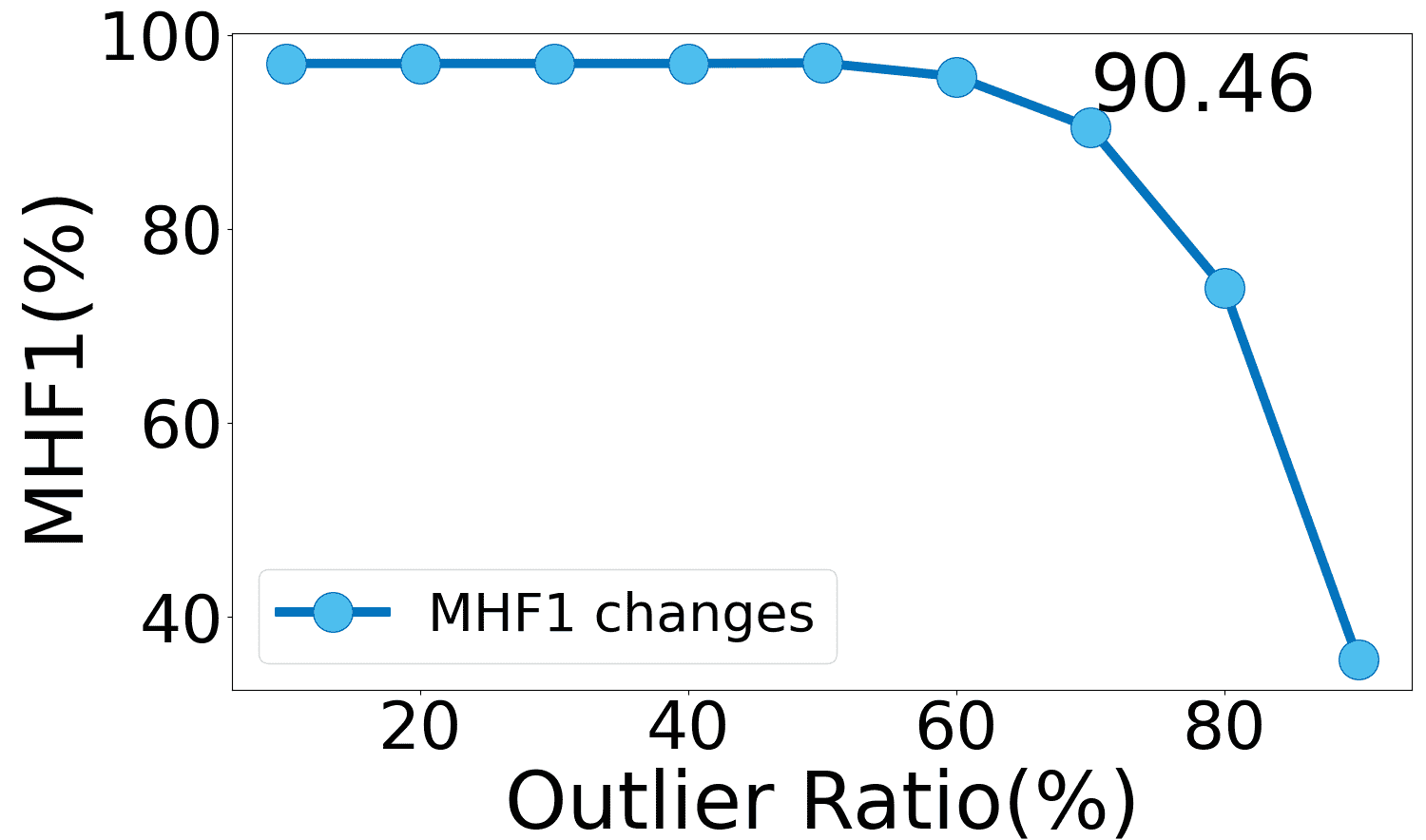}
          \caption{}
      \end{subfigure}
      \begin{subfigure}{0.21\textwidth}
        \centering   
        \includegraphics[width=\linewidth]{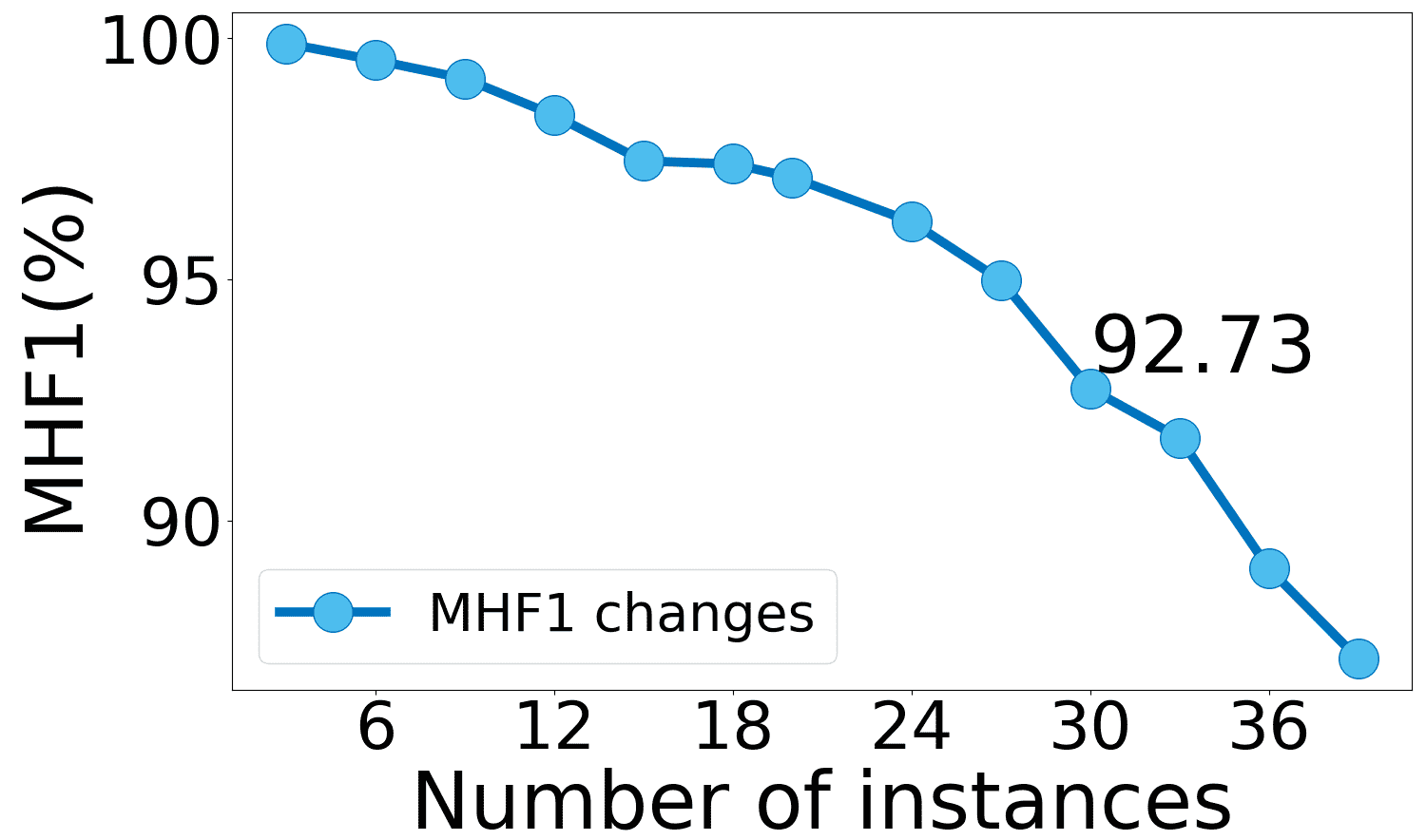}
          \caption{}
          \label{fig:multi-instance}
      \end{subfigure} % Reduce the figure size so that it is slightly narrower than the column. Don't use precise values for figure width.This setup will avoid overfull boxes.
    \caption{(a) Mean Hit F1 vs Outlier Ratio. (b) Mean Hit F1 vs Number of Instances (with a fixed outlier ratio $50\%$).}
          \label{fig:detail-mm}
    %\label{fig:multi-instance}
    \end{figure}

\paragraph{Synthetic correspondences}
In this test, we directly generate the input correspondences by mixing the ground truth and outlier ones.
Different outlier ratios were tested, $10\%\sim50\%$, $50\%\sim70\%$ and $70\%\sim90\%$. Note that the outliers were randomly sampled within a given range for each test sample. The results are shown in Table \ref{tab:mm}. As the outlier ratio increases, the performance of almost all the methods decreases, but ours drops slowly and is still significantly better than other methods. Our algorithm is 10x faster than existing methods either on CPU or on GPU.
We also plot the MHF1(Mean Hit F1) curve of our methods with different outlier ratios in $20$ instances in Figure \ref{fig:detail-mm}(a). Though the performance degrades quickly when the outlier ratio is very large, our method still achieves $90.46\%$ MHF1 with $70\%$ outlier ratio. Figure \ref{fig:detail-mm}(b) shows the MHF1 curve with different instance numbers with a fixed outlier ratio $50\%$. Our method's MHF1 is about $92.73\%$ even when $30$ instances are present.

\begin{table}[ht]
    \centering
\scriptsize
    %\resizebox{.95\columnwidth}{!}{
    \begin{tabular}{ccccc} %& $50\%~70\%$ & $70\%~90\%$
        \toprule
        % & MHR$\left( \% \right) \uparrow $ & MRRE $\left( \right) \downarrow $ & MRTE & Time\\
        \textbf{Metric}& MHR$\left( \% \right) \uparrow $ & MHP$\left( \% \right) \uparrow $ & MHF1$\left( \% \right) \uparrow $ & Time$\left( s \right) \downarrow $\\
        \hline
        \multicolumn{5}{c}{Outlier ratio : $10\%\sim50\%$} \\
        \hline
        T-Linkage & 3.05 & 14.80 & 4.65& 57.27 \\
        Progressive-X & 27.91 & 80.28 & 41.04 & 87.25\\
        CONSAC & 0.47 & 0.47 & 0.47 & 9.23  \\
        \textbf{Ours} & \textbf{96.08} & \textbf{99.73} & \textbf{97.03} & \textbf{0.62/0.30} \\ % 前gt_num个预测值的recall
        \hline
        \multicolumn{5}{c}{Outlier ratio : $50\%\sim70\%$} \\
        \hline
        %\hline
        % \textbf{metric} & MHR$\left( \% \right) \uparrow $ & RRE$\left( \degree \right) \downarrow $ & RTE$\left( m \right) \downarrow $ & t$\left( s \right) \downarrow $\\
        %\hline
        %\hline
        T-Linkage & 1.33 & 7.00 & 2.05 & 56.90 \\
        Progressive-X & 20.60 & 75.10 & 31.70 & 85.54  \\
        CONSAC & 0.49 & 0.49 & 0.49 & 9.55\\
        \textbf{Ours} & \textbf{93.99} & \textbf{99.49} & \textbf{95.51} & \textbf{0.55/0.28}\\
        \hline
        
        \multicolumn{5}{c}{Outlier ratio : $70\%\sim90\%$} \\
        \hline
        %\hline
        % \textbf{metric} & MHR$\left( \% \right) \uparrow $ & RRE$\left( \degree \right) \downarrow $ & RTE$\left( m \right) \downarrow $ & t$\left( s \right) \downarrow $\\
        %\hline
        %\hline
        T-Linkage & 0.81 & 4.42 & 1.25 & 56.89 \\
        Progressive-X & 12.88 & 62.60 & 20.73 & 84.5 \\
        CONSAC & 0.51 & 0.51 & 0.51 & 7.70  \\
        \textbf{Ours} & \textbf{60.39} & \textbf{94.42} & \textbf{69.36} &\textbf{0.50/0.24}\\

        \hline
        \multicolumn{5}{c}{Outlier ratio : $90\%\sim99\%$} \\
        \hline
        T-Linkage & 0.28 & 1.30 & 0.42 & 56.69 \\
        Progressive-X & 7.13 & 39.19 & 11.67 & 84.43\\
        CONSAC & 0.51 & 0.51 & 0.51 & 9.57  \\
        \textbf{Ours} & \textbf{14.70} & \textbf{65.20} & \textbf{22.75} & \textbf{0.47/0.21} \\ % 前gt_num个预测值的recall
        \bottomrule
        %outlier ratio & $10\%~50\%$ & $50\%~70\%$ & $70\%~90\%$\\

    \end{tabular}
    \caption{Results on synthetic correspondences with different outlier ratios. $\uparrow$ means the larger the better, while $\downarrow$ indicates the contrary. The running time on CPU/GPU of our method is presented.}
    \label{tab:mm}
    \end{table}

    \begin{figure*}[ht]
        \centering
        \begin{subfigure}{0.41\textwidth}
            \centering
            \includegraphics[height=2.8cm]{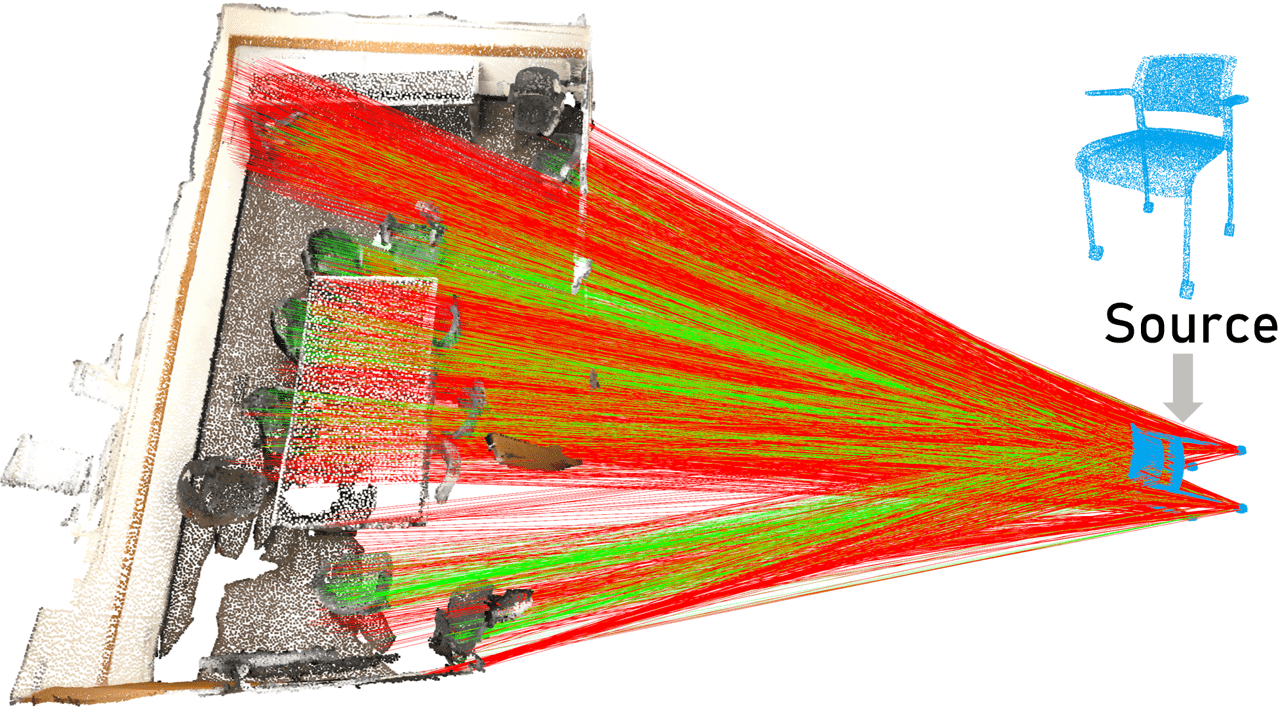}
              \caption{Input correspondences}
              \label{fig:scan2cad_cad-input-corrs}
          \end{subfigure}
          \begin{subfigure}{0.41\textwidth}
            \centering
            \includegraphics[height=2.8cm]{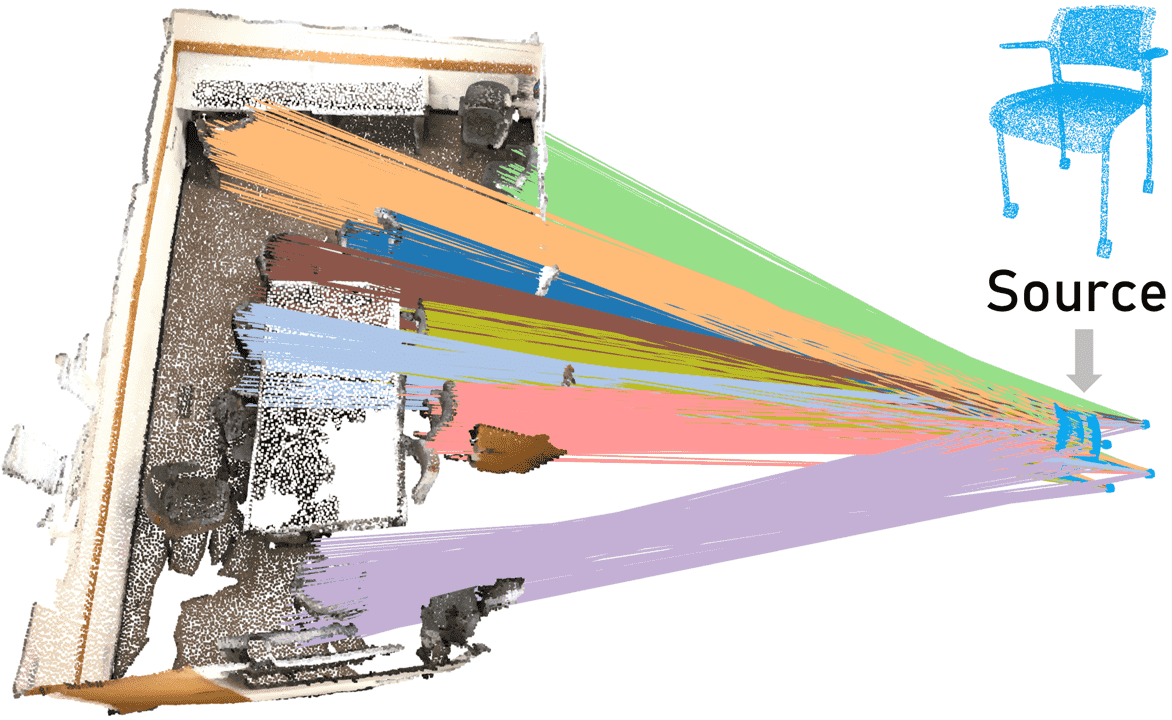}
              \caption{Our clustering result}
              \label{fig:scan2cad_cad-cluster-corrs}
          \end{subfigure}

          \begin{subfigure}{0.18\textwidth}
            \centering
            \includegraphics[height=3.3cm]{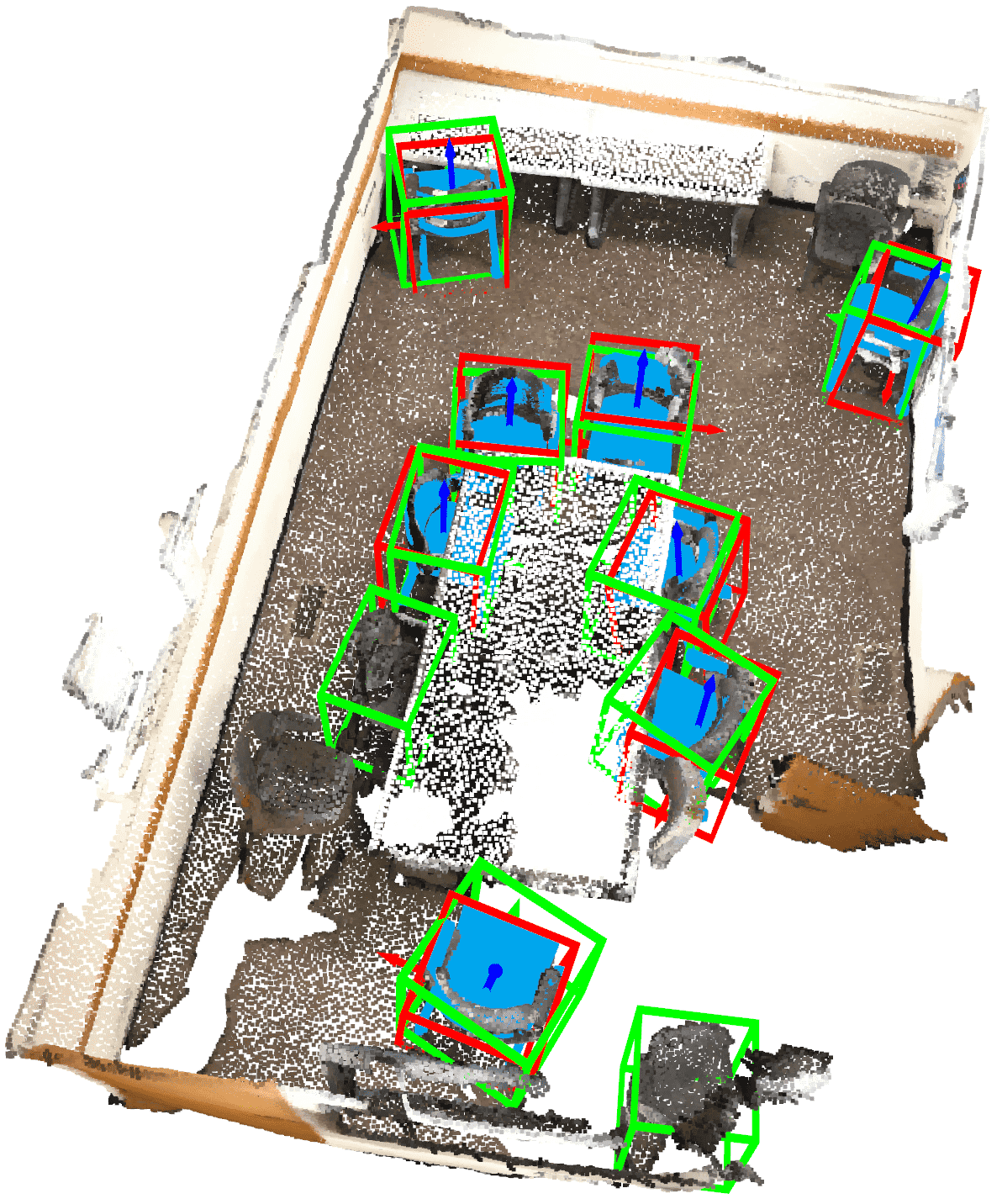}
              \caption{Ours}
              \label{fig:scan2cad_cad-result}
          \end{subfigure}
          \begin{subfigure}{0.18\textwidth}
            \centering
            \includegraphics[height=3.3cm]{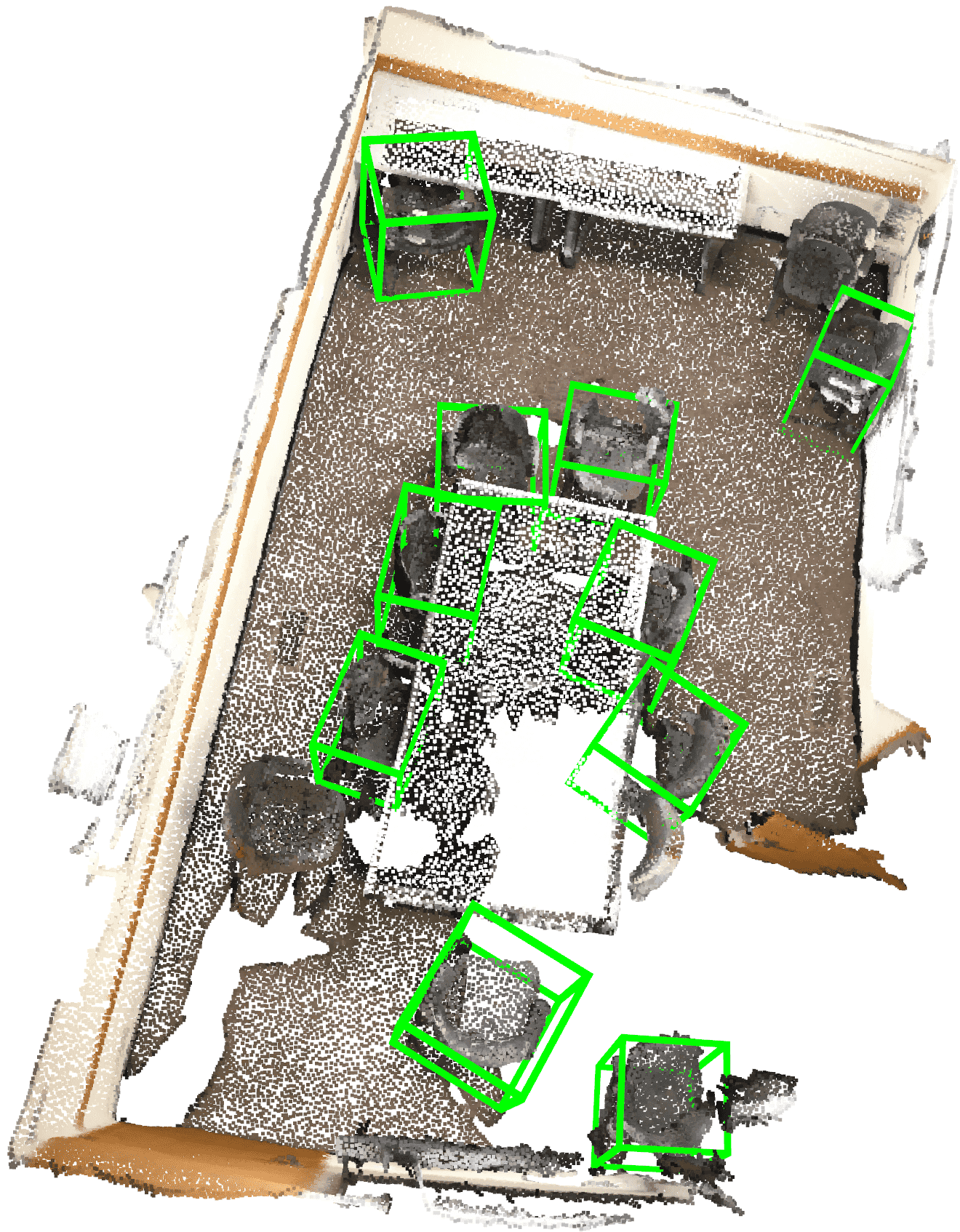}
              \caption{T-linkage(2014) \cite{Tlinkage}}
              \label{fig:scan2cad_cad-tlinkage}
          \end{subfigure}
          \begin{subfigure}{0.19\textwidth}
            \centering
            \includegraphics[height=3.3cm]{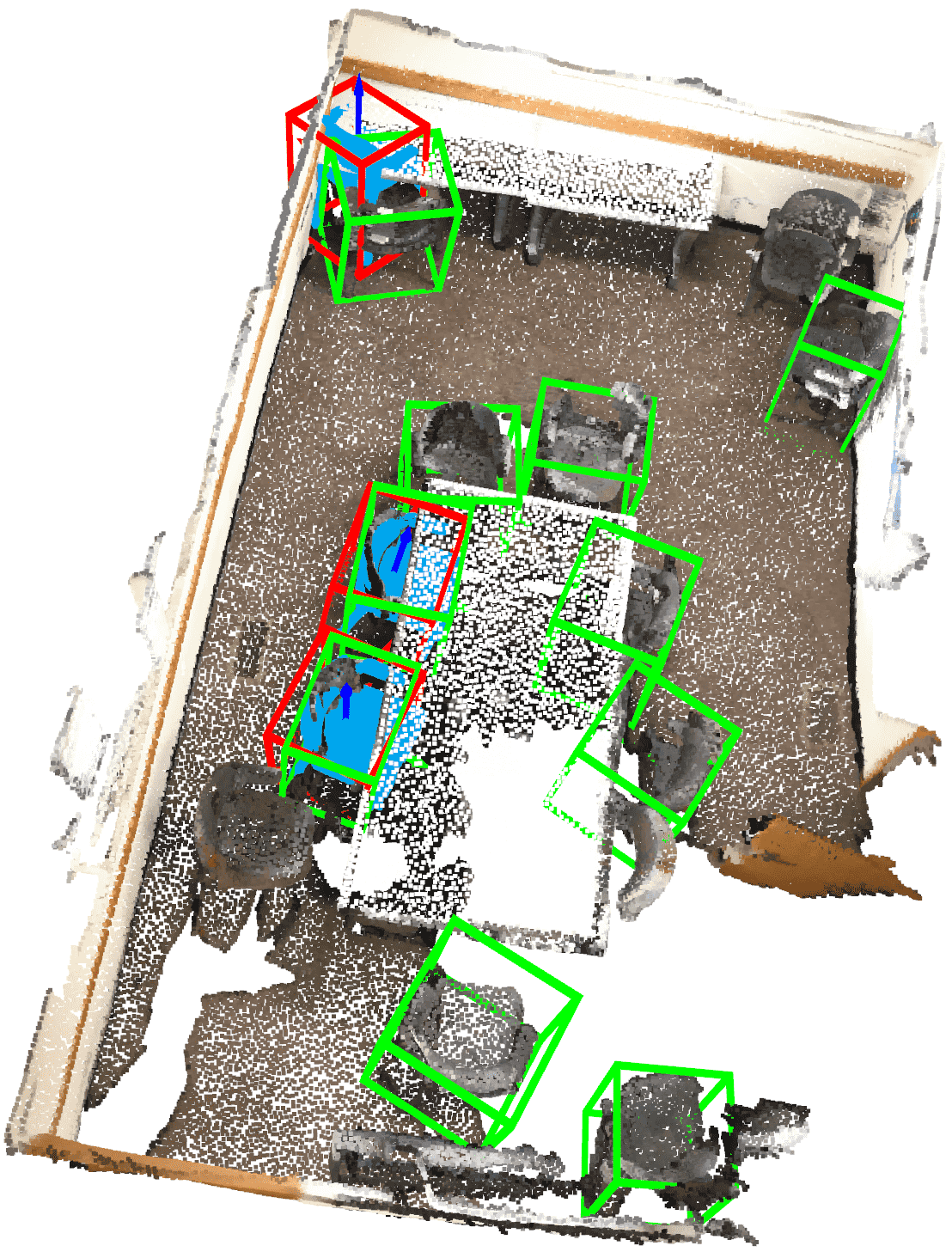}
              \caption{Progressive-X(2019) \cite{ProgressiveX}}
              \label{fig:scan2cad_cad-prox}
          \end{subfigure}
          \begin{subfigure}{0.17\textwidth}
            \centering
            \includegraphics[height=3.3cm]{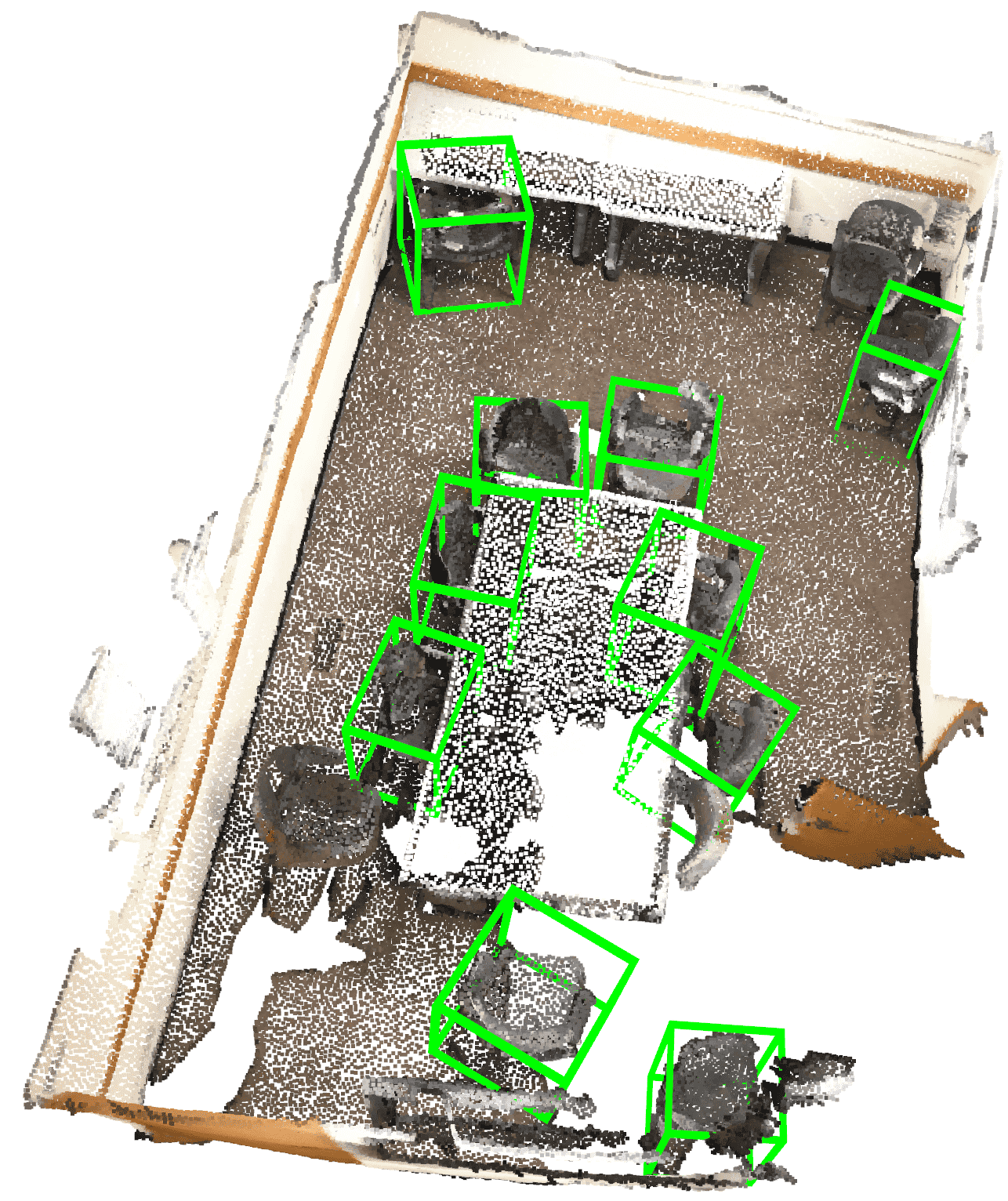}
              \caption{CONSAC(2020) \cite{CONSAC}}
              \label{fig:scan2cad_cad-consac}
          \end{subfigure}
          \begin{subfigure}{0.18\textwidth}
            \centering
            \includegraphics[height=3.3cm]{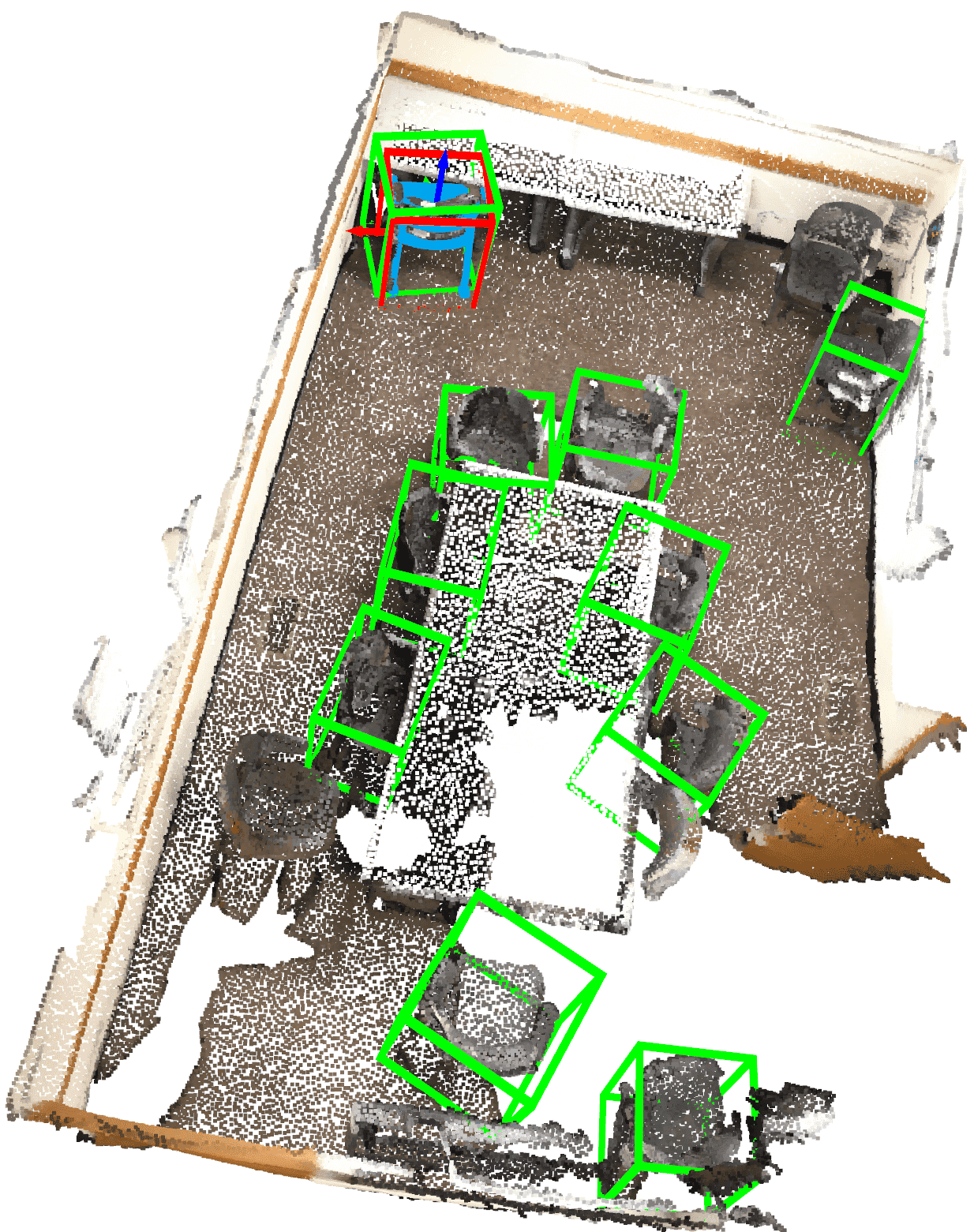}
              \caption{TEASER(2020) \cite{TEASER}}
              \label{fig:scan2cad_cad-teaser}
          \end{subfigure}
          % \begin{subfigure}{0.15\textwidth}
          %   \centering
          %   \includegraphics[height=3cm]{scan2cad-cad-ransac.png}
          %     \caption{RANSAC}
          %     \label{fig:scan2cad_cad-ransac}
          % \end{subfigure}
          \caption{\textbf{Scan2CAD results.} (a) Input correspondences by matching PREDATOR\cite{PREDATOR} features. The inlier and outliers are visualized in green and red respectively. (b) Our clustering result is visualized by different colors (only inliers are shown). In (c-g), we visualize estimated poses in red boxes and ground truth poses in green boxes. Our method (c) correctly aligns 8 instances. T-Linkage (d) and CONSAC (f) fail to register any instances. Progressive-X (e) register 3 instances. TEASER (g) registers one instance.}
        \label{fig:Scan2CAD-cadresult}
        \end{figure*}

      \begin{figure*}[ht]
          \centering
          \begin{subfigure}{0.4\textwidth}
              \centering
              \includegraphics[height=2.8cm]{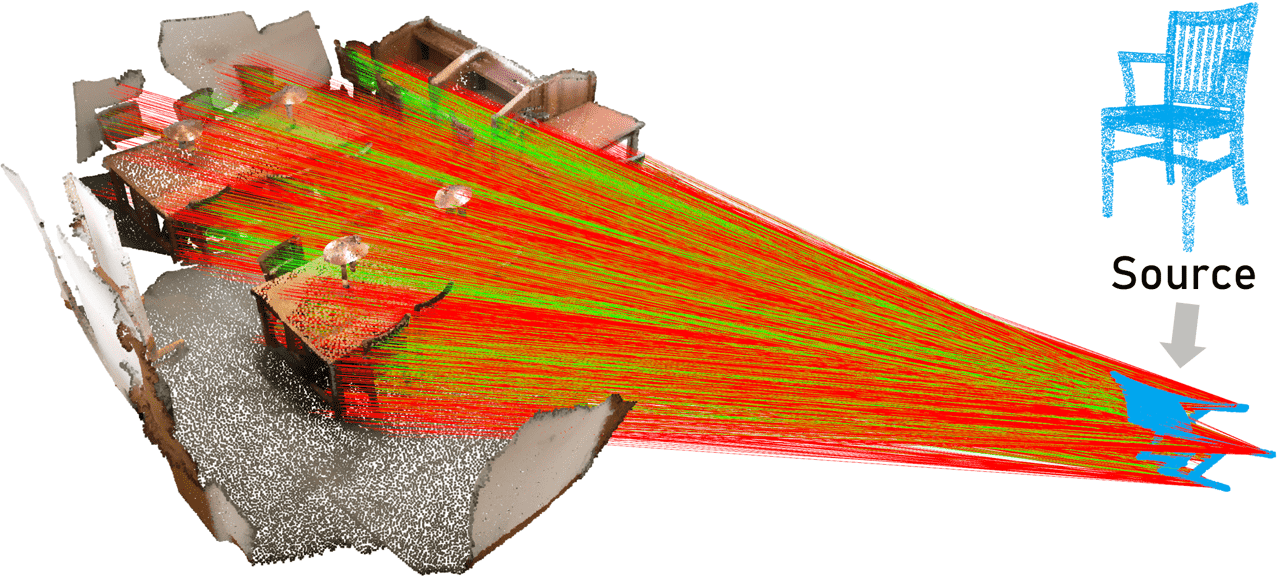}
                \caption{Input correspondences}
                \label{fig:scan2cad_cad-input-corrs1}
            \end{subfigure}
            \begin{subfigure}{0.45\textwidth}
              \centering
              \includegraphics[height=2.8cm]{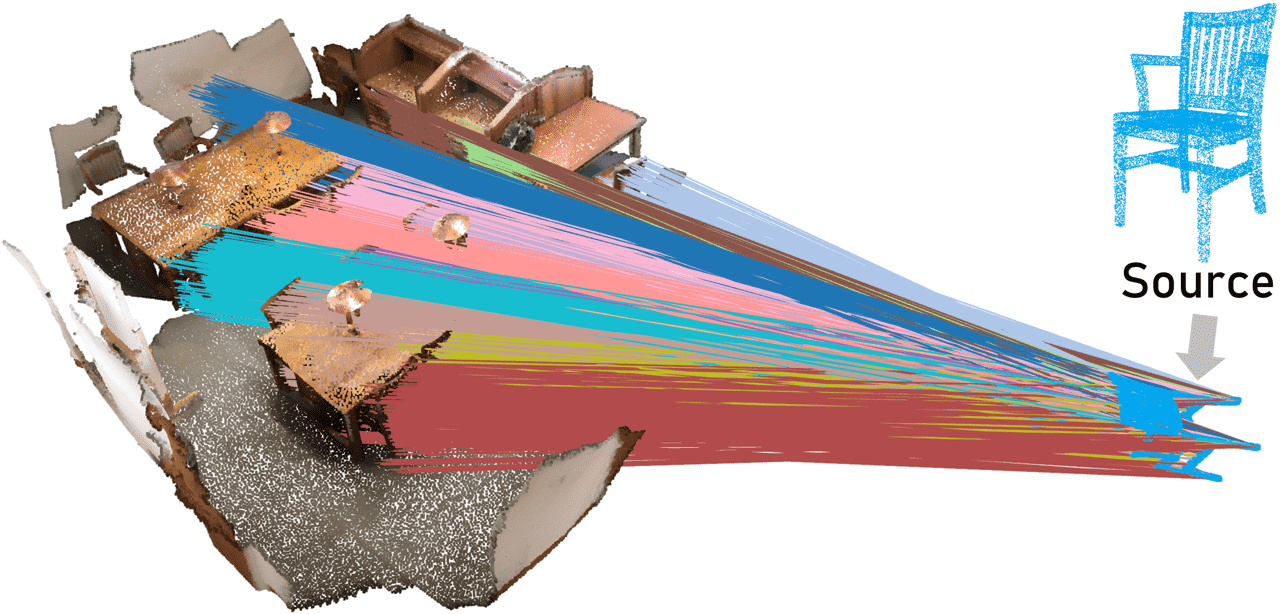}
                \caption{Our clustering result}
                \label{fig:scan2cad_cad-cluster-corrs1}
            \end{subfigure}

            \begin{subfigure}{0.18\textwidth}
              \centering
              \includegraphics[height=2.8cm]{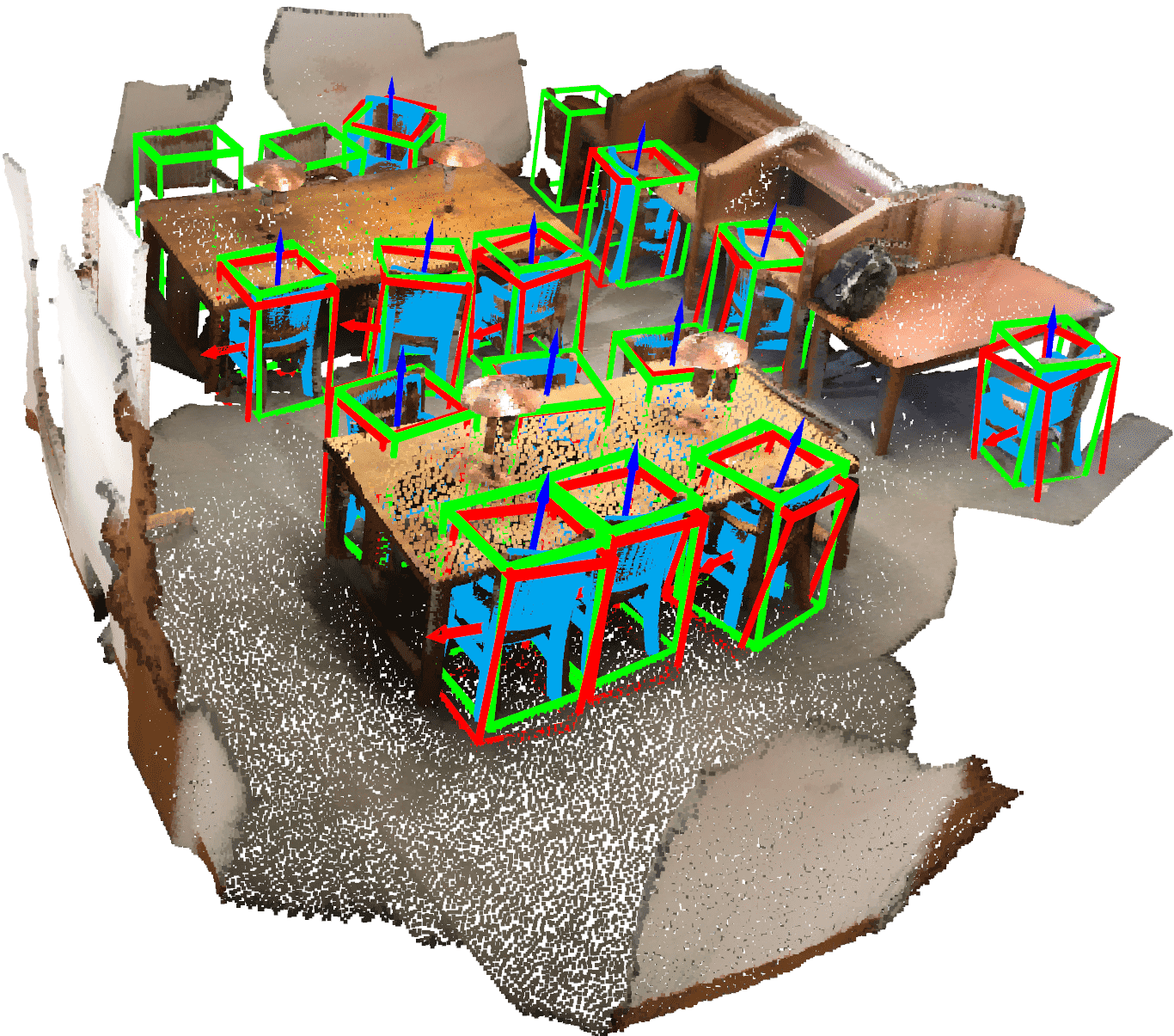}
                \caption{Ours}
                \label{fig:scan2cad_cad-result1}
            \end{subfigure}
            \begin{subfigure}{0.18\textwidth}
              \centering
              \includegraphics[height=2.8cm]{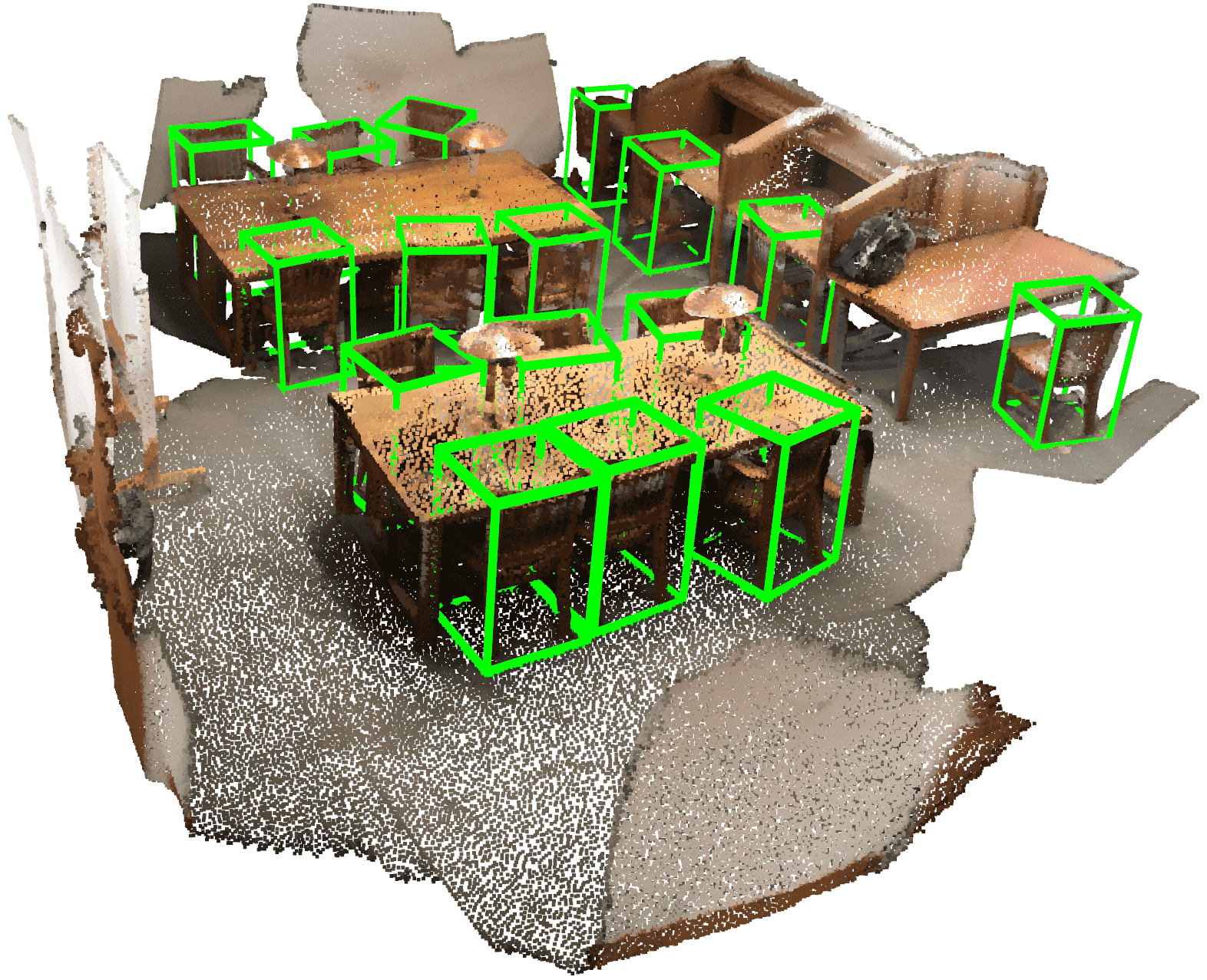}
                \caption{T-linkage(2014) \cite{Tlinkage}}
                \label{fig:scan2cad_cad-tlinkage1}
            \end{subfigure}
            \begin{subfigure}{0.2\textwidth}
              \centering
              \includegraphics[height=2.8cm]{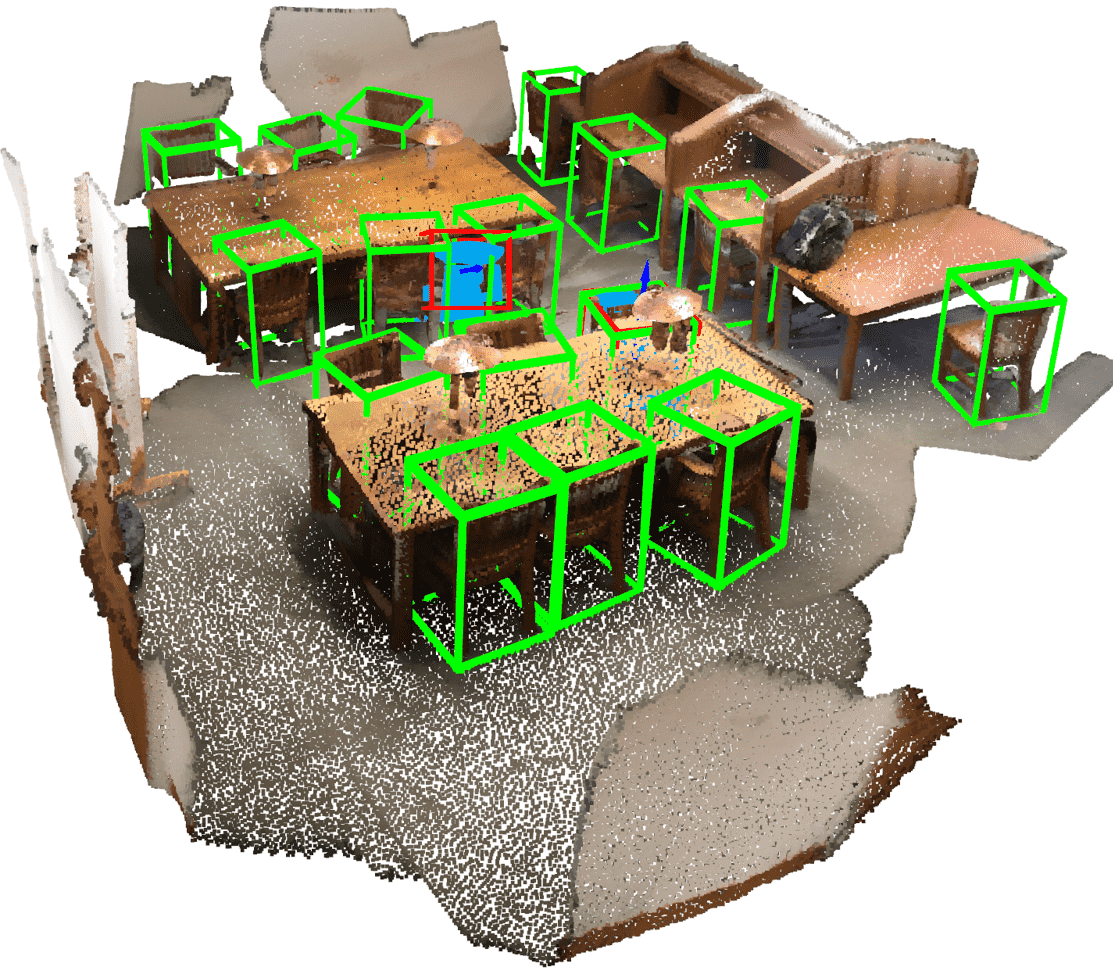}
                \caption{Progressive-X(2019) \cite{ProgressiveX}}
                \label{fig:scan2cad_cad-prox1}
            \end{subfigure}
            \begin{subfigure}{0.18\textwidth}
              \centering
              \includegraphics[height=2.8cm]{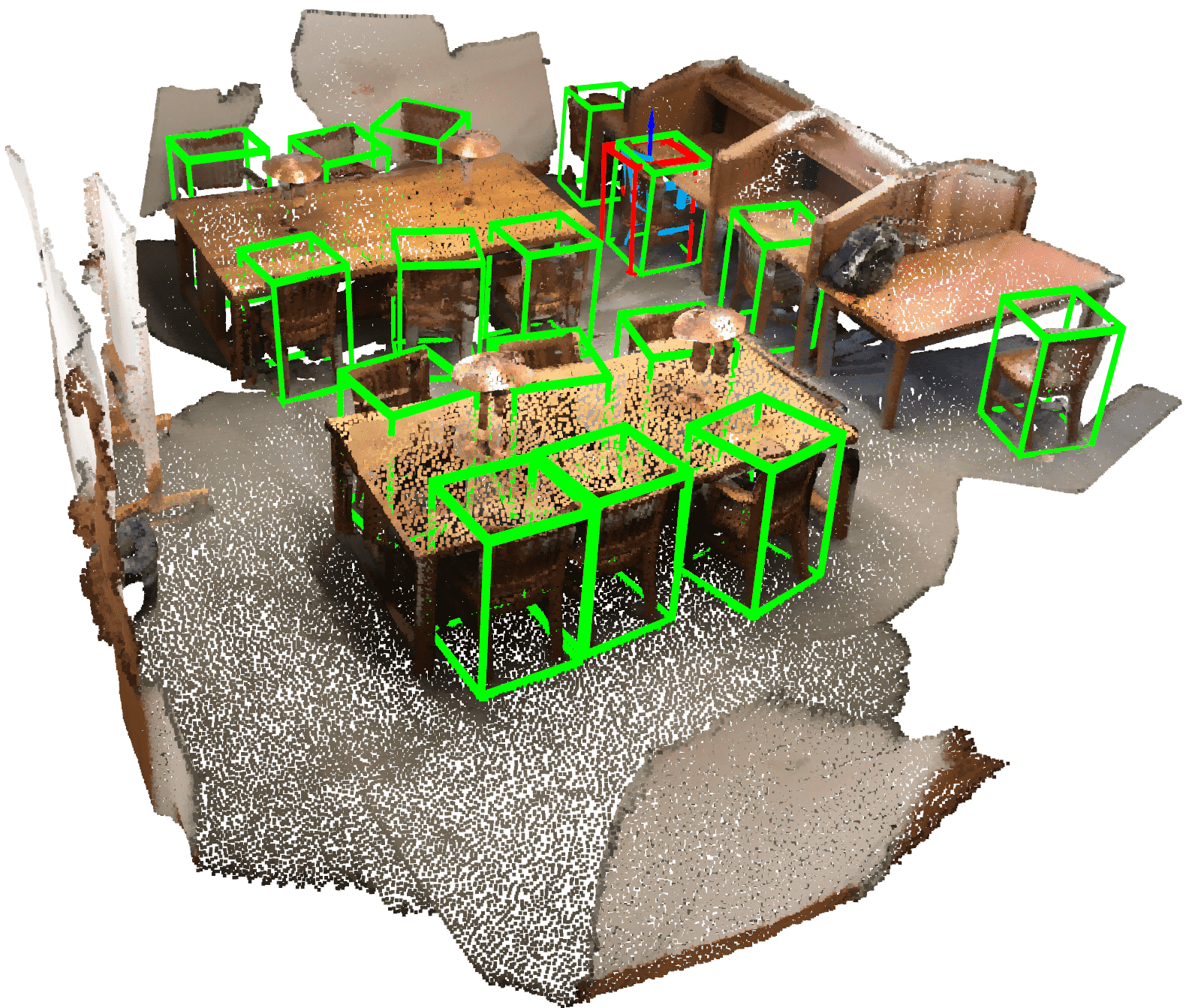}
                \caption{CONSAC(2020)\cite{CONSAC}}
                \label{fig:scan2cad_cad-consac1}
            \end{subfigure}
            \begin{subfigure}{0.18\textwidth}
              \centering
              \includegraphics[height=2.8cm]{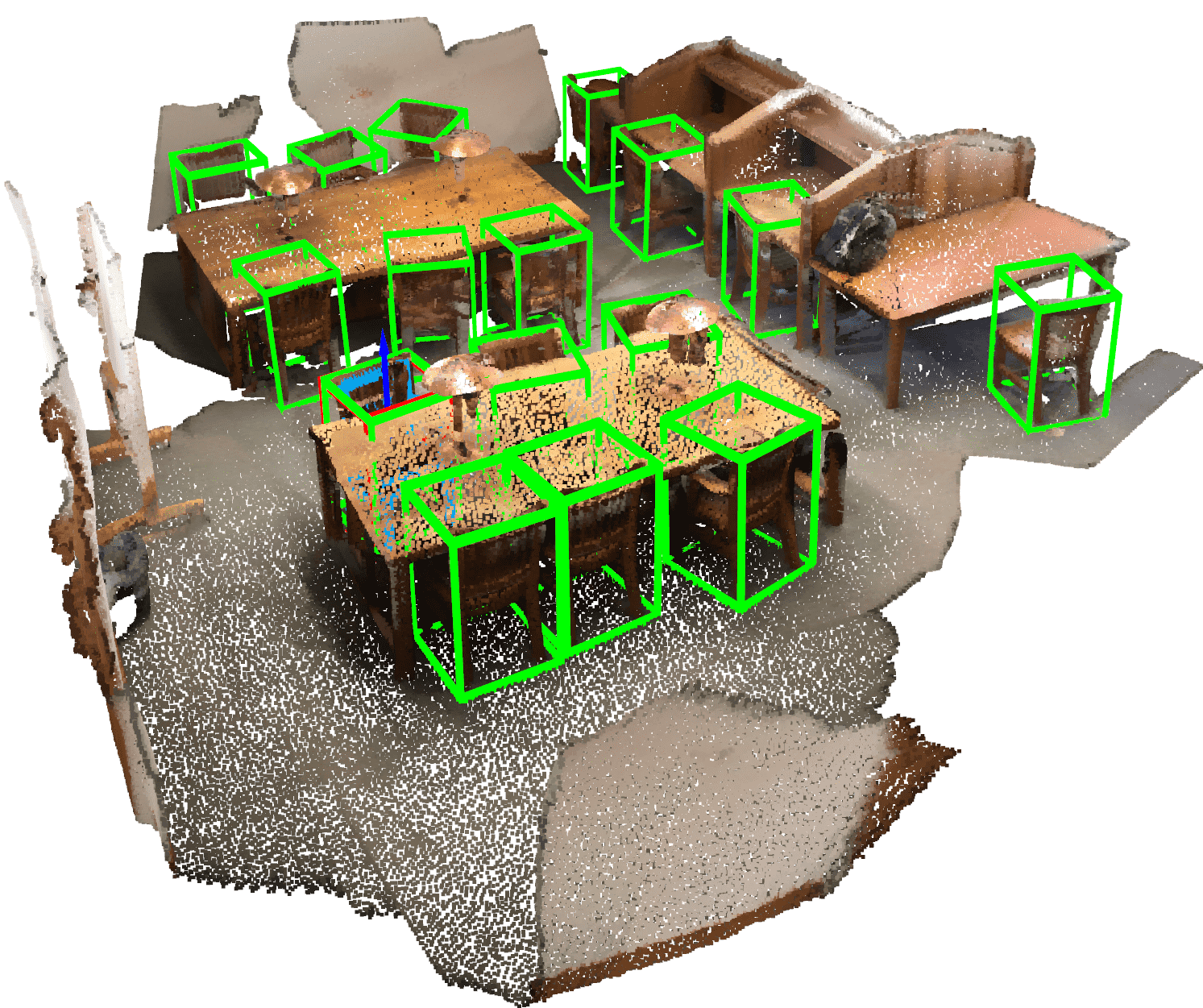}
                \caption{TEASER(2020)\cite{TEASER}}
                \label{fig:scan2cad_cad-teaser1}
            \end{subfigure}
            % \begin{subfigure}{0.18\textwidth}
            %   \centering
            %   \includegraphics[height=2.5cm]{scan2cad-cad-ransac1.png}
            %     \caption{RANSAC}
            %     \label{fig:scan2cad_cad-ransac1}
            % \end{subfigure}
            \caption{\textbf{Scan2CAD results.} Our method (c) registers $13$ instances among $16$ chairs. Progressive-X (e) registers 2 instances, but one of them has a large pose error. CONSAC (f) and TEASER (g) align one instance. T-Linkage (d) fails to register any instances.}
        \label{fig:Scan2CAD-cadresult1}
          \end{figure*}

\paragraph{Correspondences by feature matching}
In this test, we apply feature matching to obtain the point correspondences by the PREDATOR\cite{PREDATOR} and D3Feat\cite{D3Feat}. Both feature models were trained on synthetic data. The results are shown in Table \ref{tab:realcorr}.  Note that both features produce correspondences with a high outlier ratio greater than $90\%$. In such a challenging case, our method still performs well and much better than existing methods in terms of both robustness and efficiency.  The results of using D3Feat are much worse than those of using PREDATOR. The reason is that not only because of the presence of more outliers but also missing inliers as we examine the results.  We visualize some results in Figure \ref{fig:predatormm}.
% In this test, we apply feature matching to obtain the point correspondences by the PREDATOR\cite{PREDATOR} and D3Feat\cite{D3Feat}. Both models were trained on the synthetic dataset. Figure \ref{fig:predatormm}(a) illustrates an example of input correspondences, where a source point may correspond to multiple target points. The results are shown in Table \ref{tab:realcorr}.  Note that both features produce correspondences with a high outlier ratio greater than $90\%$. In such a challenging case, our method still performs well and much better than existing methods in terms of both robustness and efficiency.  The results of using D3Feat are much worse than those of using PREDATOR. The reason is not only because of the presence of more outliers but also missing inliers as we examine the results.  We visualize some results in Figure \ref{fig:predatormm}.

\begin{table}[ht]
\scriptsize
    \centering
        \begin{tabular}{ccccc} %& $50\%~70\%$ & $70\%~90\%$
            % \toprule
            % \textbf{Metric}& MHR$\left( \% \right) \uparrow $& MHP$\left( \% \right) \uparrow $& MHF1$\left( \% \right) \uparrow $ & Time$\left( s \right) \downarrow $\\
            % \hline
            % & \multicolumn{4}{c}{PREDATOR ( estimated outlier ratio : $94.32\%$)} \\
            % \hline
            % T-Linkage & 0.41 & 0.70 & 0.41 & 43.46 \\
            % Progressive-X & 12.34 & 17.94 & 13.18 & 86.39 \\
            % CONSAC & 0.71 & 0.25 & 0.33 & 7.65 \\
            % \textbf{Ours} &\textbf{51.89} & \textbf{60.15} & \textbf{46.69} & \textbf{1.28/0.48} \\
            % \hline
            
            % &\multicolumn{4}{c}{D3Feat ( estimated outlier ratio : $99.30\%$)} \\
            % \hline
            % %\hline
            % % \textbf{metric} & MHR$\left( \% \right) \uparrow $ & RRE$\left( \degree \right) \downarrow $ & RTE$\left( m \right) \downarrow $ & t$\left( s \right) \downarrow $\\
            % %\hline
            % %\hline
            % T-Linkage & 0.08 & 0.08 & 0.08 & 56.37  \\
            % Progressive-X & 3.42 & 9.12 & 4.39 & 87.22 \\
            % CONSAC & 0.71 & 0.25 & 0.33 & 9.53 \\
            % \textbf{Ours} & \textbf{9.26} & \textbf{12.34} & \textbf{9.02} & \textbf{0.68/0.30} \\
            % \bottomrule
            \toprule
            \textbf{Metric}& MHR$\left( \% \right) \uparrow $& MHP$\left( \% \right) \uparrow $& MHF1$\left( \% \right) \uparrow $ & Time$\left( s \right) \downarrow $\\
            \hline
            & \multicolumn{4}{c}{PREDATOR ( estimated outlier ratio : $94.32\%$)} \\
            \hline
            T-Linkage & 0.19 & 0.54 & 0.27 & 43.46 \\
            Progressive-X & 15.90 & 31.01 & 18.98 & 86.39 \\
            CONSAC & 0.1 & 0.07 & 0.08 & 7.65 \\
            \textbf{Ours} &\textbf{53.39} & \textbf{61.44} & \textbf{51.80} & \textbf{1.28/0.48} \\
            \hline
            
            &\multicolumn{4}{c}{D3Feat ( estimated outlier ratio : $99.30\%$)} \\
            \hline
            %\hline
            % \textbf{metric} & MHR$\left( \% \right) \uparrow $ & RRE$\left( \degree \right) \downarrow $ & RTE$\left( m \right) \downarrow $ & t$\left( s \right) \downarrow $\\
            %\hline
            %\hline
            T-Linkage & 0.07 & 0.29 & 0.1 & 56.37  \\
            Progressive-X & 4.29 & 15.28 & 5.94 & 87.22 \\
            CONSAC & 0.13 & 0.04 & 0.05 & 9.53 \\
            \textbf{Ours} & \textbf{16.98} & \textbf{27.05} & \textbf{17.91} & \textbf{0.68/0.30} \\
            \bottomrule
    \end{tabular}
    \caption{Results on synthetic data using feature matching to generate correspondences. Some results are visualized in Figure \ref{fig:predatormm}.}
    % \caption{Results on synthetic data using feature matching to generate correspondences.$\uparrow$ means the larger the better, while $\downarrow$ indicates the contrary. The running time on CPU/GPU of our method is presented. Some results are visualized in Figure \ref{fig:predatormm}.}
    \label{tab:realcorr}
    \end{table}

    \begin{table}[h]
\scriptsize 
        \centering
        %\resizebox{.95\columnwidth}{!}{
            \begin{tabular}{ccccc} %& $50\%~70\%$ & $70\%~90\%$
              \toprule
              \textbf{Metric}& MHR$\left( \% \right) \uparrow $& MHP$\left( \% \right) \uparrow $& MHF1$\left( \% \right) \uparrow $ & Time$\left( s \right) \downarrow $ \\
                \hline
                & \multicolumn{4}{c}{PREDATOR( estimated outlier ratio : $76.44\%$)} \\
                \hline
                T-Linkage & 2.46 & 3.79 & 2.71 & 1655.0 \\
                Progressive-X & 11.58 & 6.86 & 7.87 & 26.32\\
                CONSAC & 2.66 & 0.35 & 0.62 & 21.35\\
                \textbf{Ours} & \textbf{31.63} & \textbf{29.23} & \textbf{27.04} & \textbf{1.46/0.51} \\
                \hline
                
                &\multicolumn{4}{c}{ D3Feat ( estimated outlier ratio :  $97.25\%$)} \\
                \hline
                %\hline
                % \textbf{metric} & MHR$\left( \% \right) \uparrow $ & RRE$\left( \degree \right) \downarrow $ & RTE$\left( m \right) \downarrow $ & t$\left( s \right) \downarrow $\\
                %\hline
                %\hline
                T-Linkage & 0.04 & 0.22 & 0.06 & 2178.43 \\
                Progressive-X & \textbf{0.67} & \textbf{0.30} & \textbf{0.4} & 28.48 \\
                CONSAC & 0 & 0 & 0 & 21.88 \\
                \textbf{Ours} & 0.29 & 0.04 & 0.07 & \textbf{2.13/0.89} \\
                \bottomrule
        \end{tabular}
        \caption{ Results on Scan2CAD benchmark dataset.}
        % \caption{ Results on Scan2CAD benchmark dataset. $\uparrow$ means the larger the better, while $\downarrow$ indicates the contrary. The running time on CPU/GPU of our method is listed.}
        \label{tab:Scan2CAD-cad}
        \end{table}

\subsection{Benchmark dataset}
Scan2CAD\cite{Scan2cad} is a benchmark dataset that aligns the ShapeNet\cite{ShapeNet} CAD models with the object instances in ScanNet\cite{ScanNet} point clouds. Some scans have several aligned CAD models with annotated poses. We choose those scans containing multiple CAD models as the target point cloud and sample the source point cloud from the CAD model for tests. We generate 173 samples for the registration test, where most samples contain $2 \sim 5$ instances.
Note that in each point cloud, only parts of instances were annotated in Scan2CAD. It means that we cannot correctly evaluate the performance such as the precision and recall using the partially annotated poses. To address this issue, we match points only within the ground-truth bounding boxes of the annotated objects in the target point clouds to generate the correspondences. Similarly, we use PREDATOR\cite{PREDATOR} and D3Feat\cite{D3Feat} for point matching, where both are fine-tuned with 1028 training and 187 validation samples from the Scan2CAD dataset.
%For each target point, we
% selected the source point with the most similar feature as the
% correspondence.
The results are shown in Table \ref{tab:Scan2CAD-cad}. Our method performs significantly better than existing methods when using PREDATOR. Note that when using D3Feat, all the methods perform poorly. After we carefully checked the results, we found that the reason is not only the high outlier ratio ( about $97.25\%$ ) but also the lack of sufficient inliers when using D3Feat, even though the feature matching is restricted within the ground truth bounding boxes in the target point clouds. Some results are visualized in Figure \ref{fig:Scan2CAD-cadresult} and Figure \ref{fig:Scan2CAD-cadresult1}. 
We also evaluate the performance of our method by enlarging the bounding box by$1.5\times, 2.0\times$, and $4.0\times$. When the box size is adjusted to $4\times$, the target point cloud is almost the original scan. The results based on PREDATOR features are shown in Table \ref{tab:biggerbox}. When more background points are included, feature matching becomes more challenging, producing highly noisy correspondences, which makes the MHF1(Mean Hit F1) of our method decrease notably.

%\begin{figure*}[ht]
%        \centering   
%        \begin{subfigure}{0.3\textwidth}
%          \centering
%          \includegraphics[height=2.2cm]{real1.png}
%            \caption{Instant noodles}
%            \label{fig:real1}
%        \end{subfigure}
%        \begin{subfigure}{0.35\textwidth}
%          \centering
%          \includegraphics[height=2.2cm]{real2.png}
%            \caption{Watson}
%            \label{fig:real2}
%        \end{subfigure}
%        \begin{subfigure}{0.3\textwidth}
%          \centering
%          \includegraphics[height=2.2cm]{real3.png}
%            \caption{Snickers}
%            \label{fig:real3}
%        \end{subfigure}
%        \begin{subfigure}{0.3\textwidth}
%          \centering
%          \includegraphics[height=2.2cm]{real4.png}
%            \caption{Snickers}
%            \label{fig:real4}
%        \end{subfigure}
%    \caption{Real-world tests on RGB-D scans. The source point cloud is extracted from the depth scan of a single object. The target point cloud is constructed from the depth scan captured from the camera viewpoint.}
%    \label{fig:app}
%    \end{figure*}

\begin{table}[ht]
    \centering
\scriptsize
    %\resizebox{.95\columnwidth}{!}{
    \begin{tabular}{ccccc}
      \toprule
        Box Size& MHR$\left( \% \right) \uparrow $ & MHP $\left( \% \right) \uparrow $ & MHF1$\left( \% \right) \uparrow $& Time$\left( s \right) \downarrow $ \\
        \midrule  
        1.5 ($94.50\%$) & 40.58 & 12.61 & 16.64 & 0.63  \\
        2 ($95.97\%$) & 37.43 & 8.85 & 12.35 & 1.37  \\
        4 ($98.73\%$)& 7.25 & 6.28 & 4.92 & 3.86  \\
        \bottomrule
    \end{tabular}
    \caption{Results of using different sizes of bounding boxes for feature matching on Scan2CAD dataset. The estimated outlier ratio is listed in the bracket after the box size. GPU time is listed in the last column.}
    \label{tab:biggerbox}
    \end{table}

\subsection{Real-world tests}
\begin{figure}[h]
  \centering   
  \begin{subfigure}{0.24\textwidth}
    \centering
    \includegraphics[height=2.cm]{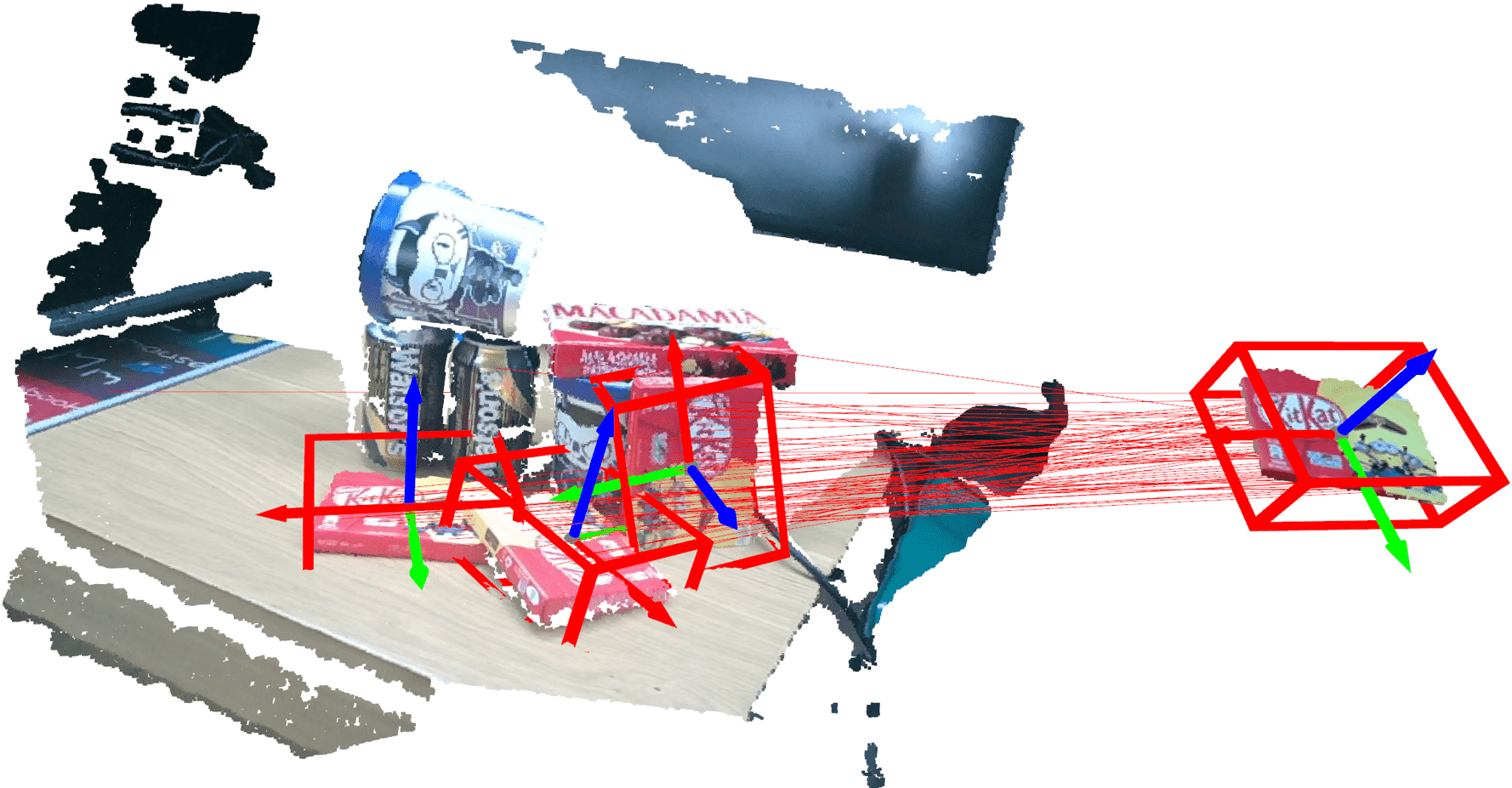}
      \caption{Kitkat}
      \label{fig:real1}
  \end{subfigure}
  \begin{subfigure}{0.21\textwidth}
    \centering
    \includegraphics[height=2.cm]{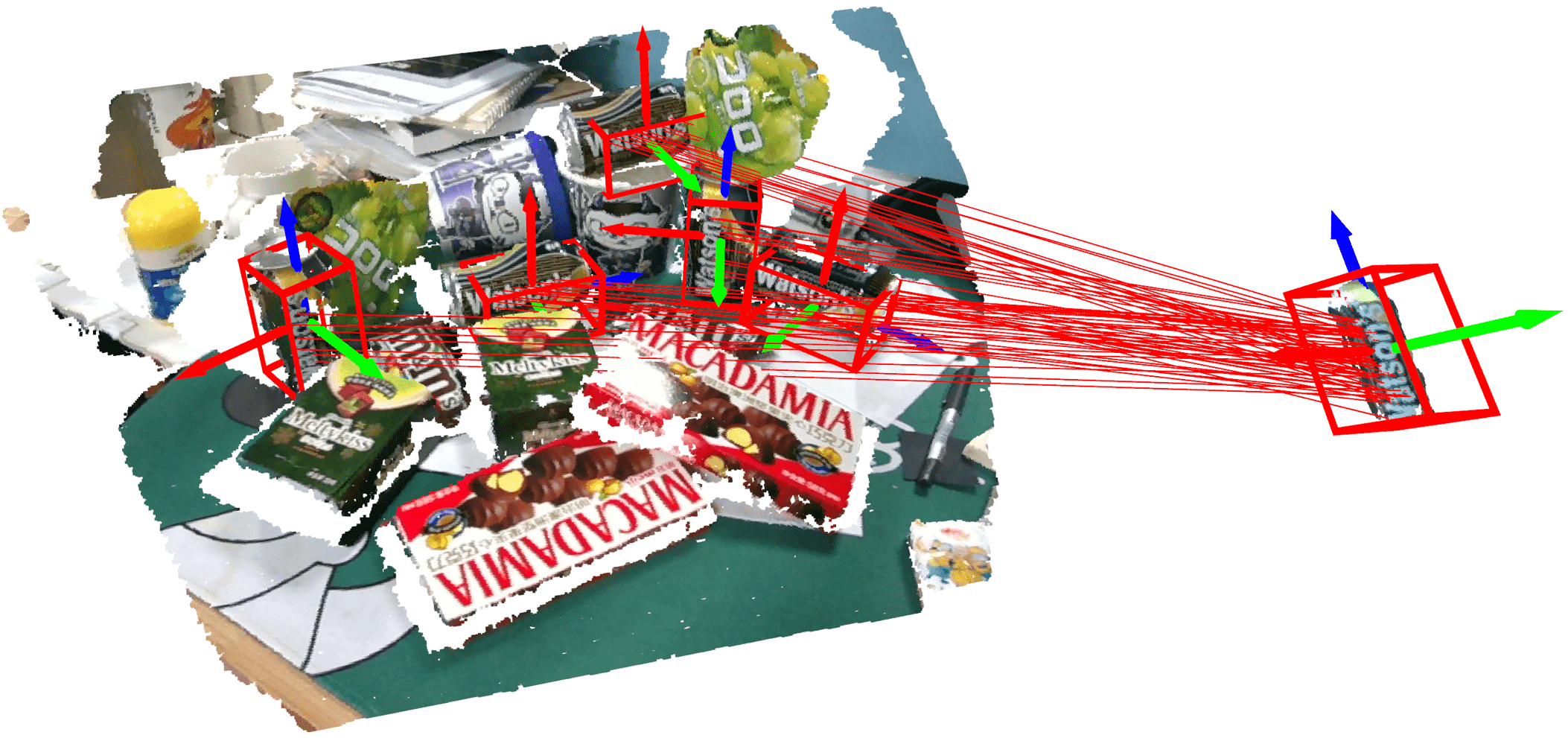}
      \caption{Watson}
      \label{fig:real2}
  \end{subfigure}

  \begin{subfigure}{0.24\textwidth}
    \centering
    \includegraphics[height=2.cm]{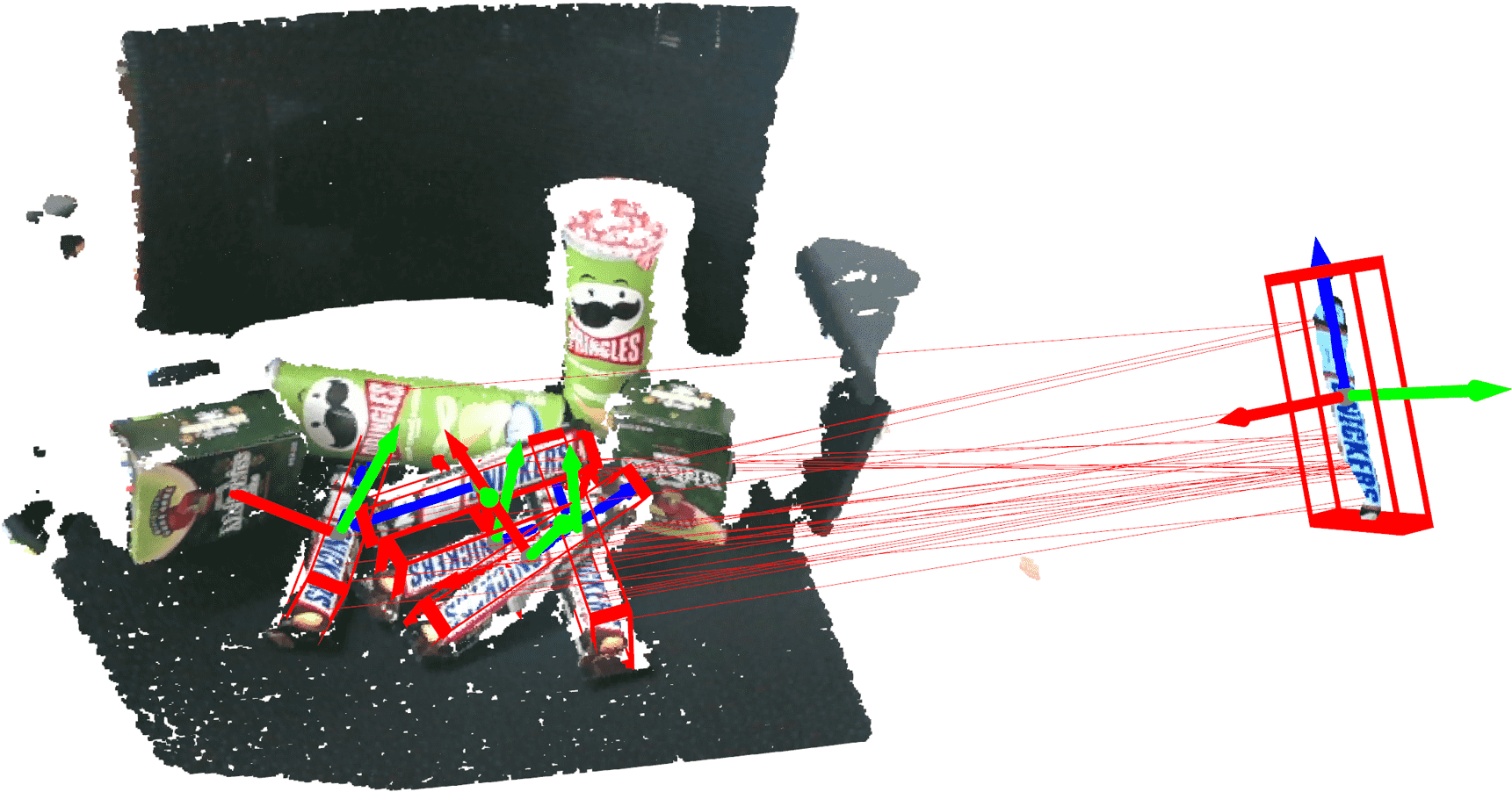}
      \caption{Snickers}
      \label{fig:real3}
  \end{subfigure}
  \begin{subfigure}{0.21\textwidth}
    \centering
    \includegraphics[height=2.cm]{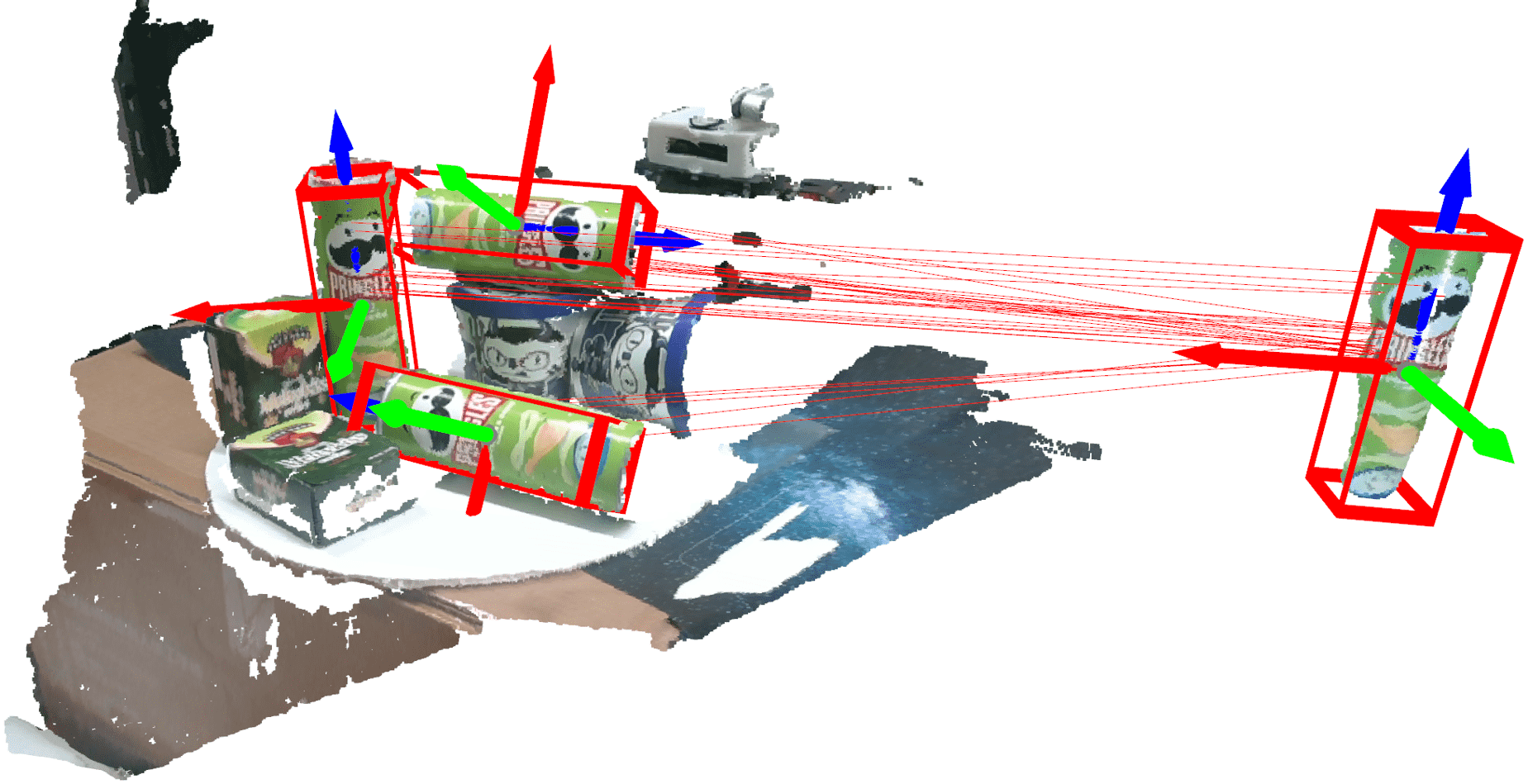}
      \caption{Crisp}
      \label{fig:real4}
  \end{subfigure}
\caption{Real-world tests on RGB-D scans. The source point cloud is extracted from the depth scan of a single object. The target point cloud is constructed from the depth scan captured from the camera viewpoint.}
\label{fig:app}
\end{figure}

We use an RGB-D camera (Intel D455) to capture a sequence of point clouds a pile of objects on the table and apply our algorithm to align a 3D scan of a particular object to its multiple instances in the target RGB-D scan. Since the color information is available, we use SIFT feature to generate 3D point correspondences. Then we apply our algorithm to extract the pose of each object. Some results are shown in Figure \ref{fig:app}. Though the table is cluttered with different objects, our methods can correctly align the source 3D scan up to more than ten instances almost in real-time (about $0.2$s per frame). More results can be found in the supplementary material.

\section{Limitation}
The performance of our method relies on the quality of point correspondences. Unfortunately, we found that using state-of-the-art 3D features such as D3Feat and PREDATOR produces unsatisfactory correspondences, although they have exhibited good performances on some benchmark tests. To improve the correspondence quality, we have to train those features for each dataset in our experiments. Even by doing so, the outlier ratios are still very high and sometimes the inliers are missing for some instances (especially when using D3Feat) which significantly reduces the recall as the experiments show. Therefore, the 3D feature is a bottleneck that requires to be significantly improved.
Another limitation is that distance invariance is a weak rule that does not hold well for noisy point clouds and sometimes is insufficient to reject outliers that are close to inliers, which may also degrade the performance of our method. One possible solution is to seek a better invariance 'feature' representing each correspondence within an end-to-end learning pipeline as \cite{PointDSC}.
\section{Conclusion}
We address the novel task of multi-instance 3D registration in this paper. We found that the column vectors of the distance invariance matrix encode rich information about the instance to which the correspondences are related. Based on this observation, we cluster the correspondences into different groups efficiently by an agglomerative algorithm and refine the result by several iterations. The results on synthetic, benchmark, and real-world datasets show  that our method outperforms existing methods significantly in terms of robustness, accuracy, and efficiency. Though our solution is still far from perfect as discussed, we hope our work could inspire future research on this topic. 

%%%%%%%%% REFERENCES
{\small
\bibliographystyle{ieee_fullname}
\bibliography{my1}
% \bibliography{egbib}
}
\clearpage
\appendix
\section{About this appendix}
We provide extra details about the proposed method, the evaluation metrics, and the experiments because of the limited space in the main draft. We also present additional qualitative results of our experiments in Fig\ref{fig:Scan2CAD-cadresult1}-\ref{fig:Scan2CAD-cadresult} (benchmark data) and Fig\ref{fig:app} (Real-world data).

\section{Definition of evaluation metrics}
Three metrics are used to evaluate the performance of multi-instance point cloud registration. They are Mean Hit Recall, Mean Hit Precision, and Mean Hit F1.

Before we start to compute those metrics, the first step is to establish the ground truth and estimation pair. 
Assuming there are $K$ ground truth transformations and $M$ estimated transformations, we construct an assignment matrix $\mathcal{F} \in \mathbb{K}\times\mathbb{M}$ and obtain the one-to-one mapping by solving the linear assignment problem \cite{hungarian}.
To construct the assignment matrix $\mathcal{F}$, we use F-norm to compute the distance between each pair of ground-truth and estimated transformation, $\mathbf{T}^{*}_i$ and $\hat{\mathbf{T}}_j$:
\begin{equation}
\mathcal{F}(i,j) = \|\mathbf{T}^*_i-\hat{\mathbf{T}}_j\|_F
\end{equation}
where the transformation matrix is defined as
\begin{equation}
\mathbf{T} = \left(
\begin{array}{cc}
\mathbf{R} & \mathbf{t} \\
\mathbf{0}^T & 1
\end{array}
\right) \in \mathbb{R}^{4\times4}.
\end{equation}
After solving the linear assignment problem, we obtain $S = \min(K,M)$ GT-estimation pairs.
We then define the Relative Rotation Error (RRE) and the Relative Translation Error (RTE)
between a GT-estimation pair $\mathbf{T}^{*}_s$ and $\hat{\mathbf{T}}_s$ as
\begin{align}\label{eq:RTE_and_RRE}
RRE_s&=\mathrm{arc}\cos\mathrm{((}tr(\mathbf{\hat{R}}_s^{T}\mathbf{R}_s^{*})-1)/2)\\
RTE_s&=\|\mathbf{t}_{s}^{*} - \mathbf{\hat{t}}_s\| \notag
\end{align}
With the two errors, we define the evaluation metrics in the following sections.

\subsection{Mean Hit Recall (MHR)}
The Mean Hit Recall between two registered point clouds is defined as 
\begin{equation}\label{eq:mhr}
    MHR=\frac{1}{K}\sum_{\mathrm{s}=1}^S{I_s}
    \end{equation}
where $I_s = \{0,1\}$ represents whether a GT-estimation pair being 'hit'. Specifically,
\begin{equation}
I_s=I(RRE_s<\tau _r)\times I(RTE_s<\tau _t)
\end{equation}
where $I\left( \cdot \right) = \{0, 1\}$ denotes an indicating function. $RRE_s$ and $RTE_s$ are the relative rotation error and relative translation error for the $s^{th}$ GT-estimation pair defined in (\ref{eq:RTE_and_RRE}). The two  thresholds $\tau _r$ and $\tau _t$ are set to be $20^{\degree}$ and $0.5m$ respectively in all our experiments.

The final Mean Hit Recall is obtained by averaging the MHR of all the point cloud pairs.

\subsection{Mean Hit Precision (MHP)}
The Mean Hit Precision between two registered point clouds is defined as 
\begin{equation}\label{eq:mhp}
    MHP=\frac{1}{M}\sum_{\mathrm{s}=1}^S{I_s}.
    \end{equation}
The final Mean Hit Precision is obtained by averaging the MHP of all the point cloud pairs.

\subsection{Mean Hit F1 (MHF1)}
The Mean Hit F1 between two registered point clouds is defined as 
\begin{equation}\label{eq:mhf1}
    MHF1=\frac{2*MHP*MHR}{MHP+MHR}
    \end{equation}
The final Mean Hit F1 is obtained by averaging the MHF1 of all the point cloud pairs.

\section{Some details about the experiments}
\subsection{Feature extraction \& one-to-many feature matching}
We extract features for all the points both in the source point cloud and the target point cloud before we start to match the corresponding points. We use the state-of-art point cloud feature extractors PREDATOR and D3Feat in our synthetic and benchmark experiments. We found the pre-trained models (trained on ModelNet40) of both PREDATOR and D3Feat perform so poorly that we train those models from scratch using the datasets used for evaluation.

To obtain multi-instance correspondences, instead of matching the source point cloud to the target point clouds, we match them in the reversed order. In other words, for each point in the target point cloud, we find the most similar one in the source point cloud by nearest neighborhood feature matching. In this way, we can associate multiple target points with a single source point.

\subsection{About estimated outlier ratios}
In Table $2\sim4$ of the main draft, we present the estimated outlier ratio for each result. Here we present how the outlier ratio is estimated.
Given the ground truth transformations from the source points to the target points, we may establish the ground truth correspondences between the source points and the target points.
The outlier correspondences are those who are different from the ground truth ones. Namely, for an estimated correspondence, we find the ground truth correspondence with the same target point and check if its source point is the same as that of the ground truth.

\section{Ablation Study}
Tab\ref{tab:albation} shows ablation study on the two successive steps of our method - \textbf{clustering} and \textbf{refinement} - on both synthetic and real-world tests (see Tab 2 and Tab 3 for results of other methods). The clustering step produces noisy clusters leading to high recall but low precision. The refinement step can remove outliers and merge duplicated clusters, achieving both high recall and high precision.

\begin{figure}[ht]
  \centering
\includegraphics[height=2.3cm]{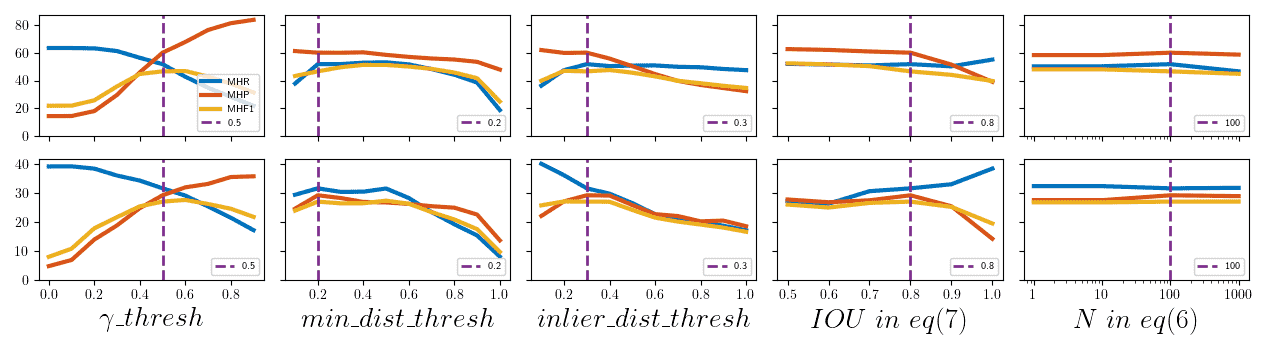}
\caption{\textbf{Top}: Results on the synthetic dataset. \textbf{Bottom:} Results on the Scan2CAD dataset.}
\label{fig:para}%(except $\gamma\_thresh$, see L469)
\end{figure}

Fig\ref{fig:para} shows performance change with five parameters, the top line are results tested on the synthetic dataset and the bottom line are those on the Scan2CAD dataset. The first three columns show the ablation study on three key parameters. The performance curves are smooth (except $\gamma\_thresh$, which is used to trade-off recall and precision) and have similar shapes across different datasets, indicating those parameters are easy to be tuned and generalize well for different datasets. The last two columns imply that the parameters in Eq\ref{eq:alpha} and Eq\ref{eq:iou} have little impact on the performance except for a high IOU threshold ($>0.9$). The values used in all our experiments are indicated by vertical dot lines.
\begin{table}[ht]
  %\scriptsize
  \centering  
  \tiny
          \begin{tabular}{cccccc} %& $50\%~70\%$ & $70\%~90\%$  
            
              \toprule
              \textbf{Step}& MHR$\left( \% \right) \uparrow $& MHP$\left( \% \right) \uparrow $& MHF1$\left( \% \right) \uparrow $ & Time & \#Clusters\\
              \hline
              & \multicolumn{4}{c}{Synthetic Dataset with PREDATOR} \\
              \hline
              
              \textbf{clustering} & \textbf{61.56} & 6.09 & 10.70 & 0.23s & 76.63\\
              \textbf{clustering+refinement} & 53.39 & \textbf{61.44} & \textbf{51.80} & 0.48s & 8.10\\
              \hline
              
              &\multicolumn{4}{c}{Scan2CAD Dataset with PREDATOR} \\
              \hline
              
              \textbf{clustering} & \textbf{41.66} & 4.19 & 7.19 & 0.36s & 51.38\\
              \textbf{clustering+refinement} & 31.63 & \textbf{29.23} & \textbf{27.04} & 0.51s & 5.86\\
              \bottomrule
      \end{tabular}
       \caption{Ablation study on \textbf{clustering} and \textbf{refinement}.}
      \label{tab:albation}
      \end{table}
\section{Qualitative result}
\begin{figure*}[ht]
  \centering
  \begin{subfigure}{0.3\textwidth}
      \centering
      \includegraphics[height=2.8cm]{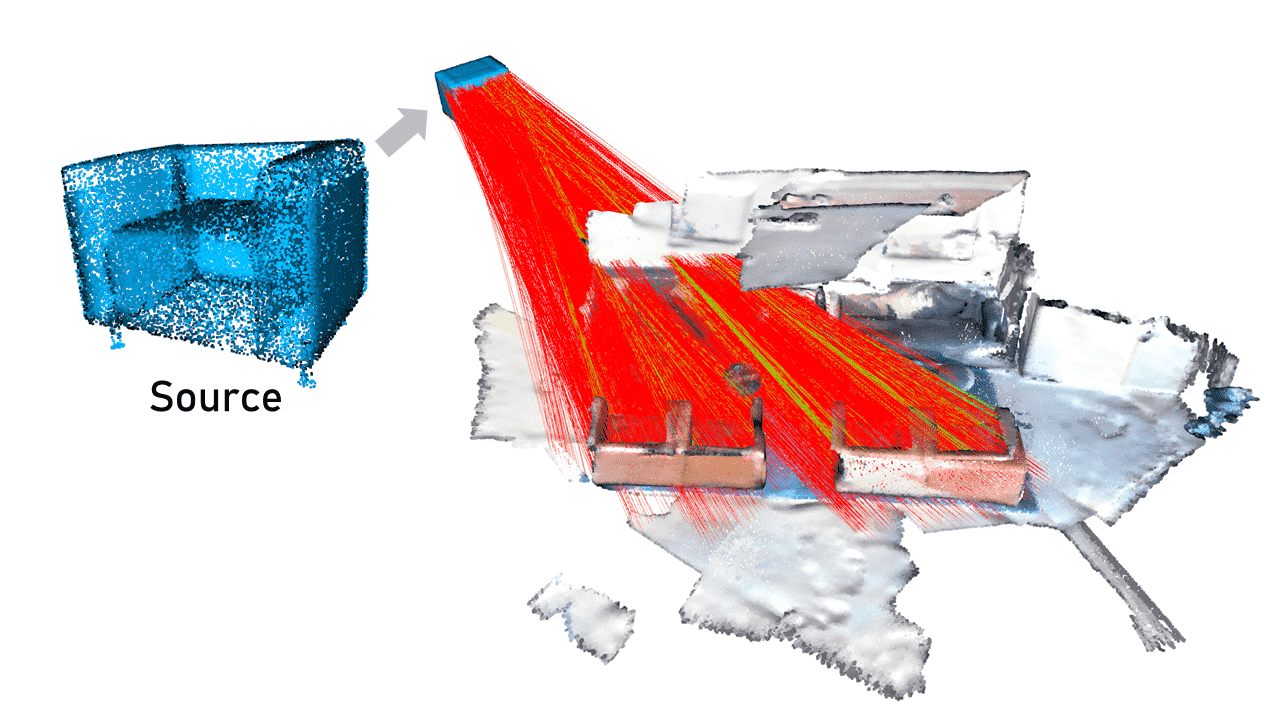}
        \caption{Input correspondences}
        \label{fig:ascan2cad_cad-input-corrs1}
    \end{subfigure}\hfill
    \begin{subfigure}{0.3\textwidth}
      \centering
      \includegraphics[height=2.8cm]{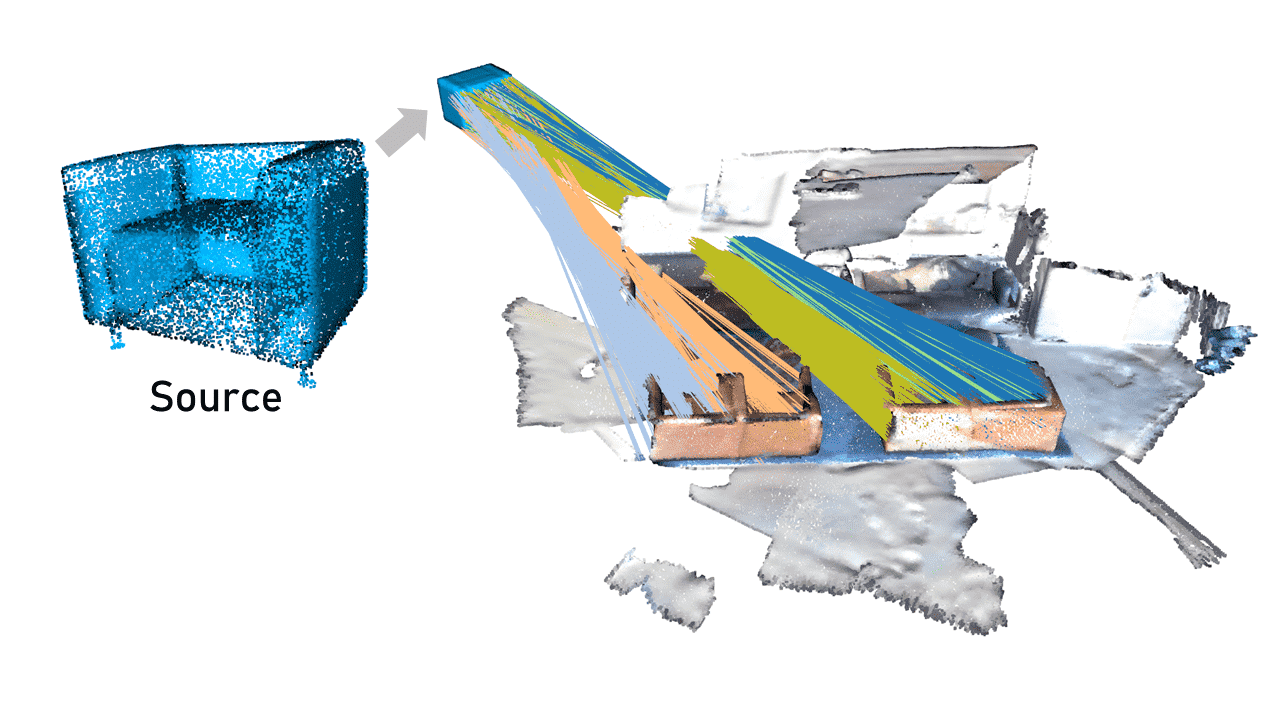}
        \caption{Our clustering result}
        \label{fig:ascan2cad_cad-cluster-corrs1}
    \end{subfigure}\hfill
    \begin{subfigure}{0.3\textwidth}
      \centering
      \includegraphics[height=2.8cm]{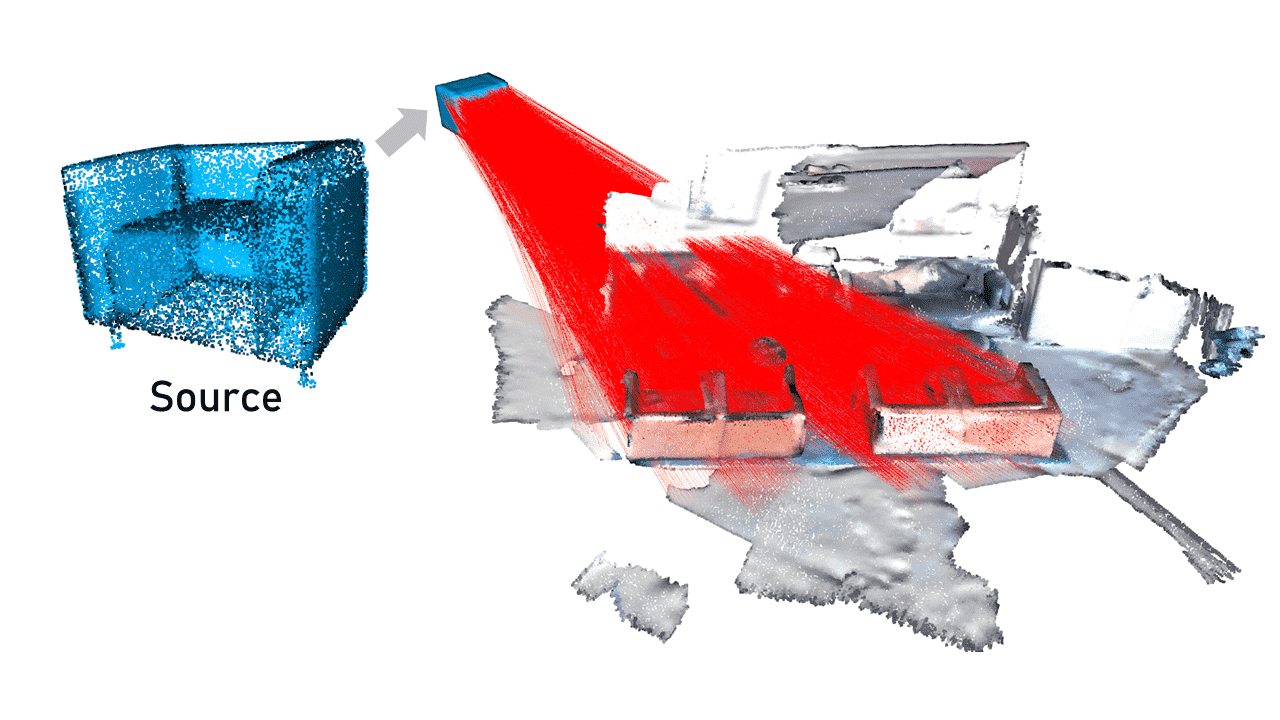}
        \caption{Our reject outliers}
        \label{fig:ascan2cad_cad-reject-corrs1}
    \end{subfigure}

    \begin{subfigure}{0.23\textwidth}
      \centering
      \includegraphics[height=2.5cm]{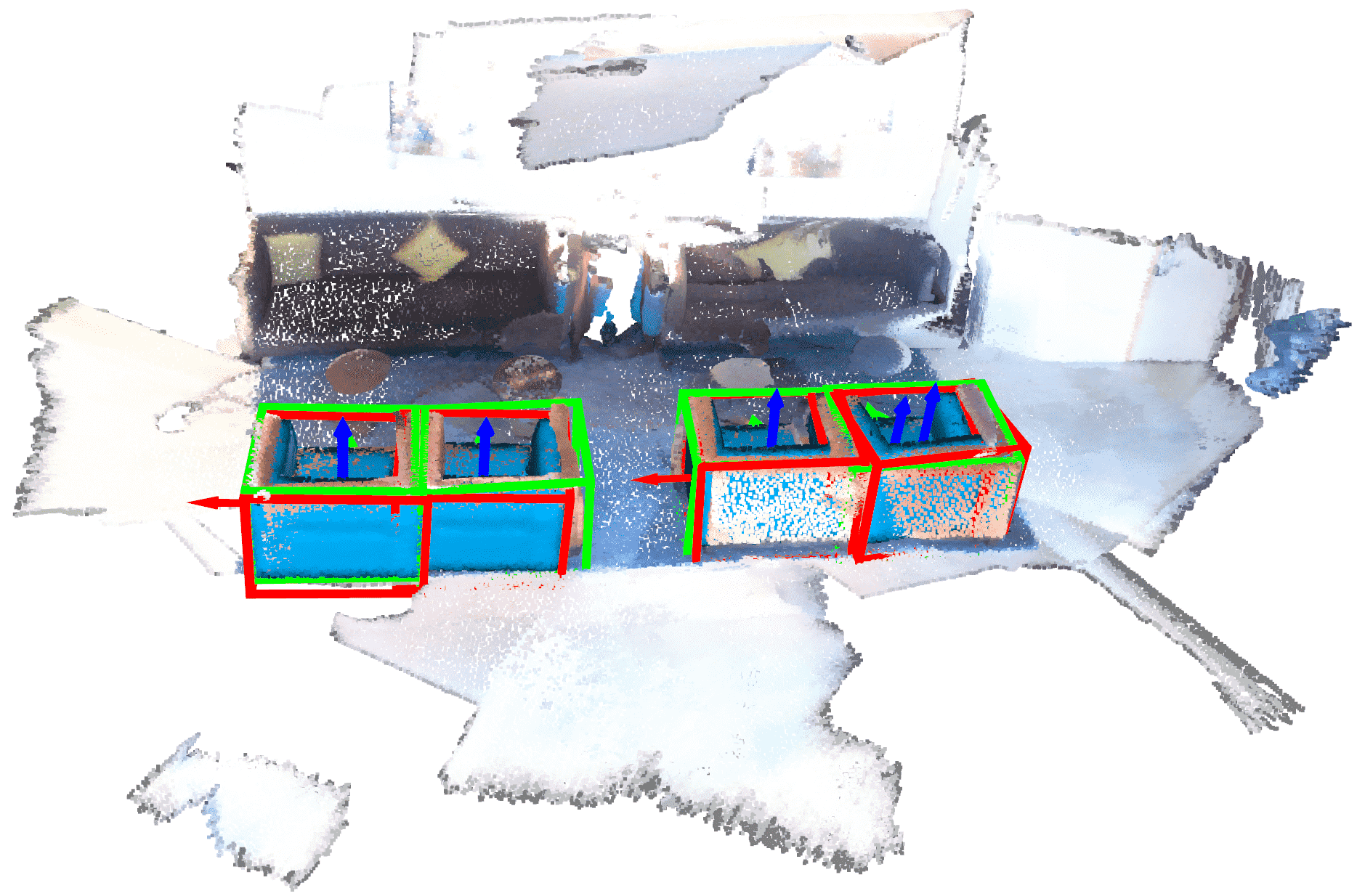}
        \caption{Ours}
        \label{fig:ascan2cad_cad-result1}
    \end{subfigure}\hfill
    \begin{subfigure}{0.23\textwidth}
      \centering
      \includegraphics[height=2.5cm]{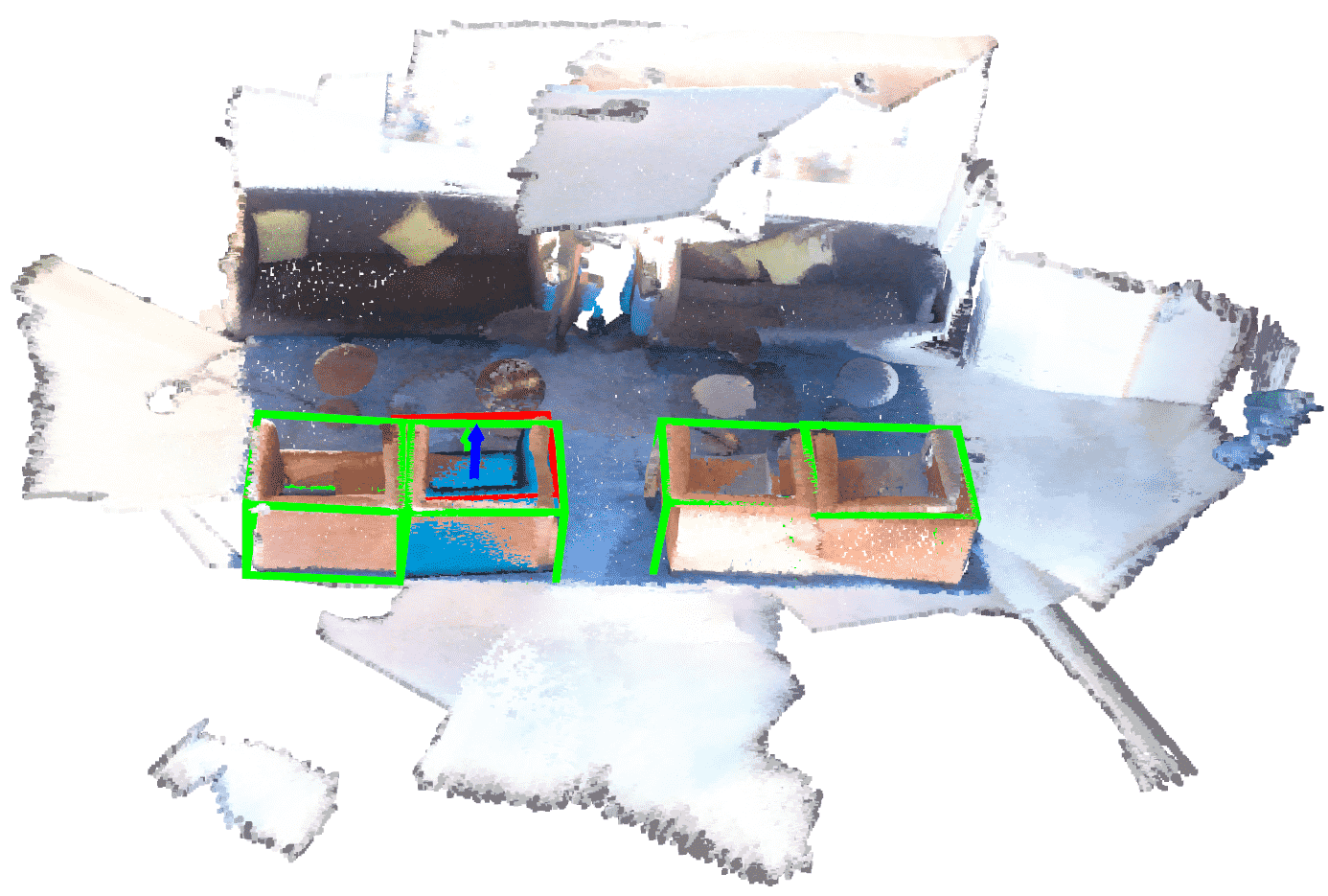}
        \caption{Progressive-X(2019) \cite{ProgressiveX}}
        \label{fig:ascan2cad_cad-prox1}
    \end{subfigure}\hfill
    \begin{subfigure}{0.23\textwidth}
      \centering
      \includegraphics[height=2.5cm]{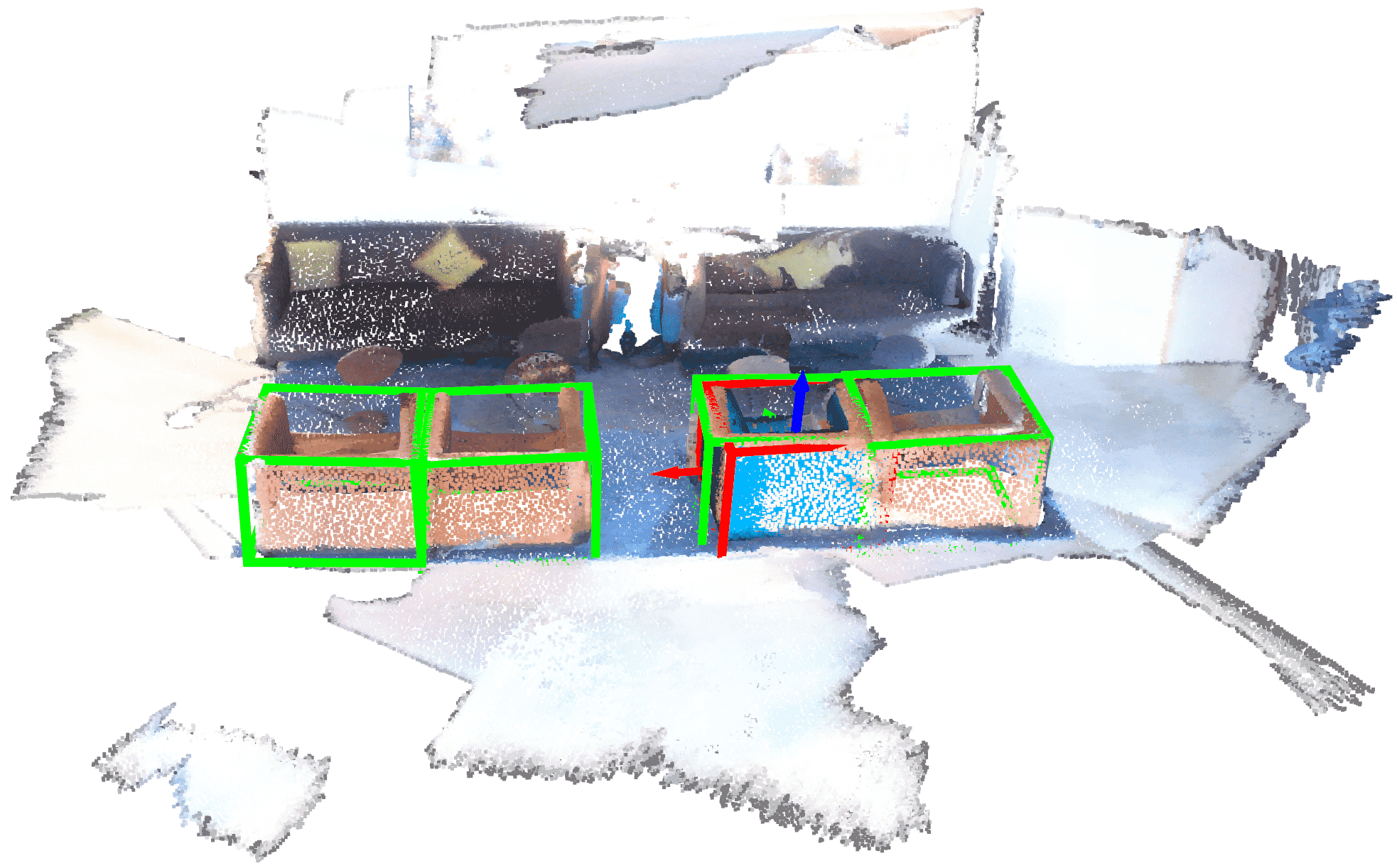}
        \caption{TEASER(2020)\cite{TEASER}}
        \label{fig:ascan2cad_cad-teaser1}
    \end{subfigure}
    \begin{subfigure}{0.23\textwidth}
      \centering
      \includegraphics[height=2.5cm]{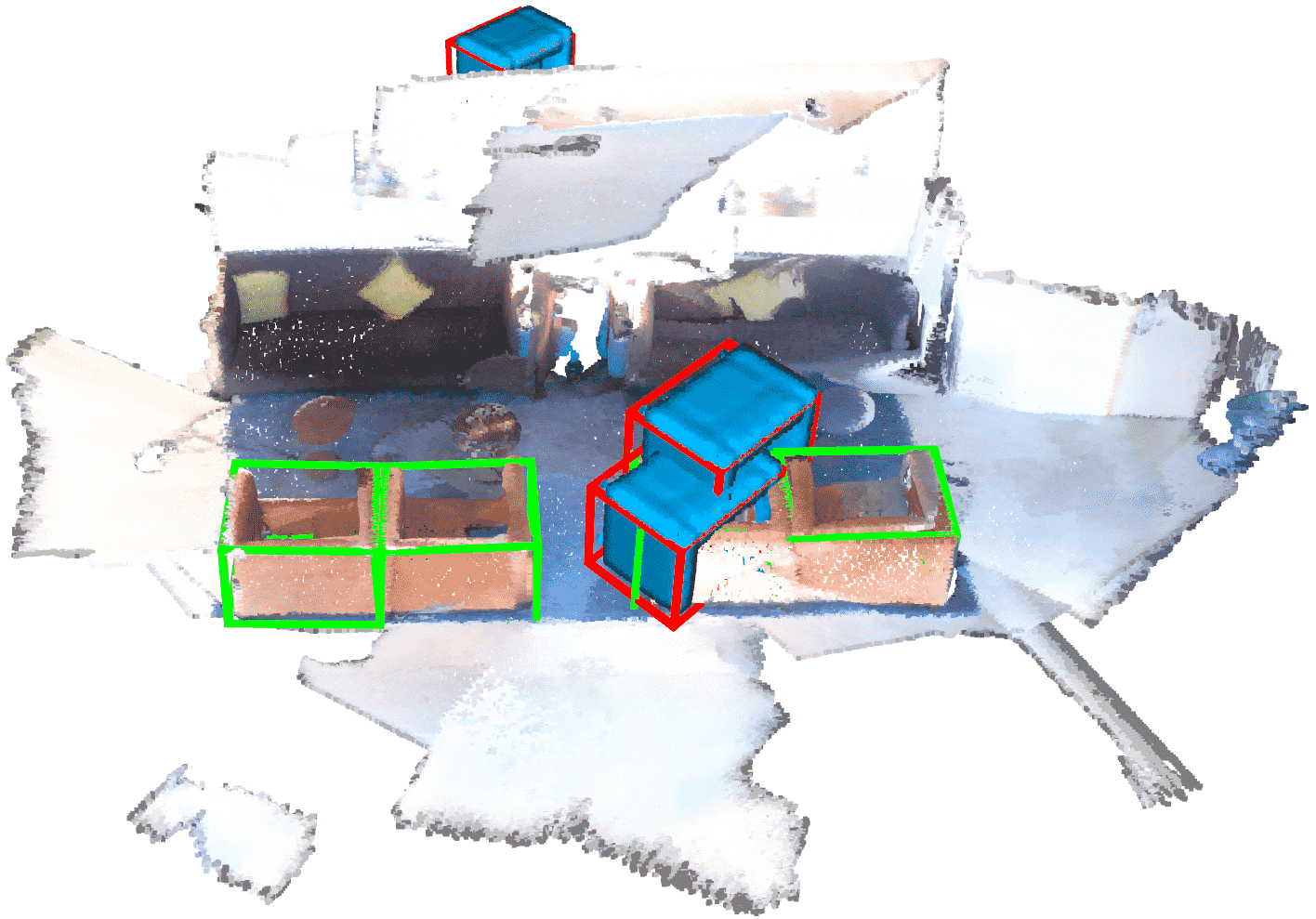}
        \caption{CONSAC(2020)\cite{CONSAC}}
        \label{fig:ascan2cad_cad-consac1}
    \end{subfigure}\hfill
    \caption{\textbf{Scan2CAD results.}}
\label{fig:aScan2CAD-cadresult1}
  \end{figure*}

\begin{figure*}[ht]
  \centering
  \begin{subfigure}{0.3\textwidth}
      \centering
      \includegraphics[height=2.8cm]{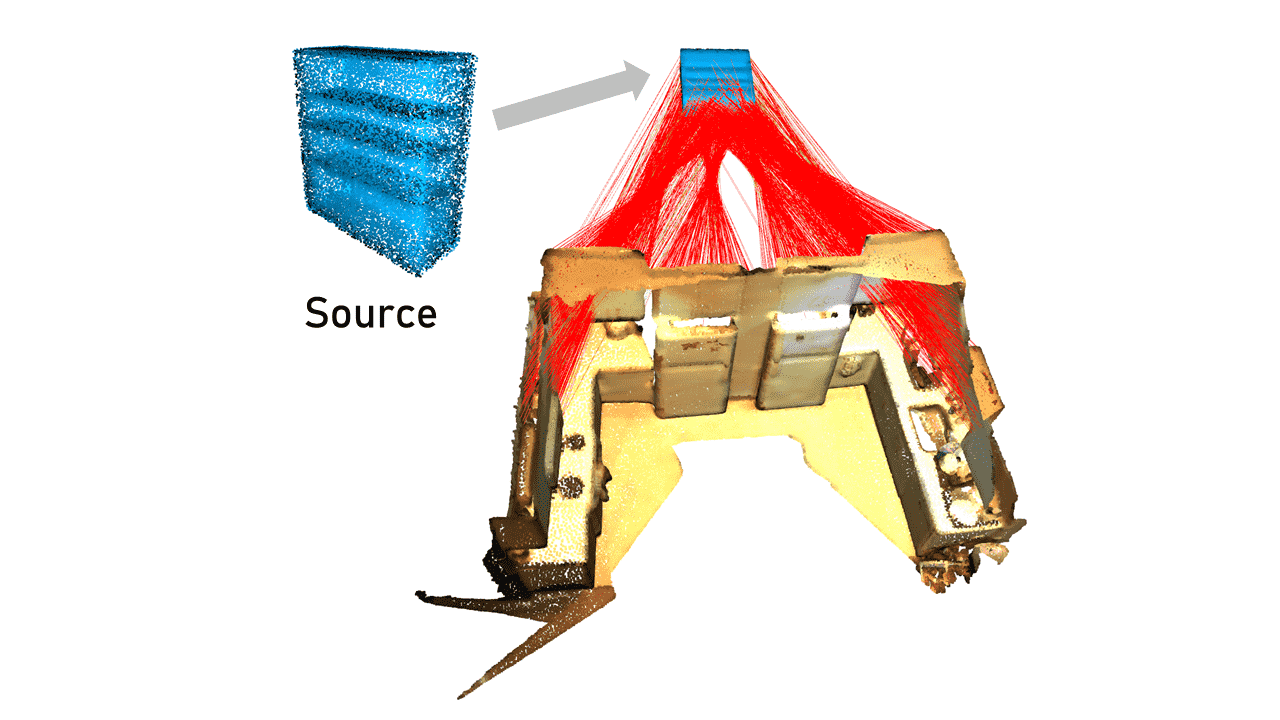}
        \caption{Input correspondences}
        \label{fig:scan2cad_cad-input-corrs2}
    \end{subfigure}\hfill
    \begin{subfigure}{0.3\textwidth}
      \centering
      \includegraphics[height=2.8cm]{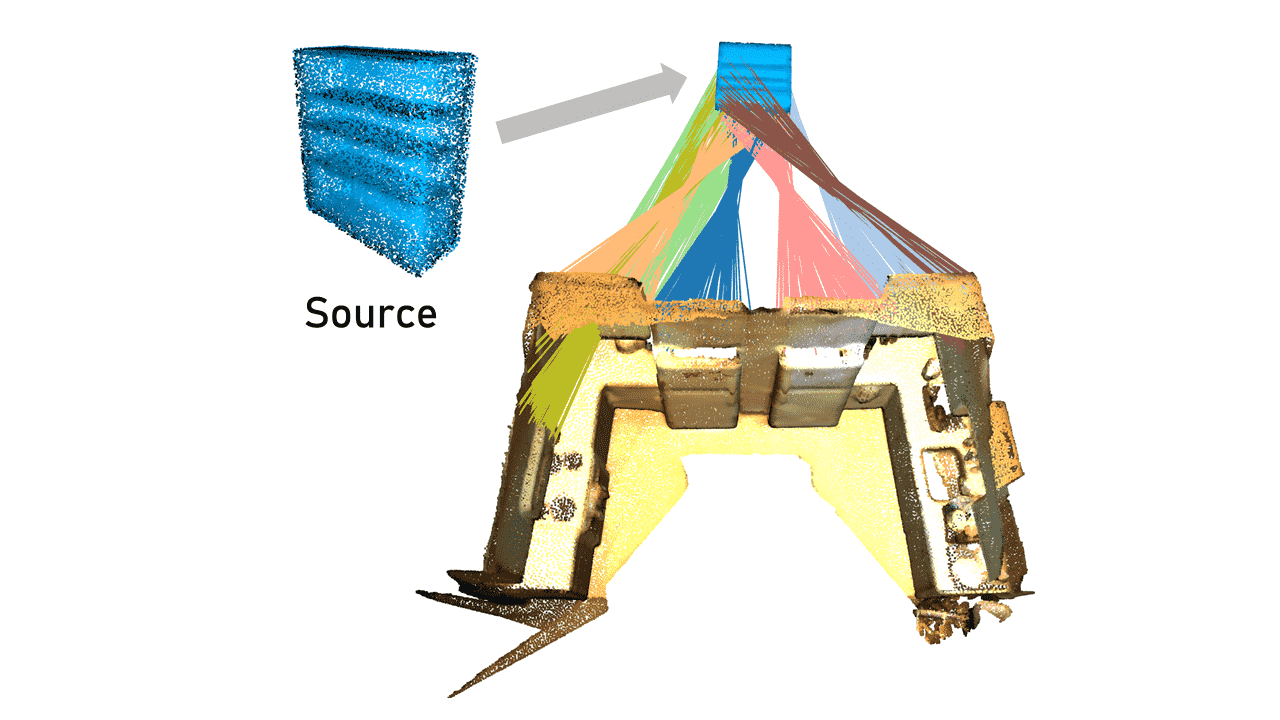}
        \caption{Our clustering result}
        \label{fig:scan2cad_cad-cluster-corrs2}
    \end{subfigure}\hfill
    \begin{subfigure}{0.3\textwidth}
      \centering
      \includegraphics[height=2.8cm]{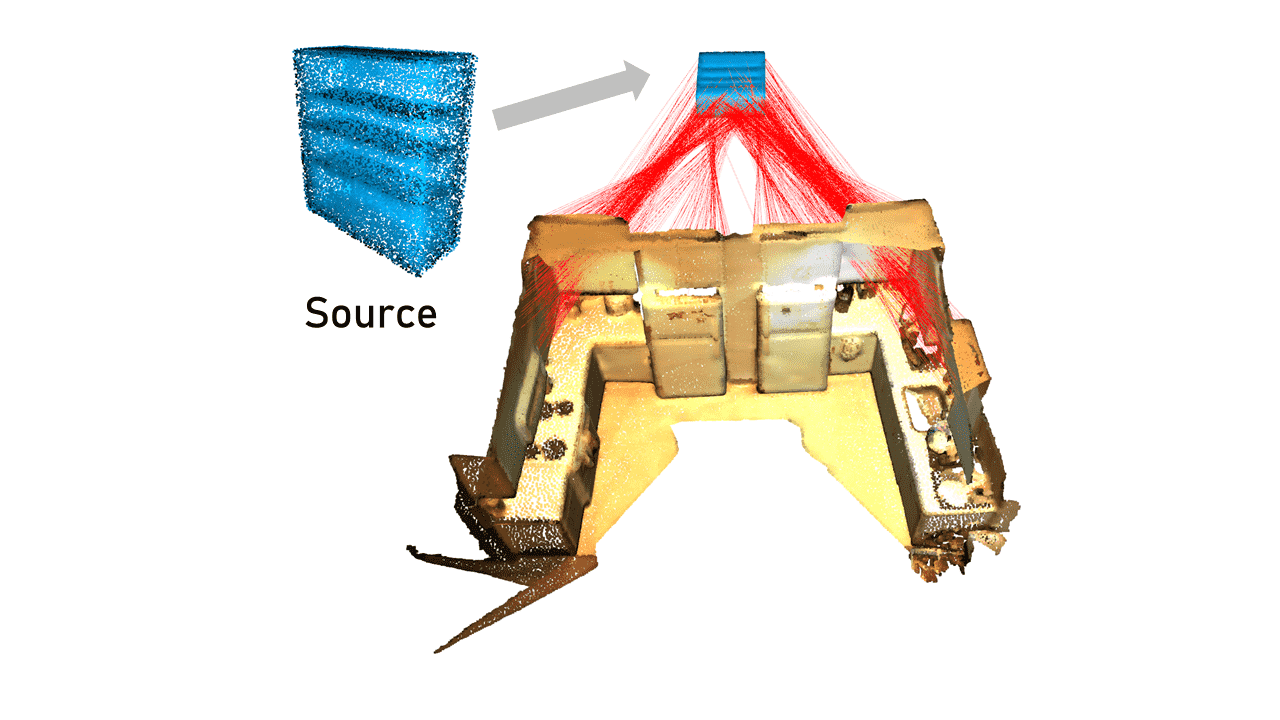}
        \caption{Our reject outliers}
        \label{fig:scan2cad_cad-reject-corrs2}
    \end{subfigure}

    \begin{subfigure}{0.23\textwidth}
      \centering
      \includegraphics[height=2.8cm]{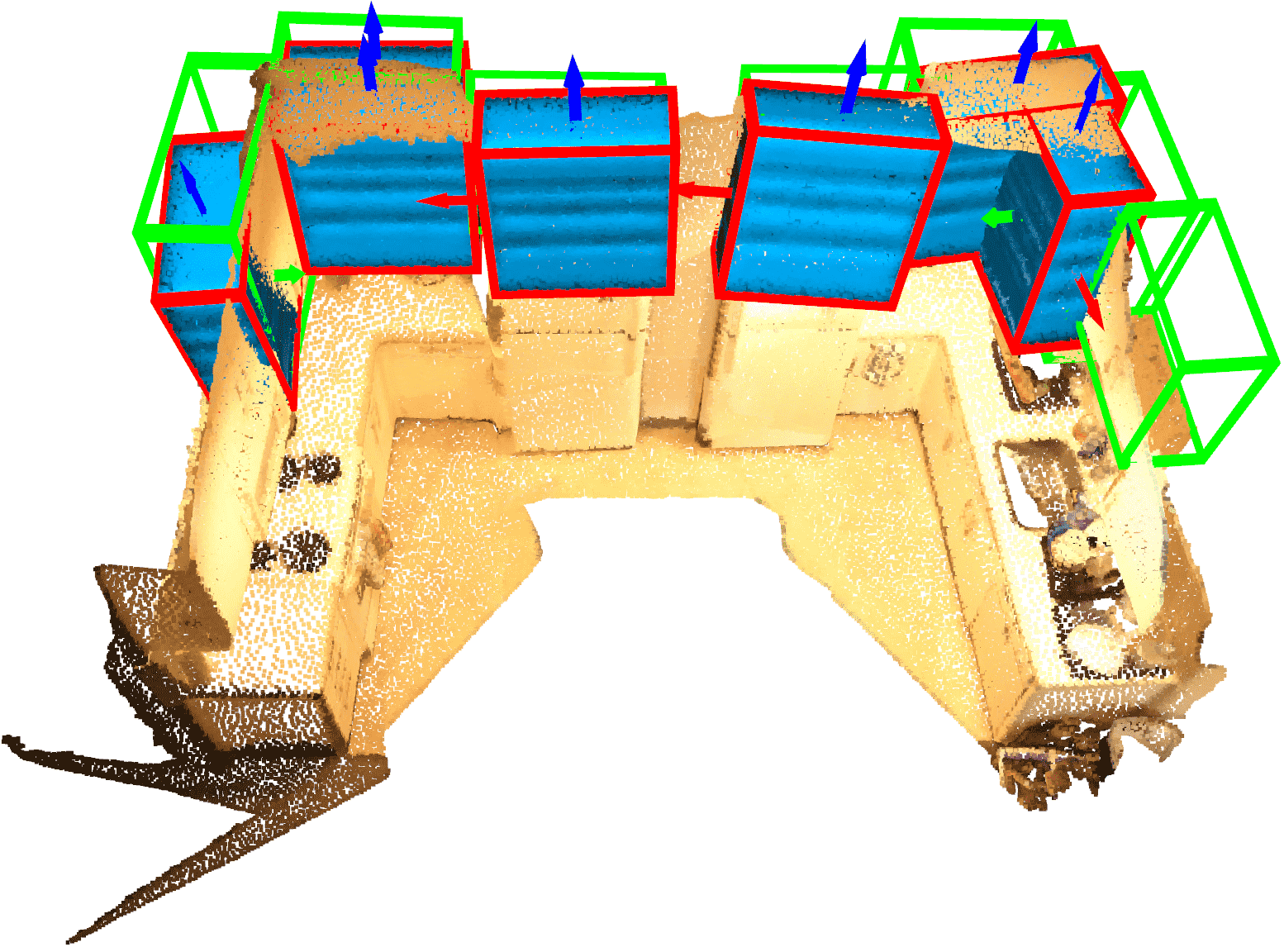}
        \caption{Ours}
        \label{fig:scan2cad_cad-result2}
    \end{subfigure}\hfill
    \begin{subfigure}{0.23\textwidth}
      \centering
      \includegraphics[height=2.8cm]{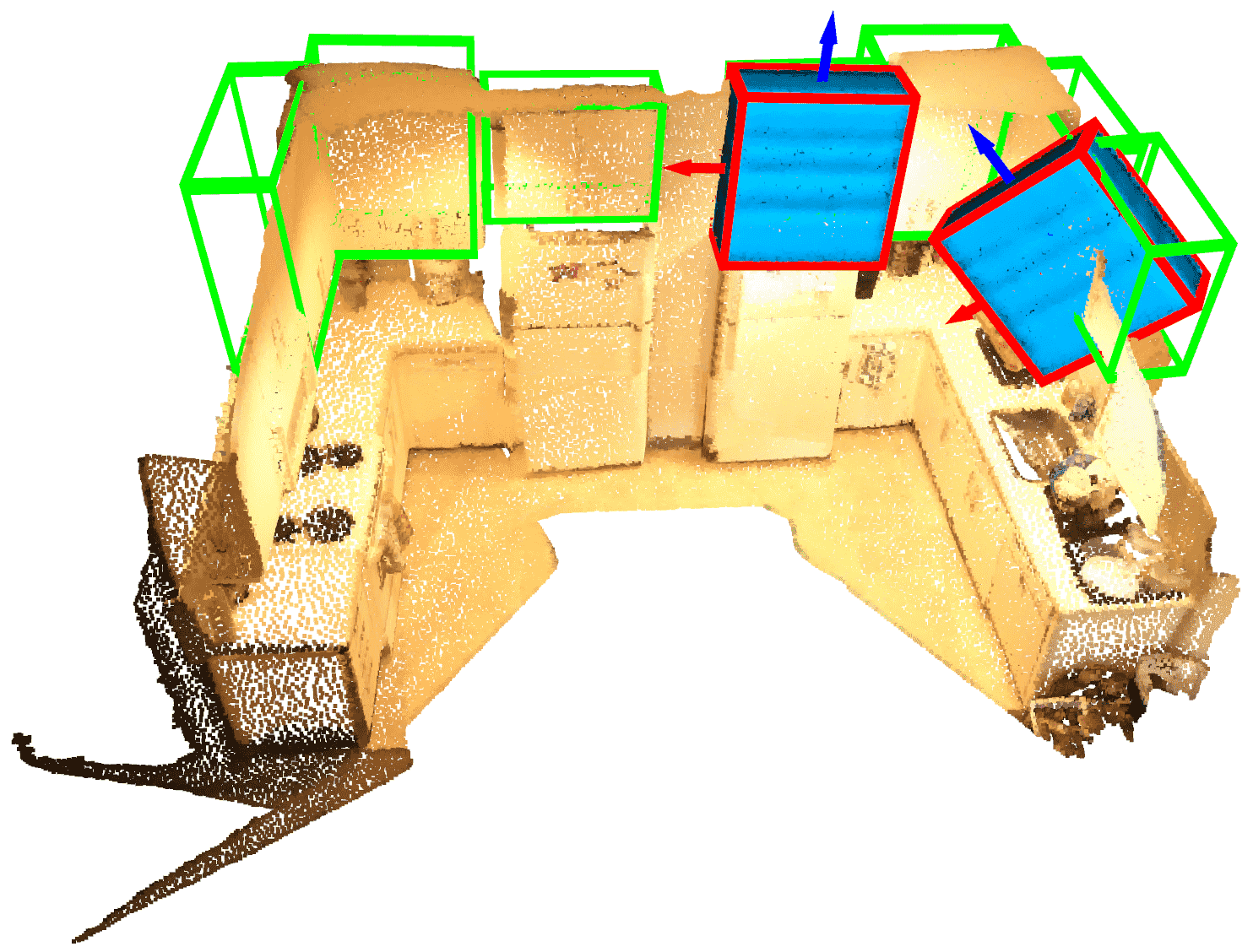}
        \caption{Progressive-X(2019) \cite{ProgressiveX}}
        \label{fig:scan2cad_cad-prox2}
    \end{subfigure}\hfill
    \begin{subfigure}{0.23\textwidth}
      \centering
      \includegraphics[height=2.8cm]{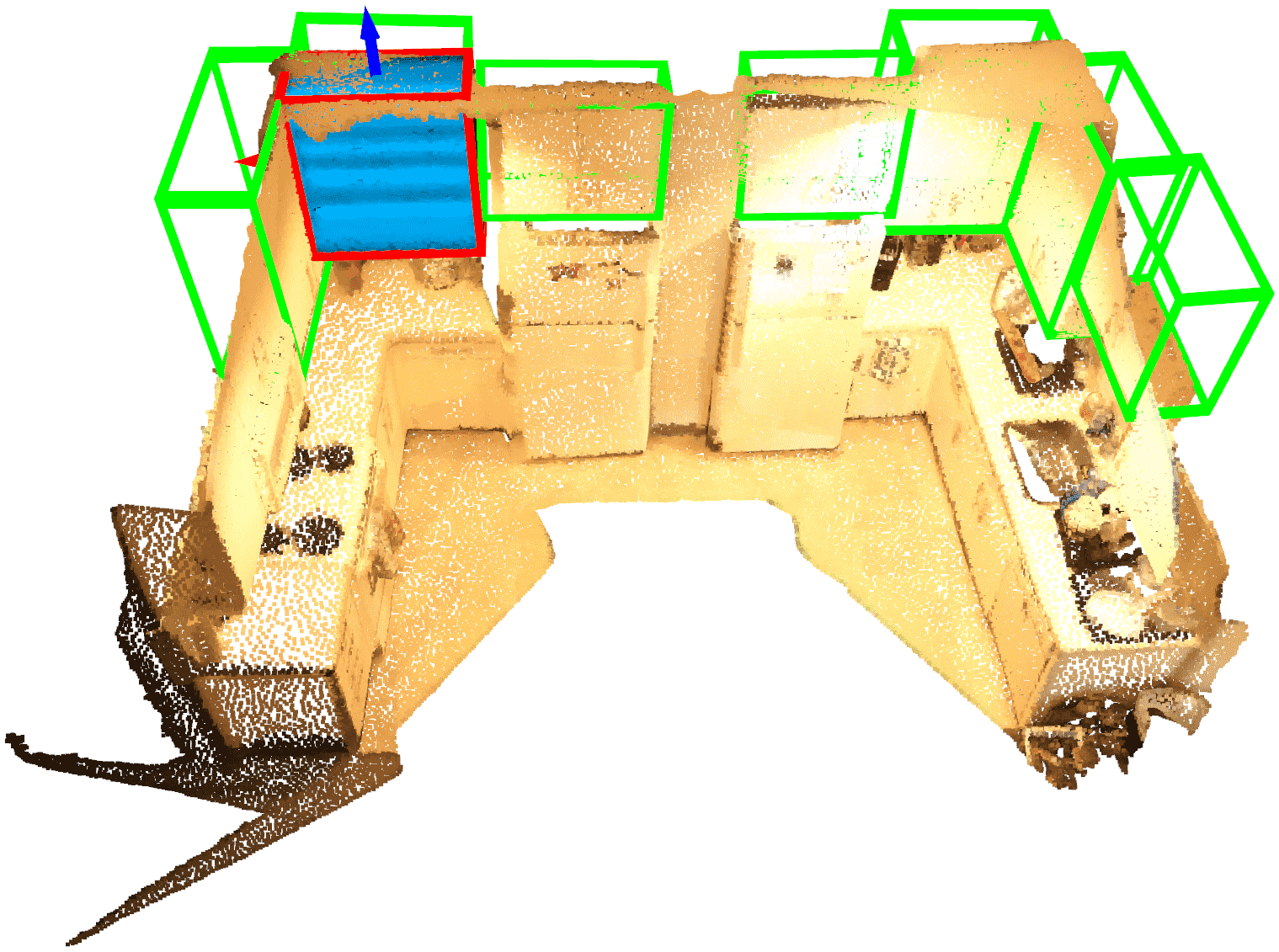}
        \caption{TEASER(2020)\cite{TEASER}}
        \label{fig:scan2cad_cad-teaser2}
    \end{subfigure}
    \begin{subfigure}{0.23\textwidth}
      \centering
      \includegraphics[height=2.8cm]{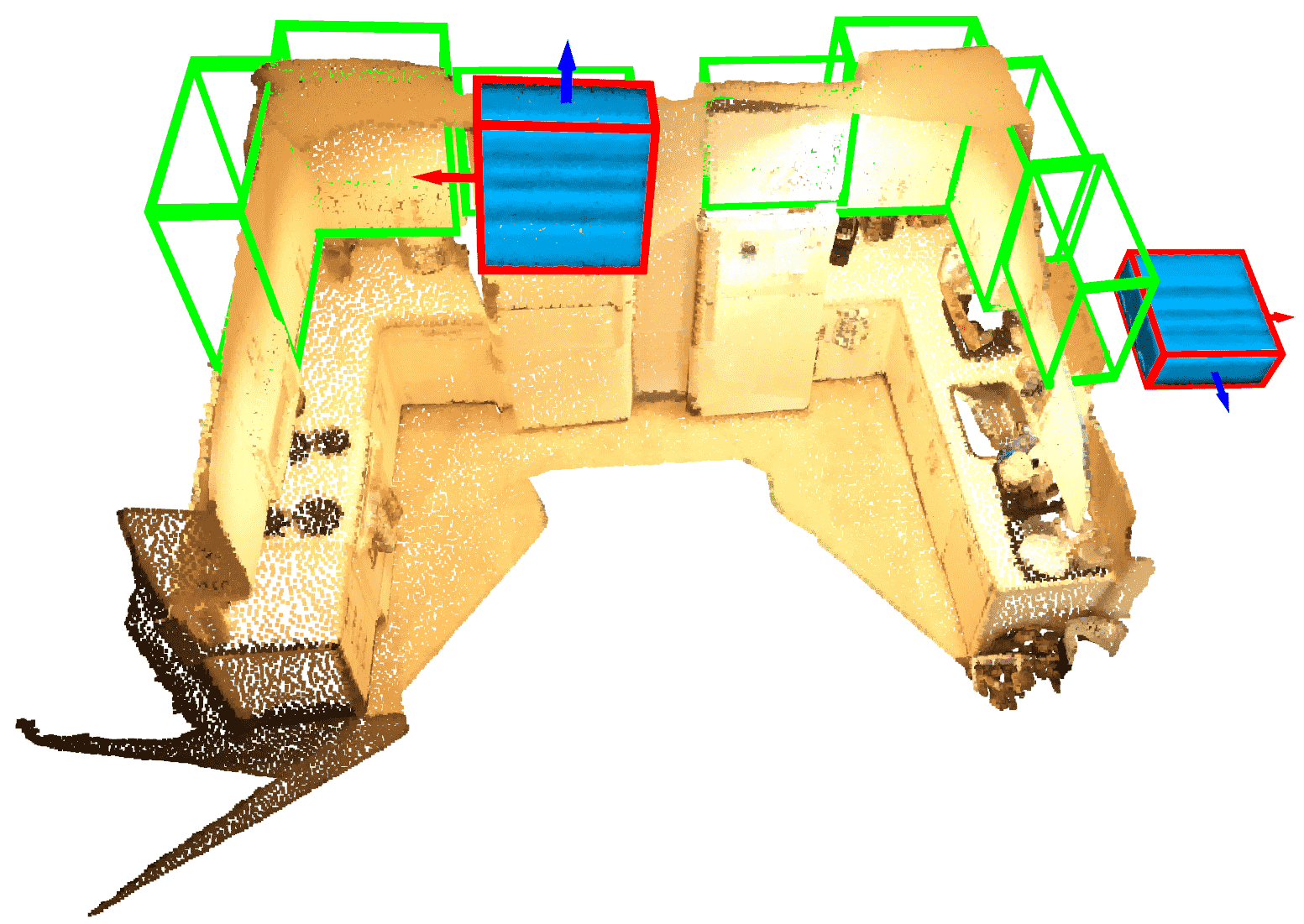}
        \caption{CONSAC(2020)\cite{CONSAC}}
        \label{fig:scan2cad_cad-consac2}
    \end{subfigure}\hfill
    \caption{\textbf{Scan2CAD results.}}
\label{fig:Scan2CAD-cadresult2}
  \end{figure*}

\begin{figure*}[ht]
  \centering
  \begin{subfigure}{0.3\textwidth}
      \centering
      \includegraphics[height=2.8cm]{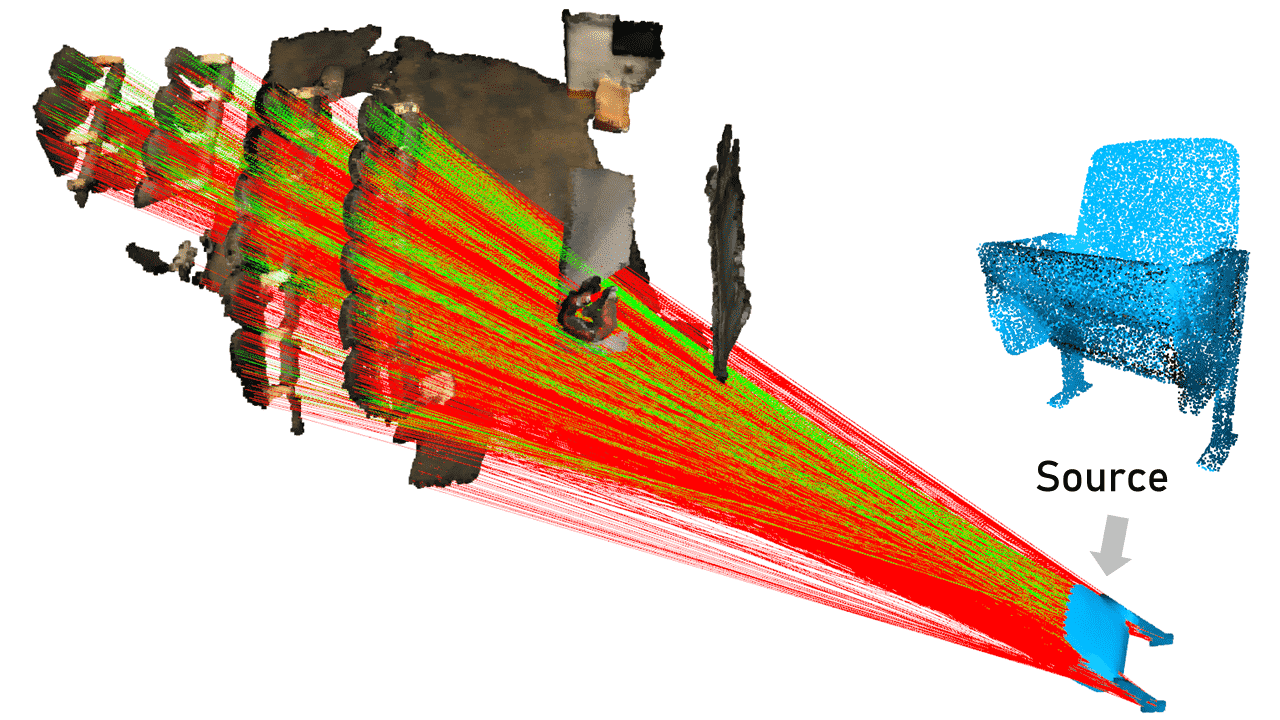}
        \caption{Input correspondences}
        \label{fig:scan2cad_cad-input-corrs19}
    \end{subfigure}\hfill
    \begin{subfigure}{0.3\textwidth}
      \centering
      \includegraphics[height=2.8cm]{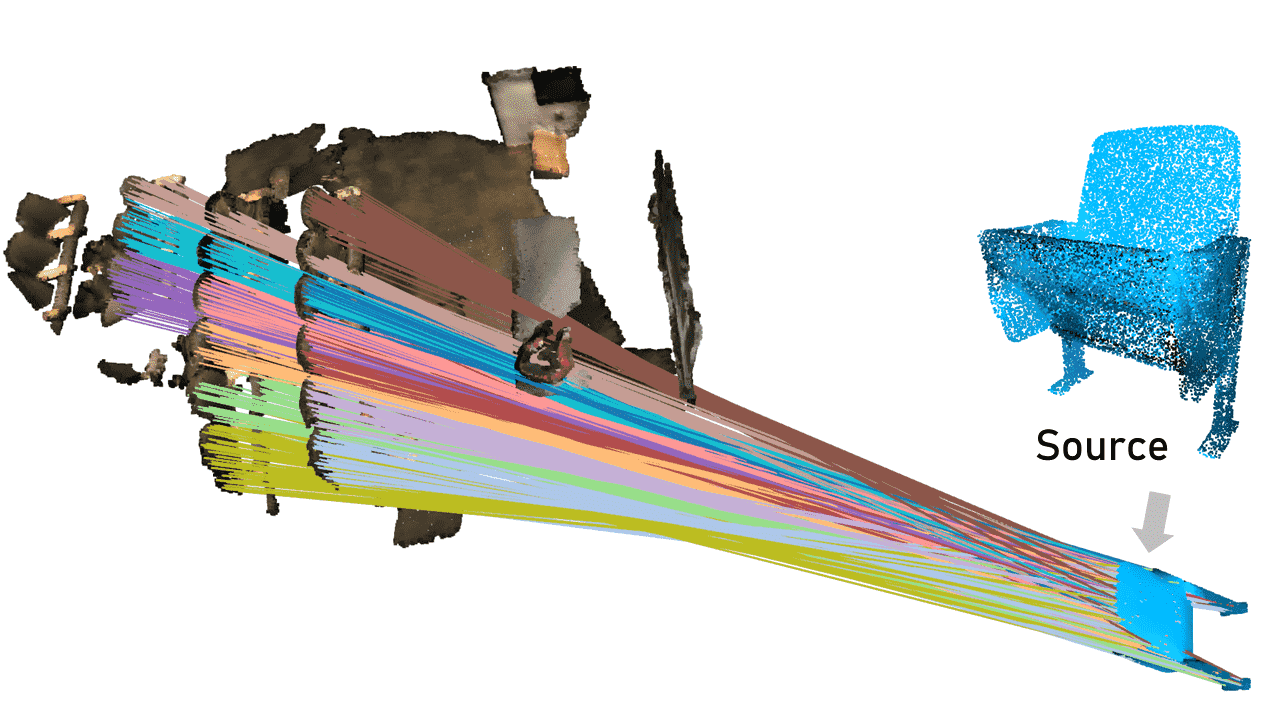}
        \caption{Our clustering result}
        \label{fig:scan2cad_cad-cluster-corrs19}
    \end{subfigure}\hfill
    \begin{subfigure}{0.3\textwidth}
      \centering
      \includegraphics[height=2.8cm]{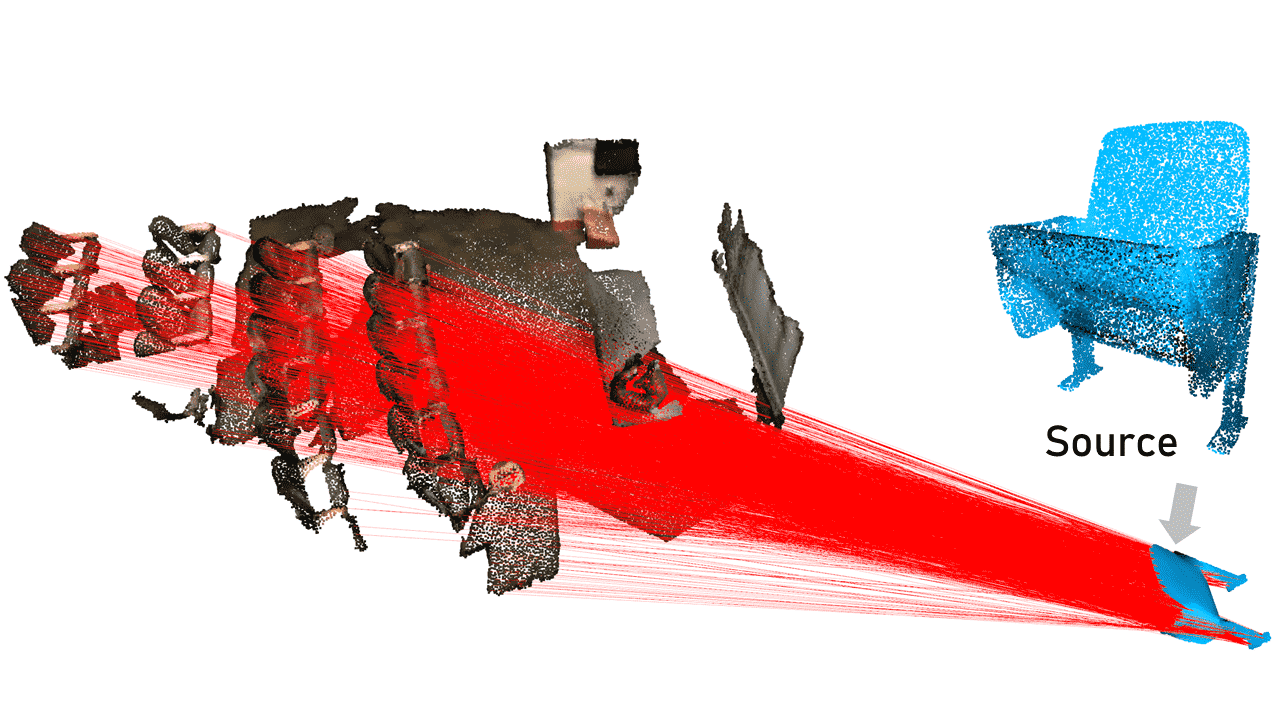}
        \caption{Our reject outliers}
        \label{fig:scan2cad_cad-reject-corrs19}
    \end{subfigure}

    \begin{subfigure}{0.23\textwidth}
      \centering
      \includegraphics[height=2.8cm]{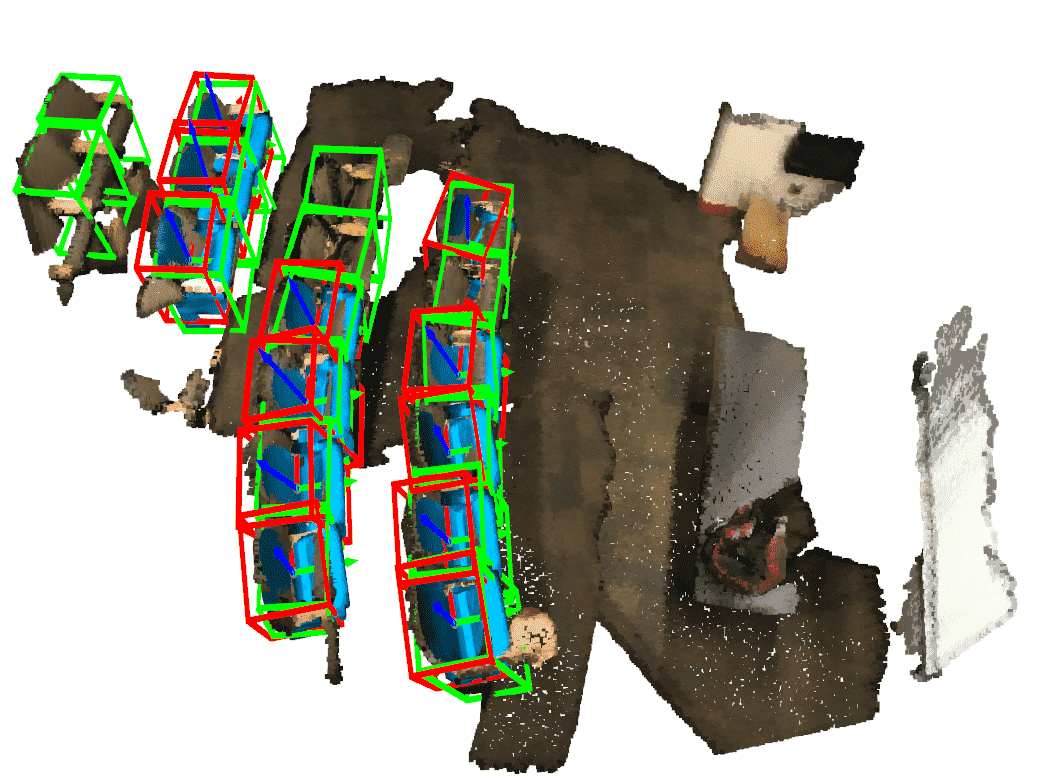}
        \caption{Ours}
        \label{fig:scan2cad_cad-result19}
    \end{subfigure}\hfill
    \begin{subfigure}{0.23\textwidth}
      \centering
      \includegraphics[height=2.8cm]{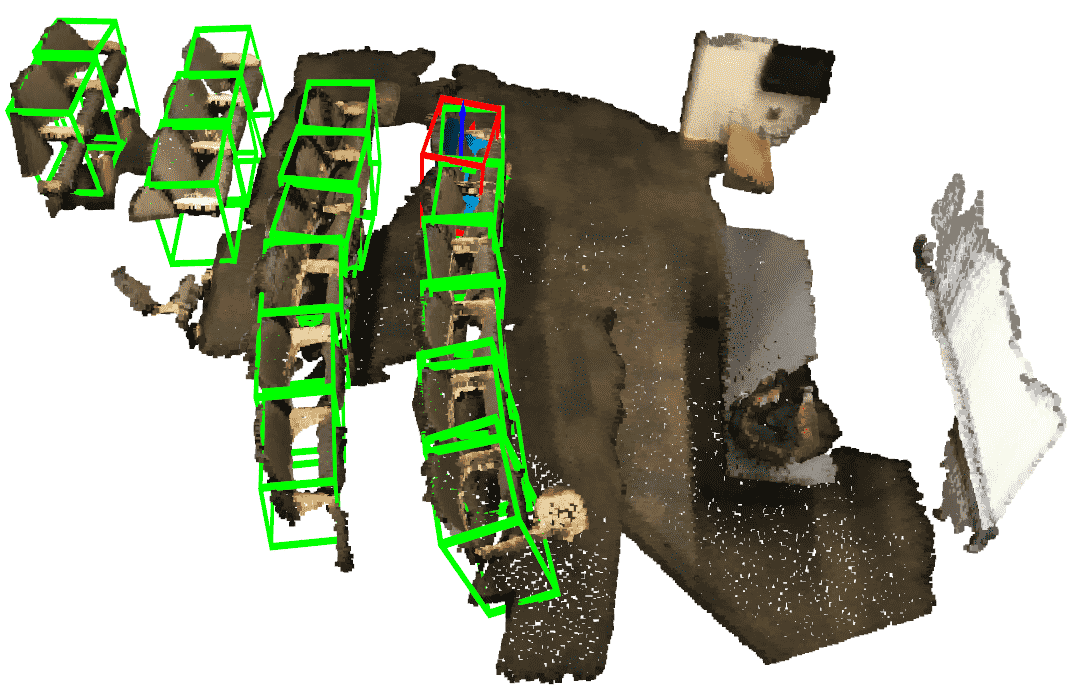}
        \caption{Progressive-X(2019) \cite{ProgressiveX}}
        \label{fig:scan2cad_cad-prox19}
    \end{subfigure}\hfill
    \begin{subfigure}{0.23\textwidth}
      \centering
      \includegraphics[height=2.8cm]{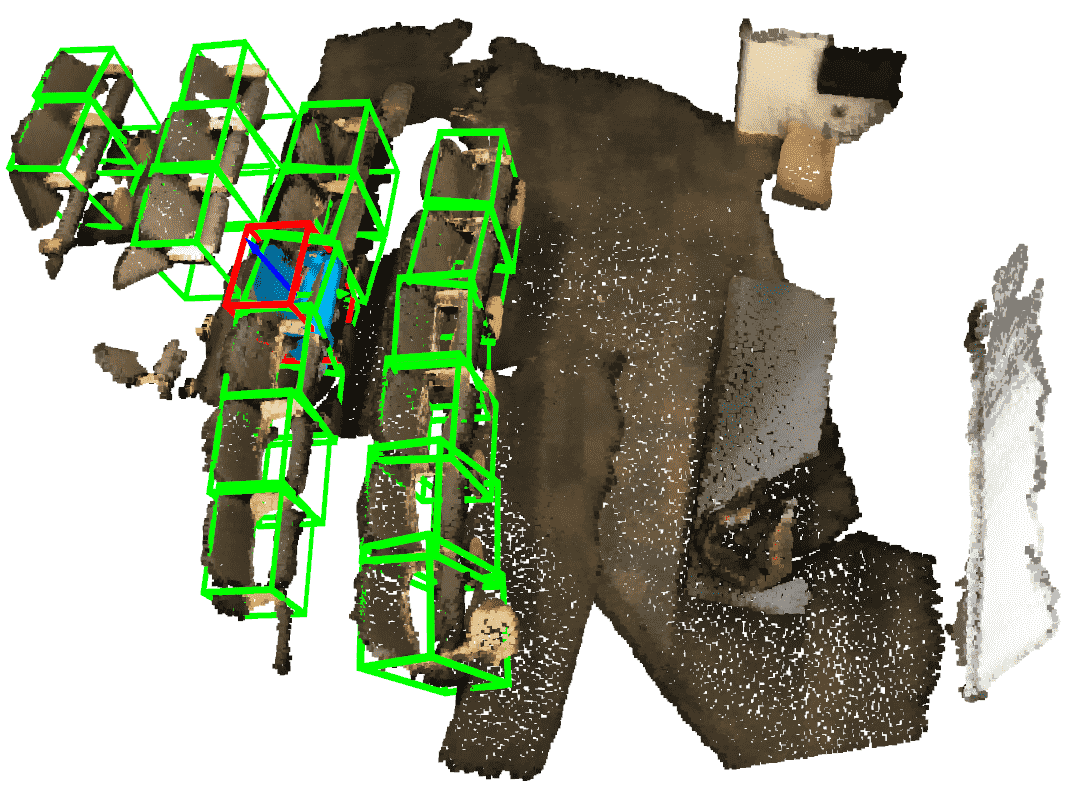}
        \caption{TEASER(2020)\cite{TEASER}}
        \label{fig:scan2cad_cad-teaser19}
    \end{subfigure}
    \begin{subfigure}{0.23\textwidth}
      \centering
      \includegraphics[height=2.8cm]{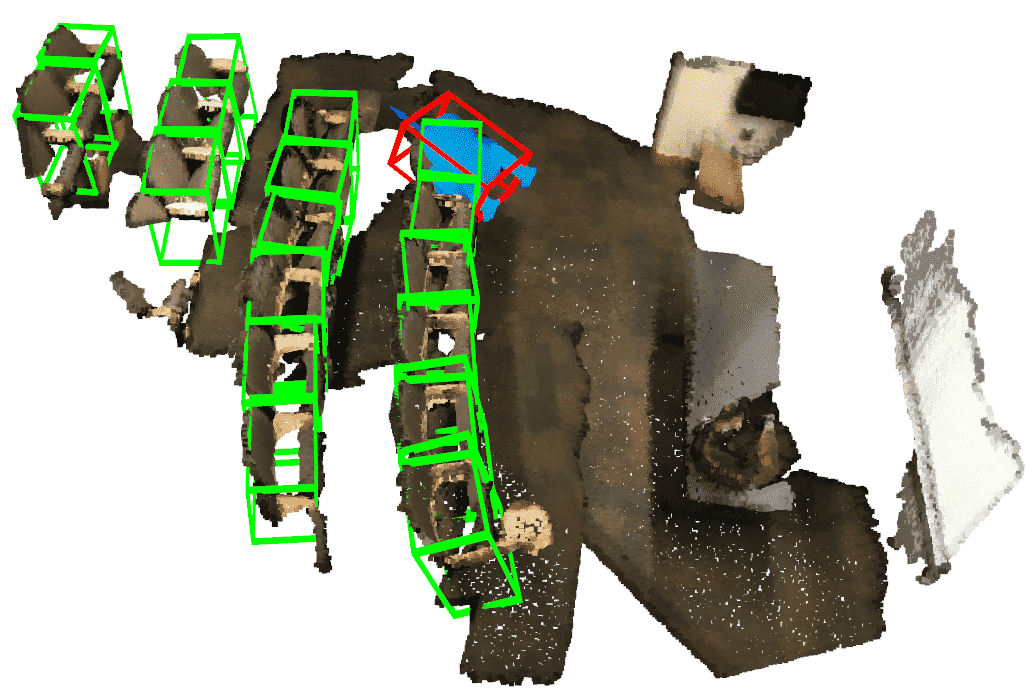}
        \caption{CONSAC(2020)\cite{CONSAC}}
        \label{fig:scan2cad_cad-consac19}
    \end{subfigure}\hfill
    \caption{\textbf{Scan2CAD results.}}
\label{fig:Scan2CAD-cadresult19}
  \end{figure*}

\begin{figure*}[ht]
  \centering
  \begin{subfigure}{0.3\textwidth}
      \centering
      \includegraphics[height=2.8cm]{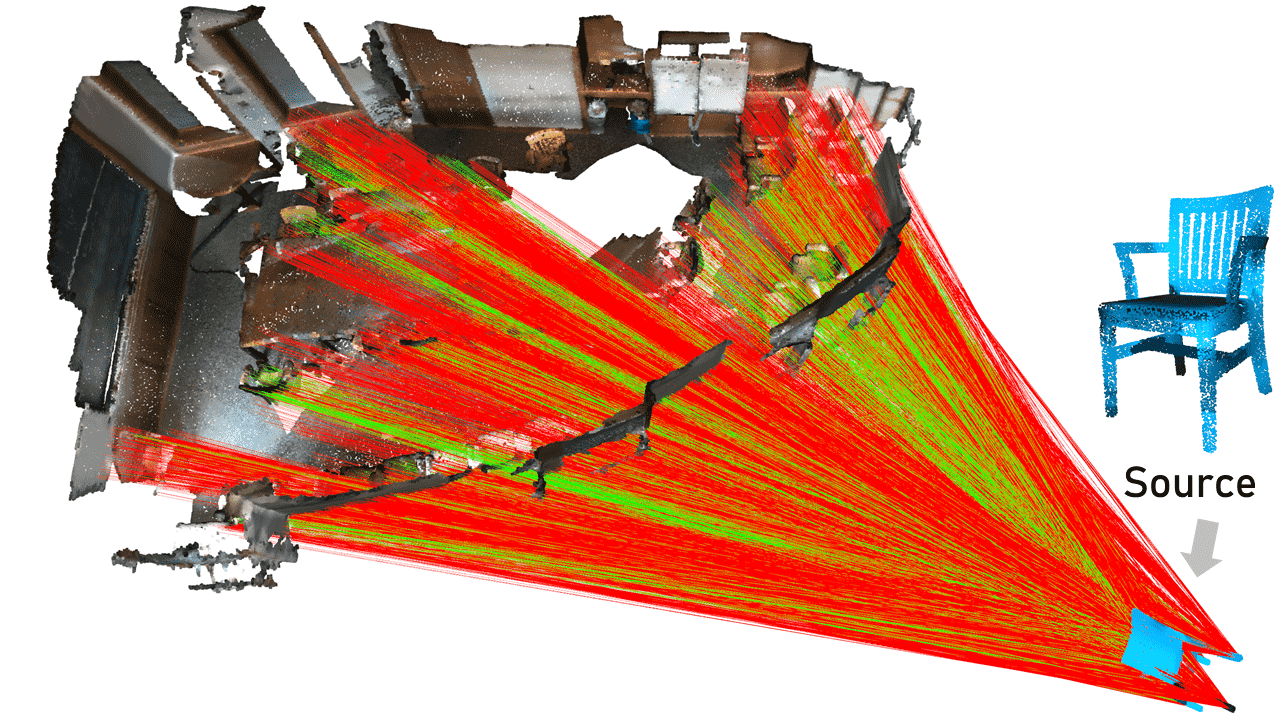}
        \caption{Input correspondences}
        \label{fig:scan2cad_cad-input-corrs20}
    \end{subfigure}\hfill
    \begin{subfigure}{0.3\textwidth}
      \centering
      \includegraphics[height=2.8cm]{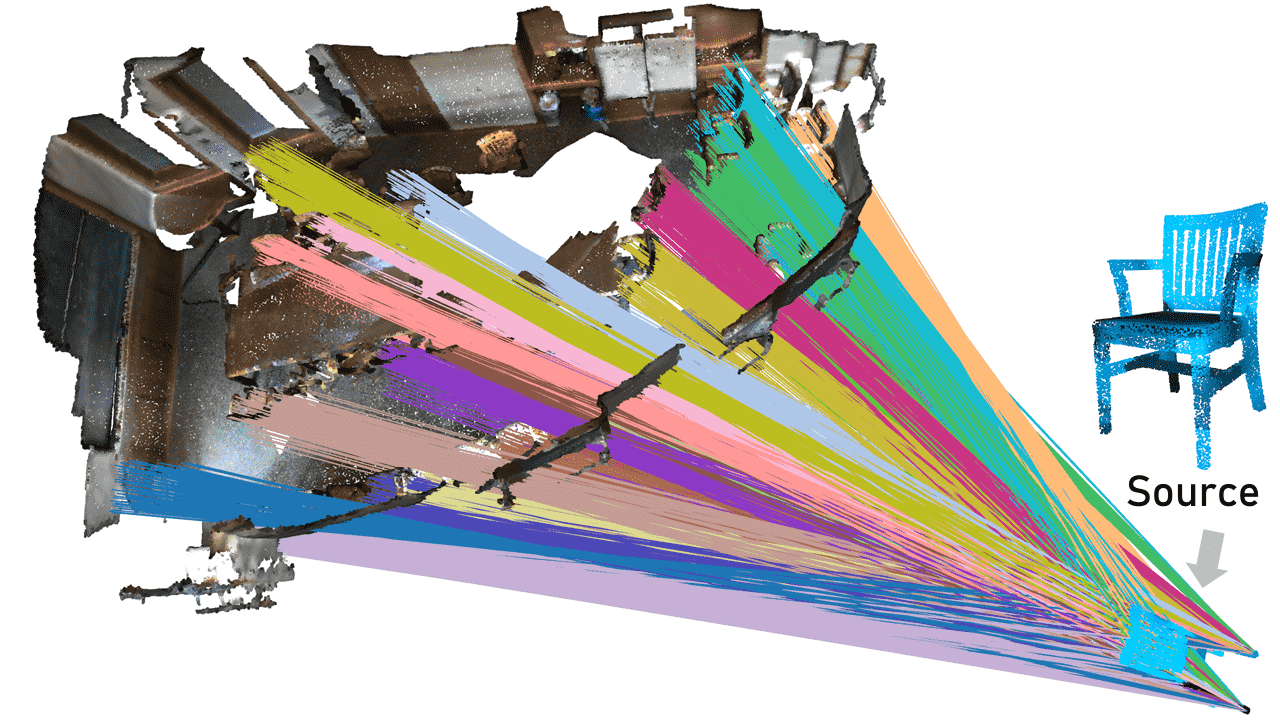}
        \caption{Our clustering result}
        \label{fig:scan2cad_cad-cluster-corrs20}
    \end{subfigure}\hfill
    \begin{subfigure}{0.3\textwidth}
      \centering
      \includegraphics[height=2.8cm]{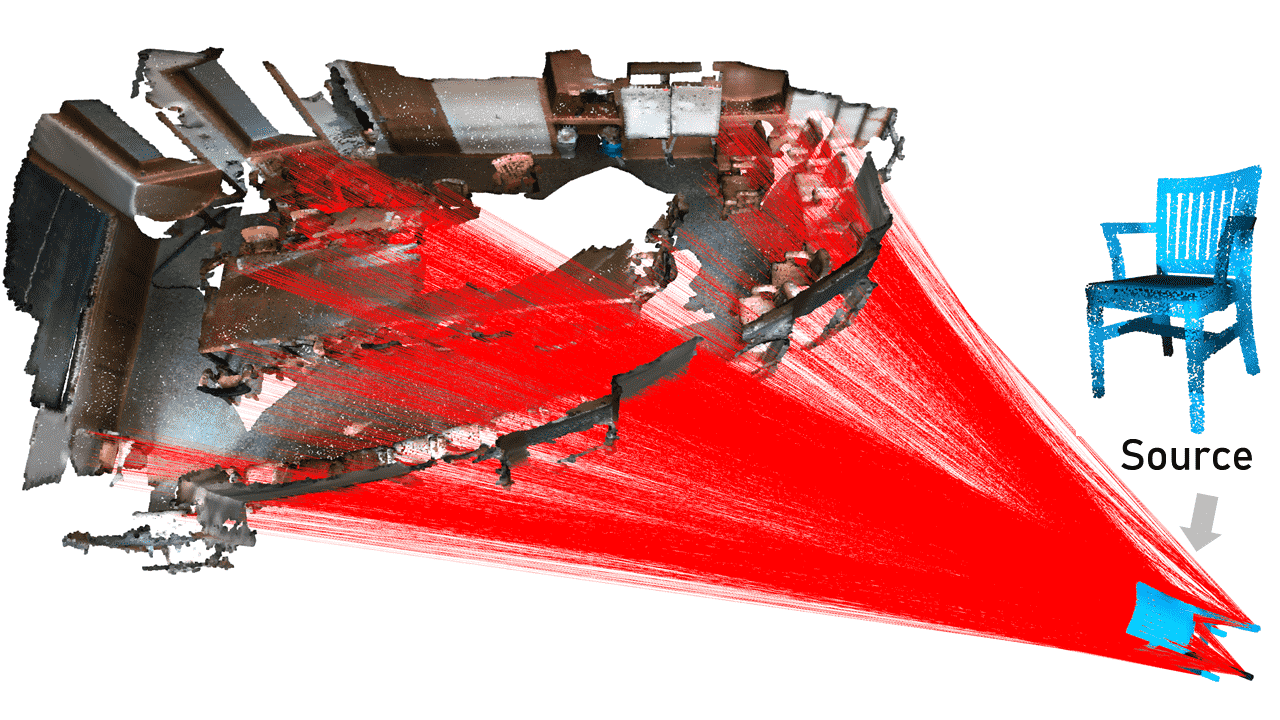}
        \caption{Our reject outliers}
        \label{fig:scan2cad_cad-reject-corrs20}
    \end{subfigure}

    \begin{subfigure}{0.23\textwidth}
      \centering
      \includegraphics[height=2.8cm]{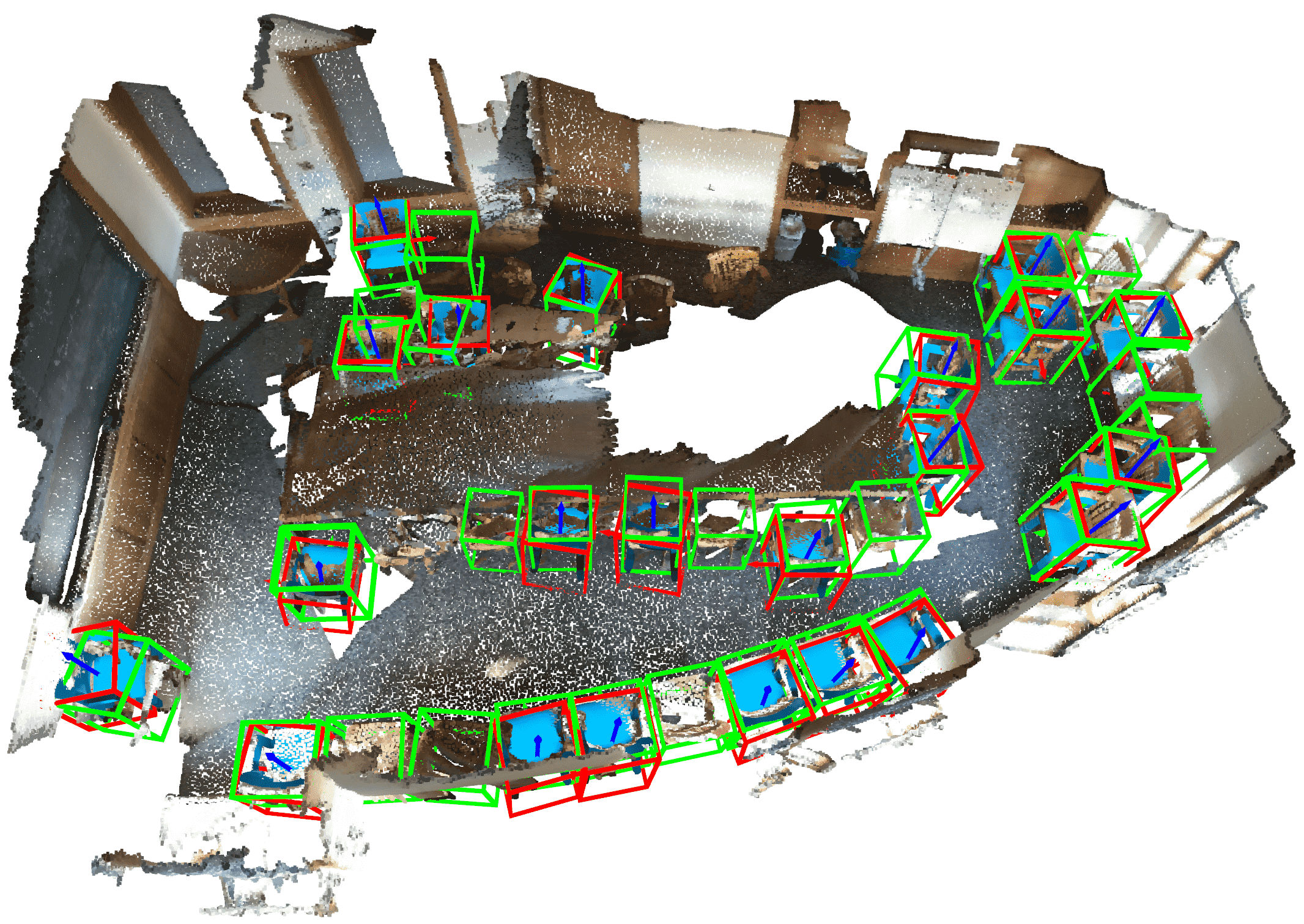}
        \caption{Ours}
        \label{fig:scan2cad_cad-result20}
    \end{subfigure}\hfill
    \begin{subfigure}{0.23\textwidth}
      \centering
      \includegraphics[height=2.8cm]{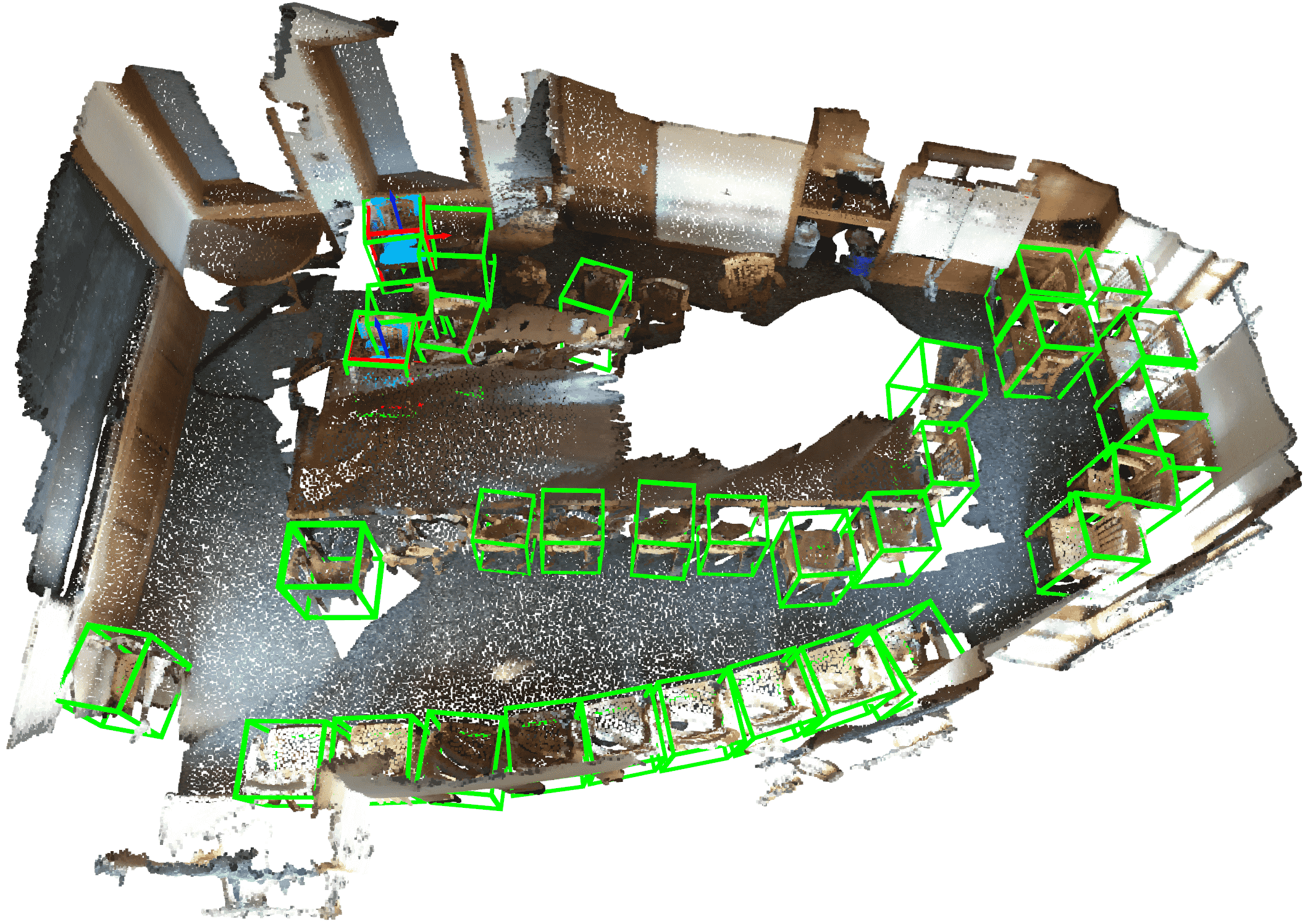}
        \caption{Progressive-X(2019) \cite{ProgressiveX}}
        \label{fig:scan2cad_cad-prox20}
    \end{subfigure}\hfill
    \begin{subfigure}{0.23\textwidth}
      \centering
      \includegraphics[height=2.8cm]{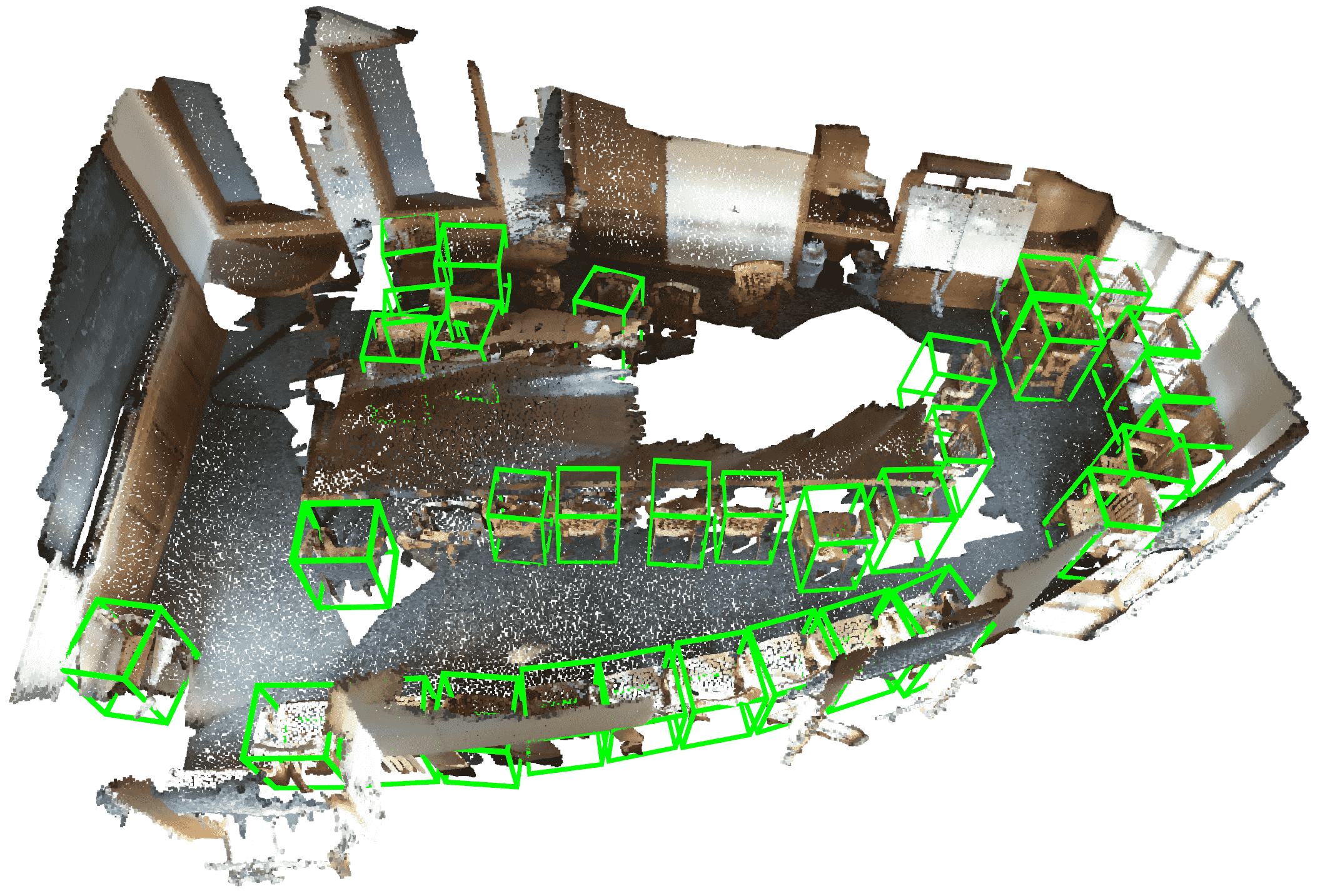}
        \caption{TEASER(2020)\cite{TEASER}}
        \label{fig:scan2cad_cad-teaser20}
    \end{subfigure}
    \begin{subfigure}{0.23\textwidth}
      \centering
      \includegraphics[height=2.8cm]{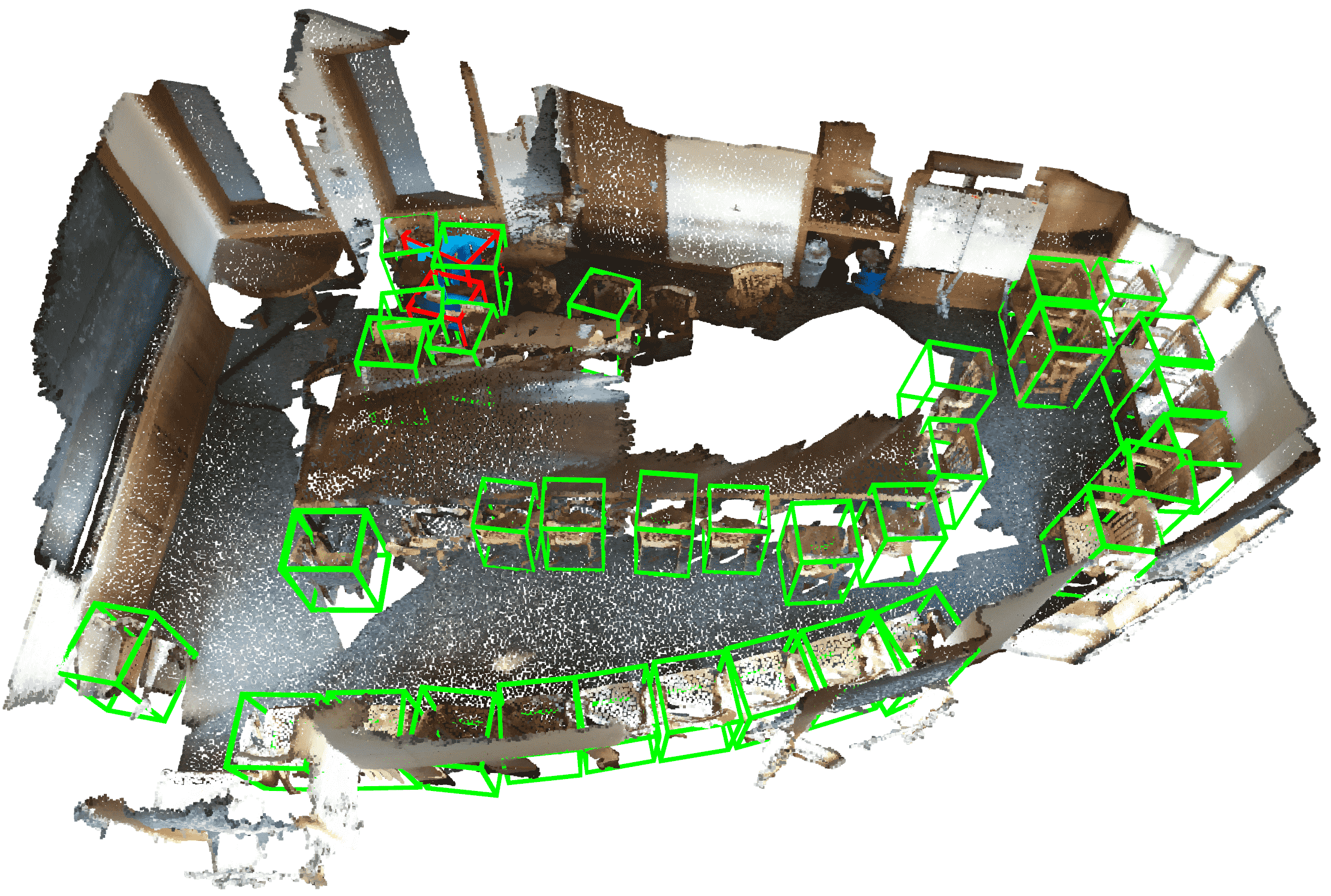}
        \caption{CONSAC(2020)\cite{CONSAC}}
        \label{fig:scan2cad_cad-consac20}
    \end{subfigure}\hfill
    \caption{\textbf{Scan2CAD results.}}
\label{fig:Scan2CAD-cadresult20}
  \end{figure*}

\begin{figure*}[ht]
  \centering
  \begin{subfigure}{0.3\textwidth}
      \centering
      \includegraphics[height=2.8cm]{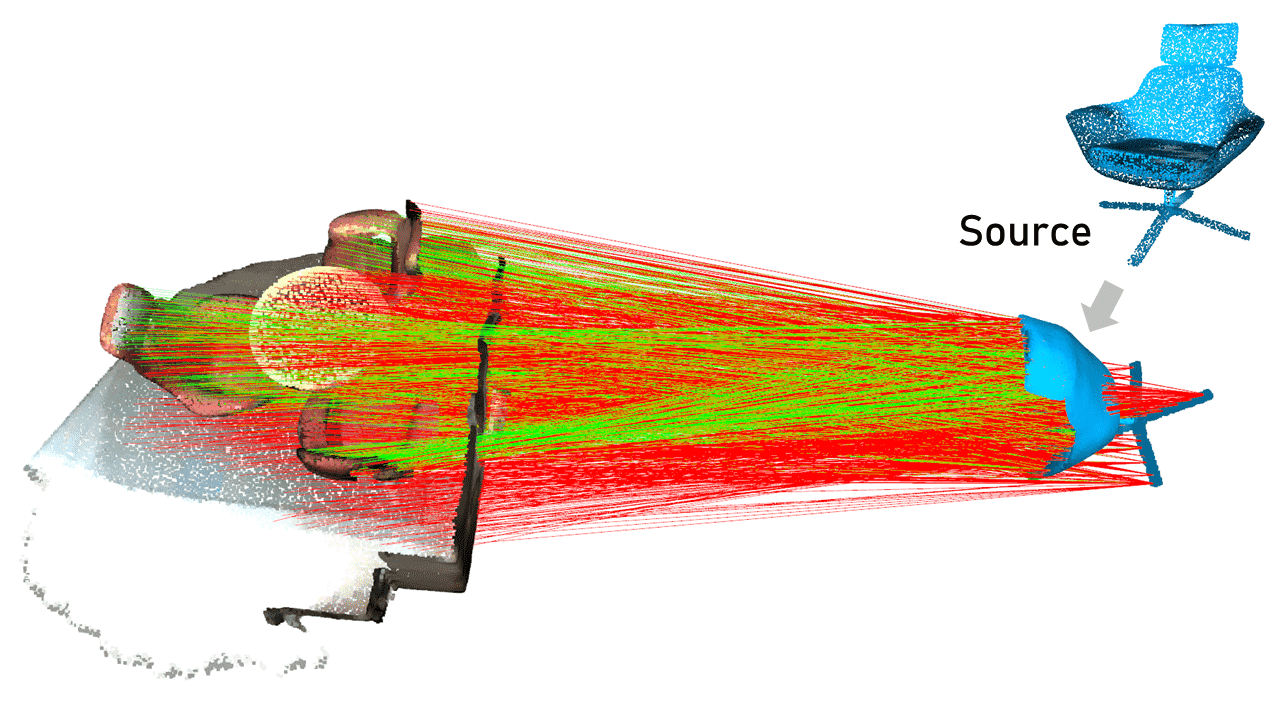}
        \caption{Input correspondences}
        \label{fig:scan2cad_cad-input-corrs11}
    \end{subfigure}\hfill
    \begin{subfigure}{0.3\textwidth}
      \centering
      \includegraphics[height=2.8cm]{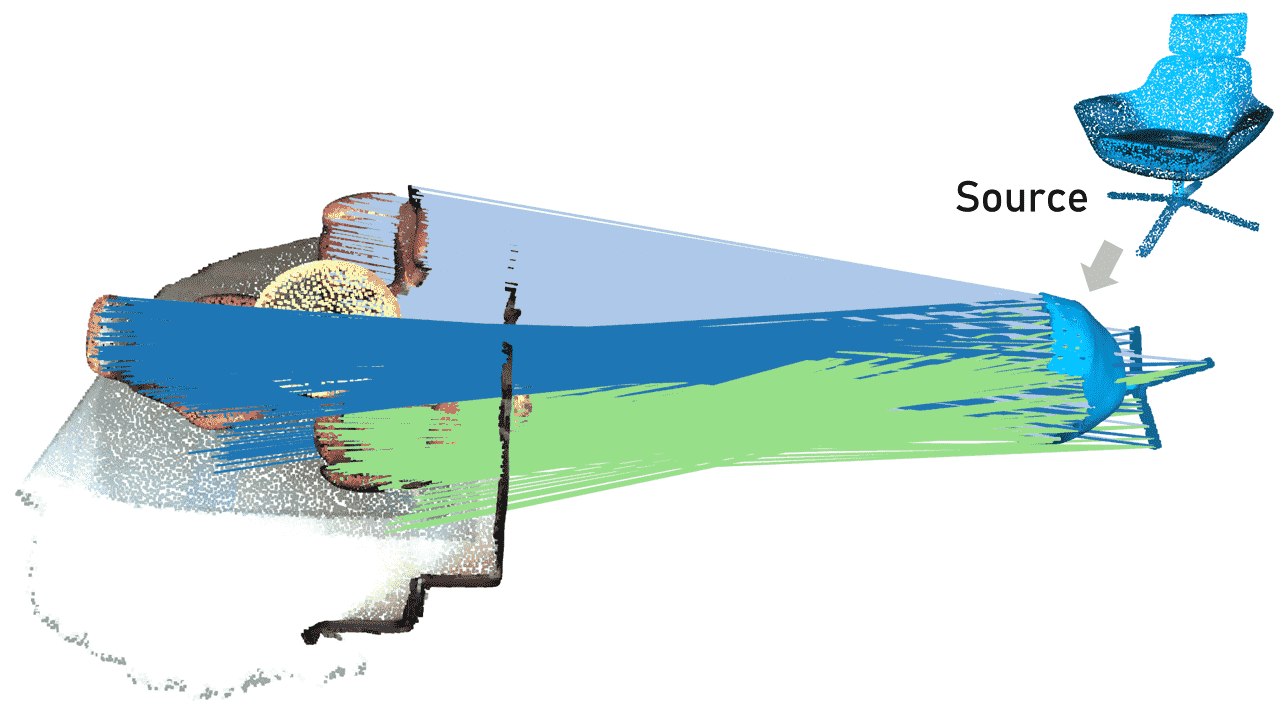}
        \caption{Our clustering result}
        \label{fig:scan2cad_cad-cluster-corrs11}
    \end{subfigure}\hfill
    \begin{subfigure}{0.3\textwidth}
      \centering
      \includegraphics[height=2.8cm]{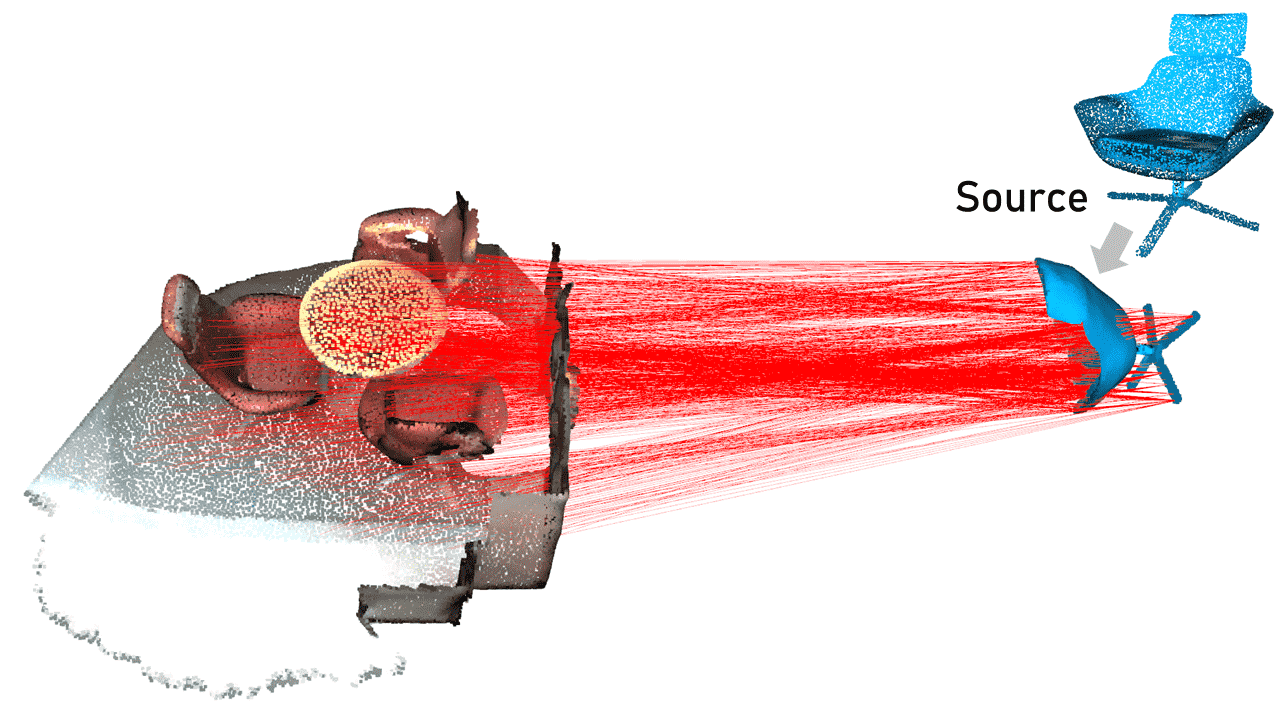}
        \caption{Our reject outliers}
        \label{fig:scan2cad_cad-reject-corrs11}
    \end{subfigure}

    \begin{subfigure}{0.23\textwidth}
      \centering
      \includegraphics[height=2.8cm]{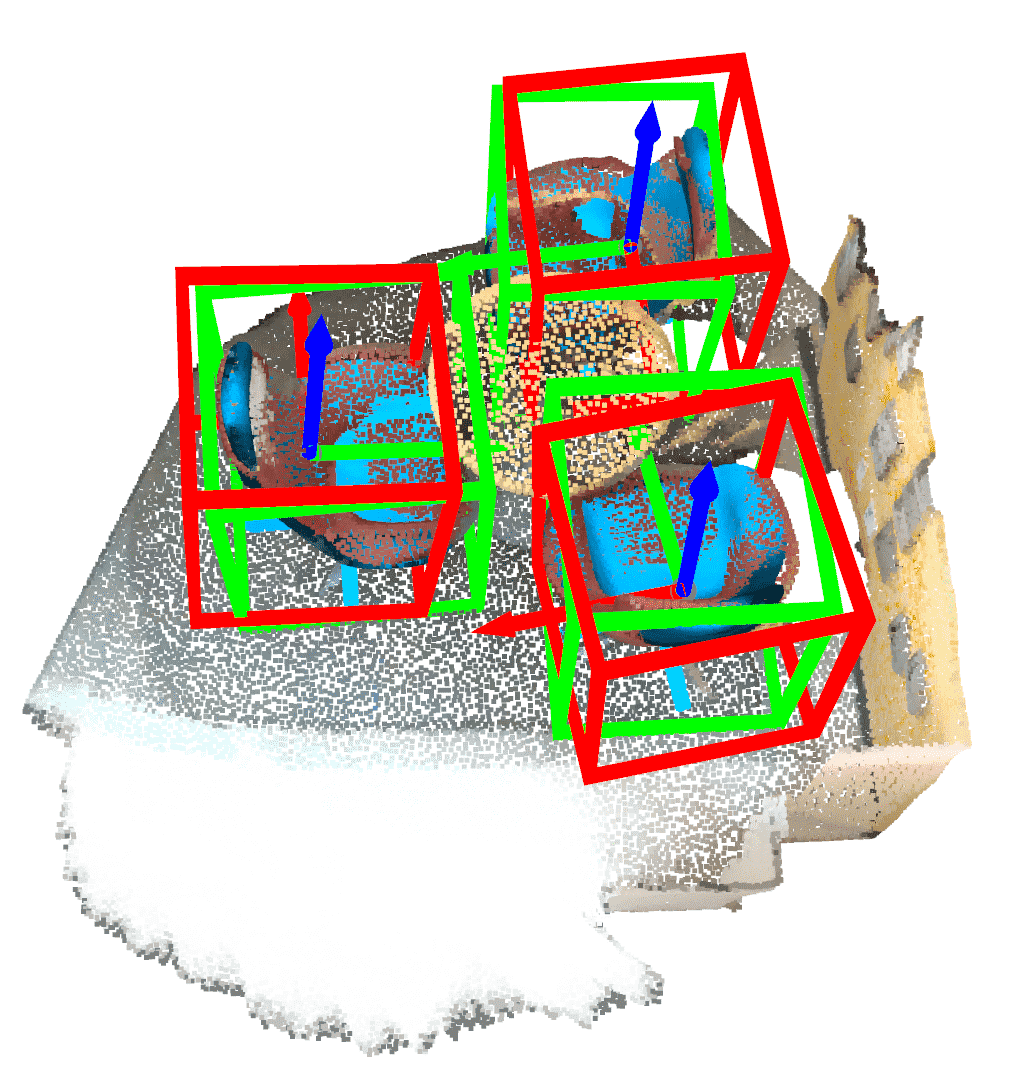}
        \caption{Ours}
        \label{fig:scan2cad_cad-result11}
    \end{subfigure}\hfill
    \begin{subfigure}{0.23\textwidth}
      \centering
      \includegraphics[height=2.8cm]{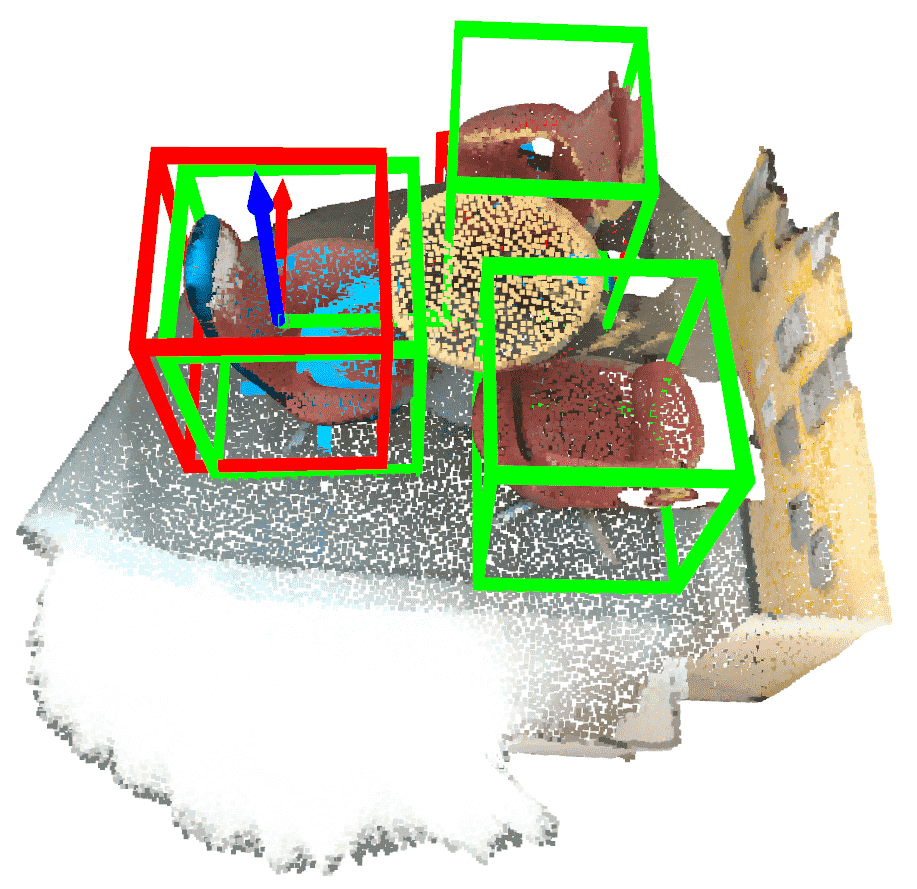}
        \caption{Progressive-X(2019) \cite{ProgressiveX}}
        \label{fig:scan2cad_cad-prox11}
    \end{subfigure}\hfill
    \begin{subfigure}{0.23\textwidth}
      \centering
      \includegraphics[height=2.8cm]{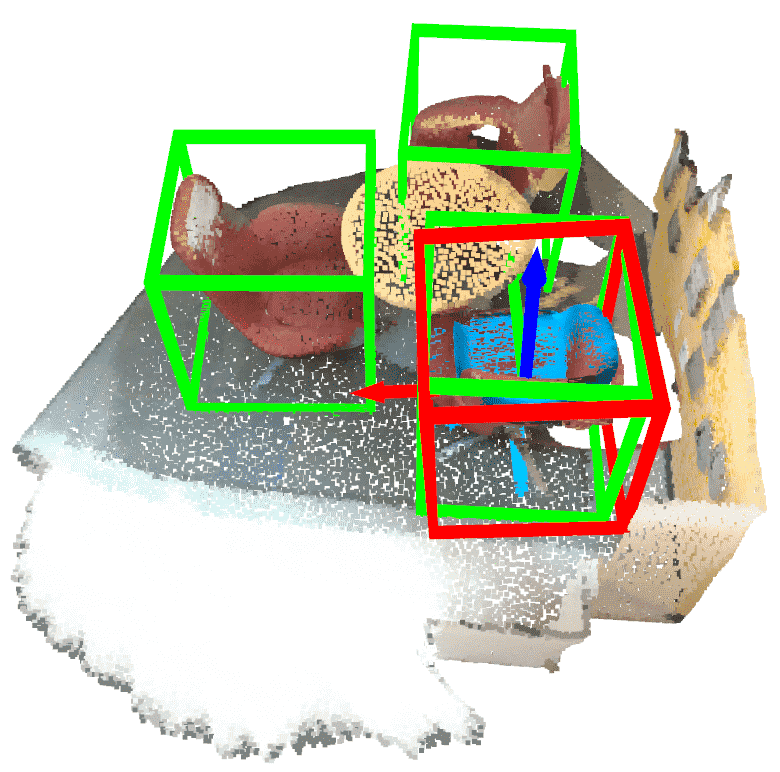}
        \caption{TEASER(2020)\cite{TEASER}}
        \label{fig:scan2cad_cad-teaser11}
    \end{subfigure}
    \begin{subfigure}{0.23\textwidth}
      \centering
      \includegraphics[height=2.8cm]{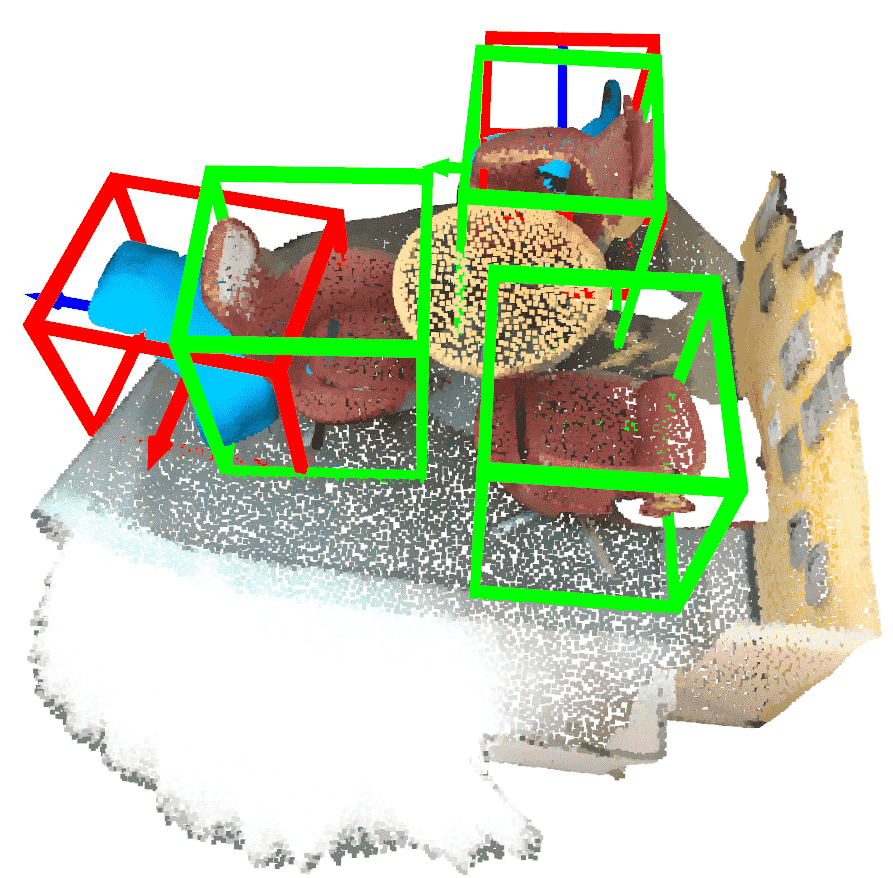}
        \caption{CONSAC(2020)\cite{CONSAC}}
        \label{fig:scan2cad_cad-consac11}
    \end{subfigure}\hfill
    \caption{\textbf{Scan2CAD results.}}
\label{fig:Scan2CAD-cadresult11}
  \end{figure*}

\begin{figure*}[ht]
  \centering
  \begin{subfigure}{0.3\textwidth}
      \centering
      \includegraphics[height=2.8cm]{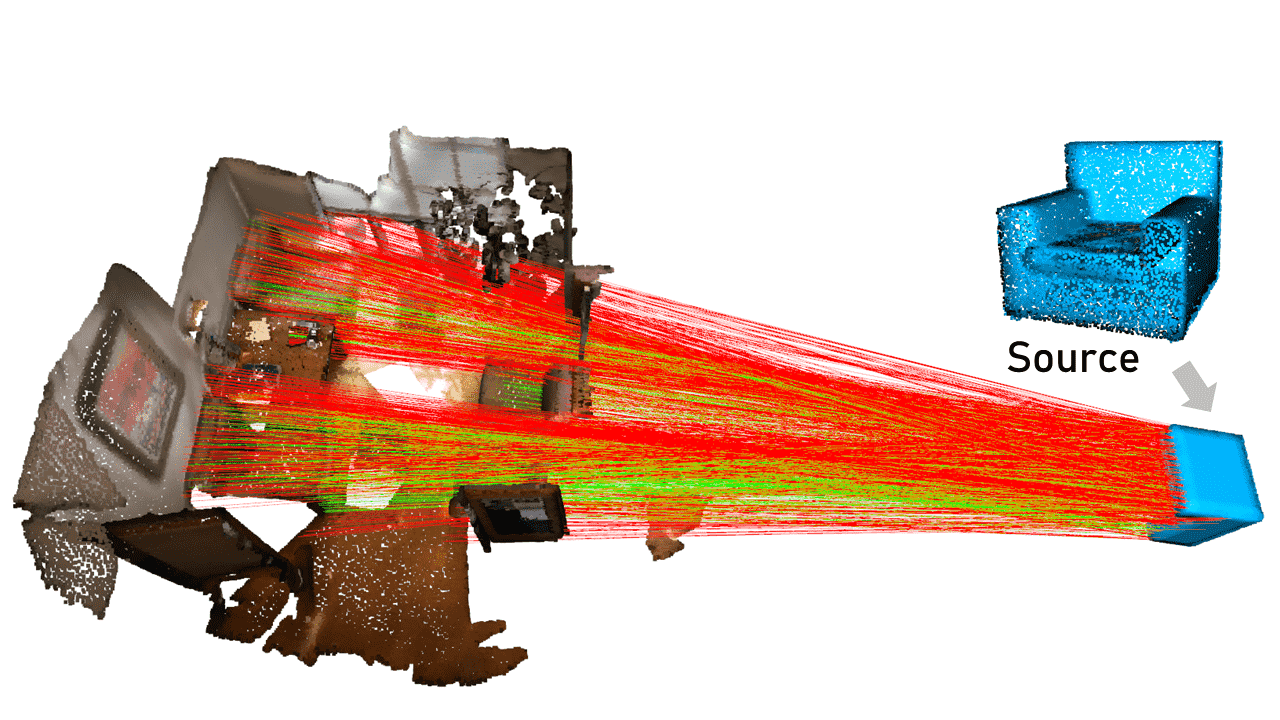}
        \caption{Input correspondences}
        \label{fig:scan2cad_cad-input-corrs12}
    \end{subfigure}\hfill
    \begin{subfigure}{0.3\textwidth}
      \centering
      \includegraphics[height=2.8cm]{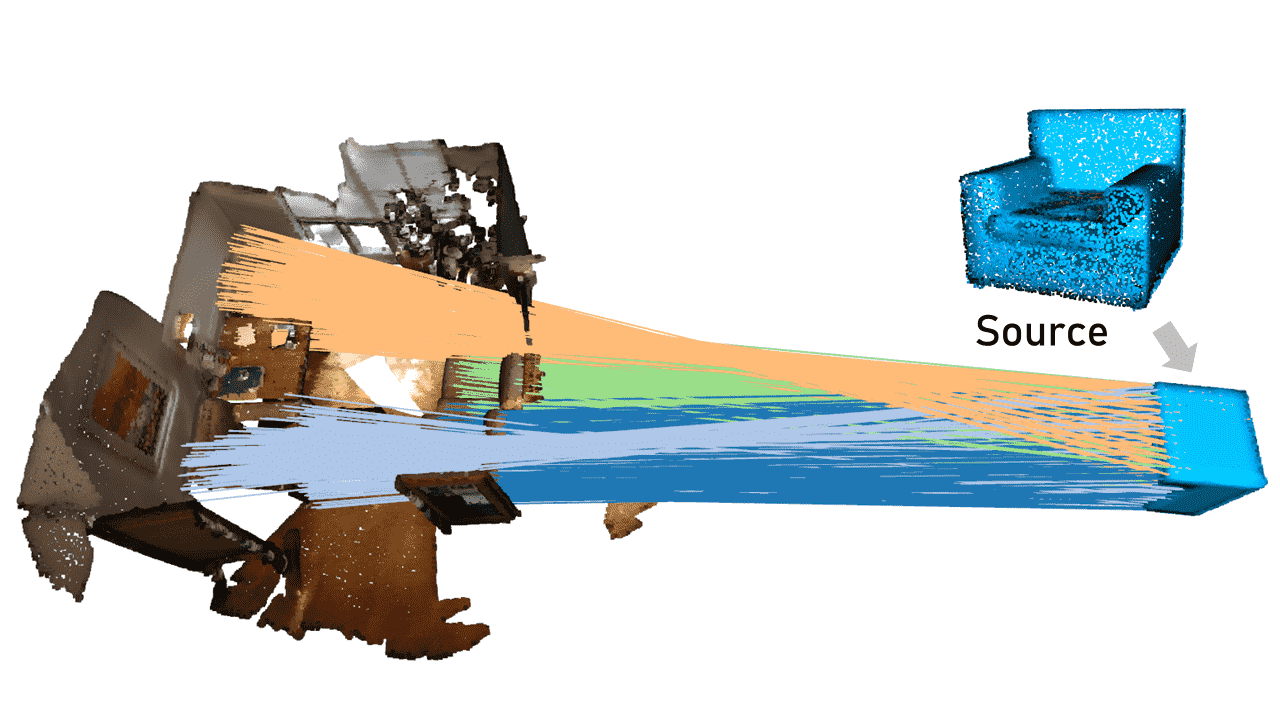}
        \caption{Our clustering result}
        \label{fig:scan2cad_cad-cluster-corrs12}
    \end{subfigure}\hfill
    \begin{subfigure}{0.3\textwidth}
      \centering
      \includegraphics[height=2.8cm]{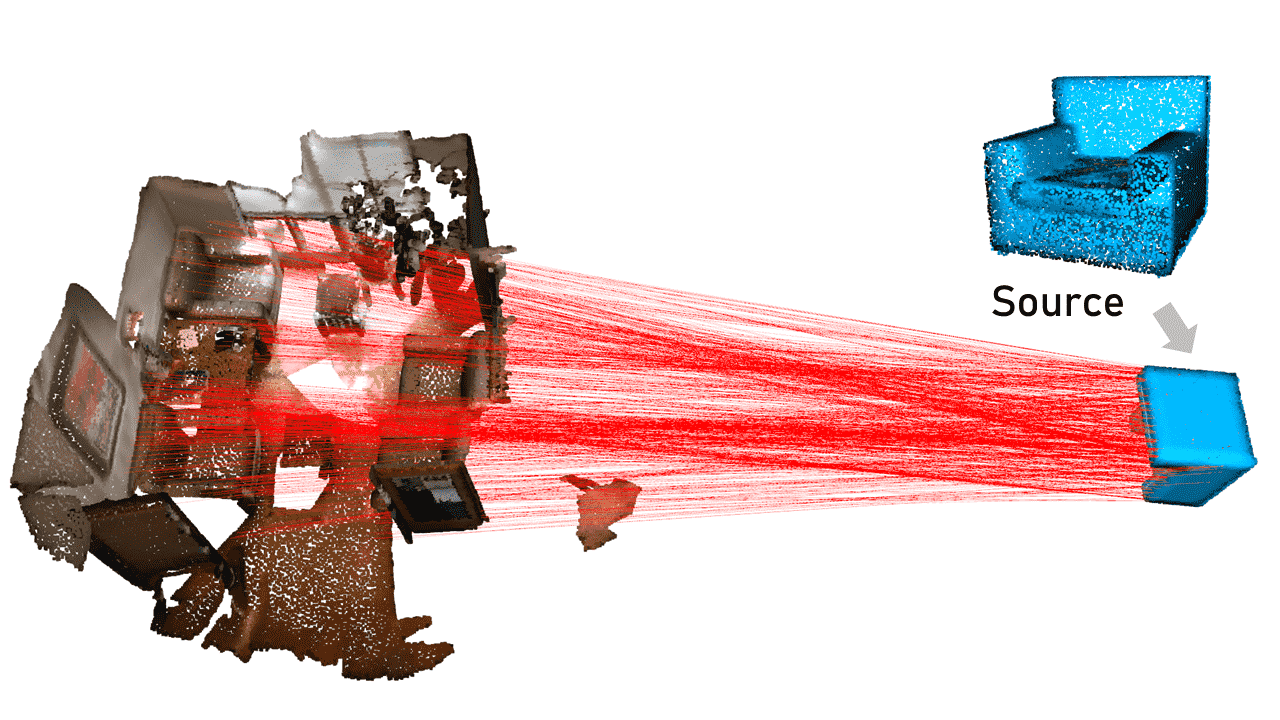}
        \caption{Our reject outliers}
        \label{fig:scan2cad_cad-reject-corrs12}
    \end{subfigure}

    \begin{subfigure}{0.23\textwidth}
      \centering
      \includegraphics[height=2.8cm]{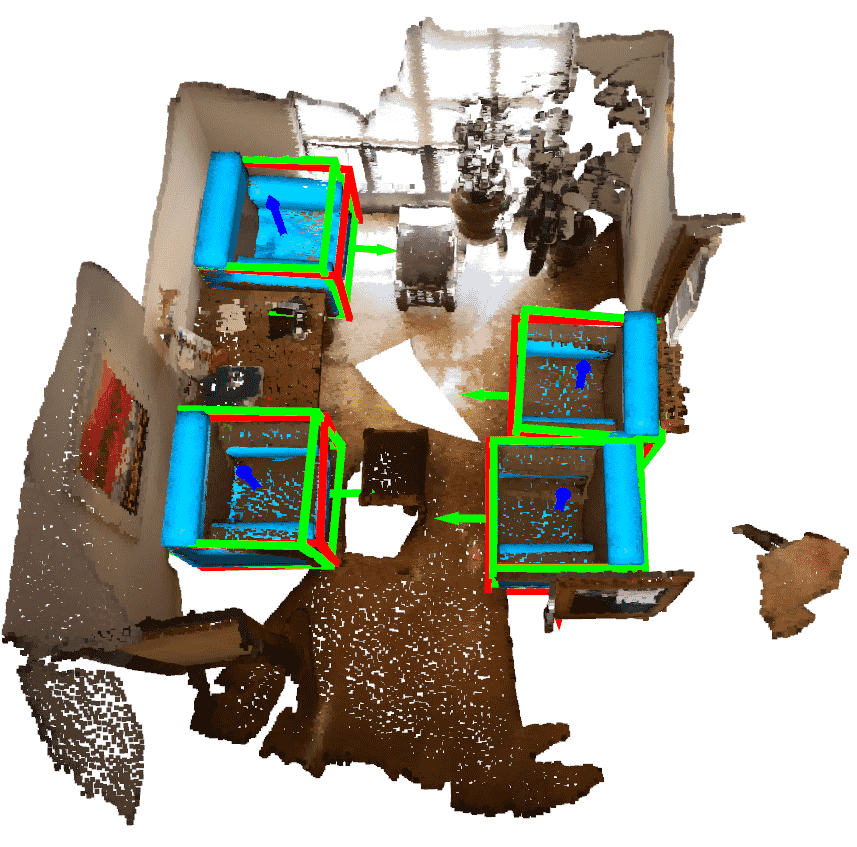}
        \caption{Ours}
        \label{fig:scan2cad_cad-result12}
    \end{subfigure}\hfill
    \begin{subfigure}{0.23\textwidth}
      \centering
      \includegraphics[height=2.8cm]{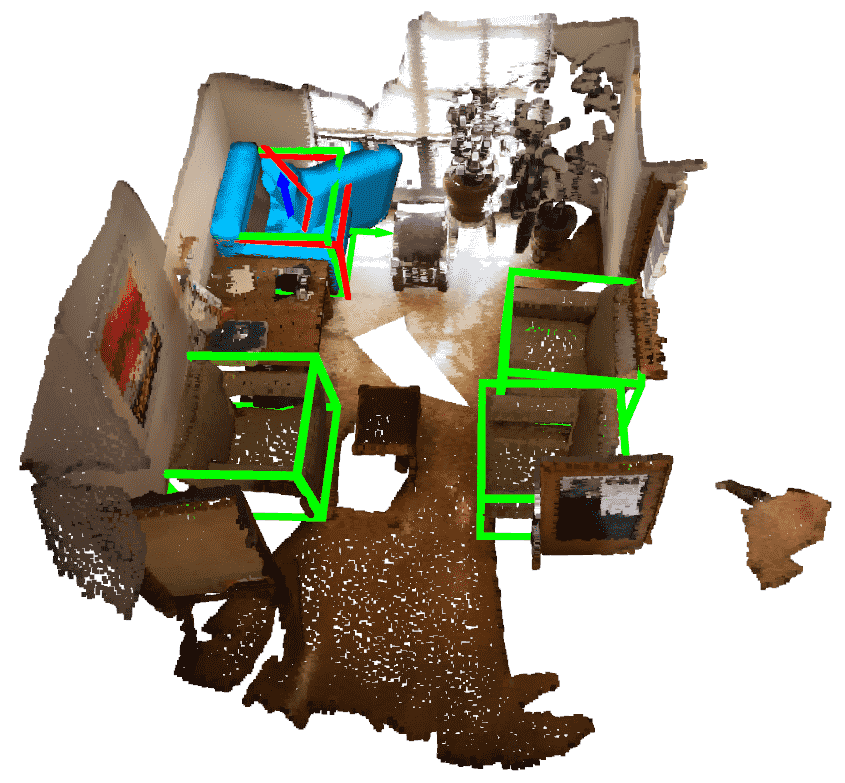}
        \caption{Progressive-X(2019) \cite{ProgressiveX}}
        \label{fig:scan2cad_cad-prox12}
    \end{subfigure}\hfill
    \begin{subfigure}{0.23\textwidth}
      \centering
      \includegraphics[height=2.8cm]{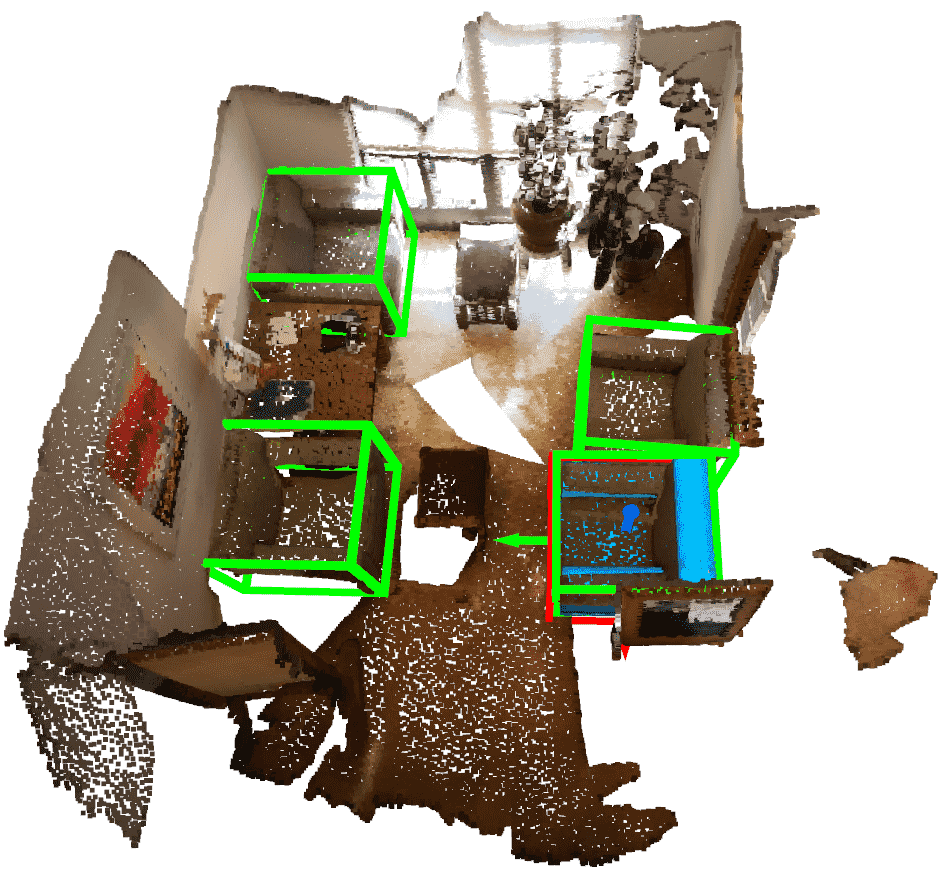}
        \caption{TEASER(2020)\cite{TEASER}}
        \label{fig:scan2cad_cad-teaser12}
    \end{subfigure}
    \begin{subfigure}{0.23\textwidth}
      \centering
      \includegraphics[height=2.8cm]{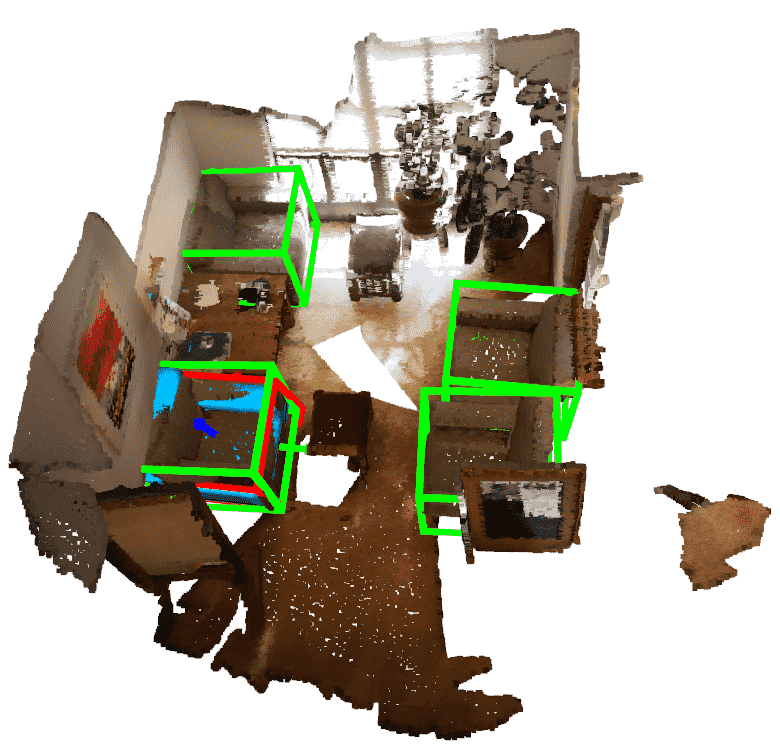}
        \caption{CONSAC(2020)\cite{CONSAC}}
        \label{fig:scan2cad_cad-consac12}
    \end{subfigure}\hfill
    \caption{\textbf{Scan2CAD results.}}
\label{fig:Scan2CAD-cadresult12}
  \end{figure*}

\begin{figure*}[ht]
  \centering
  \begin{subfigure}{0.3\textwidth}
      \centering
      \includegraphics[height=2.8cm]{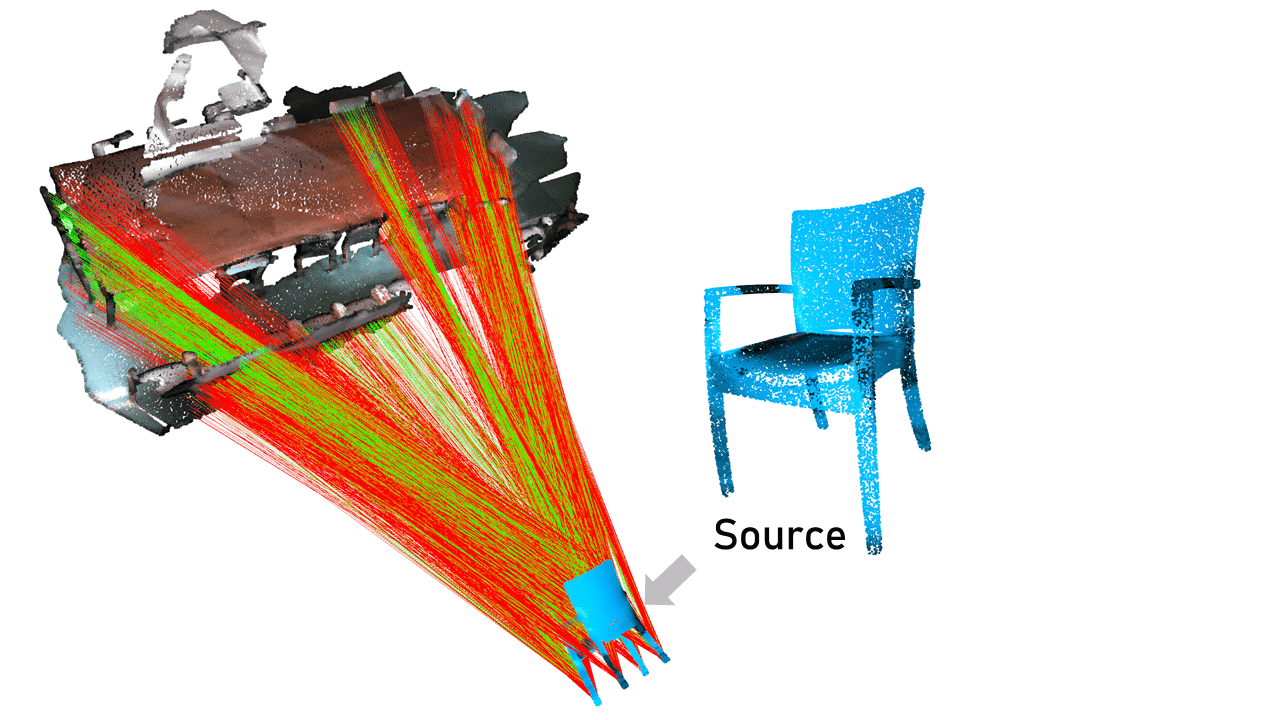}
        \caption{Input correspondences}
        \label{fig:scan2cad_cad-input-corrs17}
    \end{subfigure}\hfill
    \begin{subfigure}{0.3\textwidth}
      \centering
      \includegraphics[height=2.8cm]{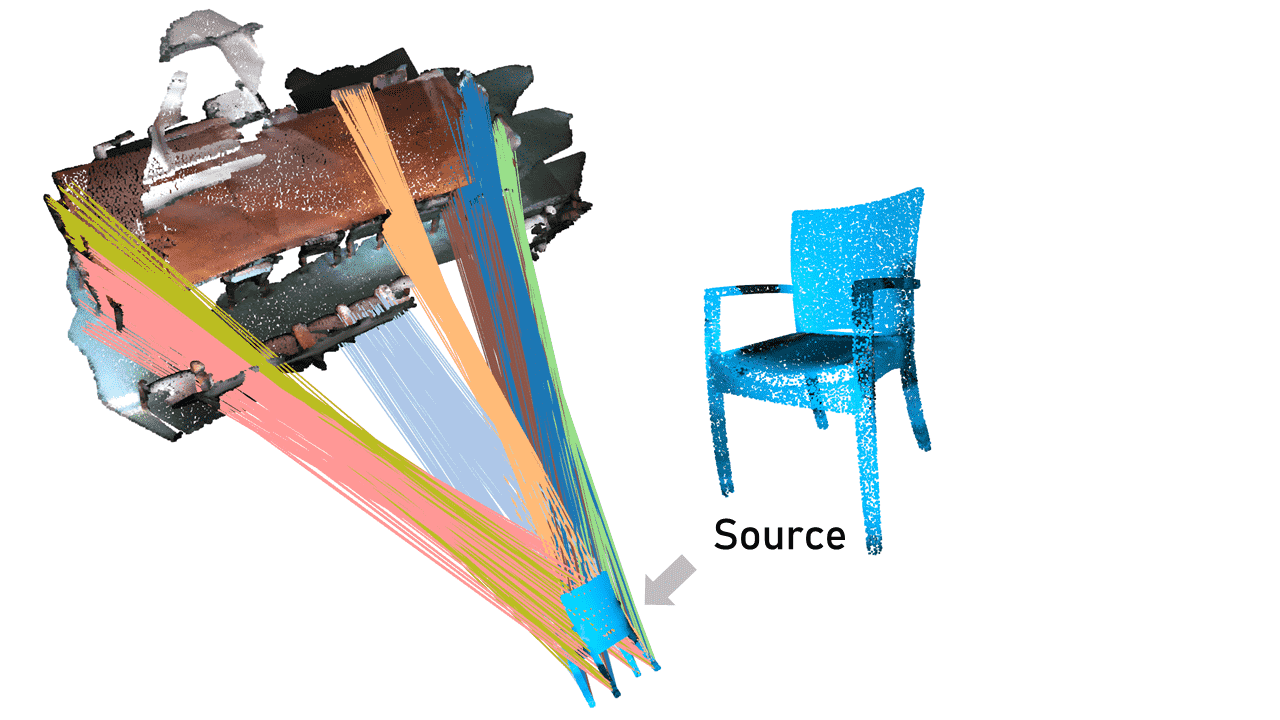}
        \caption{Our clustering result}
        \label{fig:scan2cad_cad-cluster-corrs17}
    \end{subfigure}\hfill
    \begin{subfigure}{0.3\textwidth}
      \centering
      \includegraphics[height=2.8cm]{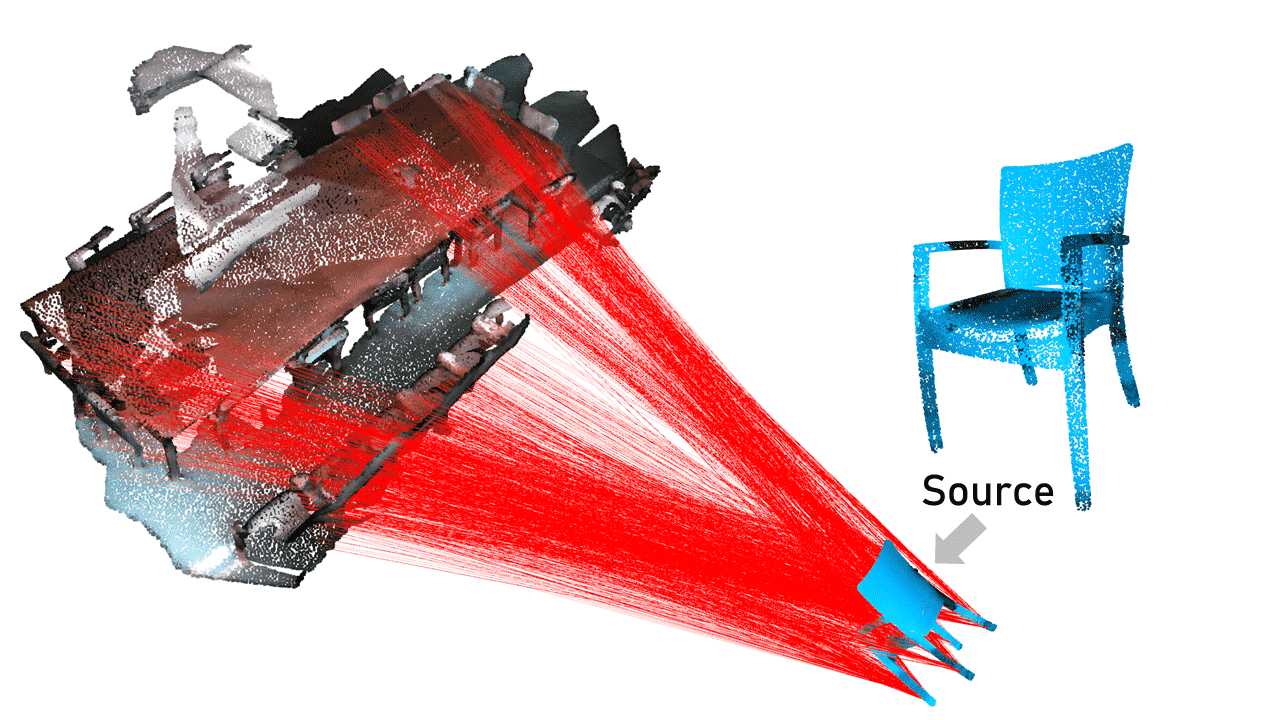}
        \caption{Our reject outliers}
        \label{fig:scan2cad_cad-reject-corrs17}
    \end{subfigure}

    \begin{subfigure}{0.23\textwidth}
      \centering
      \includegraphics[height=2.8cm]{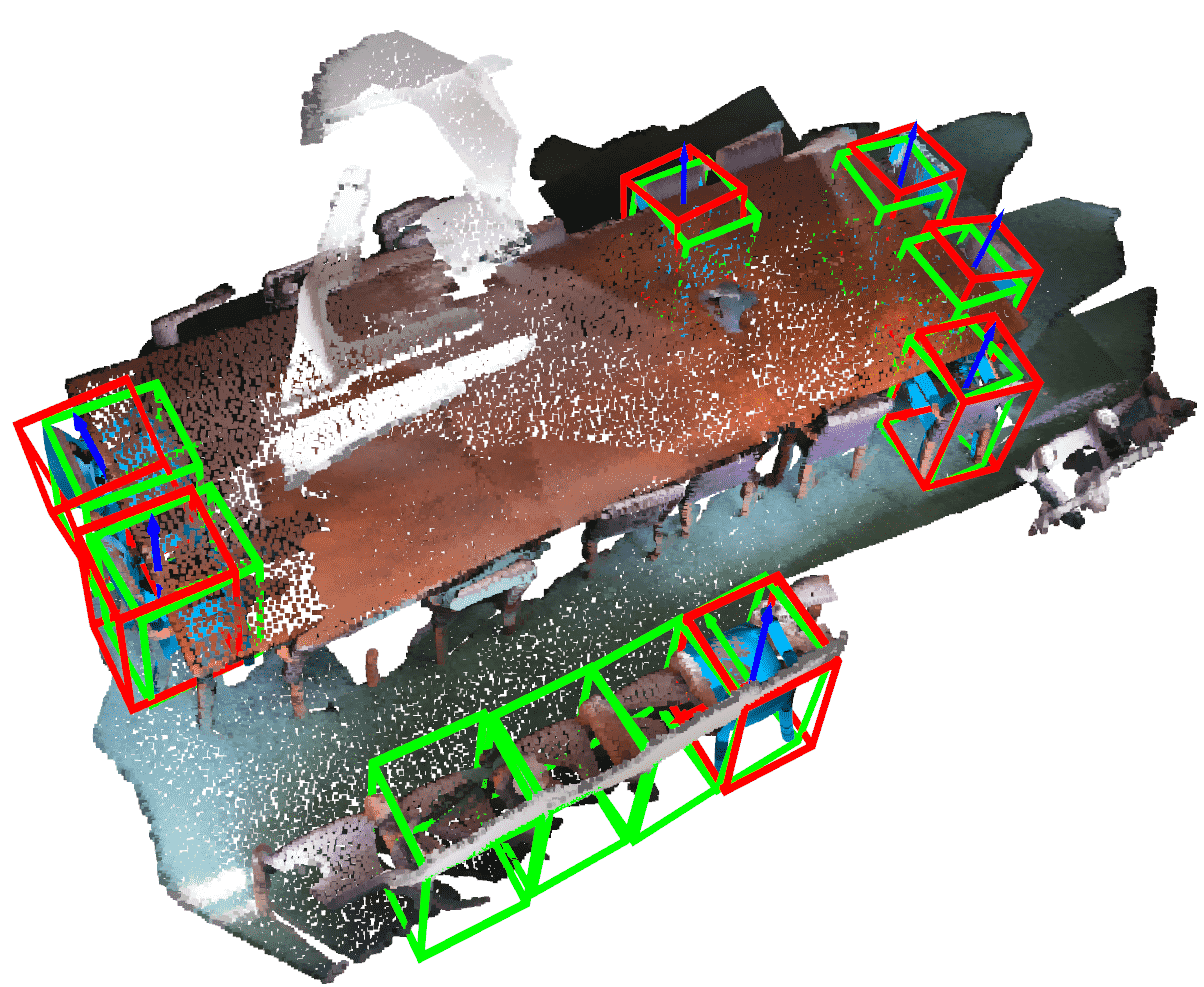}
        \caption{Ours}
        \label{fig:scan2cad_cad-result17}
    \end{subfigure}\hfill
    \begin{subfigure}{0.23\textwidth}
      \centering
      \includegraphics[height=2.8cm]{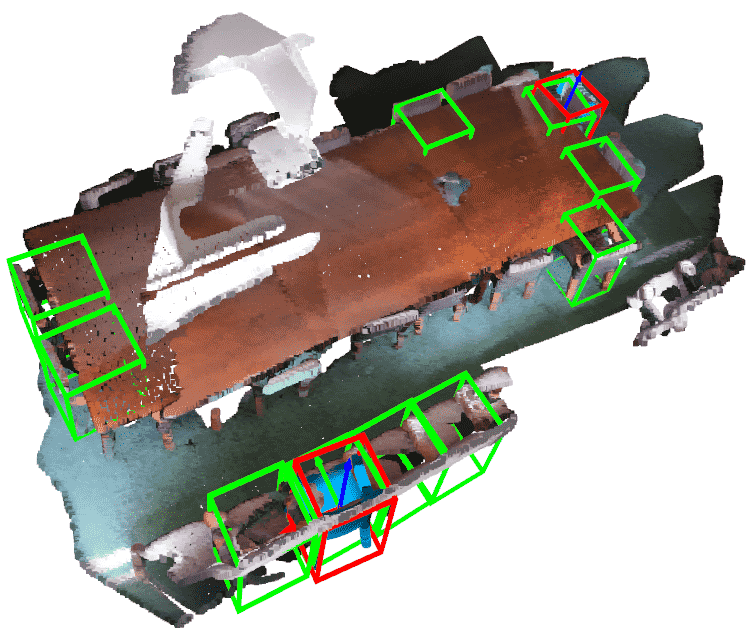}
        \caption{Progressive-X(2019) \cite{ProgressiveX}}
        \label{fig:scan2cad_cad-prox17}
    \end{subfigure}\hfill
    \begin{subfigure}{0.23\textwidth}
      \centering
      \includegraphics[height=2.8cm]{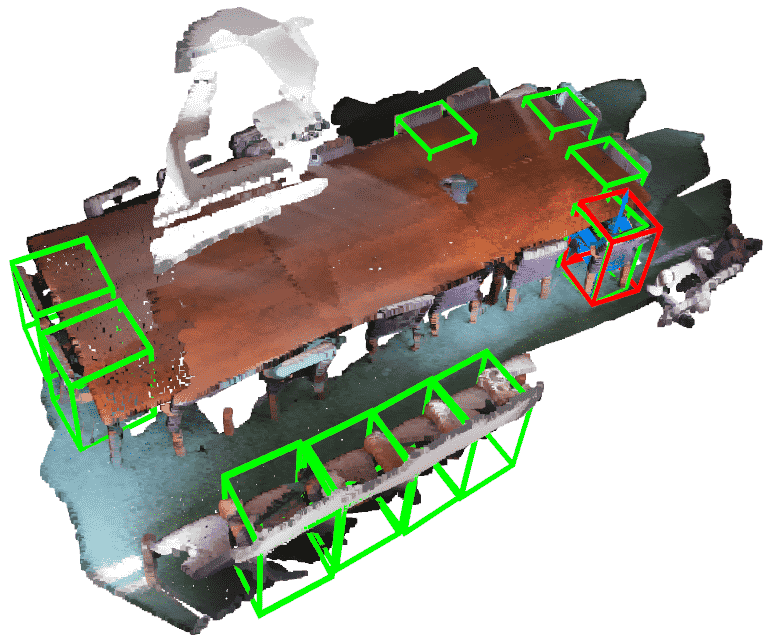}
        \caption{TEASER(2020)\cite{TEASER}}
        \label{fig:scan2cad_cad-teaser17}
    \end{subfigure}
    \begin{subfigure}{0.23\textwidth}
      \centering
      \includegraphics[height=2.8cm]{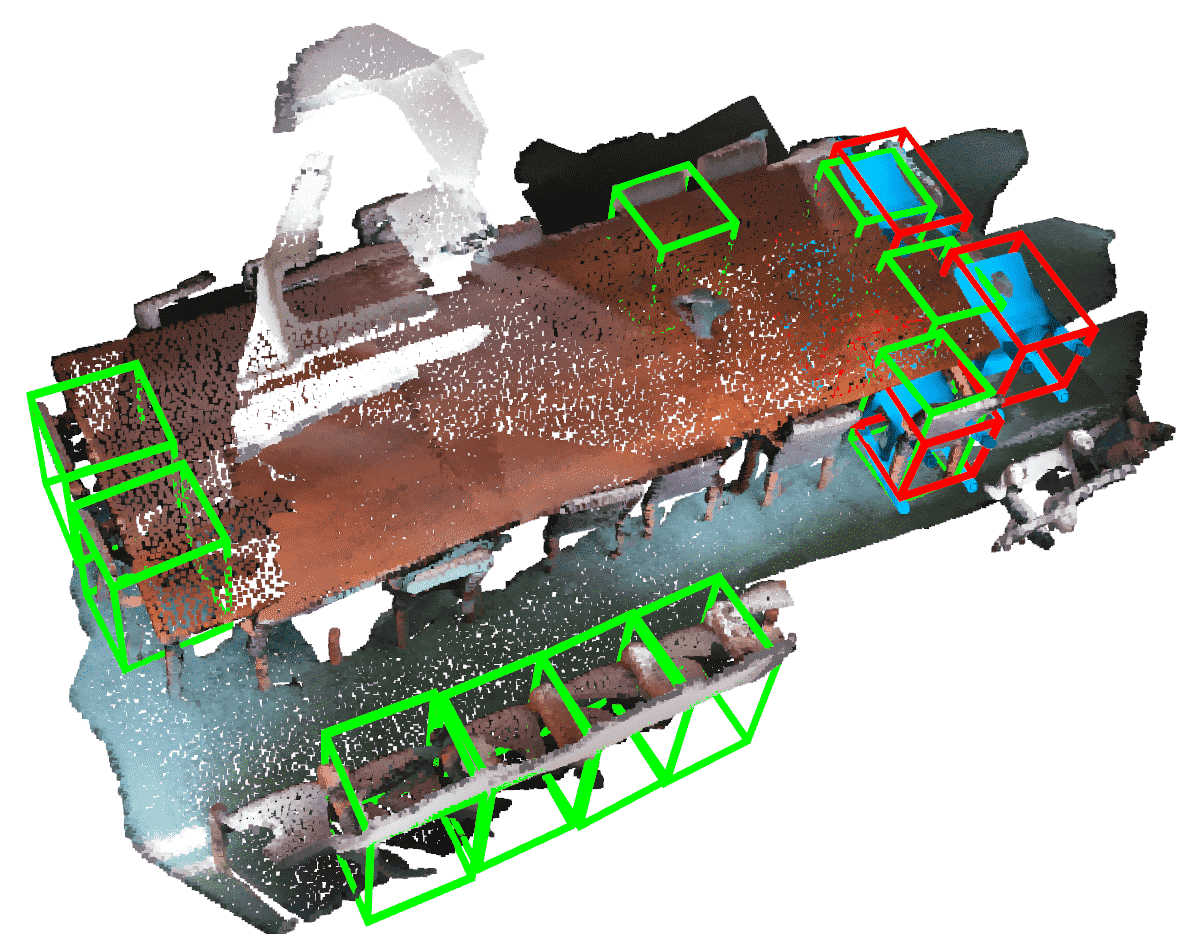}
        \caption{CONSAC(2020)\cite{CONSAC}}
        \label{fig:scan2cad_cad-consac17}
    \end{subfigure}\hfill
    \caption{\textbf{Scan2CAD results.}}
\label{fig:Scan2CAD-cadresult17}
  \end{figure*}

\begin{figure*}[ht]
  \centering
  \begin{subfigure}{0.3\textwidth}
      \centering
      \includegraphics[height=2.8cm]{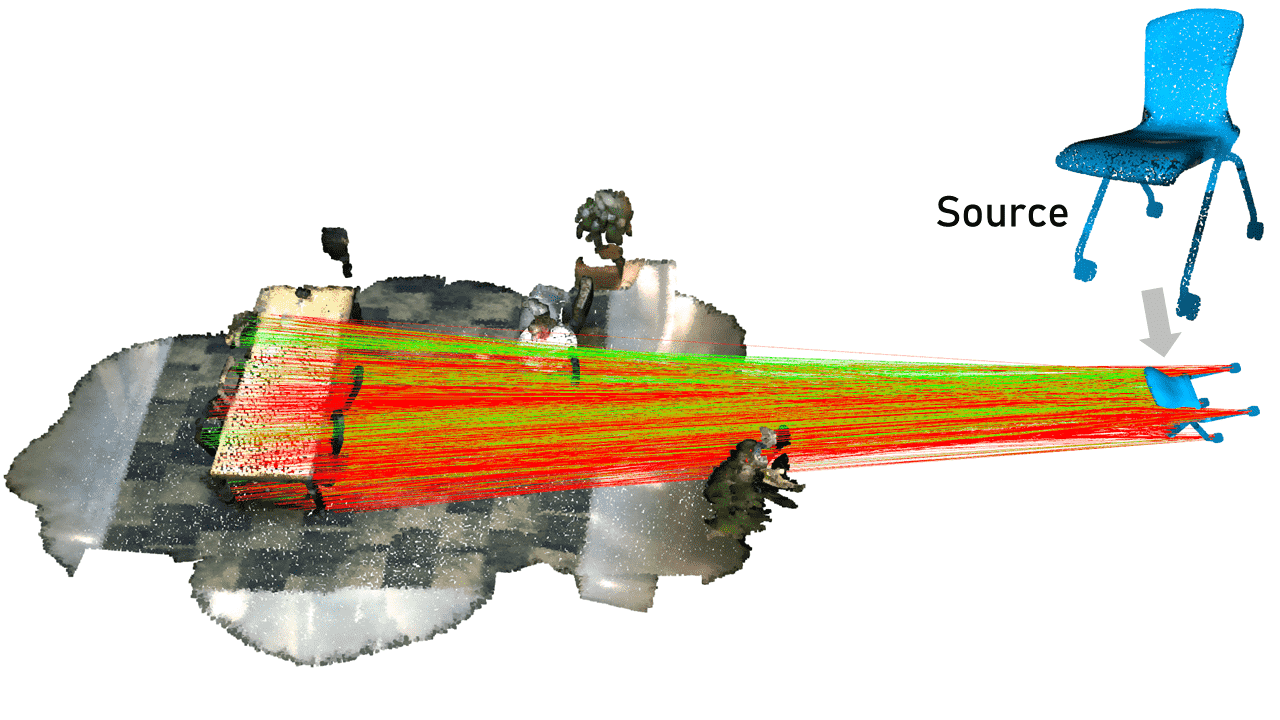}
        \caption{Input correspondences}
        \label{fig:scan2cad_cad-input-corrs22}
    \end{subfigure}\hfill
    \begin{subfigure}{0.3\textwidth}
      \centering
      \includegraphics[height=2.8cm]{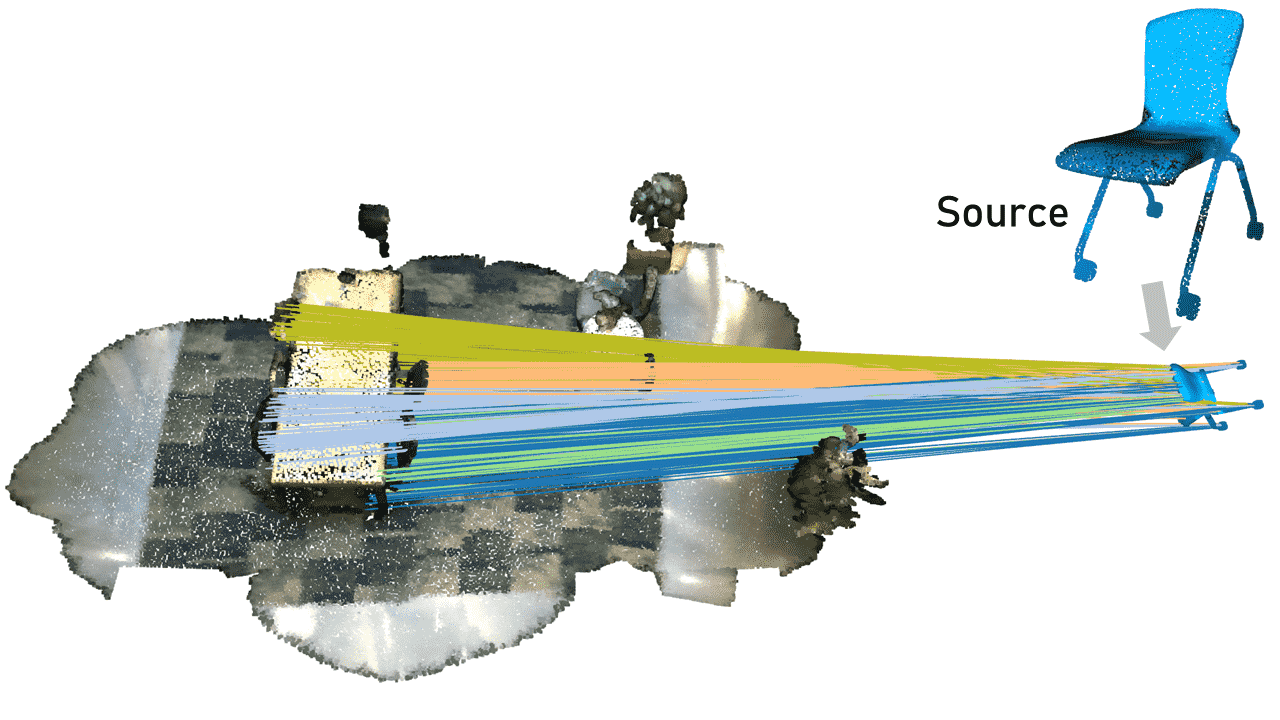}
        \caption{Our clustering result}
        \label{fig:scan2cad_cad-cluster-corrs22}
    \end{subfigure}\hfill
    \begin{subfigure}{0.3\textwidth}
      \centering
      \includegraphics[height=2.8cm]{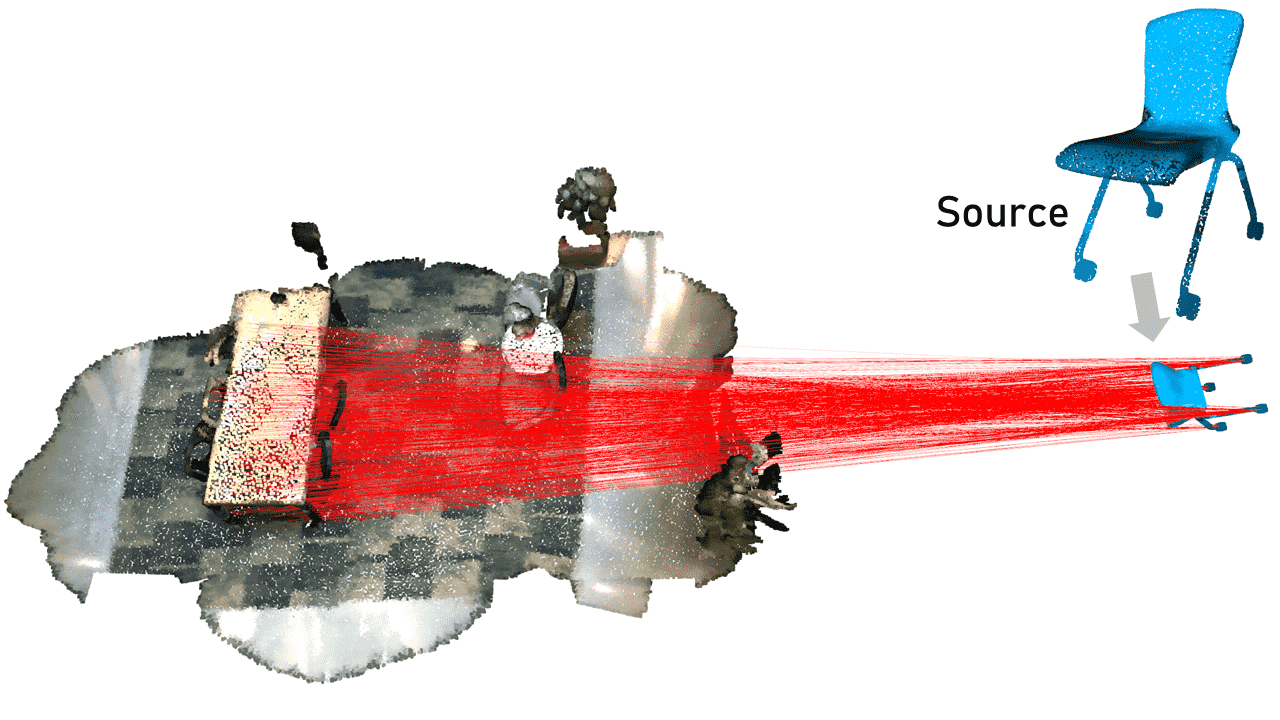}
        \caption{Our reject outliers}
        \label{fig:scan2cad_cad-reject-corrs22}
    \end{subfigure}

    \begin{subfigure}{0.23\textwidth}
      \centering
      \includegraphics[height=2.1cm]{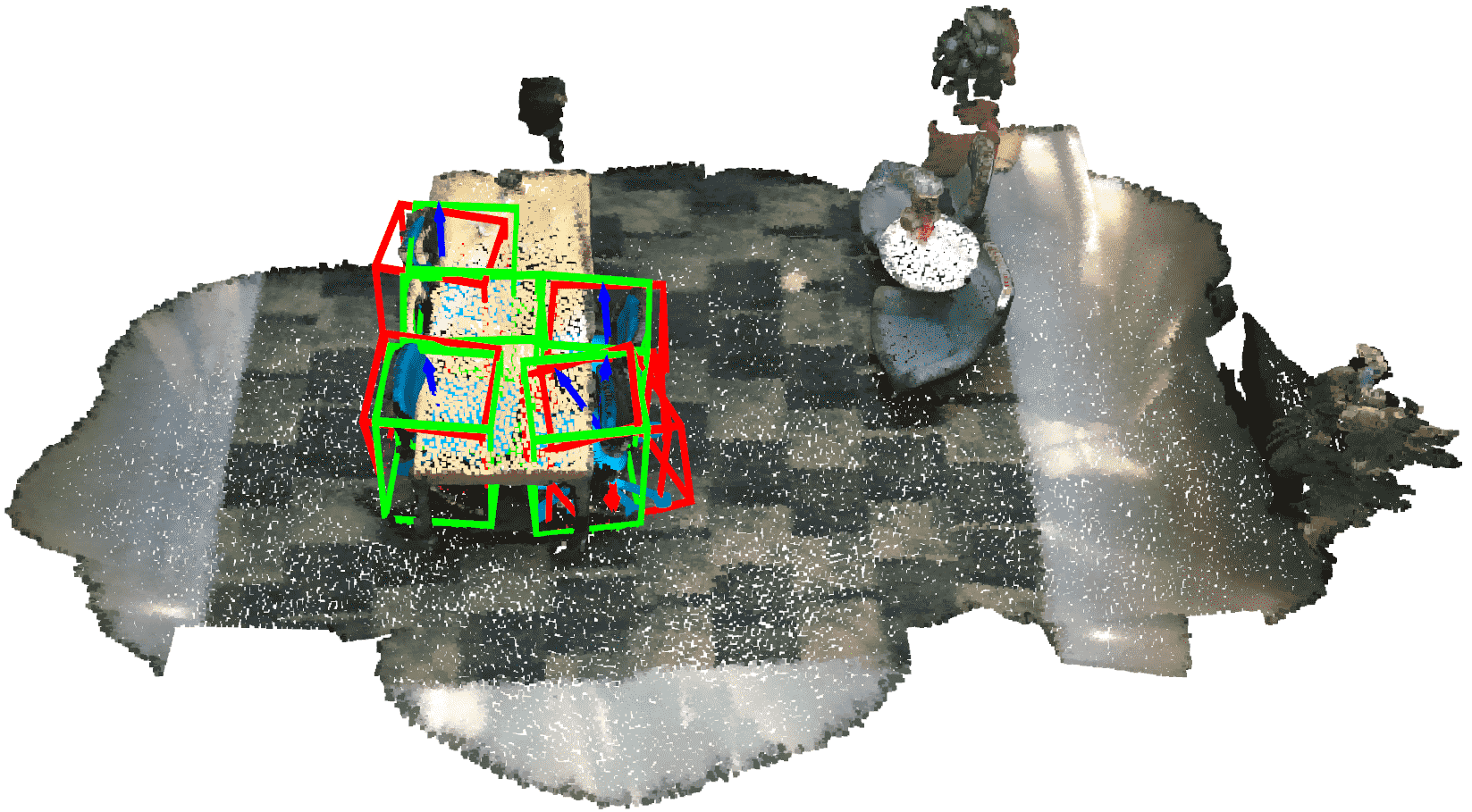}
        \caption{Ours}
        \label{fig:scan2cad_cad-result22}
    \end{subfigure}\hfill
    \begin{subfigure}{0.23\textwidth}
      \centering
      \includegraphics[height=2.1cm]{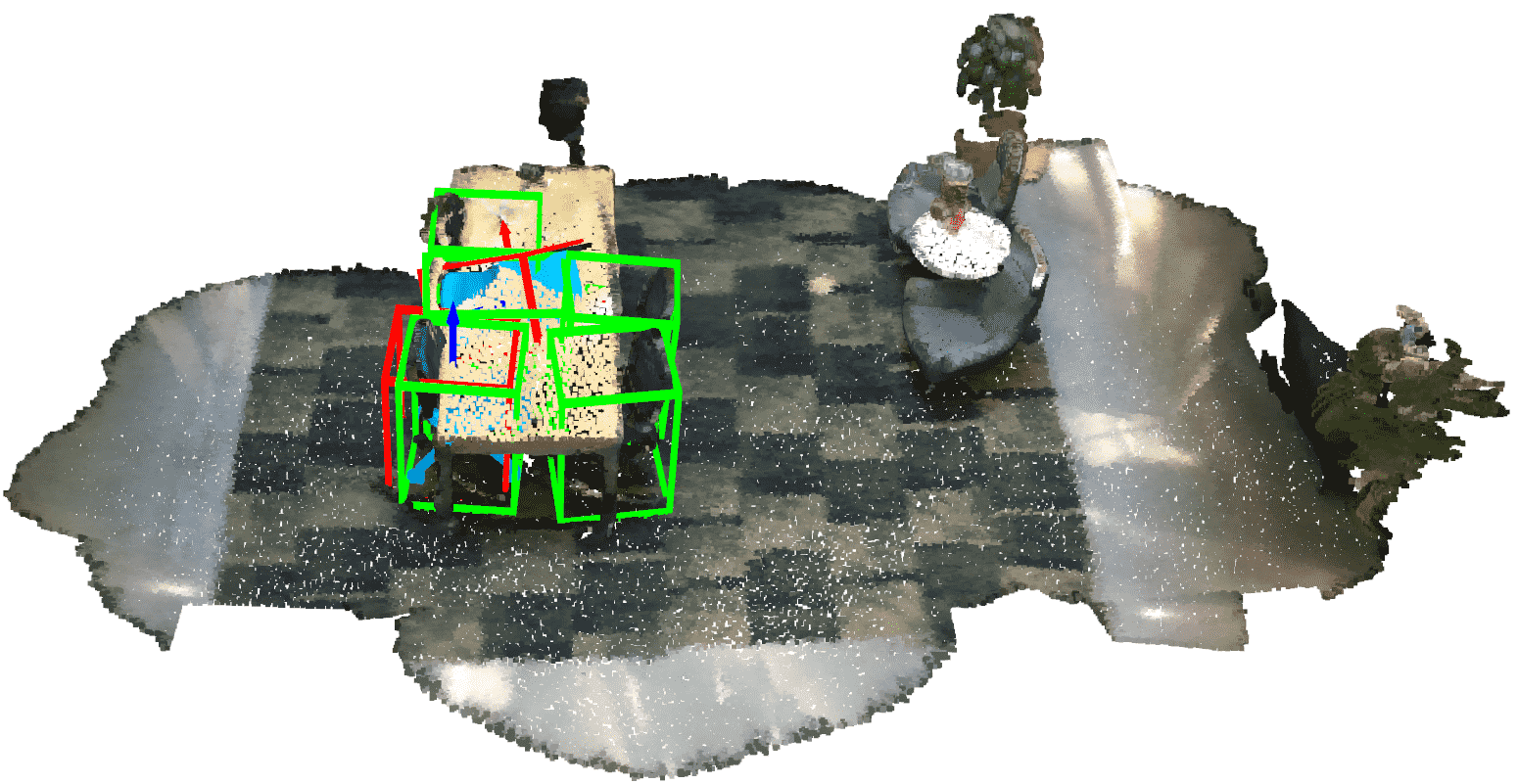}
        \caption{Progressive-X(2019) \cite{ProgressiveX}}
        \label{fig:scan2cad_cad-prox22}
    \end{subfigure}\hfill
    \begin{subfigure}{0.23\textwidth}
      \centering
      \includegraphics[height=2.1cm]{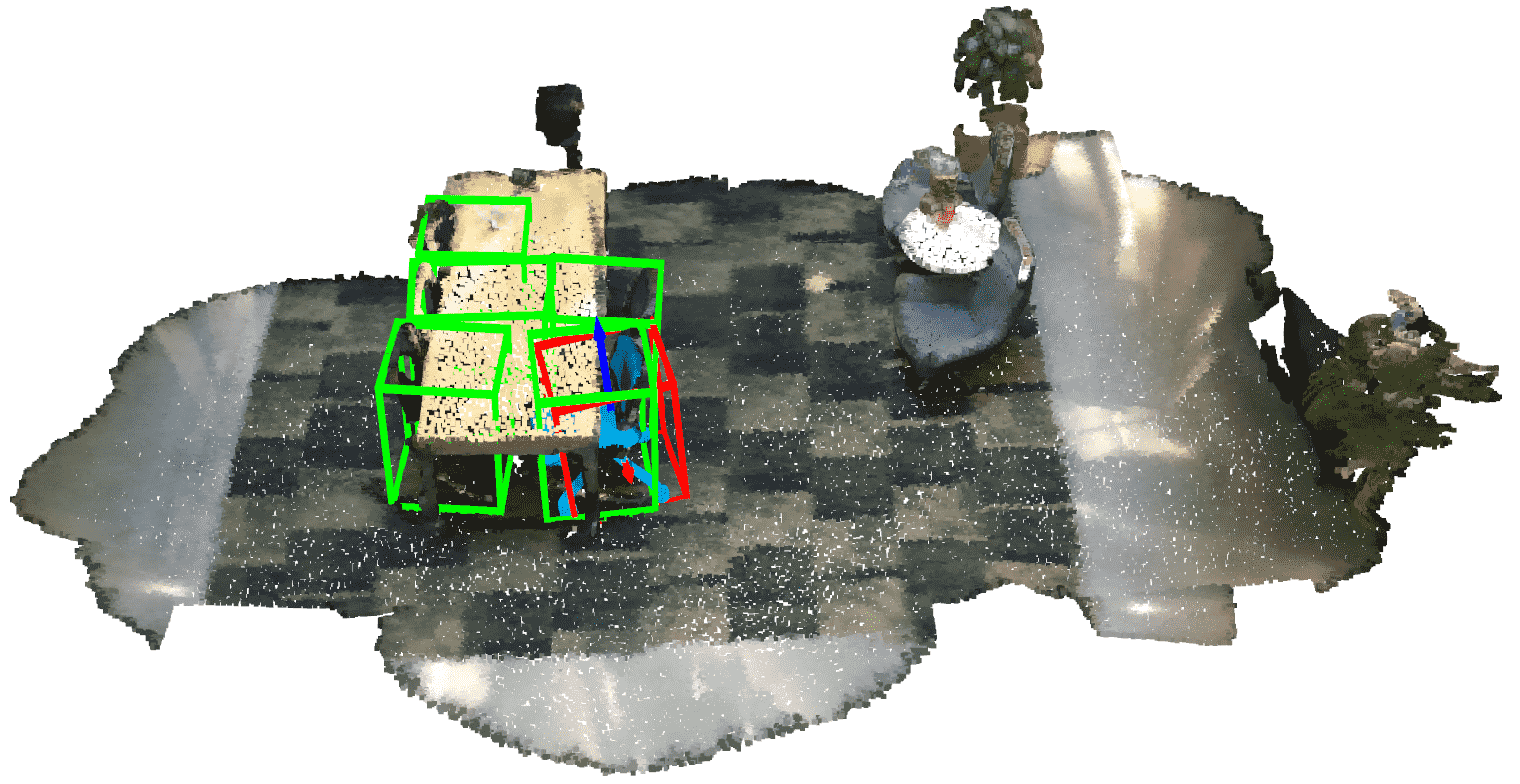}
        \caption{TEASER(2020)\cite{TEASER}}
        \label{fig:scan2cad_cad-teaser22}
    \end{subfigure}
    \begin{subfigure}{0.23\textwidth}
      \centering
      \includegraphics[height=2.1cm]{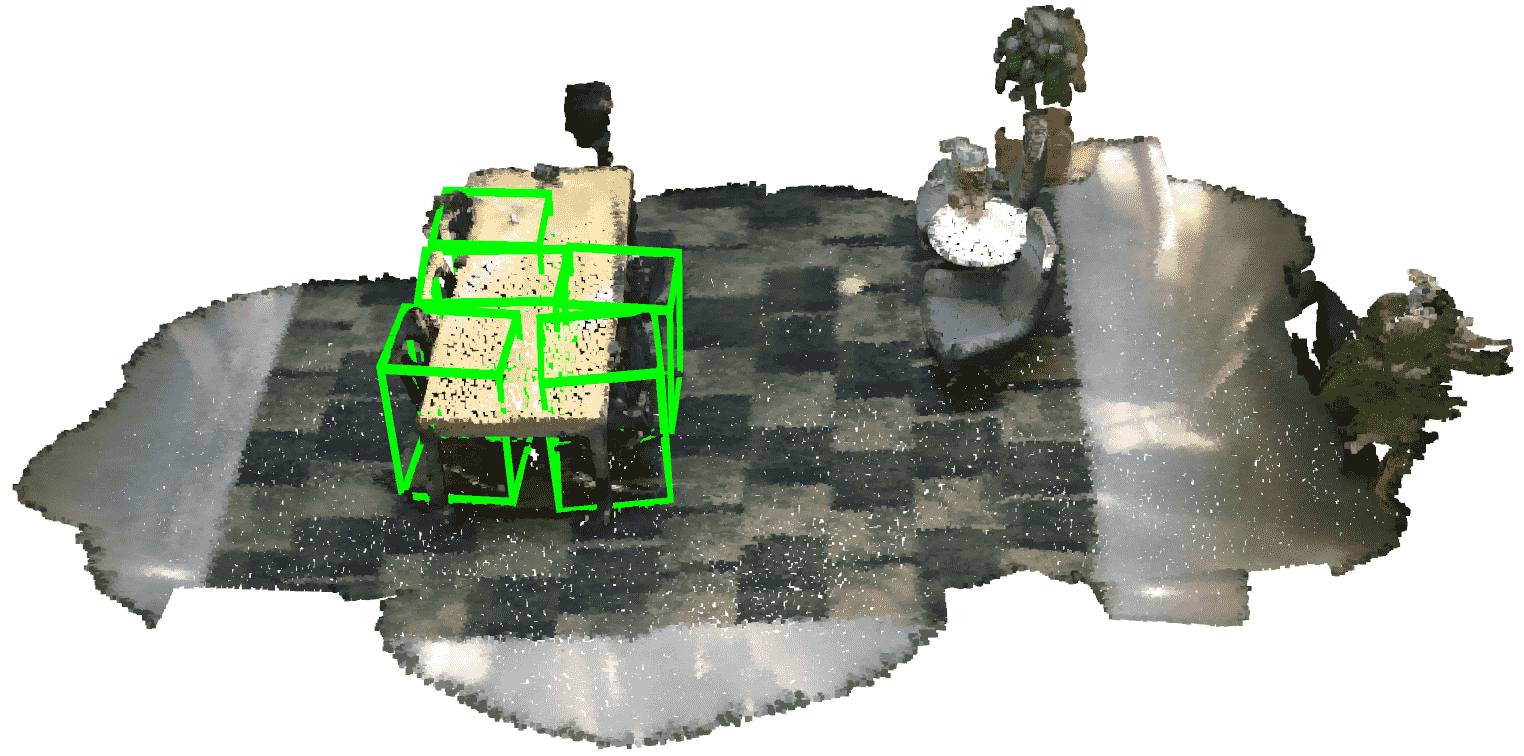}
        \caption{CONSAC(2020)\cite{CONSAC}}
        \label{fig:scan2cad_cad-consac22}
    \end{subfigure}\hfill
    \caption{\textbf{Scan2CAD results.}}
\label{fig:Scan2CAD-cadresult22}
  \end{figure*}

\begin{figure*}[ht]
  \centering
  \begin{subfigure}{0.3\textwidth}
      \centering
      \includegraphics[height=2.8cm]{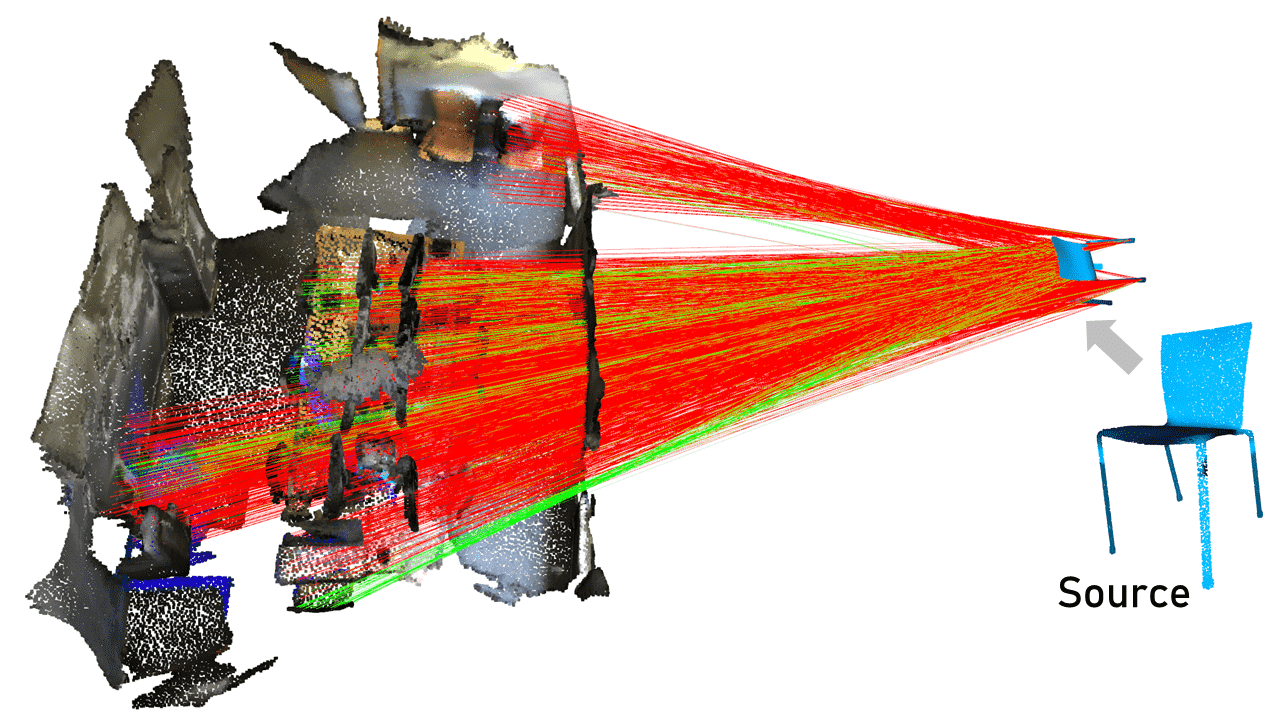}
        \caption{Input correspondences}
        \label{fig:scan2cad_cad-input-corrs24}
    \end{subfigure}\hfill
    \begin{subfigure}{0.3\textwidth}
      \centering
      \includegraphics[height=2.8cm]{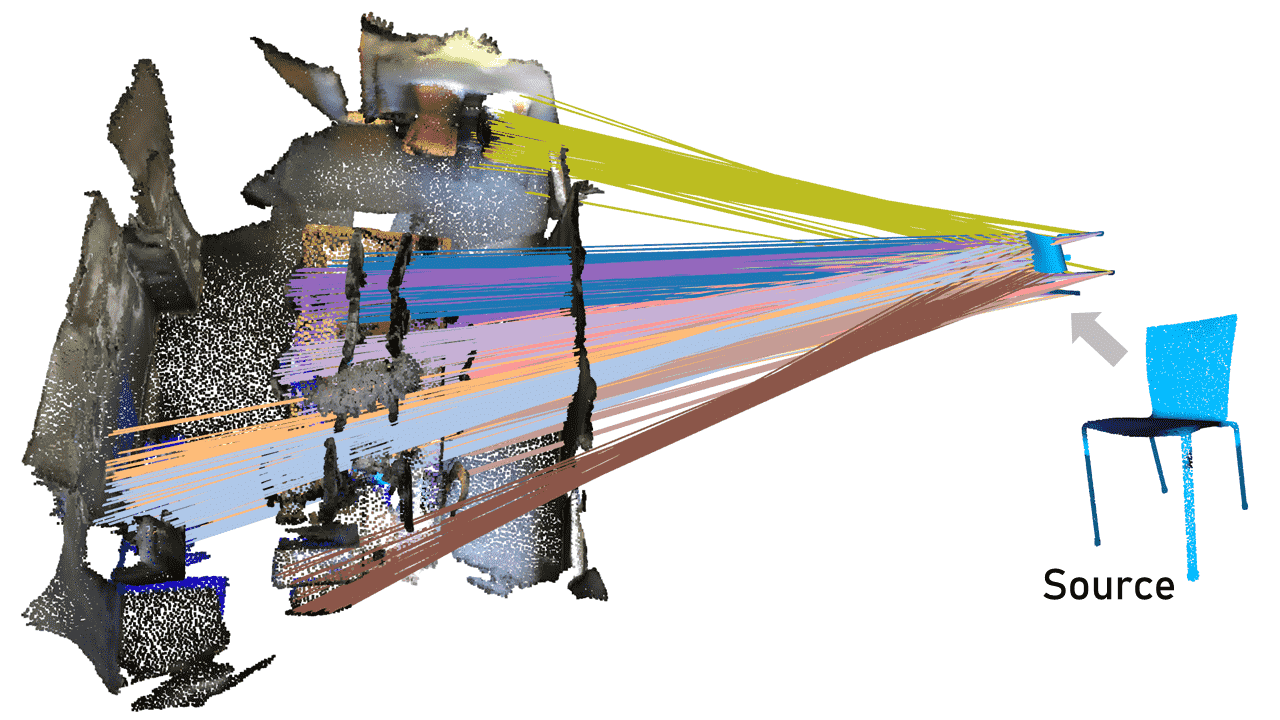}
        \caption{Our clustering result}
        \label{fig:scan2cad_cad-cluster-corrs24}
    \end{subfigure}\hfill
    \begin{subfigure}{0.3\textwidth}
      \centering
      \includegraphics[height=2.8cm]{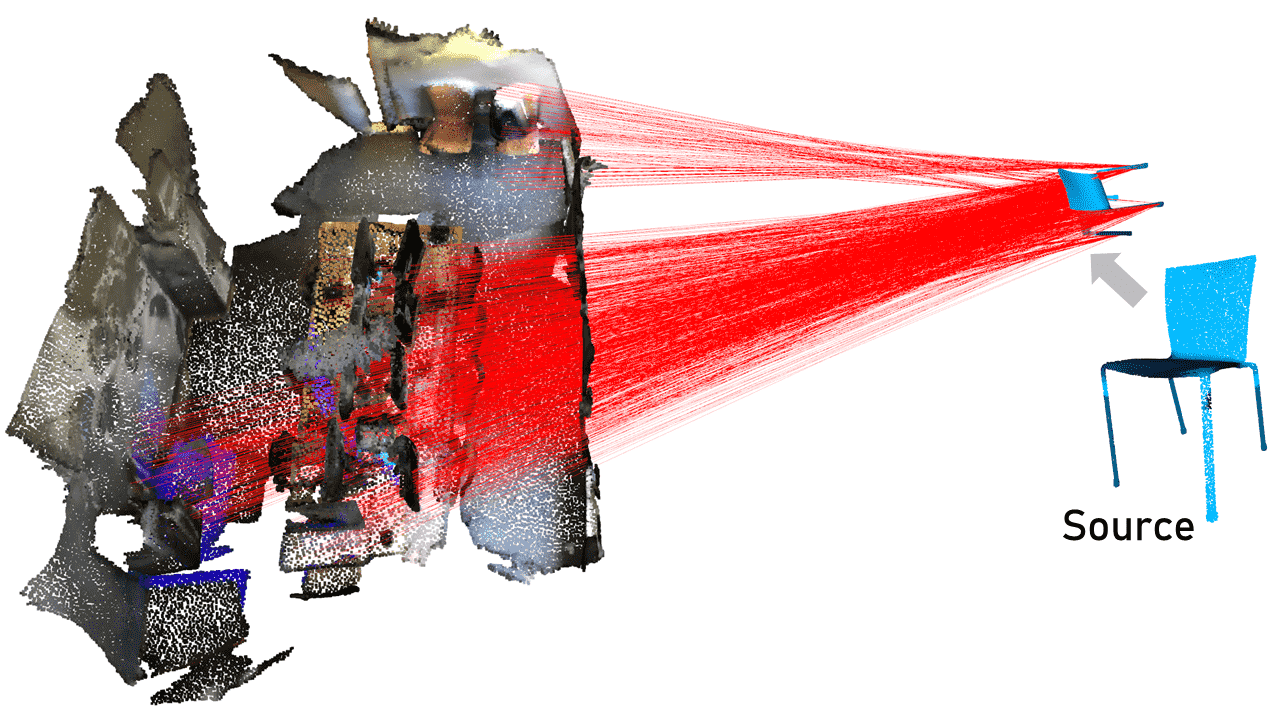}
        \caption{Our reject outliers}
        \label{fig:scan2cad_cad-reject-corrs24}
    \end{subfigure}

    \begin{subfigure}{0.23\textwidth}
      \centering
      \includegraphics[height=2.8cm]{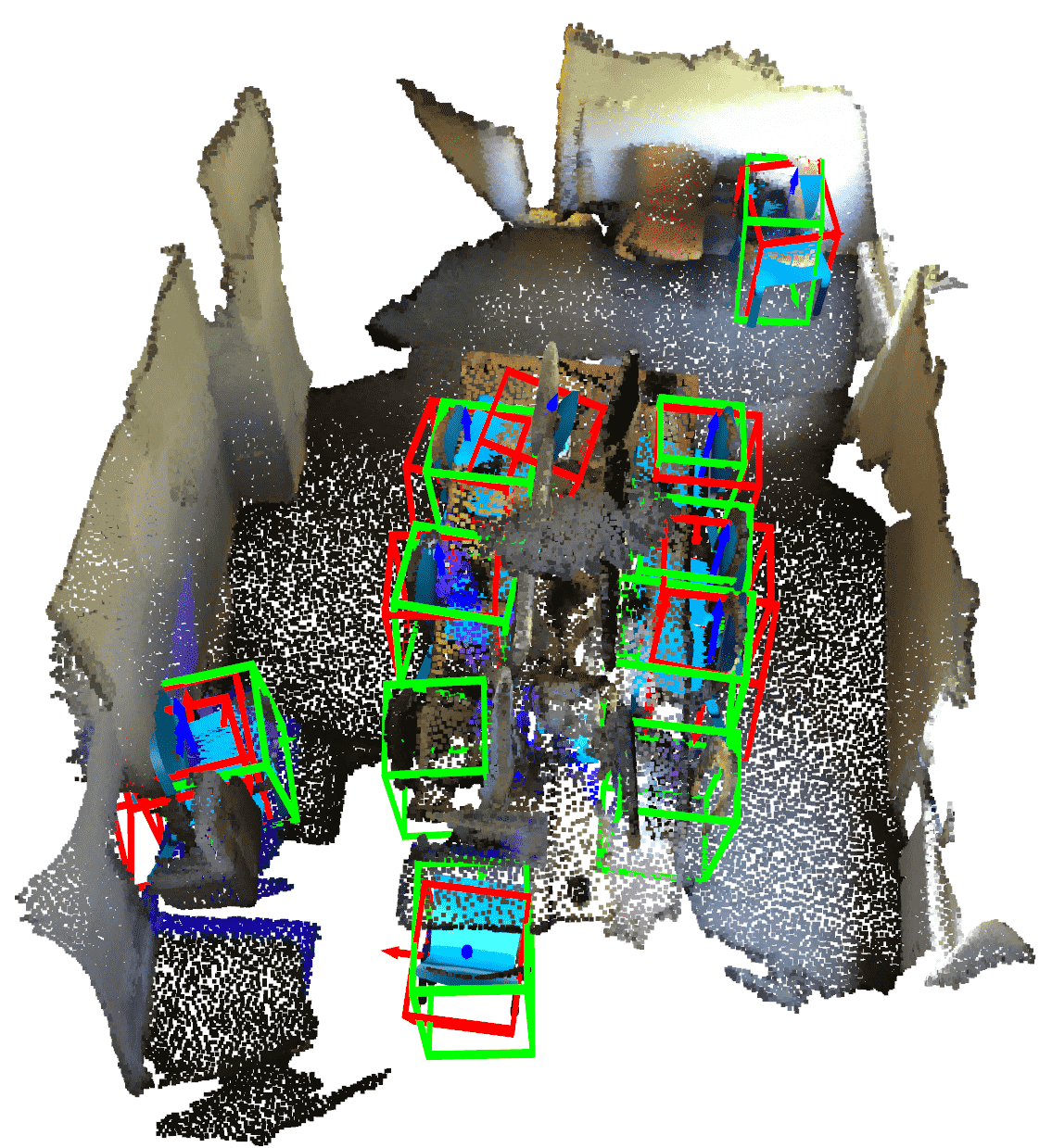}
        \caption{Ours}
        \label{fig:scan2cad_cad-result24}
    \end{subfigure}\hfill
    \begin{subfigure}{0.23\textwidth}
      \centering
      \includegraphics[height=2.8cm]{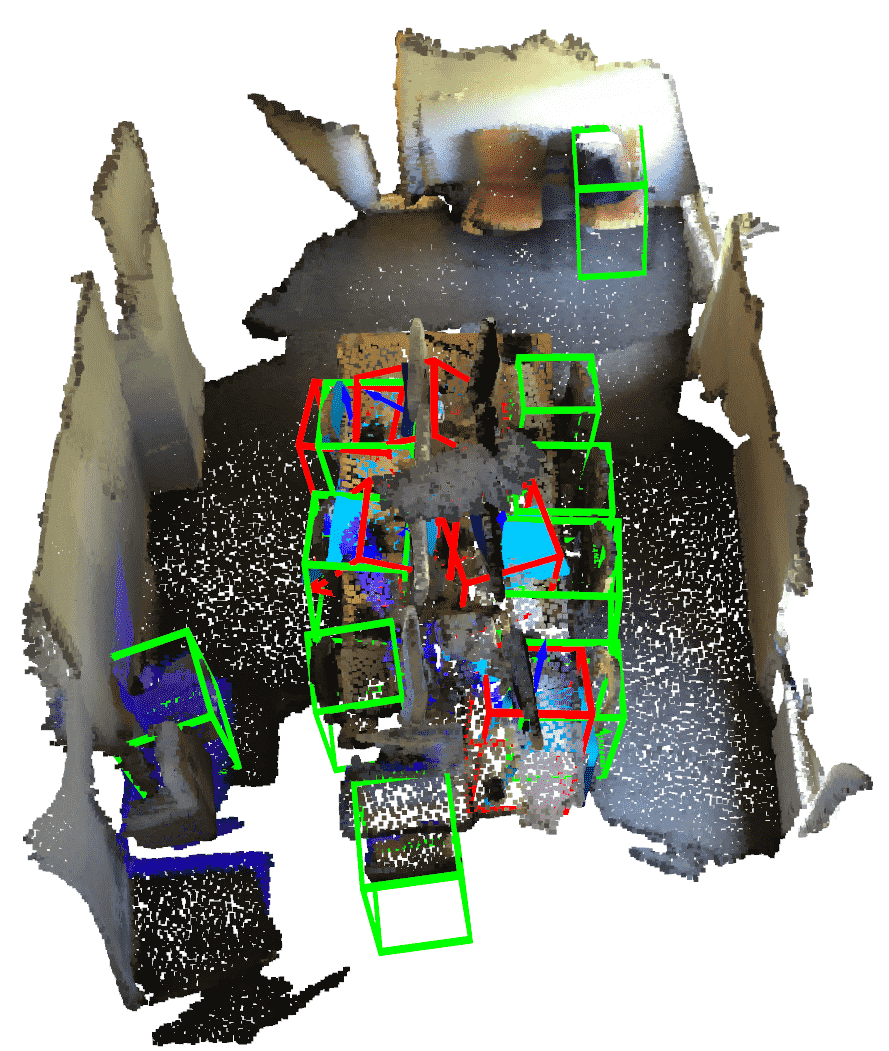}
        \caption{Progressive-X(2019) \cite{ProgressiveX}}
        \label{fig:scan2cad_cad-prox24}
    \end{subfigure}\hfill
    \begin{subfigure}{0.23\textwidth}
      \centering
      \includegraphics[height=2.8cm]{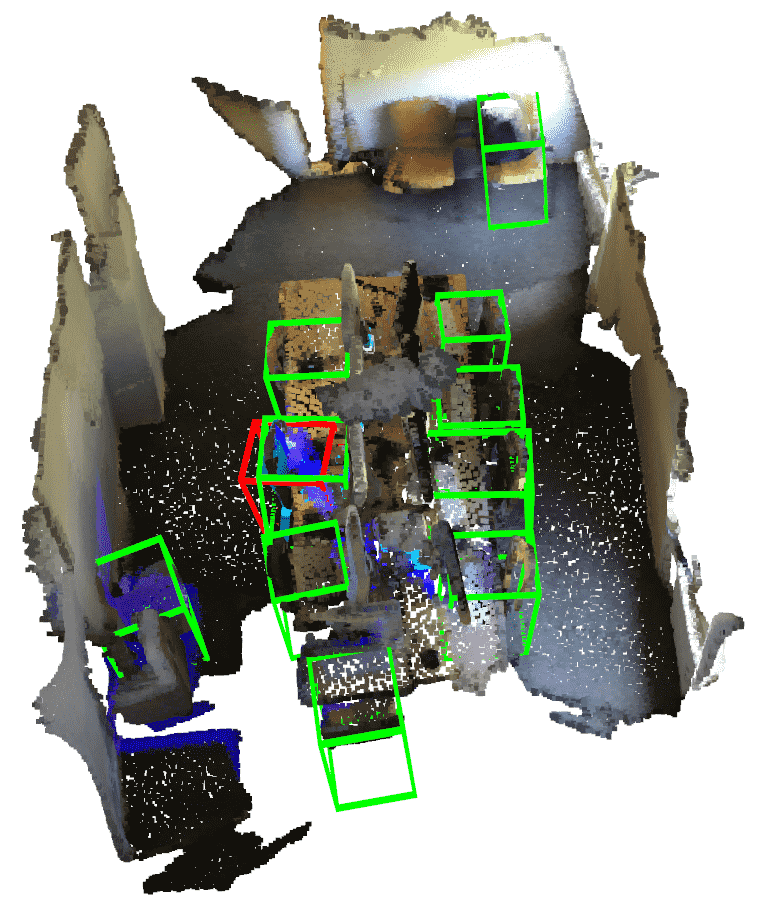}
        \caption{TEASER(2020)\cite{TEASER}}
        \label{fig:scan2cad_cad-teaser24}
    \end{subfigure}
    \begin{subfigure}{0.23\textwidth}
      \centering
      \includegraphics[height=2.8cm]{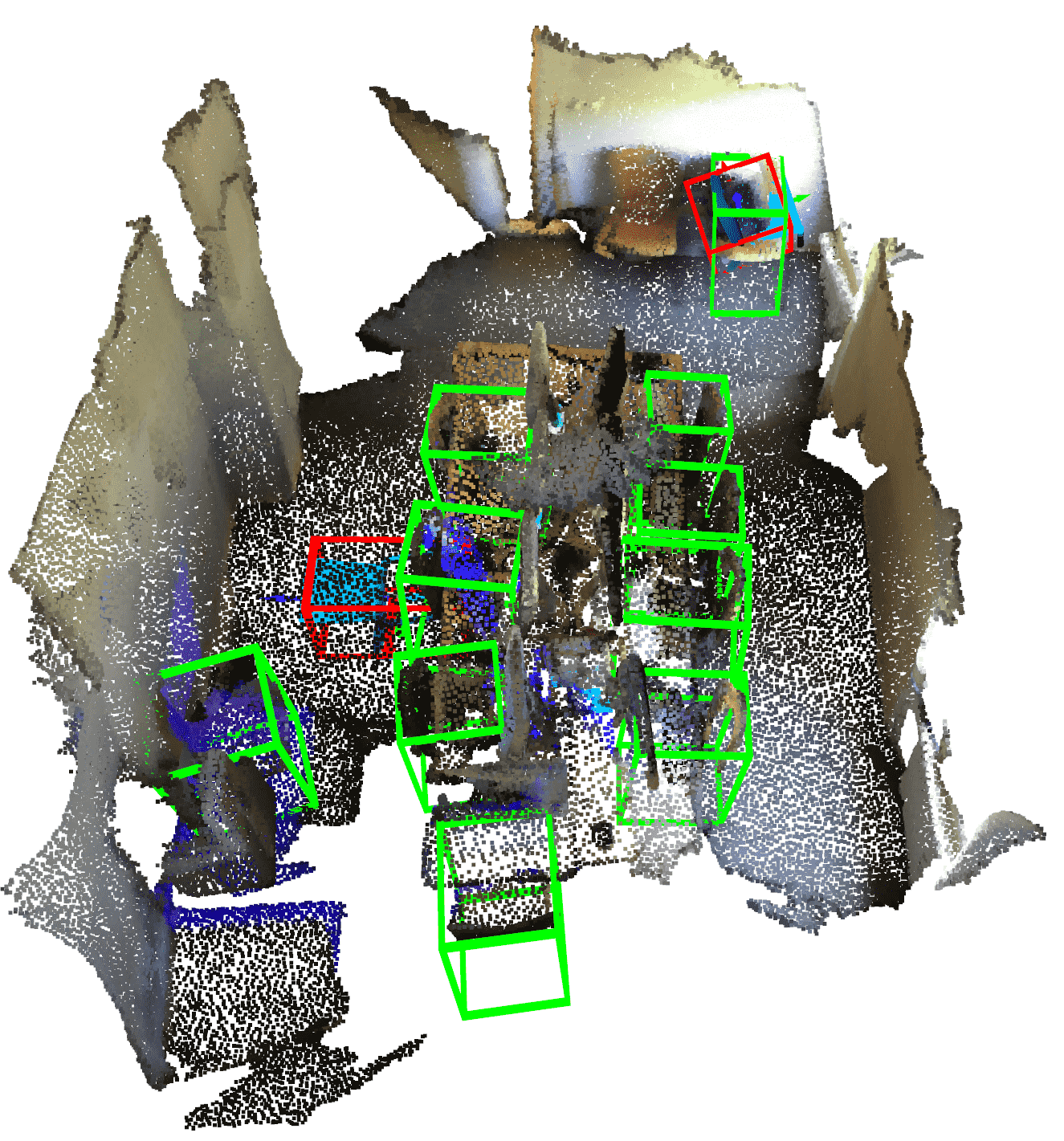}
        \caption{CONSAC(2020)\cite{CONSAC}}
        \label{fig:scan2cad_cad-consac24}
    \end{subfigure}\hfill
    \caption{\textbf{Scan2CAD results.}}
\label{fig:Scan2CAD-cadresult24}
  \end{figure*}

\begin{figure*}[ht]
  \centering
  \begin{subfigure}{0.3\textwidth}
      \centering
      \includegraphics[height=2.8cm]{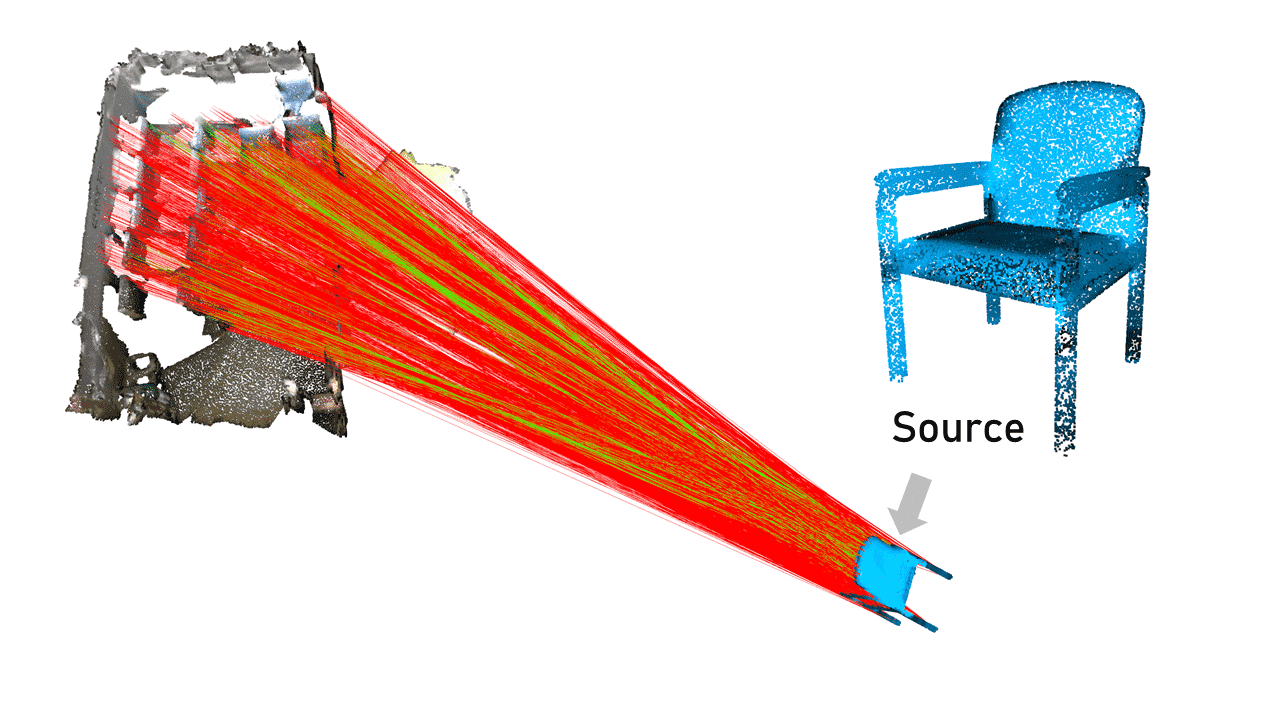}
        \caption{Input correspondences}
        \label{fig:scan2cad_cad-input-corrs4}
    \end{subfigure}\hfill
    \begin{subfigure}{0.3\textwidth}
      \centering
      \includegraphics[height=2.8cm]{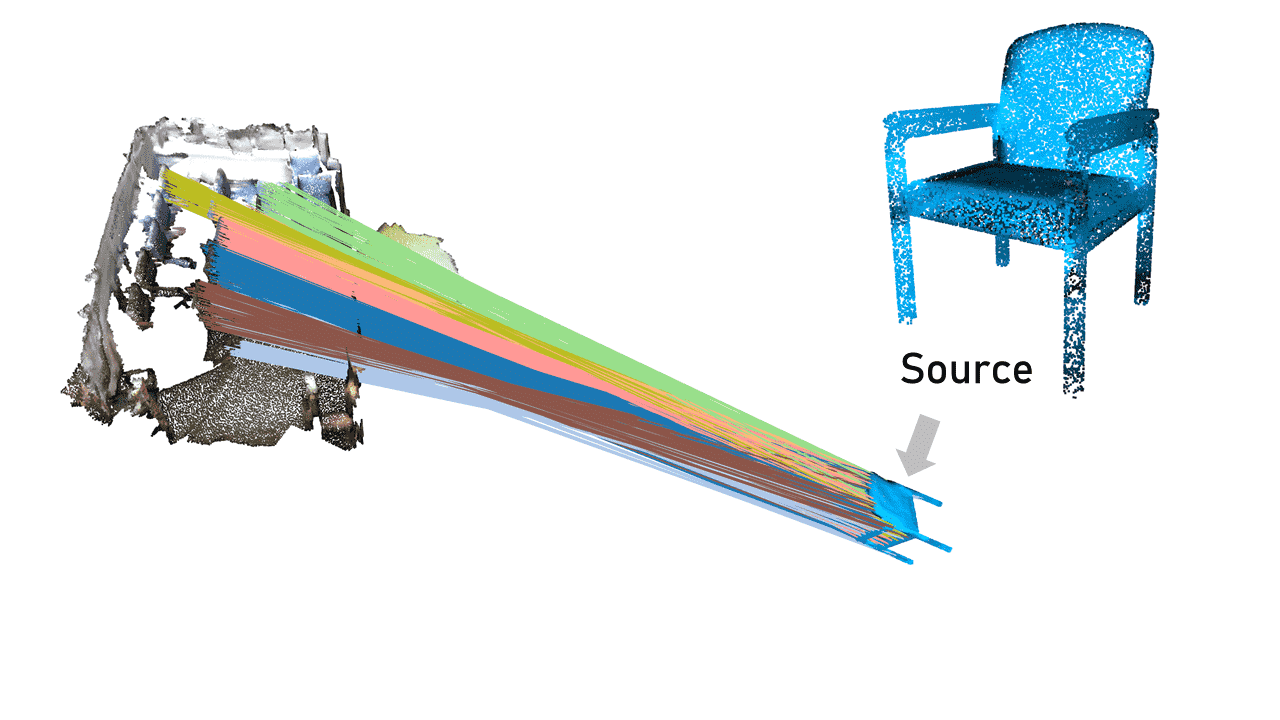}
        \caption{Our clustering result}
        \label{fig:scan2cad_cad-cluster-corrs4}
    \end{subfigure}\hfill
    \begin{subfigure}{0.3\textwidth}
      \centering
      \includegraphics[height=2.8cm]{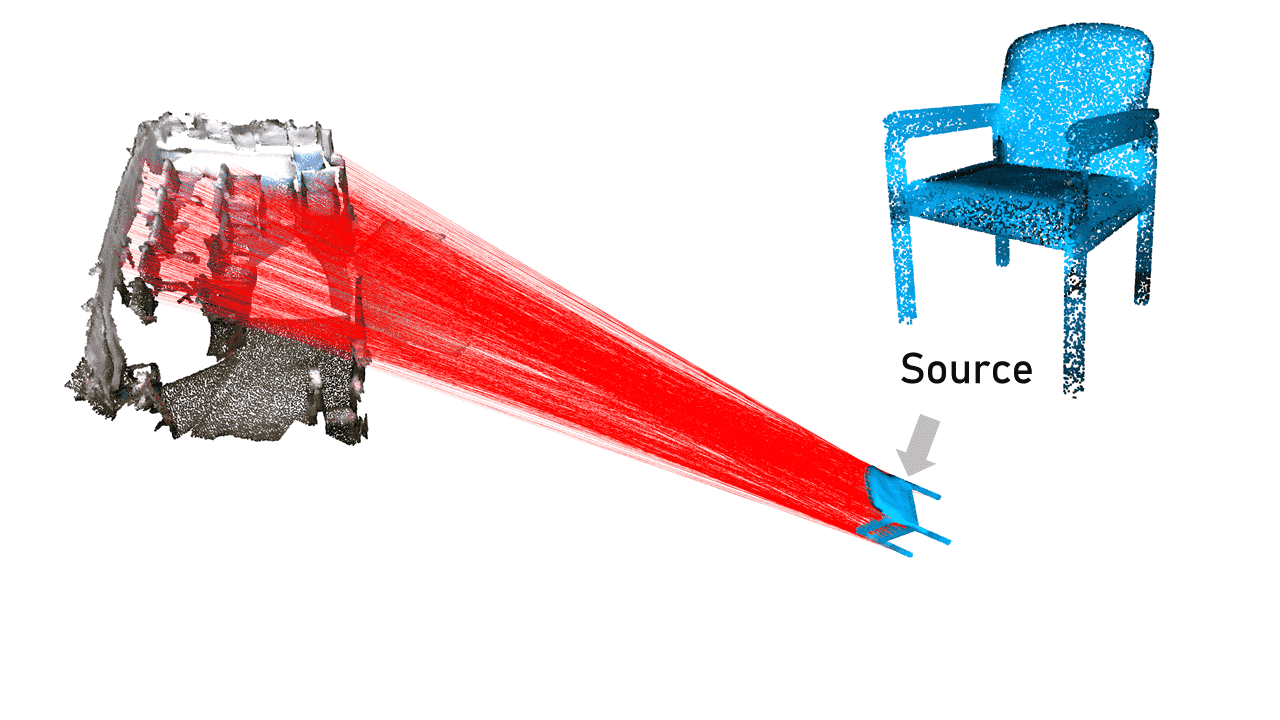}
        \caption{Our reject outliers}
        \label{fig:scan2cad_cad-reject-corrs4}
    \end{subfigure}

    \begin{subfigure}{0.23\textwidth}
      \centering
      \includegraphics[height=2.8cm]{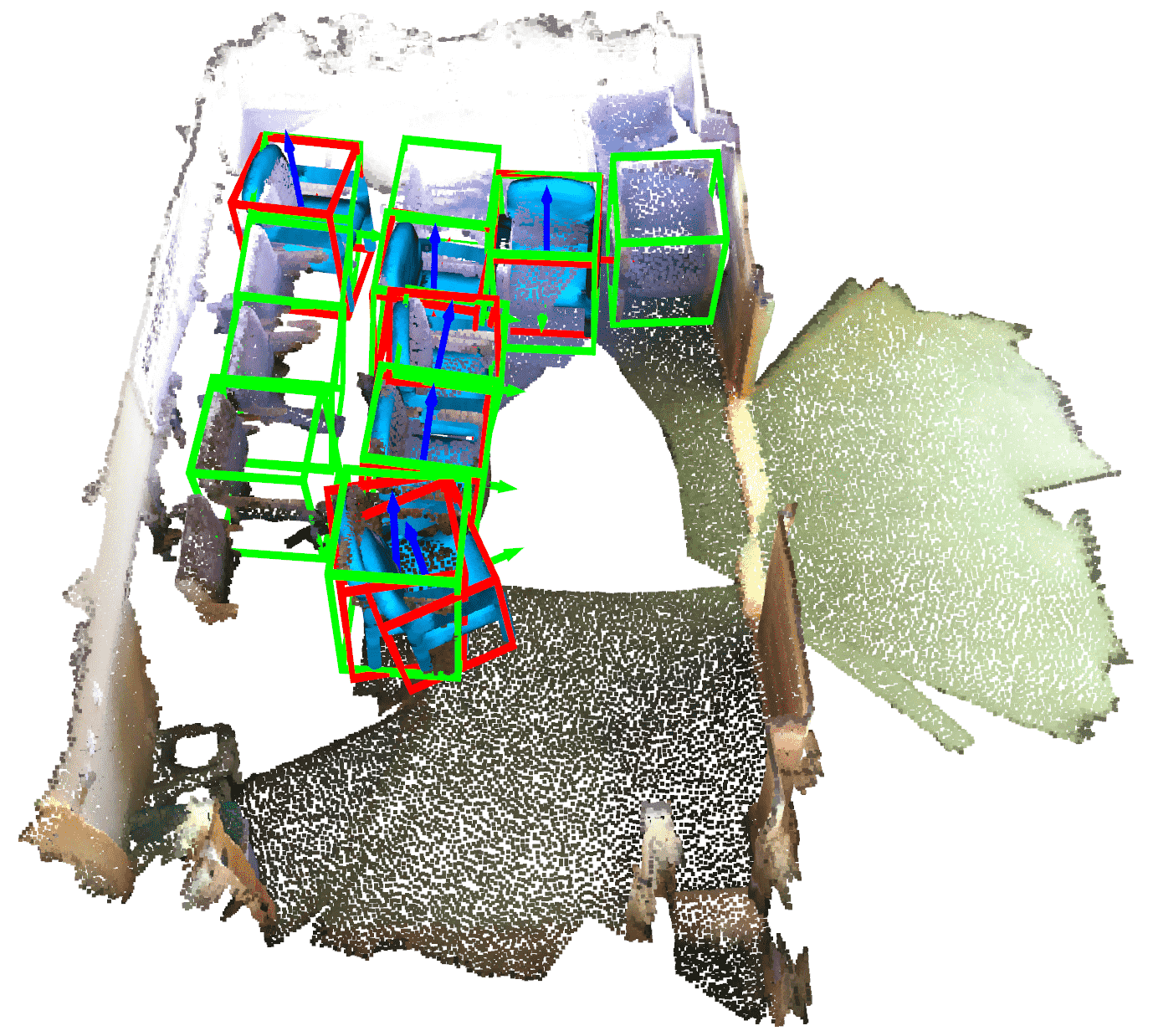}
        \caption{Ours}
        \label{fig:scan2cad_cad-result4}
    \end{subfigure}\hfill
    \begin{subfigure}{0.23\textwidth}
      \centering
      \includegraphics[height=2.8cm]{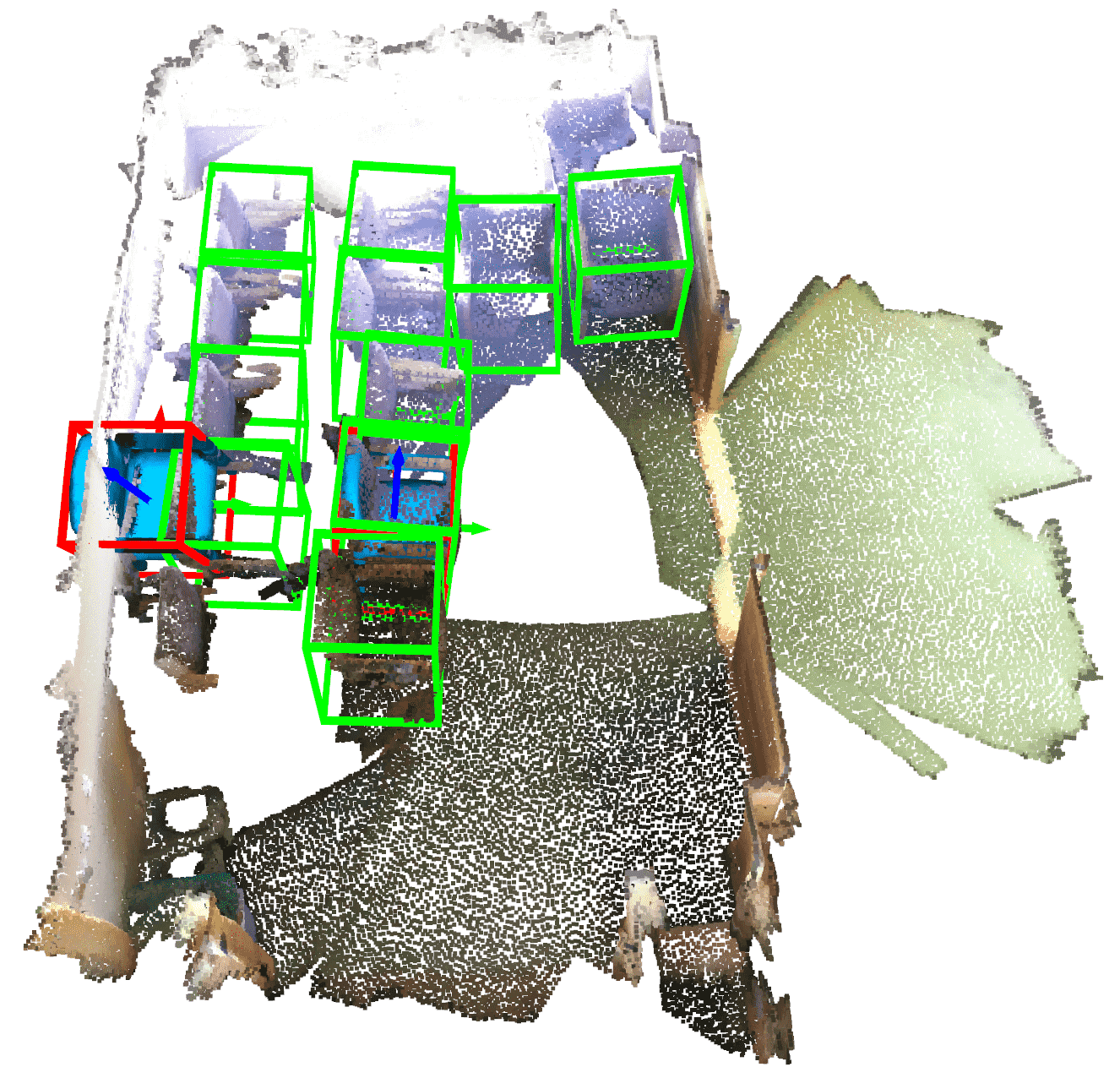}
        \caption{Progressive-X(2019) \cite{ProgressiveX}}
        \label{fig:scan2cad_cad-prox4}
    \end{subfigure}\hfill
    \begin{subfigure}{0.23\textwidth}
      \centering
      \includegraphics[height=2.8cm]{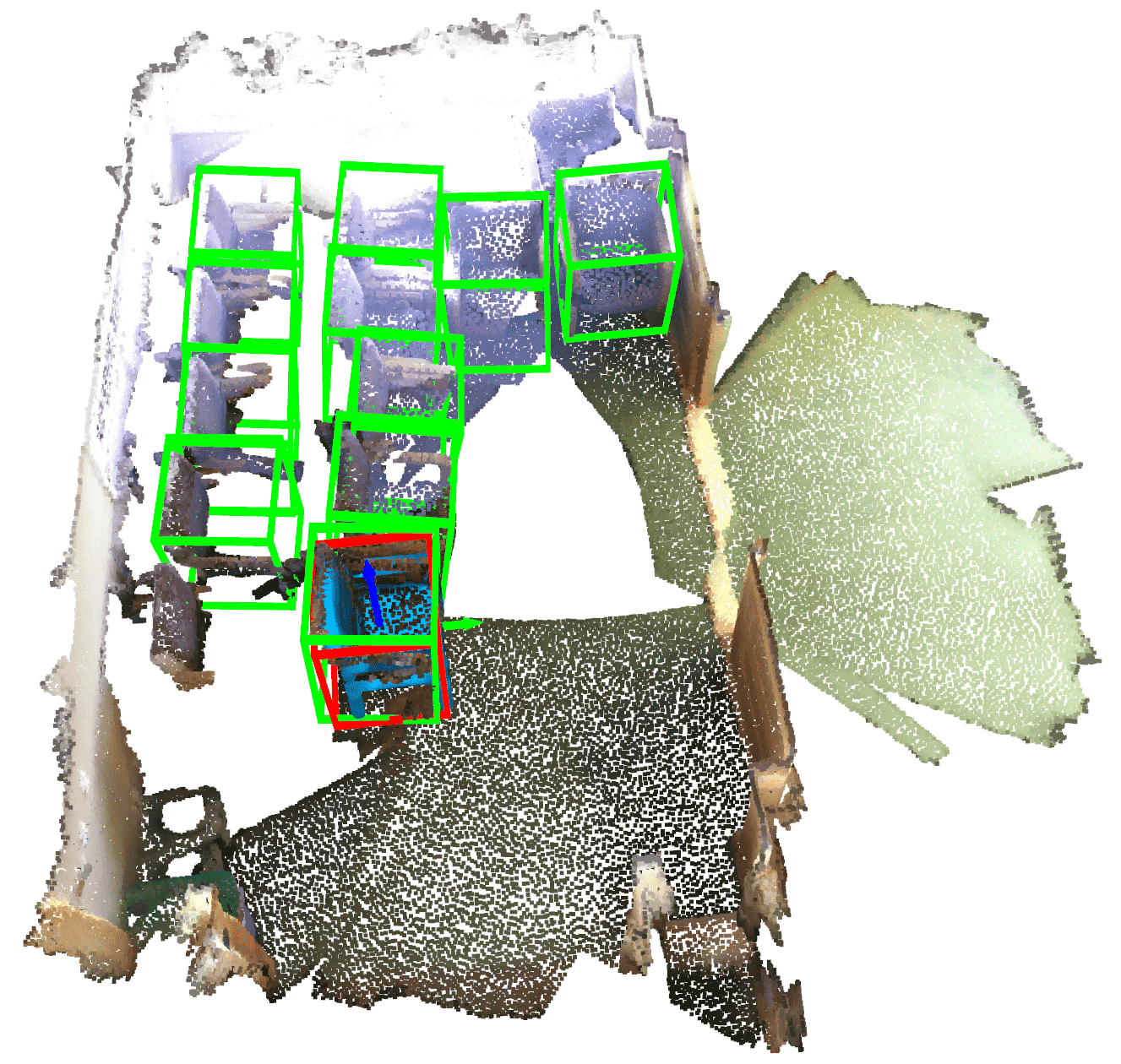}
        \caption{TEASER(2020)\cite{TEASER}}
        \label{fig:scan2cad_cad-teaser4}
    \end{subfigure}
    \begin{subfigure}{0.23\textwidth}
      \centering
      \includegraphics[height=2.8cm]{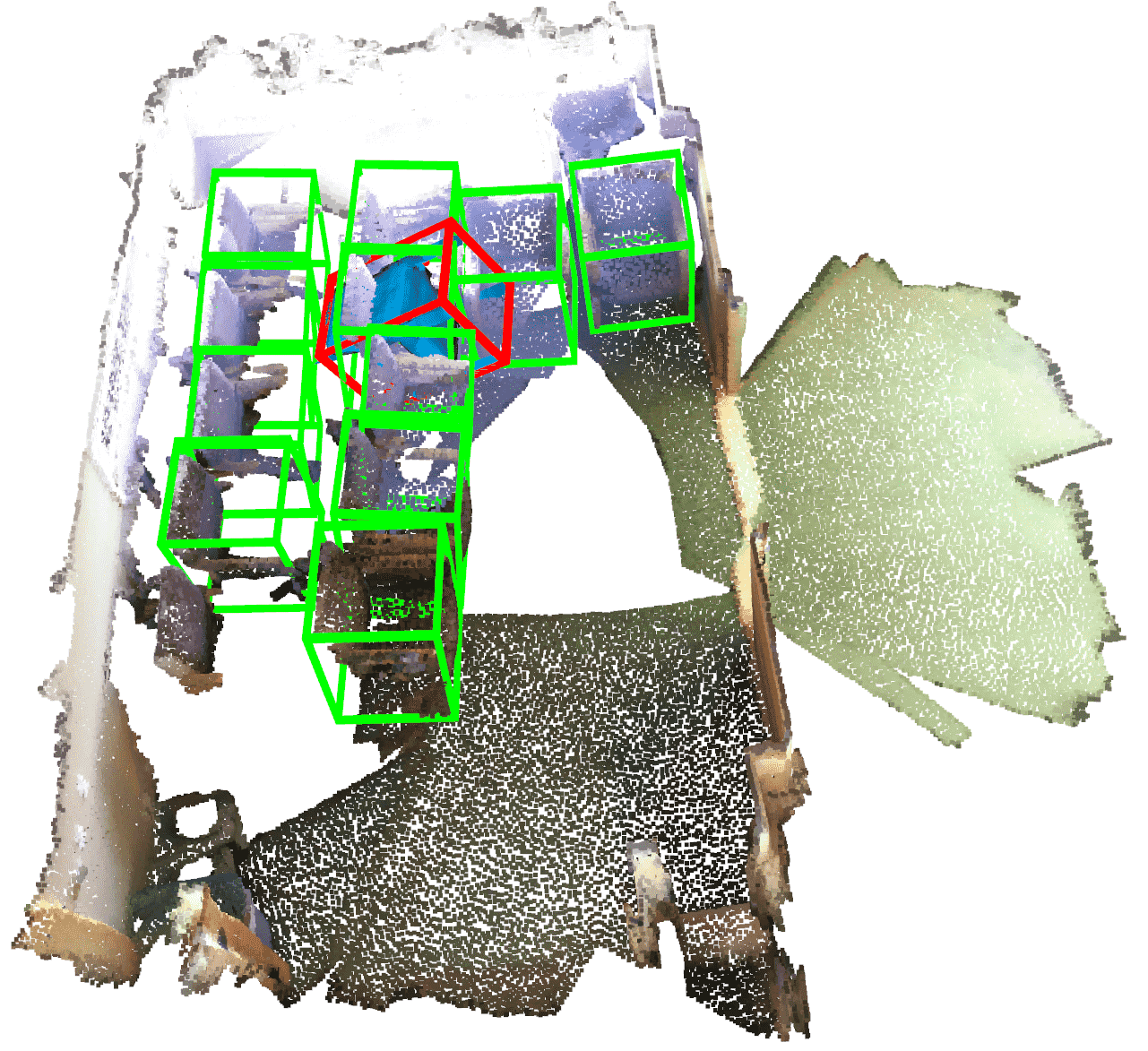}
        \caption{CONSAC(2020)\cite{CONSAC}}
        \label{fig:scan2cad_cad-consac4}
    \end{subfigure}\hfill
    \caption{\textbf{Scan2CAD results.}}
\label{fig:Scan2CAD-cadresult4}
  \end{figure*}

\begin{figure*}[ht]
  \centering
  \begin{subfigure}{0.3\textwidth}
      \centering
      \includegraphics[height=2.8cm]{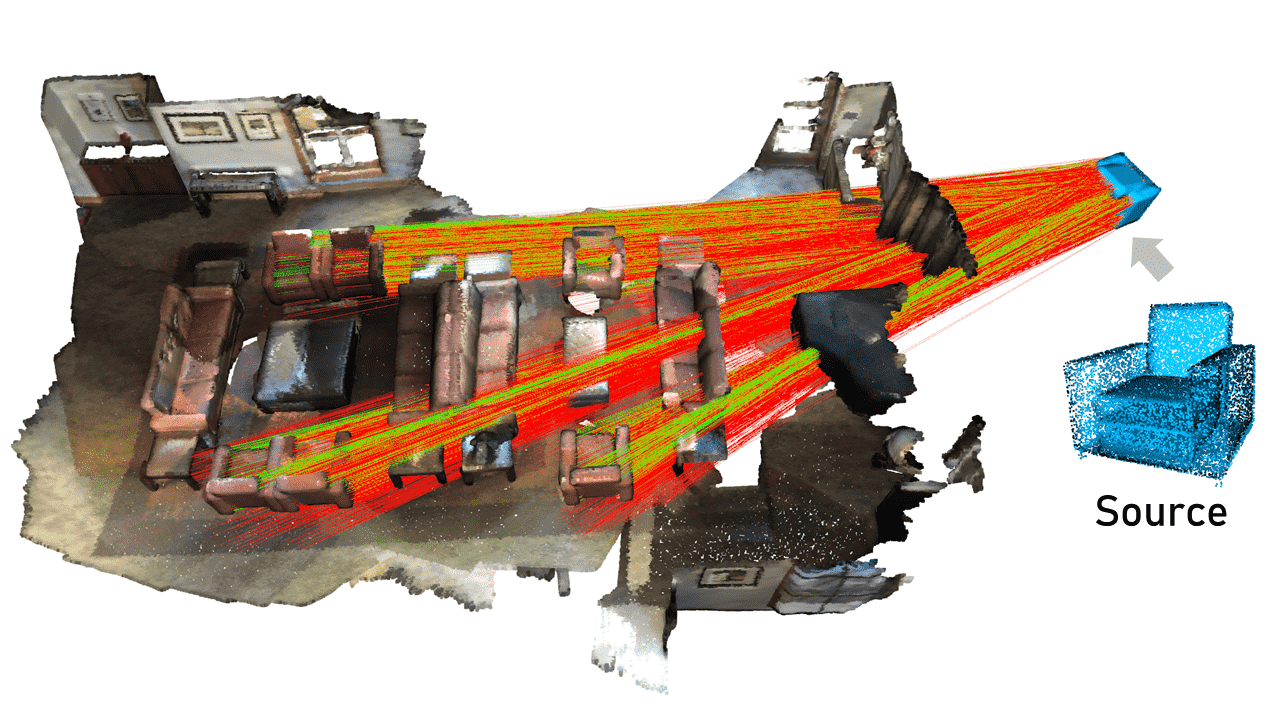}
        \caption{Input correspondences}
        \label{fig:scan2cad_cad-input-corrs27}
    \end{subfigure}\hfill
    \begin{subfigure}{0.3\textwidth}
      \centering
      \includegraphics[height=2.8cm]{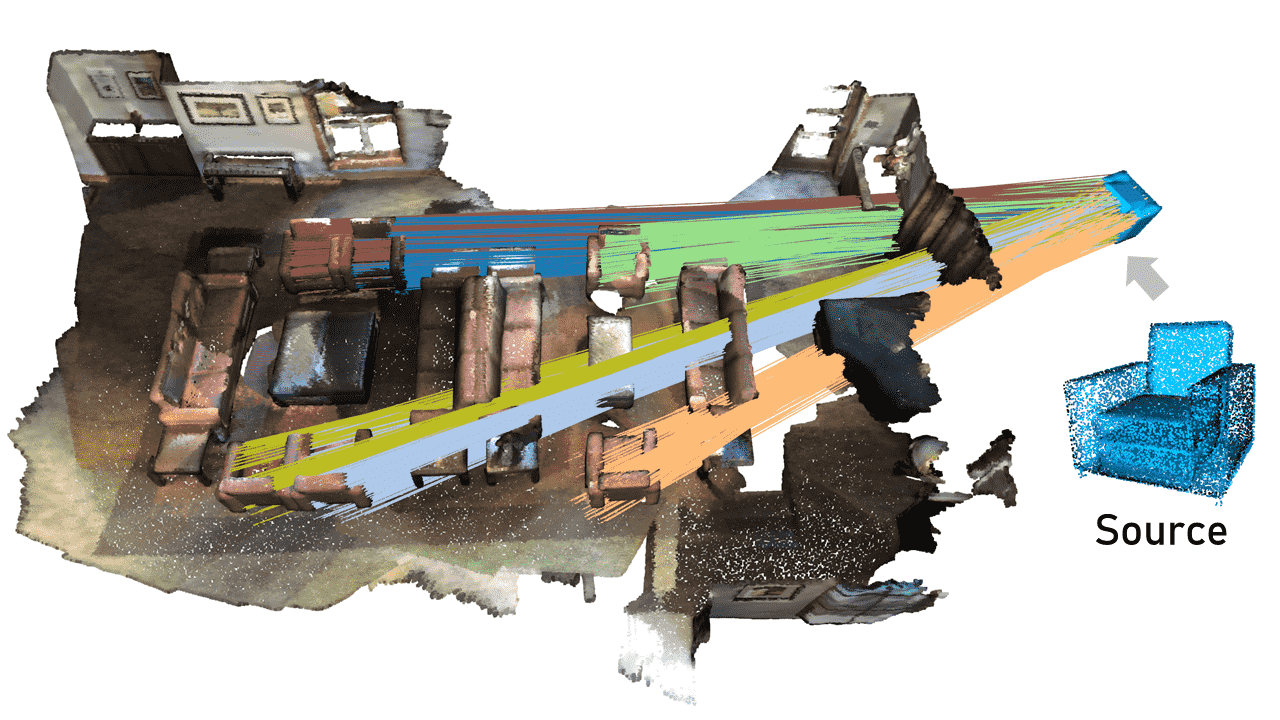}
        \caption{Our clustering result}
        \label{fig:scan2cad_cad-cluster-corrs27}
    \end{subfigure}\hfill
    \begin{subfigure}{0.3\textwidth}
      \centering
      \includegraphics[height=2.8cm]{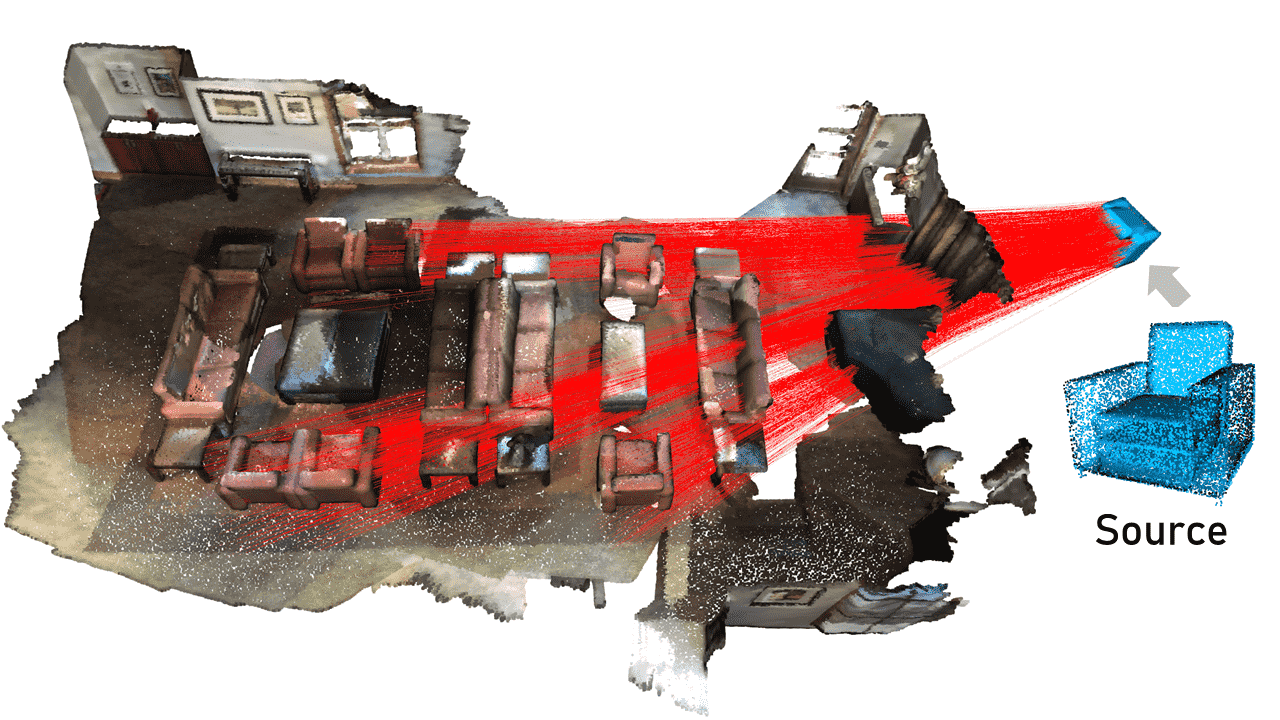}
        \caption{Our reject outliers}
        \label{fig:scan2cad_cad-reject-corrs27}
    \end{subfigure}

    \begin{subfigure}{0.23\textwidth}
      \centering
      \includegraphics[height=2.8cm]{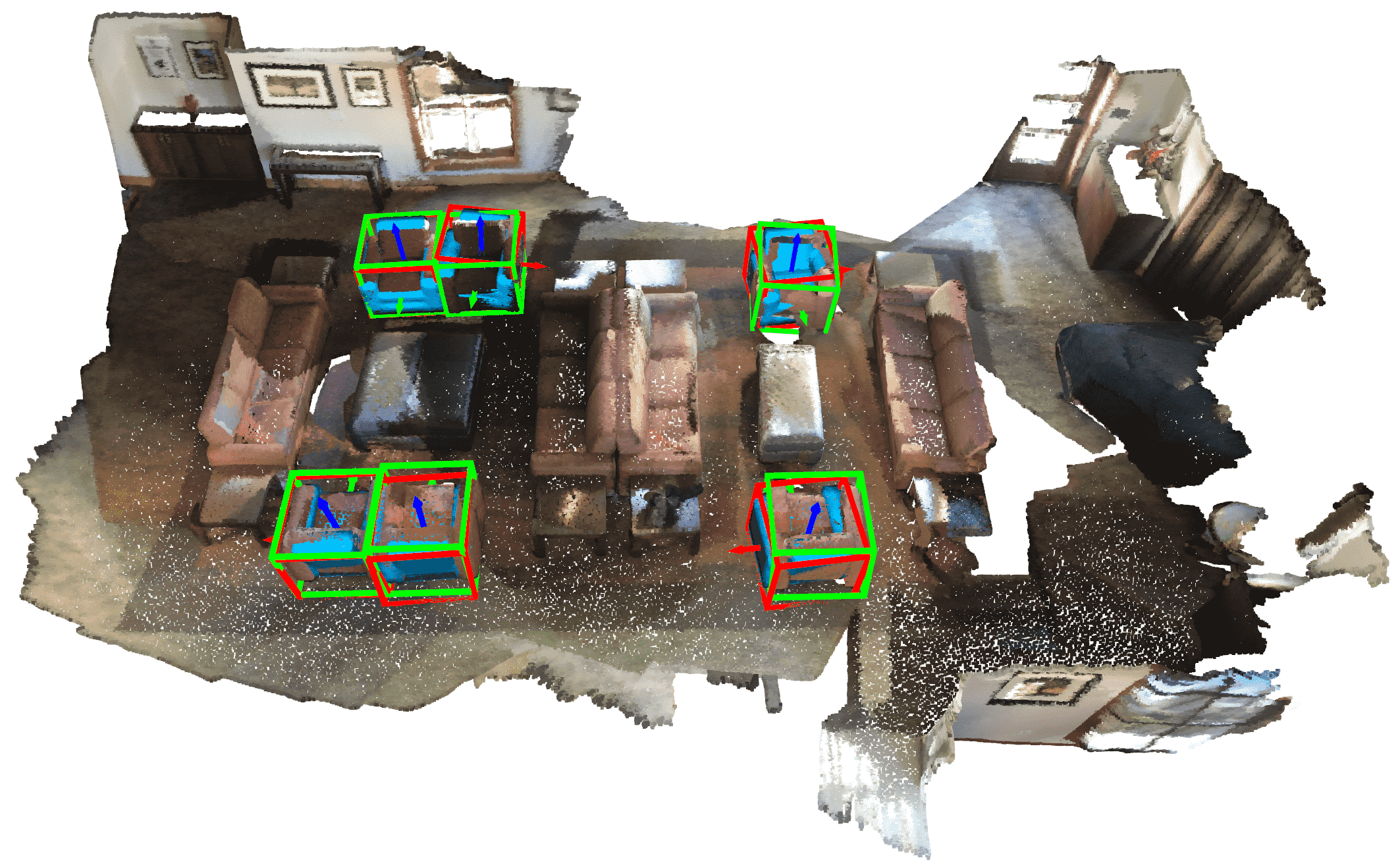}
        \caption{Ours}
        \label{fig:scan2cad_cad-result27}
    \end{subfigure}\hfill
    \begin{subfigure}{0.23\textwidth}
      \centering
      \includegraphics[height=2.8cm]{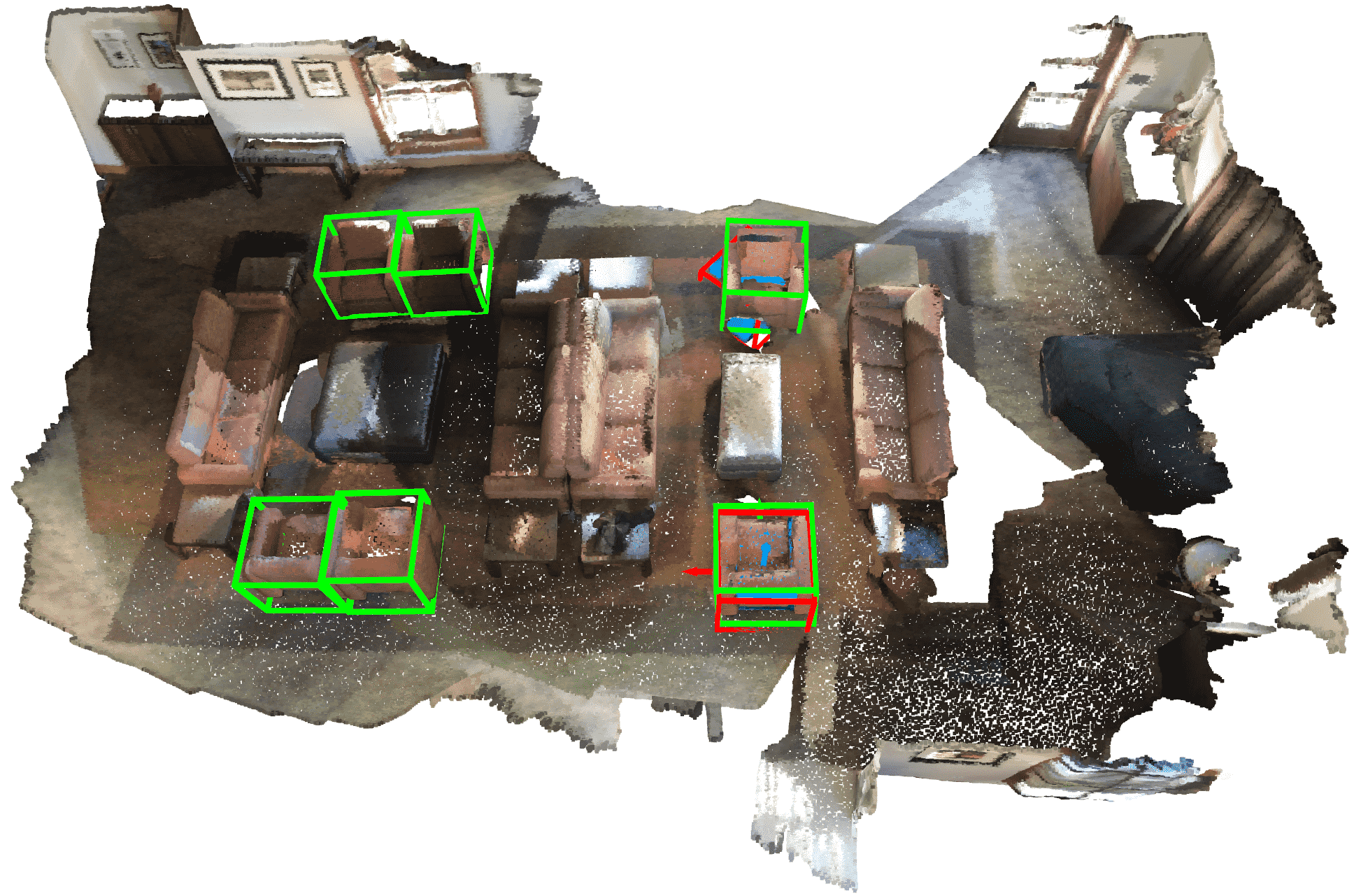}
        \caption{Progressive-X(2019) \cite{ProgressiveX}}
        \label{fig:scan2cad_cad-prox27}
    \end{subfigure}\hfill
    \begin{subfigure}{0.23\textwidth}
      \centering
      \includegraphics[height=2.8cm]{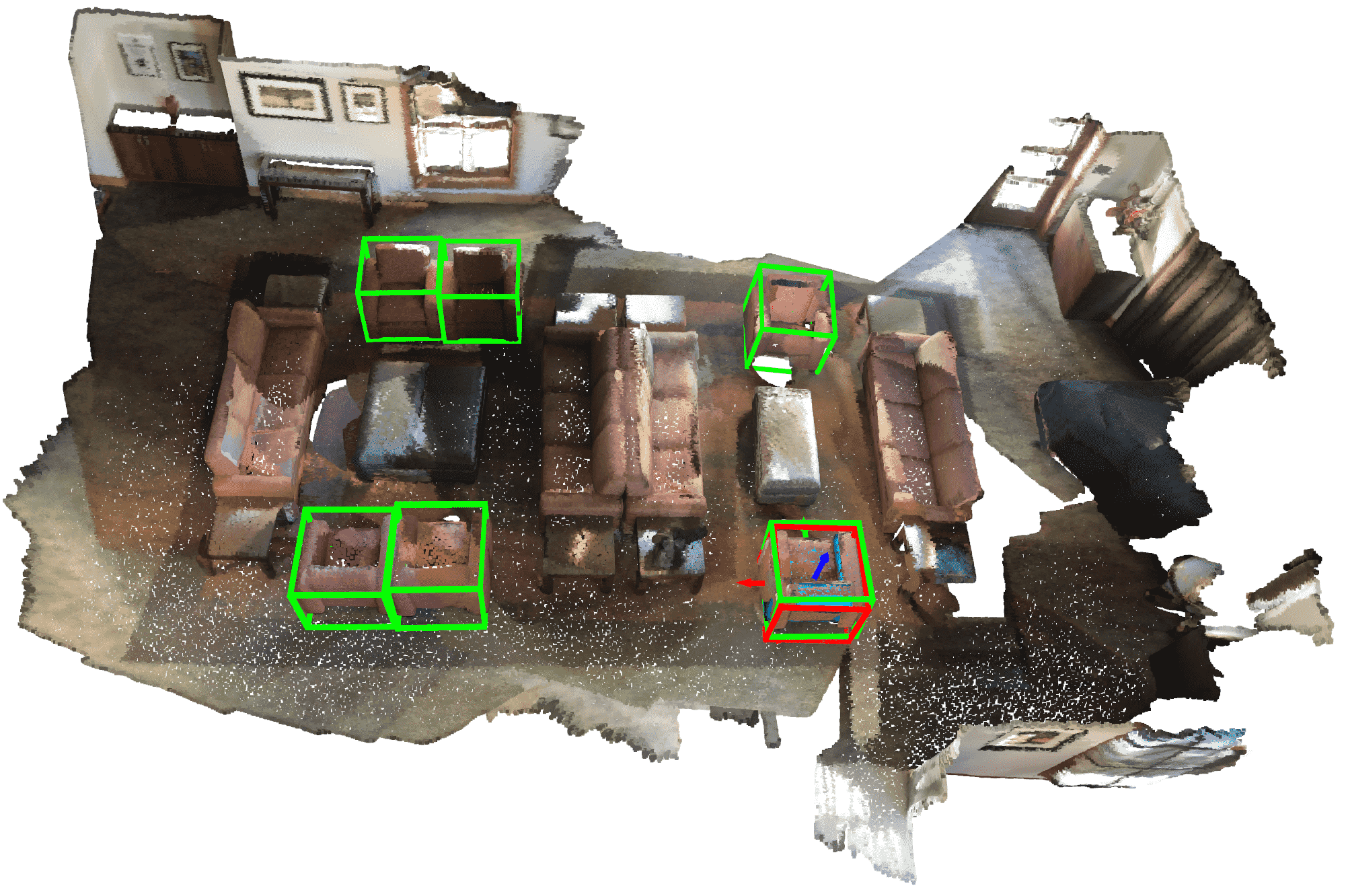}
        \caption{TEASER(2020)\cite{TEASER}}
        \label{fig:scan2cad_cad-teaser27}
    \end{subfigure}
    \begin{subfigure}{0.23\textwidth}
      \centering
      \includegraphics[height=2.8cm]{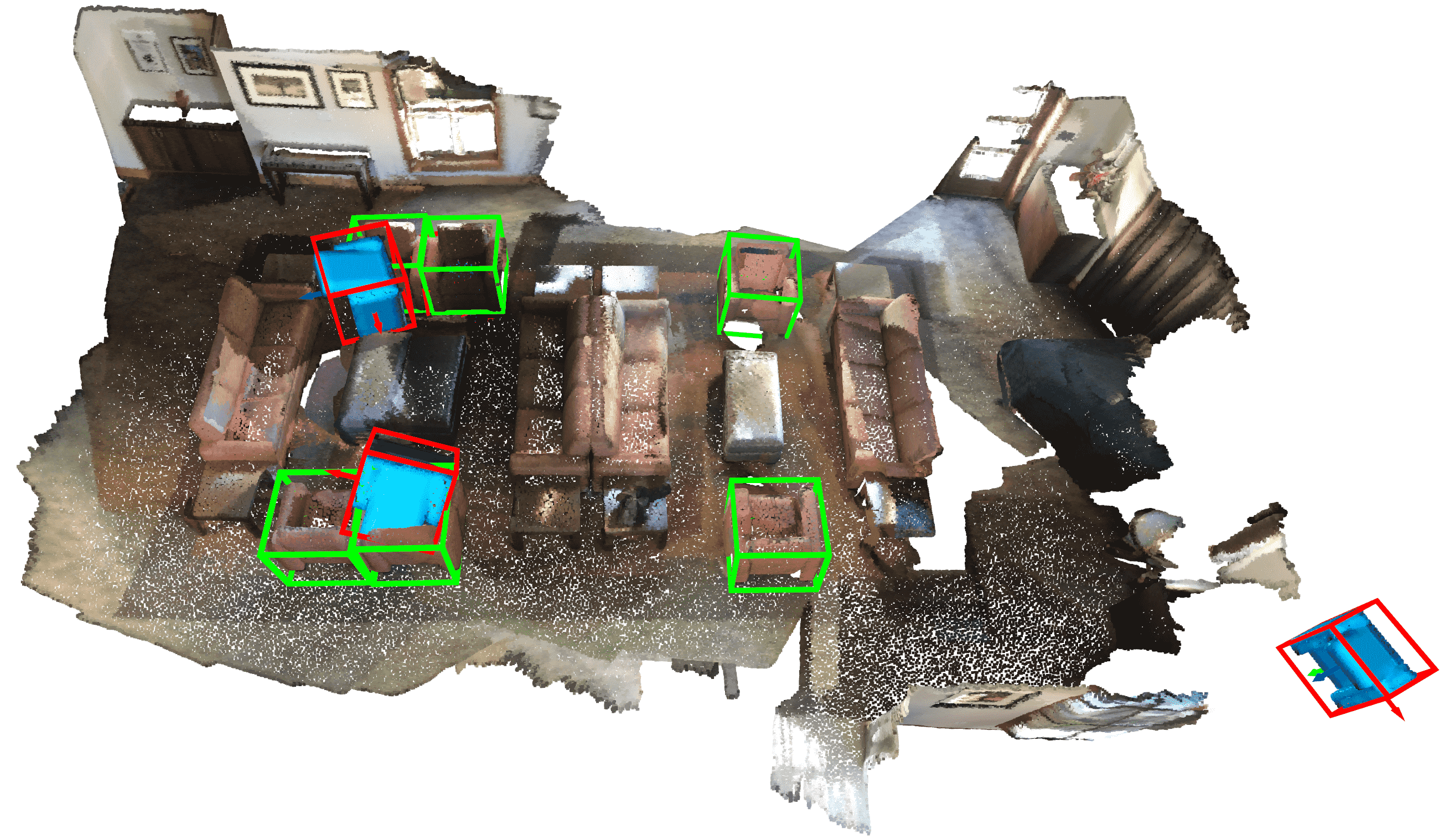}
        \caption{CONSAC(2020)\cite{CONSAC}}
        \label{fig:scan2cad_cad-consac27}
    \end{subfigure}\hfill
    
    \caption{\textbf{Scan2CAD results.}}
\label{fig:Scan2CAD-cadresult27}
  \end{figure*}

\begin{figure*}[ht]
  \centering
  \begin{subfigure}{0.3\textwidth}
      \centering
      \includegraphics[height=2.8cm]{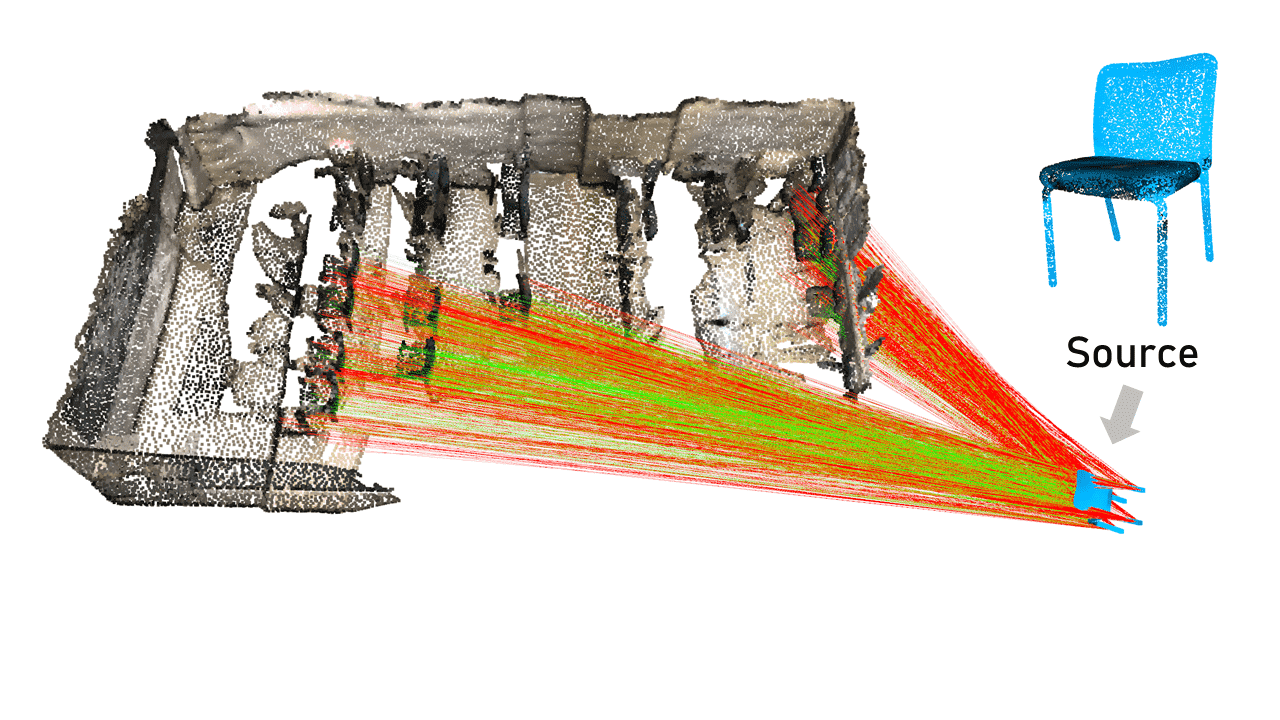}
        \caption{Input correspondences}
        \label{fig:scan2cad_cad-input-corrs3}
    \end{subfigure}\hfill
    \begin{subfigure}{0.3\textwidth}
      \centering
      \includegraphics[height=2.8cm]{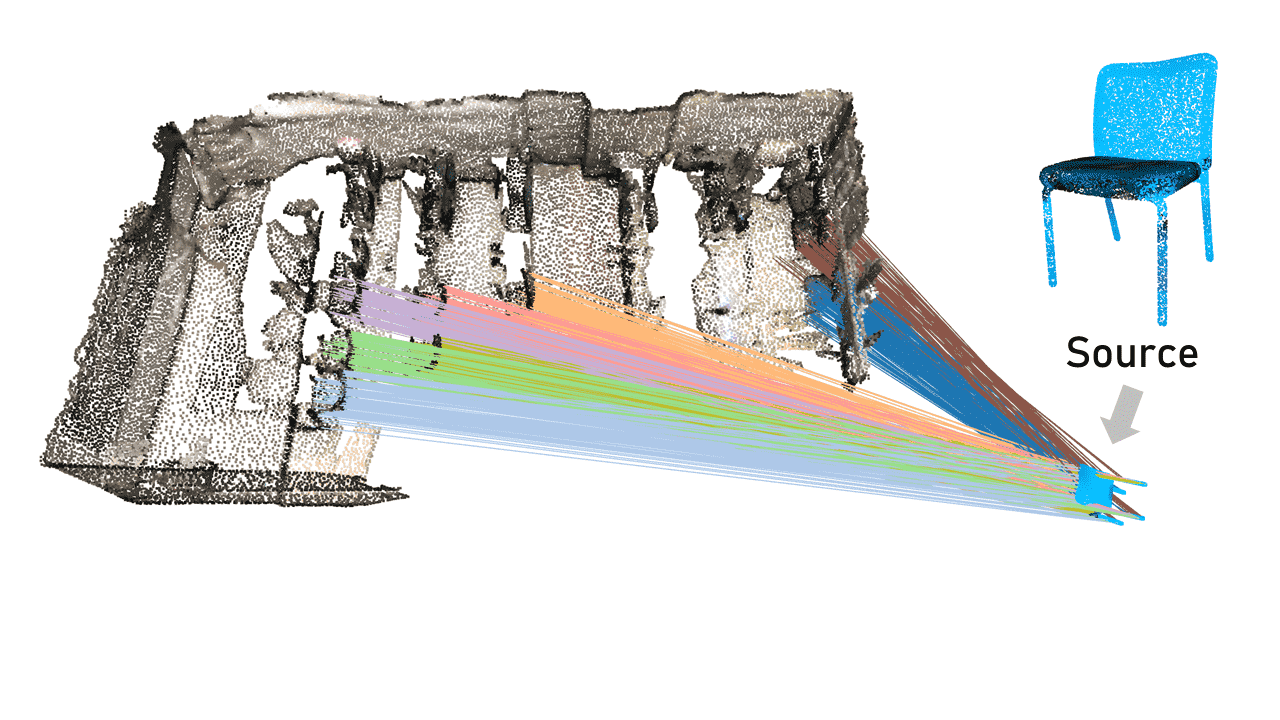}
        \caption{Our clustering result}
        \label{fig:scan2cad_cad-cluster-corrs3}
    \end{subfigure}\hfill
    \begin{subfigure}{0.3\textwidth}
      \centering
      \includegraphics[height=2.8cm]{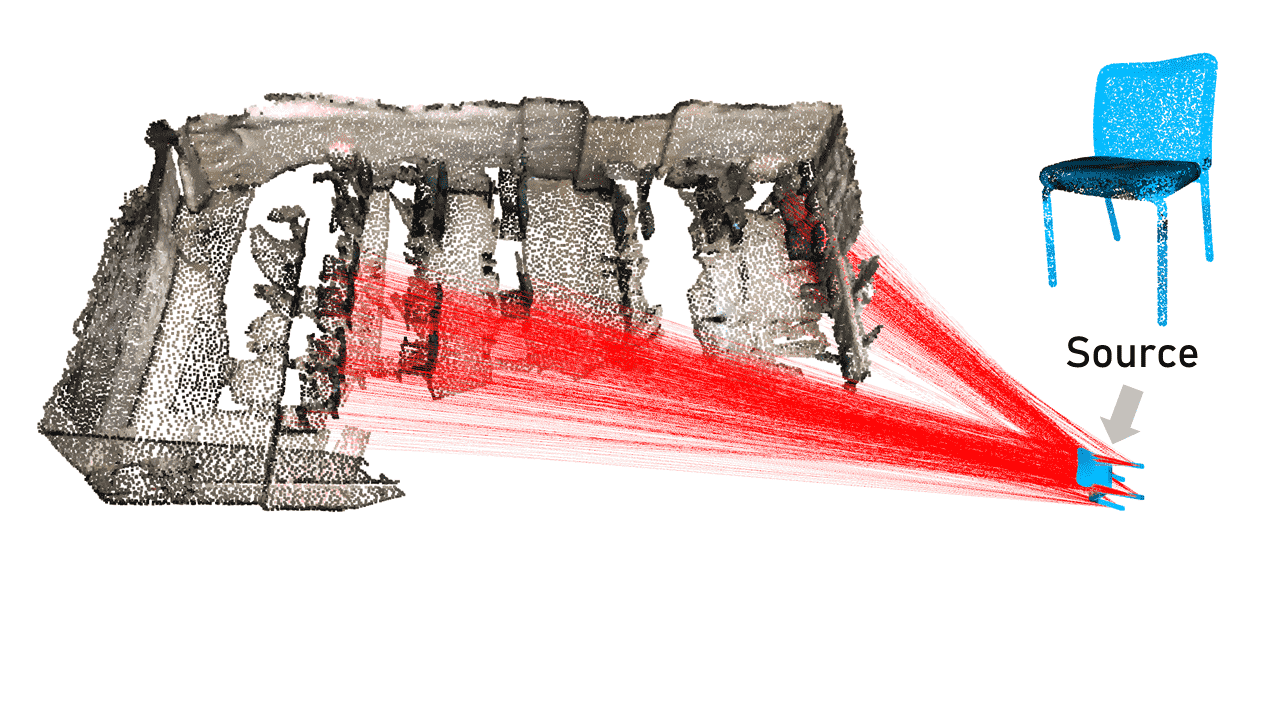}
        \caption{Our reject outliers}
        \label{fig:scan2cad_cad-reject-corrs3}
    \end{subfigure}

    \begin{subfigure}{0.23\textwidth}
      \centering
      \includegraphics[height=2.0cm]{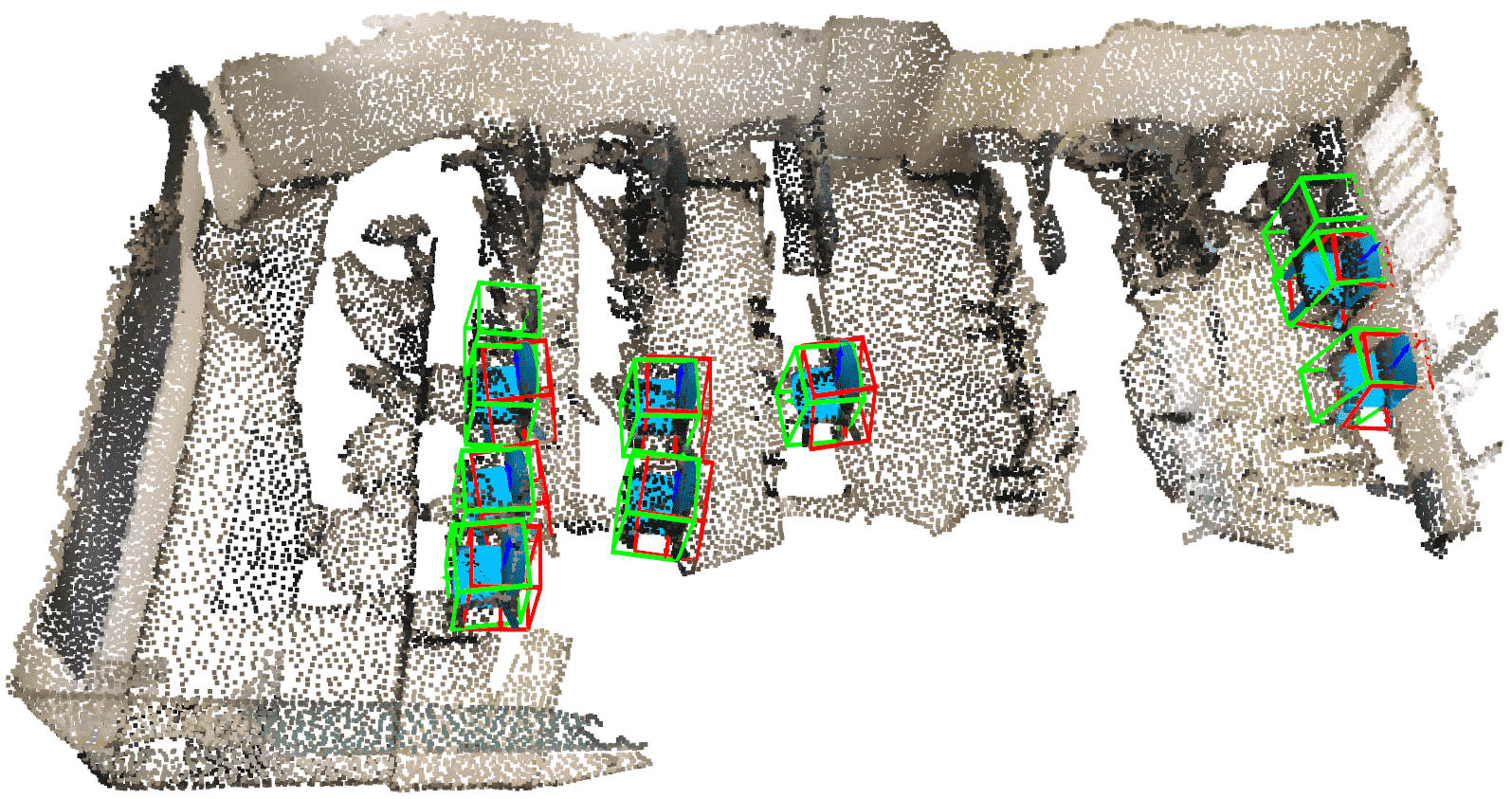}
        \caption{Ours}
        \label{fig:scan2cad_cad-result3}
    \end{subfigure}\hfill
    \begin{subfigure}{0.23\textwidth}
      \centering
      \includegraphics[height=2.0cm]{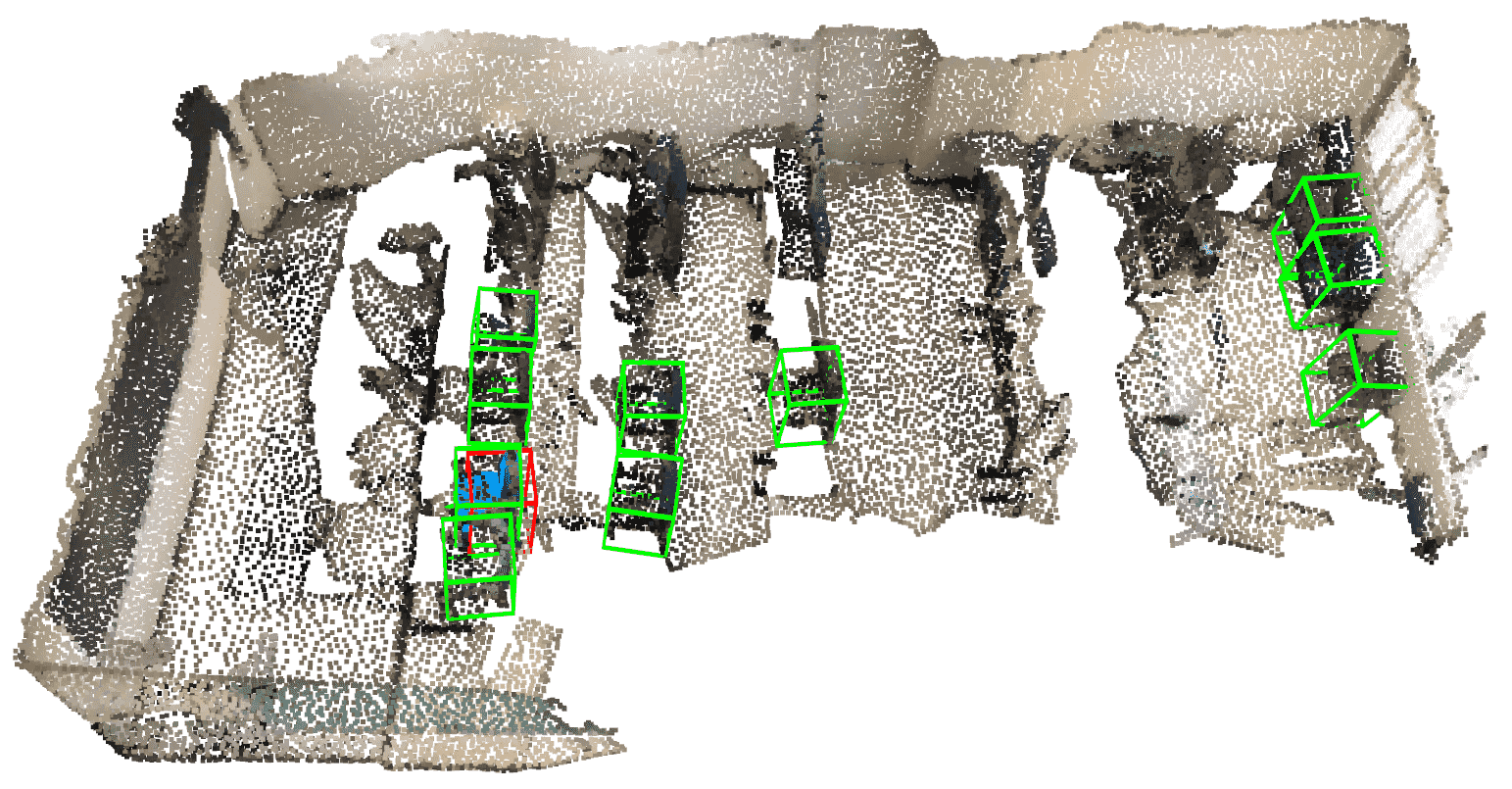}
        \caption{Progressive-X(2019) \cite{ProgressiveX}}
        \label{fig:scan2cad_cad-prox3}
    \end{subfigure}\hfill
    \begin{subfigure}{0.23\textwidth}
      \centering
      \includegraphics[height=2.0cm]{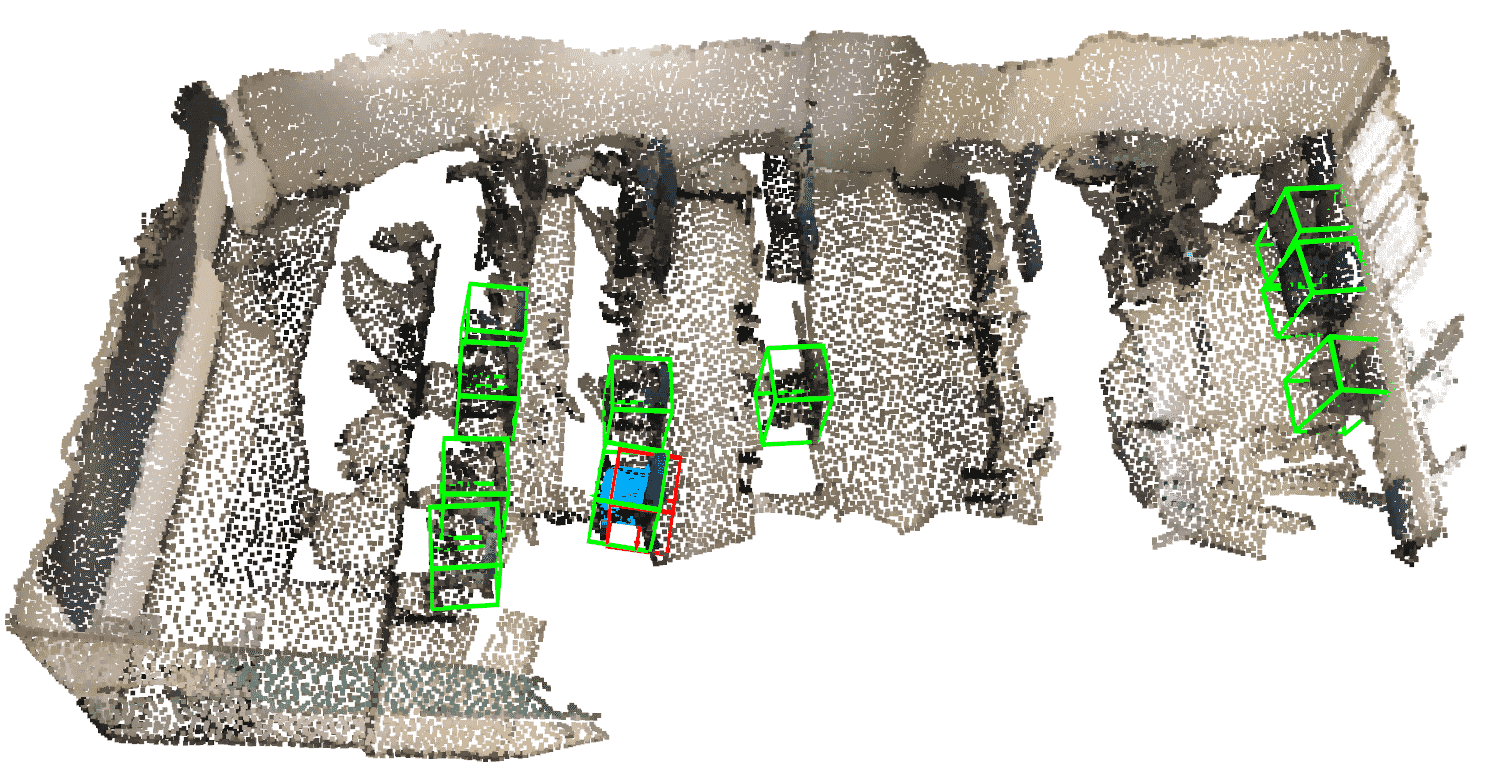}
        \caption{TEASER(2020)\cite{TEASER}}
        \label{fig:scan2cad_cad-teaser3}
    \end{subfigure}
    \begin{subfigure}{0.23\textwidth}
      \centering
      \includegraphics[height=2.0cm]{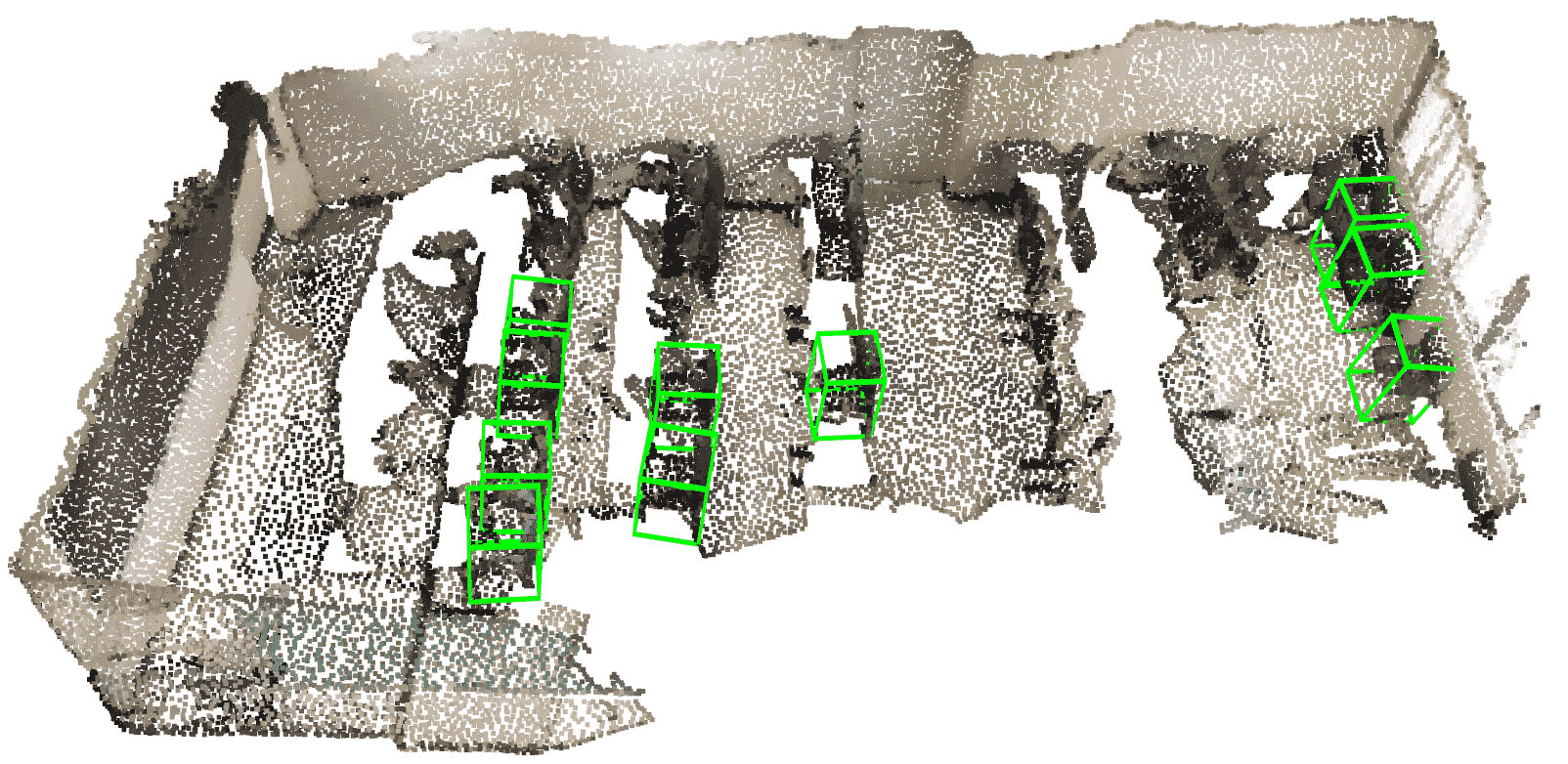}
        \caption{CONSAC(2020)\cite{CONSAC}}
        \label{fig:scan2cad_cad-consac3}
    \end{subfigure}\hfill
    \caption{\textbf{Scan2CAD results.}}
\label{fig:Scan2CAD-cadresult3}
  \end{figure*}

\begin{figure*}[ht]
  \centering
\begin{subfigure}{0.3\textwidth}
  \centering
  \includegraphics[height=2.8cm]{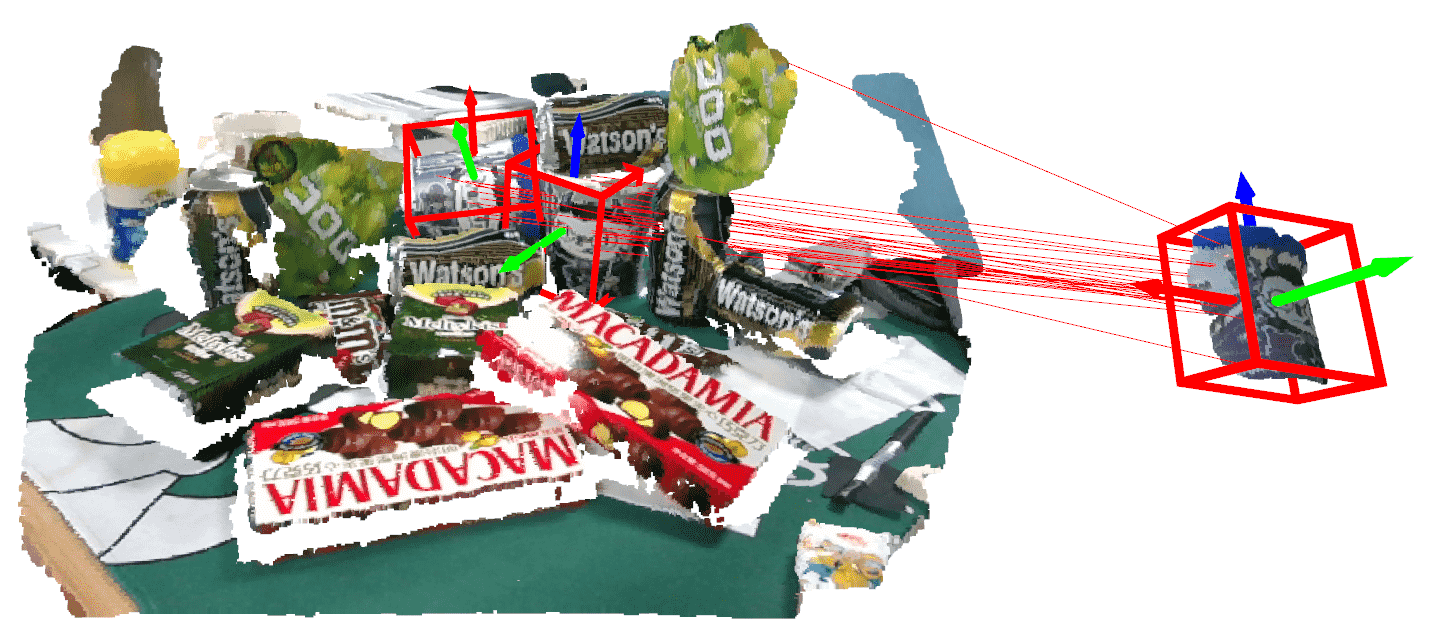}
    \caption{Ours}
    \label{fig:real-result1}
\end{subfigure}\hfill
\begin{subfigure}{0.3\textwidth}
  \centering
  \includegraphics[height=2.8cm]{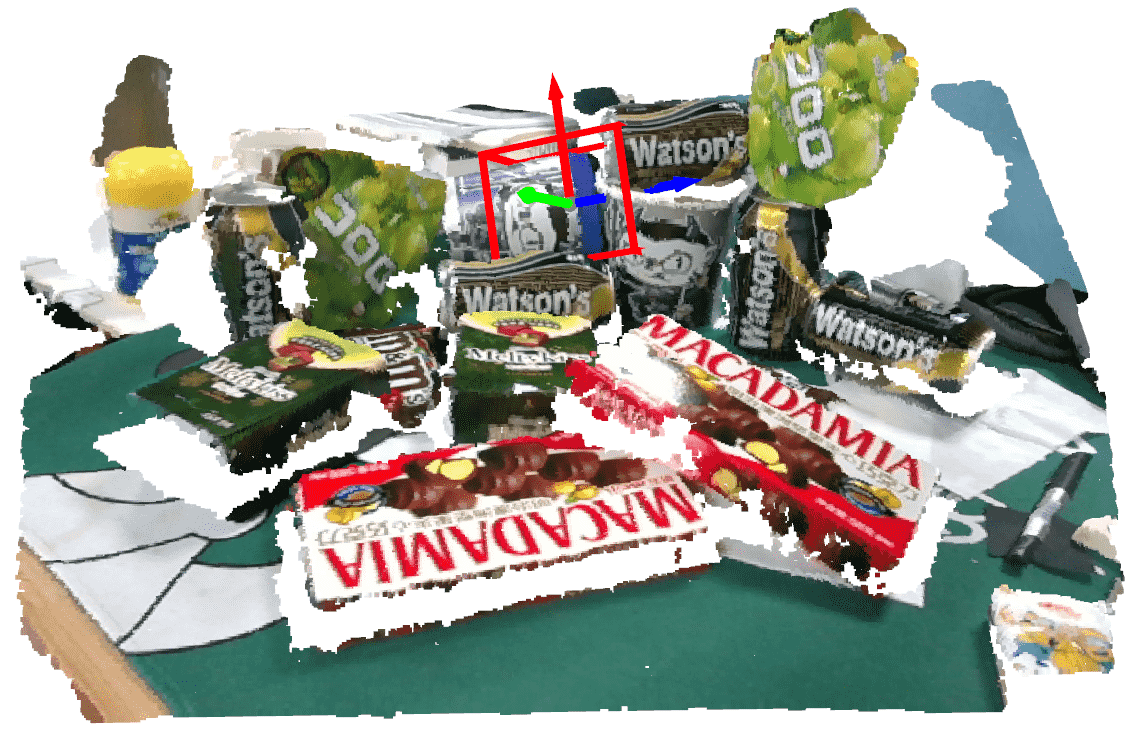}
    \caption{Progressive-X(2019) \cite{ProgressiveX}}
    \label{fig:real-prox1}
\end{subfigure}\hfill
\begin{subfigure}{0.3\textwidth}
  \centering
  \includegraphics[height=2.8cm]{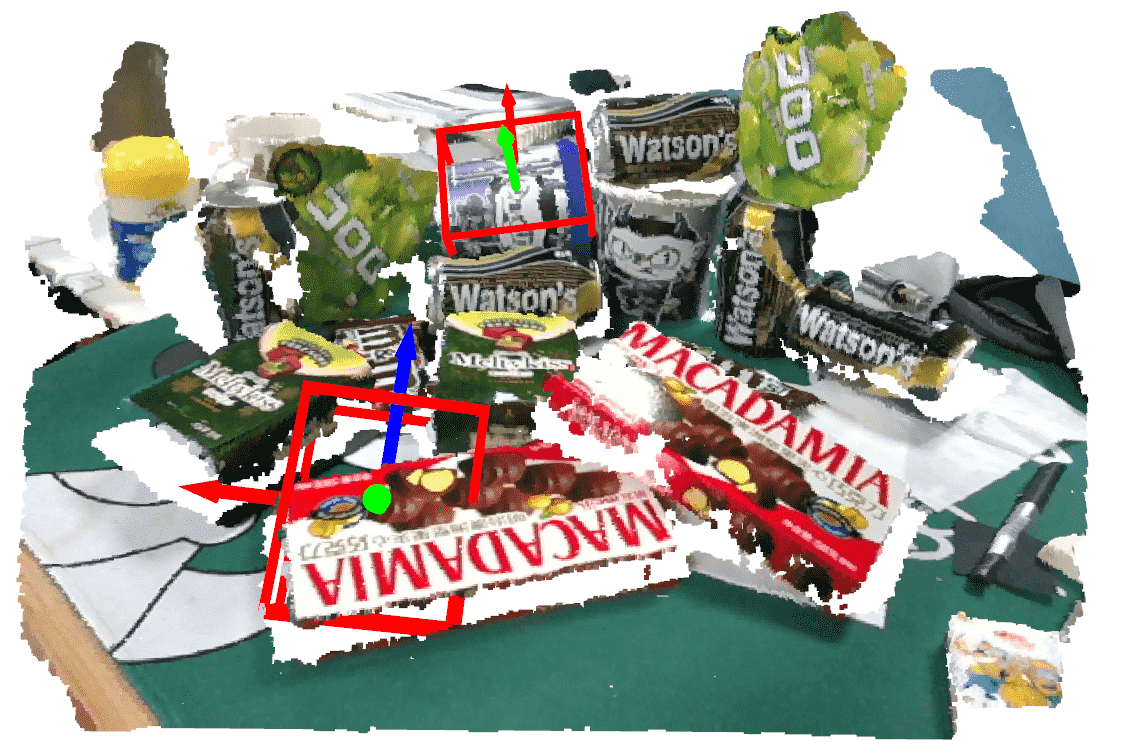}
    \caption{CONSAC(2020)\cite{CONSAC}}
    \label{fig:real-consac1}
\end{subfigure}\hfill
\caption{\textbf{Real-world tests on RGB-D scans.}}
\label{fig:Real1}
\end{figure*}

\begin{figure*}[ht]
  \centering
\begin{subfigure}{0.3\textwidth}
  \centering
  \includegraphics[height=2.8cm]{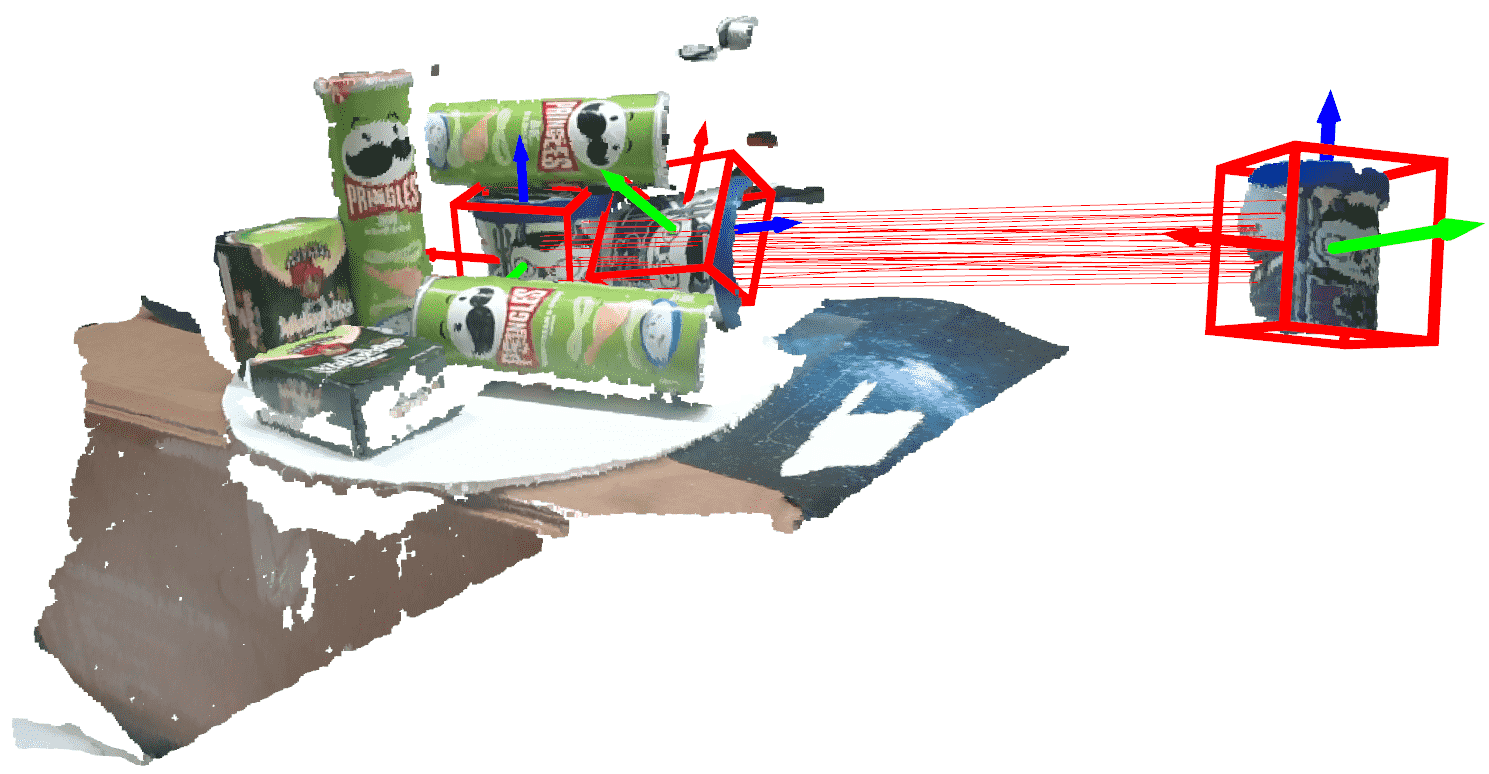}
    \caption{Ours}
    \label{fig:real-result2}
\end{subfigure}\hfill
\begin{subfigure}{0.3\textwidth}
  \centering
  \includegraphics[height=2.8cm]{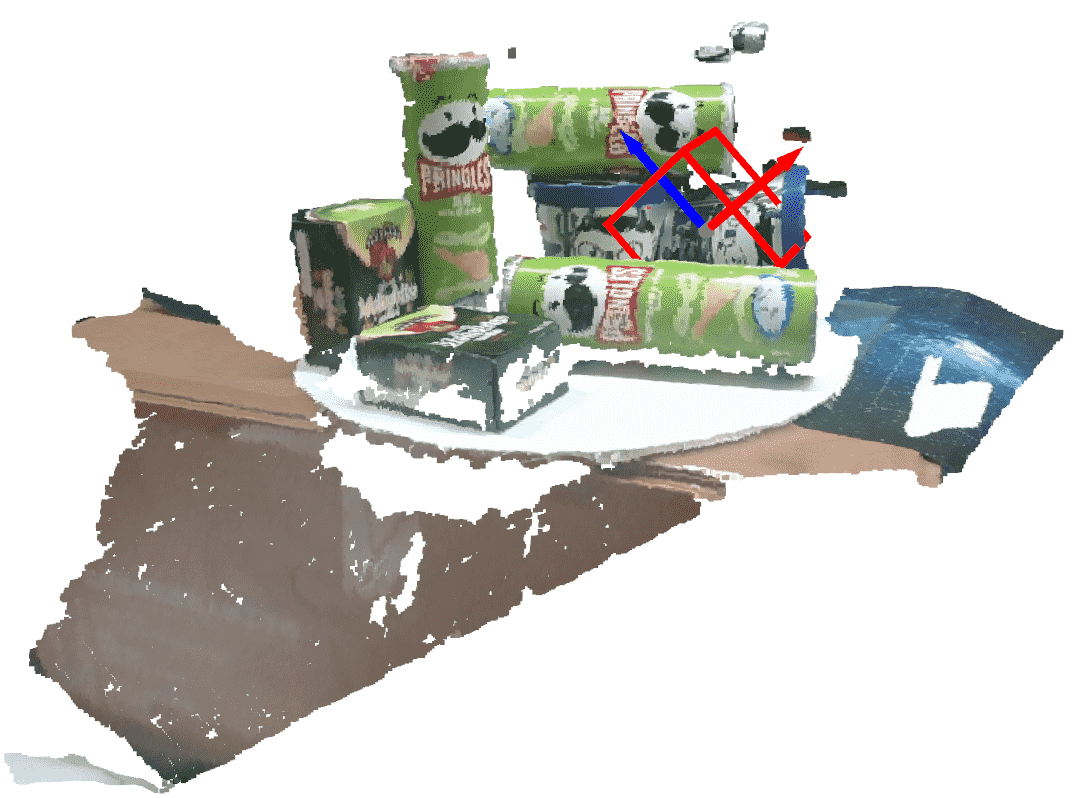}
    \caption{Progressive-X(2019) \cite{ProgressiveX}}
    \label{fig:real-prox2}
\end{subfigure}\hfill
\begin{subfigure}{0.3\textwidth}
  \centering
  \includegraphics[height=2.8cm]{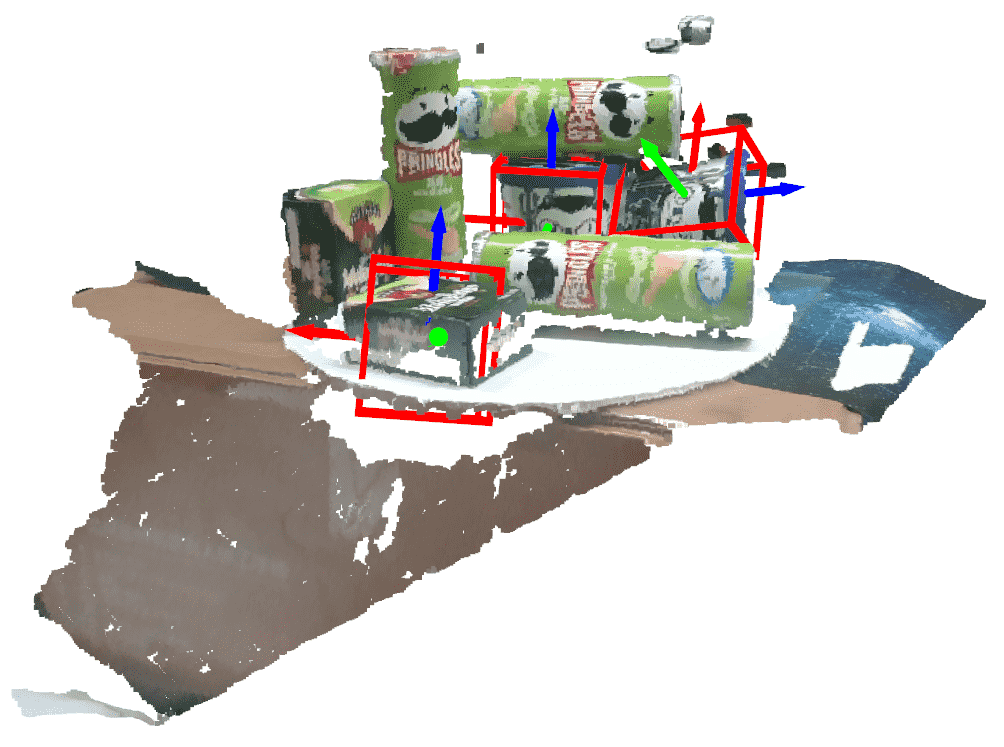}
    \caption{CONSAC(2020)\cite{CONSAC}}
    \label{fig:real-consac2}
\end{subfigure}\hfill
\caption{\textbf{Real-world tests on RGB-D scans.}}
\label{fig:Real2}
\end{figure*}

\begin{figure*}[ht]
  \centering
\begin{subfigure}{0.3\textwidth}
  \centering
  \includegraphics[height=2.8cm]{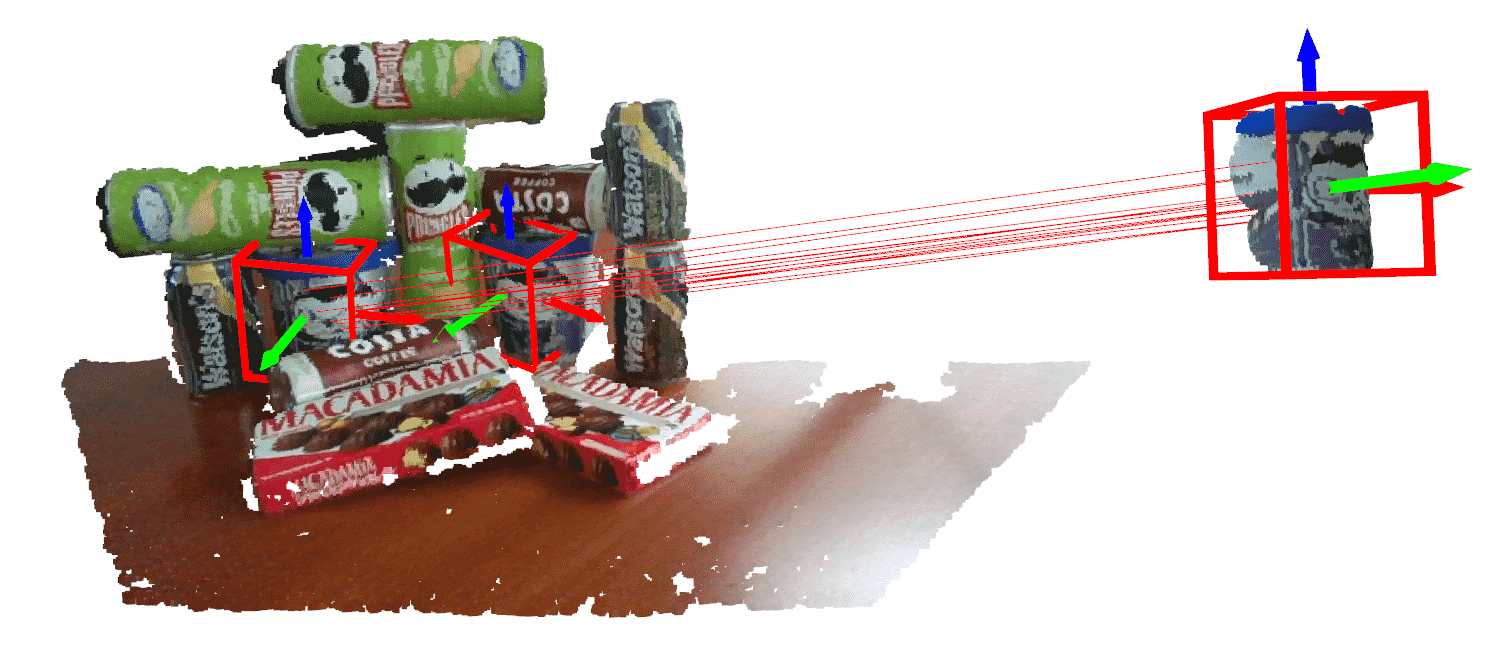}
    \caption{Ours}
    \label{fig:real-result3}
\end{subfigure}\hfill
\begin{subfigure}{0.3\textwidth}
  \centering
  \includegraphics[height=2.8cm]{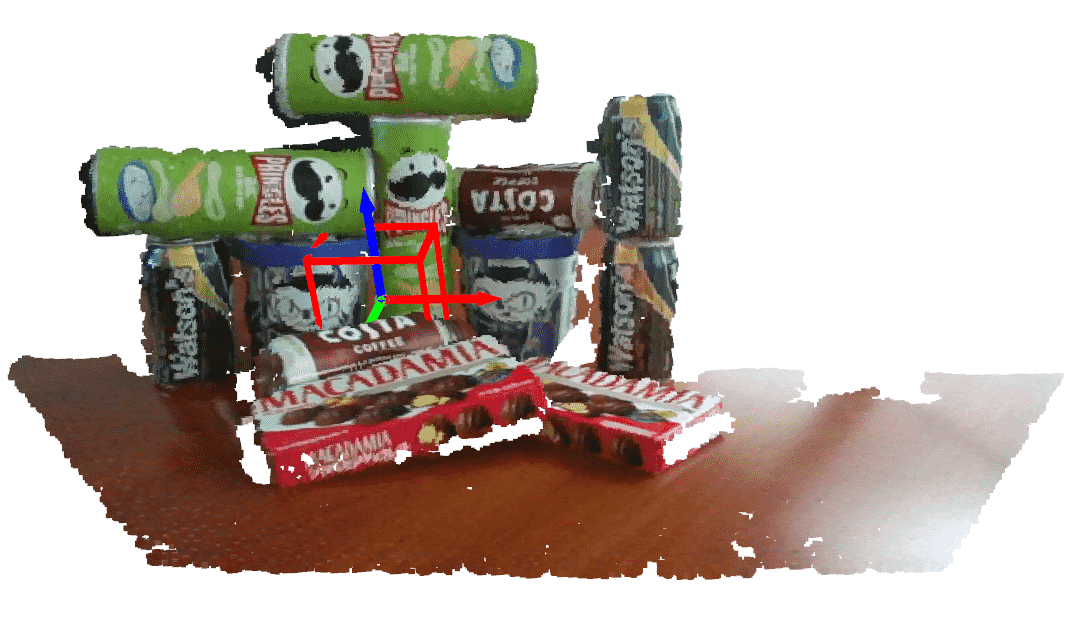}
    \caption{Progressive-X(2019) \cite{ProgressiveX}}
    \label{fig:real-prox3}
\end{subfigure}\hfill
\begin{subfigure}{0.3\textwidth}
  \centering
  \includegraphics[height=2.8cm]{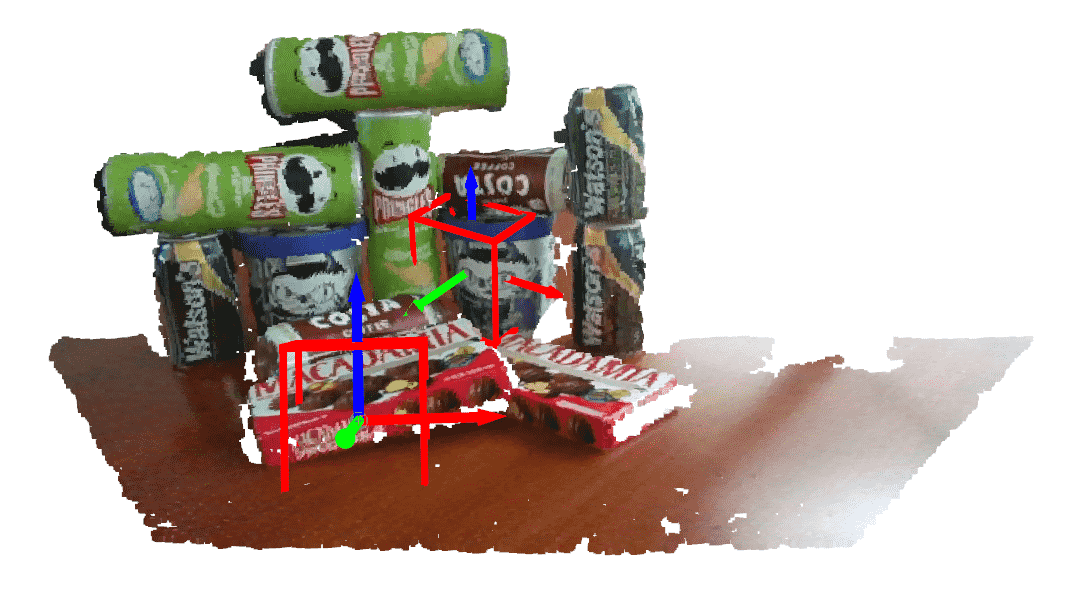}
    \caption{CONSAC(2020)\cite{CONSAC}}
    \label{fig:real-consac3}
\end{subfigure}\hfill
\caption{\textbf{Real-world tests on RGB-D scans.}}
\label{fig:Real3}
\end{figure*}

\begin{figure*}[ht]
  \centering
\begin{subfigure}{0.4\textwidth}
  \centering
  \includegraphics[height=2.8cm]{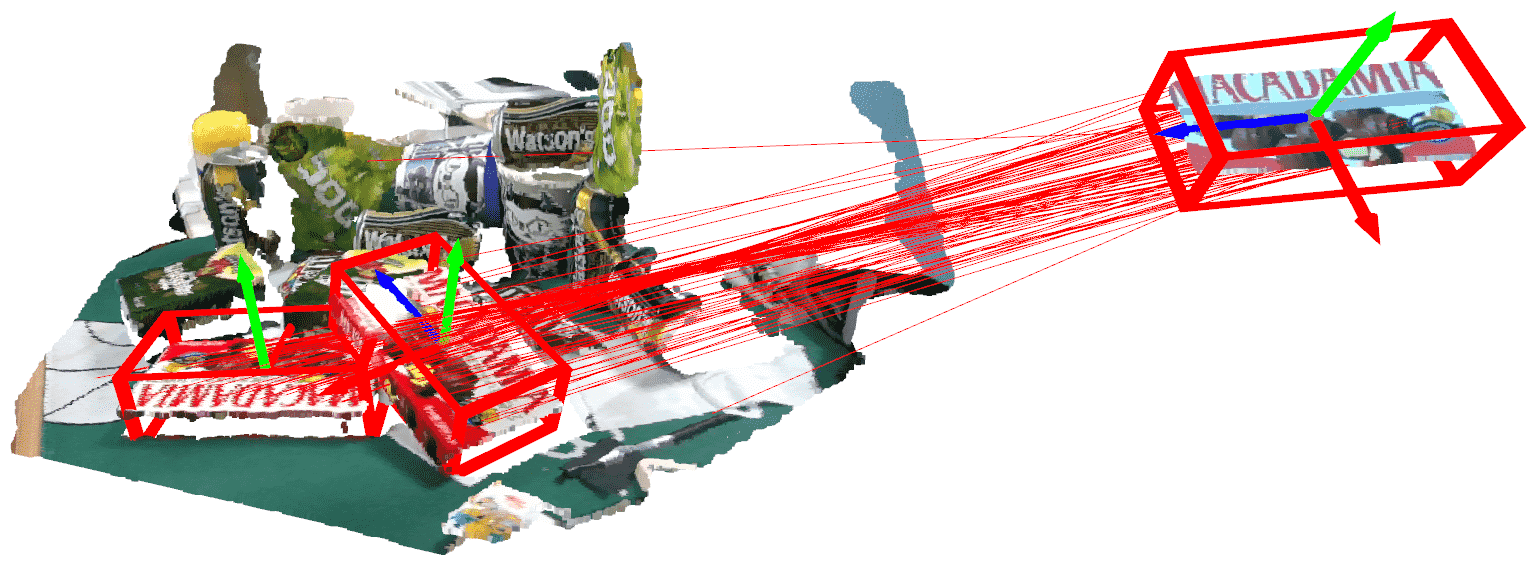}
    \caption{Ours}
    \label{fig:real-result4}
\end{subfigure}\hfill
\begin{subfigure}{0.25\textwidth}
  \centering
  \includegraphics[height=2.8cm]{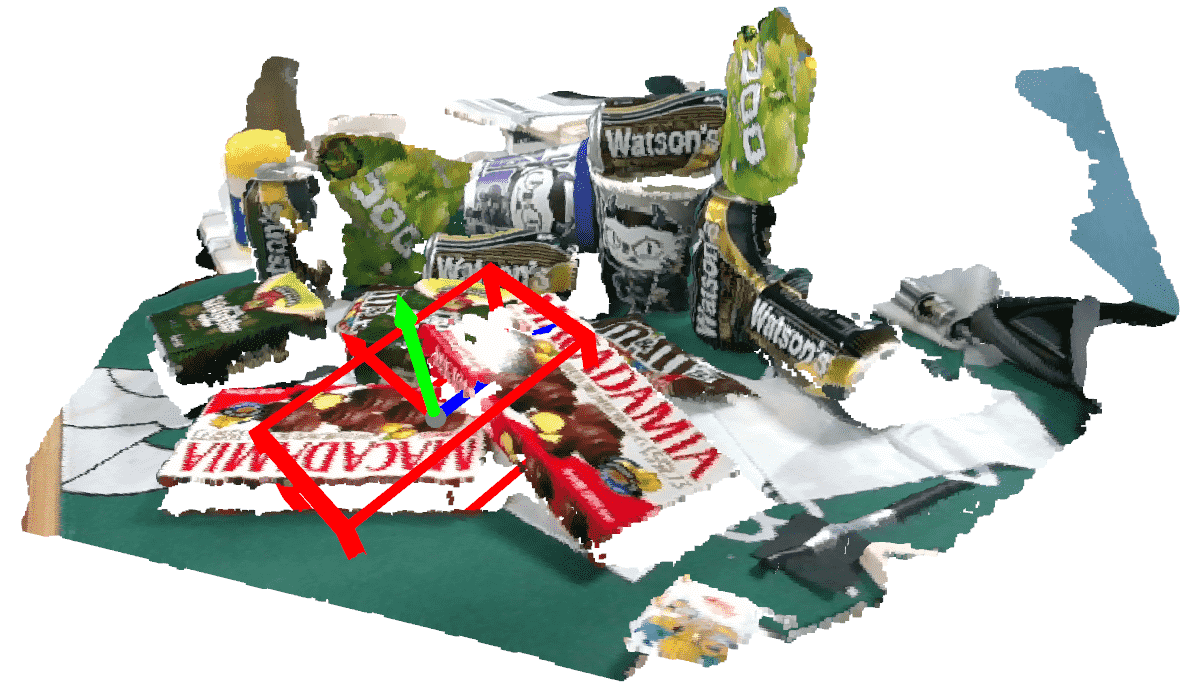}
    \caption{Progressive-X(2019) \cite{ProgressiveX}}
    \label{fig:real-prox4}
\end{subfigure}\hfill
\begin{subfigure}{0.25\textwidth}
  \centering
  \includegraphics[height=2.8cm]{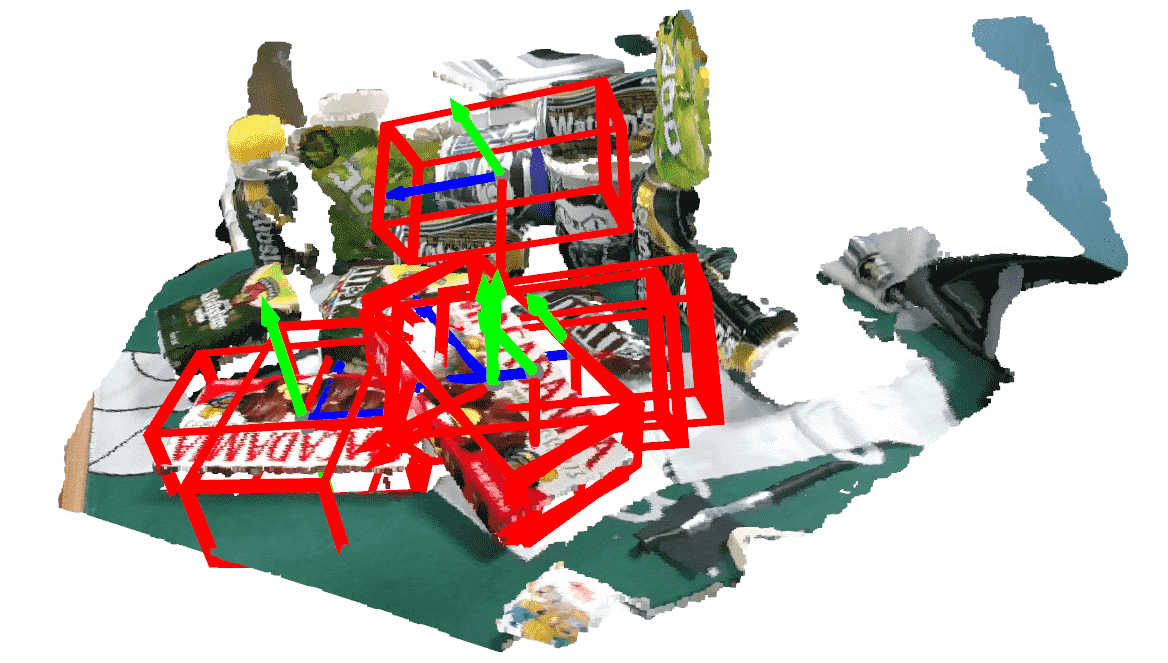}
    \caption{CONSAC(2020)\cite{CONSAC}}
    \label{fig:real-consac4}
\end{subfigure}\hfill
\caption{\textbf{Real-world tests on RGB-D scans.}}
\label{fig:Real4}
\end{figure*}

\begin{figure*}[ht]
  \centering
\begin{subfigure}{0.3\textwidth}
  \centering
  \includegraphics[height=2.8cm]{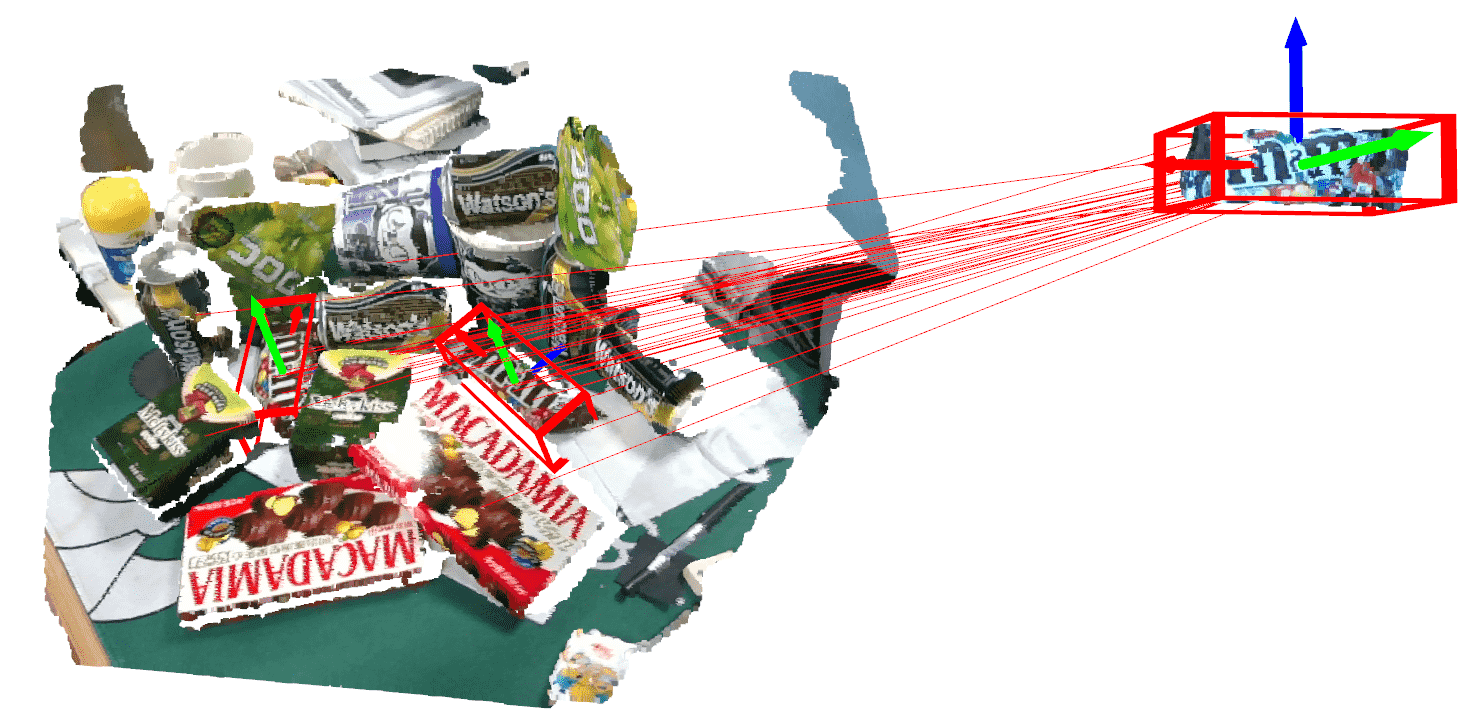}
    \caption{Ours}
    \label{fig:real-result6}
\end{subfigure}\hfill
\begin{subfigure}{0.3\textwidth}
  \centering
  \includegraphics[height=2.8cm]{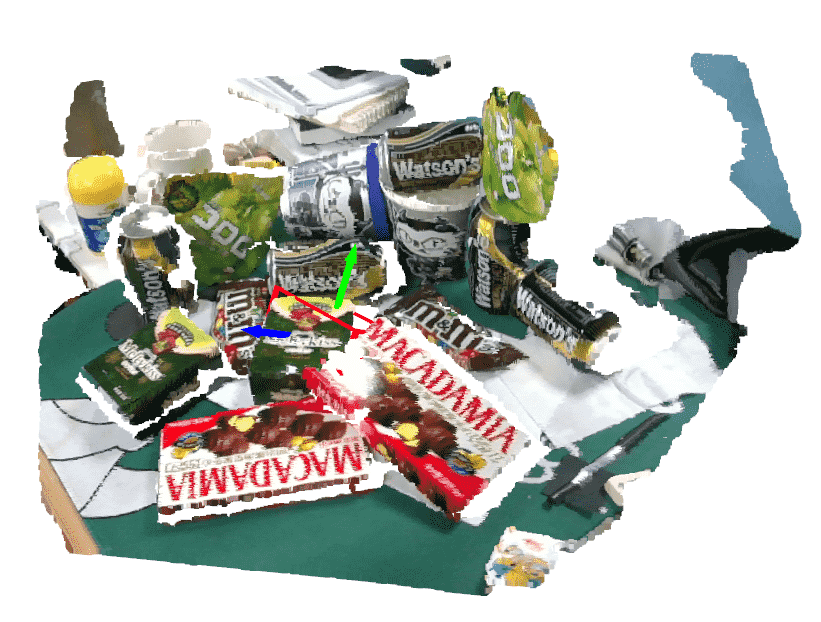}
    \caption{Progressive-X(2019) \cite{ProgressiveX}}
    \label{fig:real-prox6}
\end{subfigure}\hfill
\begin{subfigure}{0.3\textwidth}
  \centering
  \includegraphics[height=2.8cm]{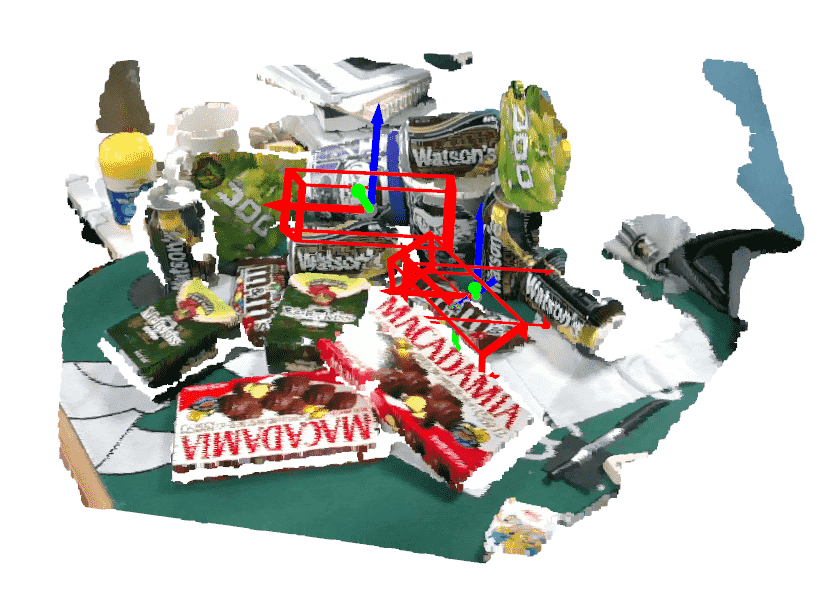}
    \caption{CONSAC(2020)\cite{CONSAC}}
    \label{fig:real-consac6}
\end{subfigure}\hfill
\caption{\textbf{Real-world tests on RGB-D scans.}}
\label{fig:Real6}
\end{figure*}

\begin{figure*}[ht]
  \centering
\begin{subfigure}{0.3\textwidth}
  \centering
  \includegraphics[height=2.8cm]{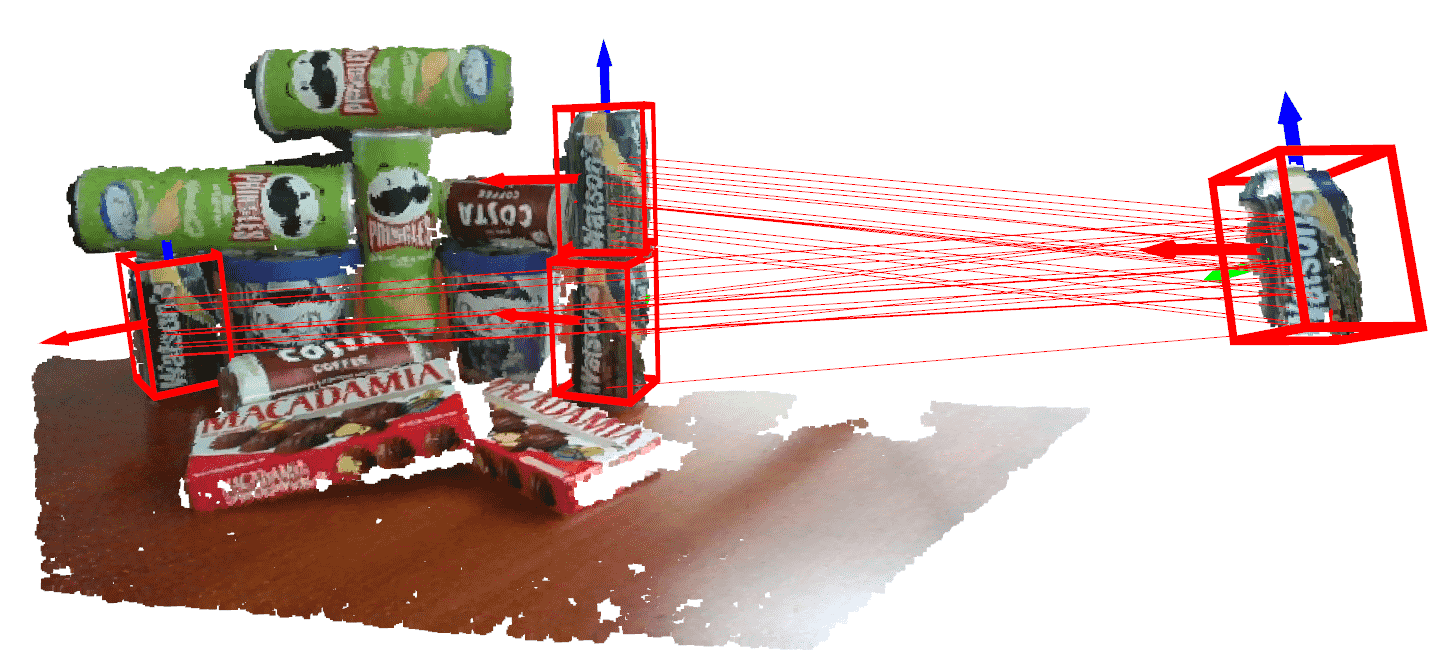}
    \caption{Ours}
    \label{fig:real-result10}
\end{subfigure}\hfill
\begin{subfigure}{0.3\textwidth}
  \centering
  \includegraphics[height=2.8cm]{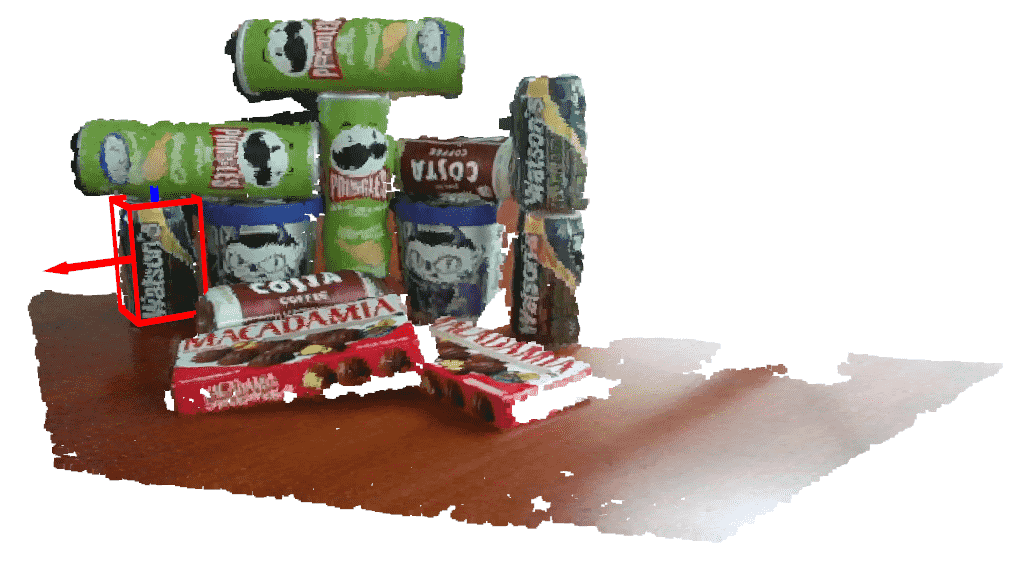}
    \caption{Progressive-X(2019) \cite{ProgressiveX}}
    \label{fig:real-prox10}
\end{subfigure}\hfill
\begin{subfigure}{0.3\textwidth}
  \centering
  \includegraphics[height=2.8cm]{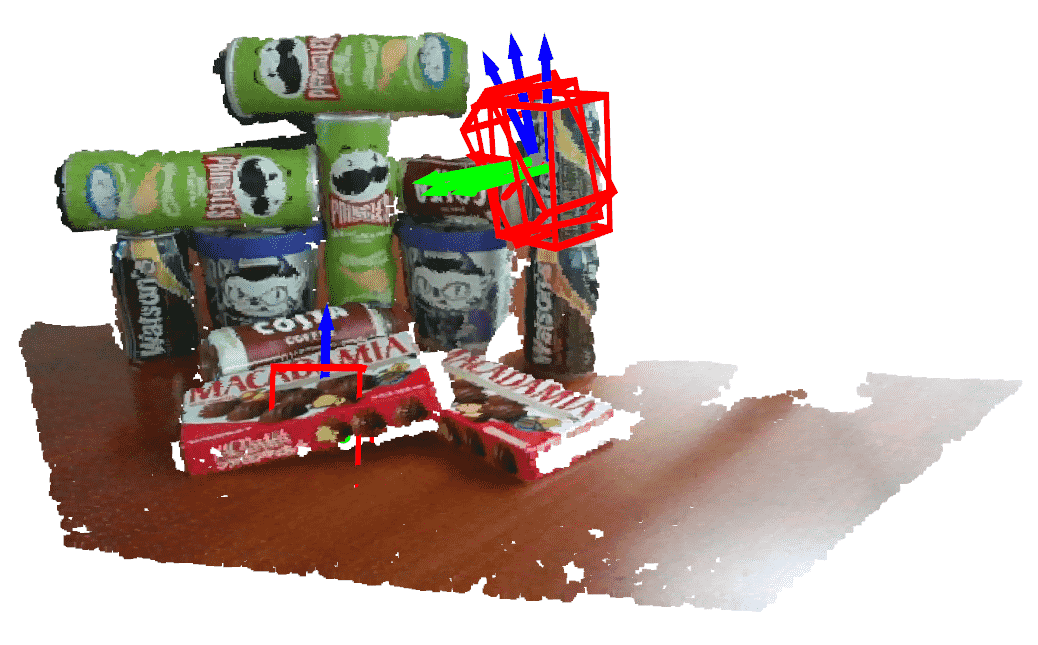}
    \caption{CONSAC(2020)\cite{CONSAC}}
    \label{fig:real-consac10}
\end{subfigure}\hfill
\caption{\textbf{Real-world tests on RGB-D scans.}}
\label{fig:Real10}
\end{figure*}

\begin{figure*}[ht]
  \centering
\begin{subfigure}{0.3\textwidth}
  \centering
  \includegraphics[height=2.8cm]{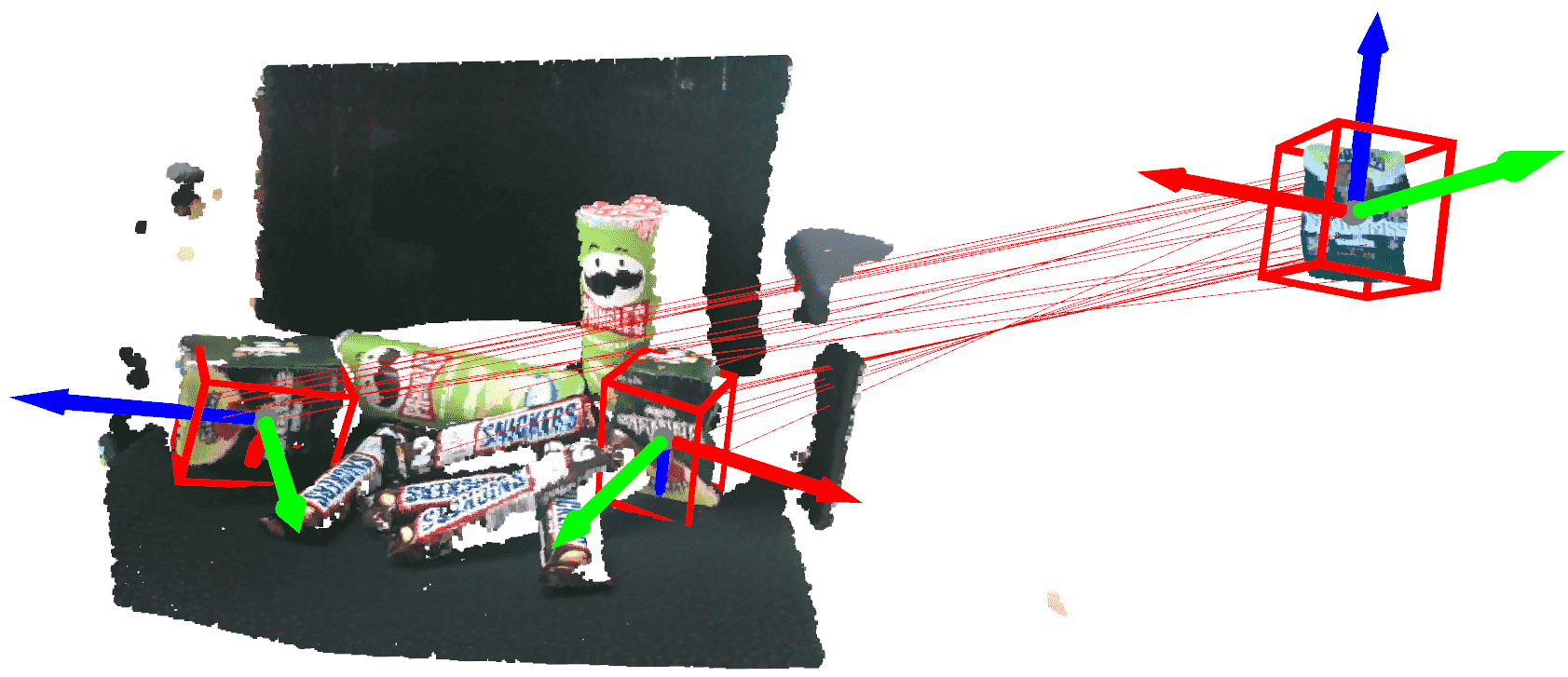}
    \caption{Ours}
    \label{fig:real-result13}
\end{subfigure}\hfill
\begin{subfigure}{0.3\textwidth}
  \centering
  \includegraphics[height=2.8cm]{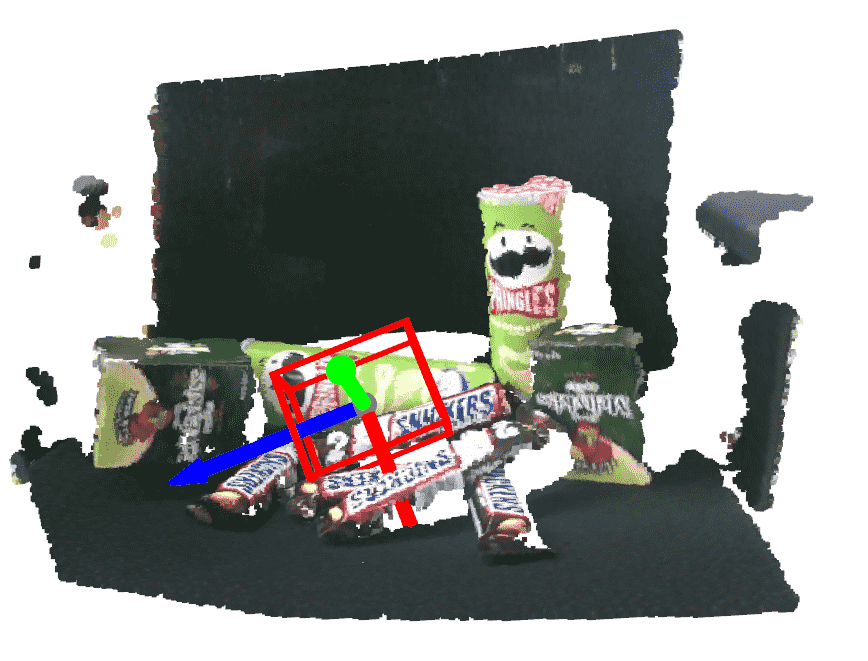}
    \caption{Progressive-X(2019) \cite{ProgressiveX}}
    \label{fig:real-prox13}
\end{subfigure}\hfill
\begin{subfigure}{0.3\textwidth}
  \centering
  \includegraphics[height=2.8cm]{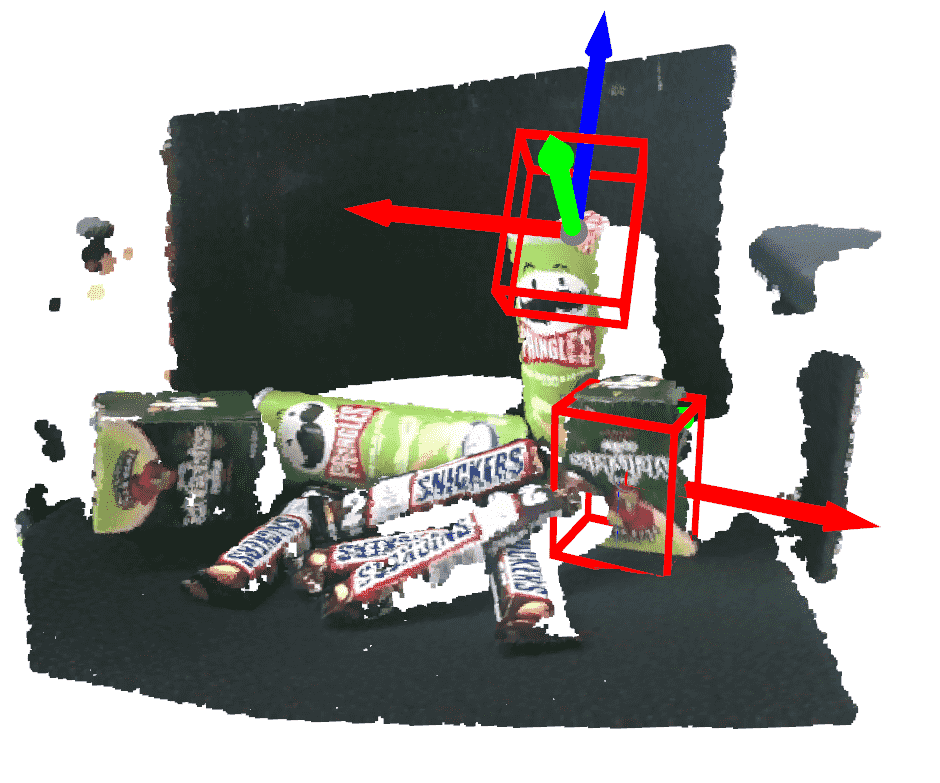}
    \caption{CONSAC(2020)\cite{CONSAC}}
    \label{fig:real-consac13}
\end{subfigure}\hfill
\caption{\textbf{Real-world tests on RGB-D scans.}}
\label{fig:Real13}
\end{figure*}

\begin{figure*}[ht]
  \centering
\begin{subfigure}{0.3\textwidth}
  \centering
  \includegraphics[height=2.8cm]{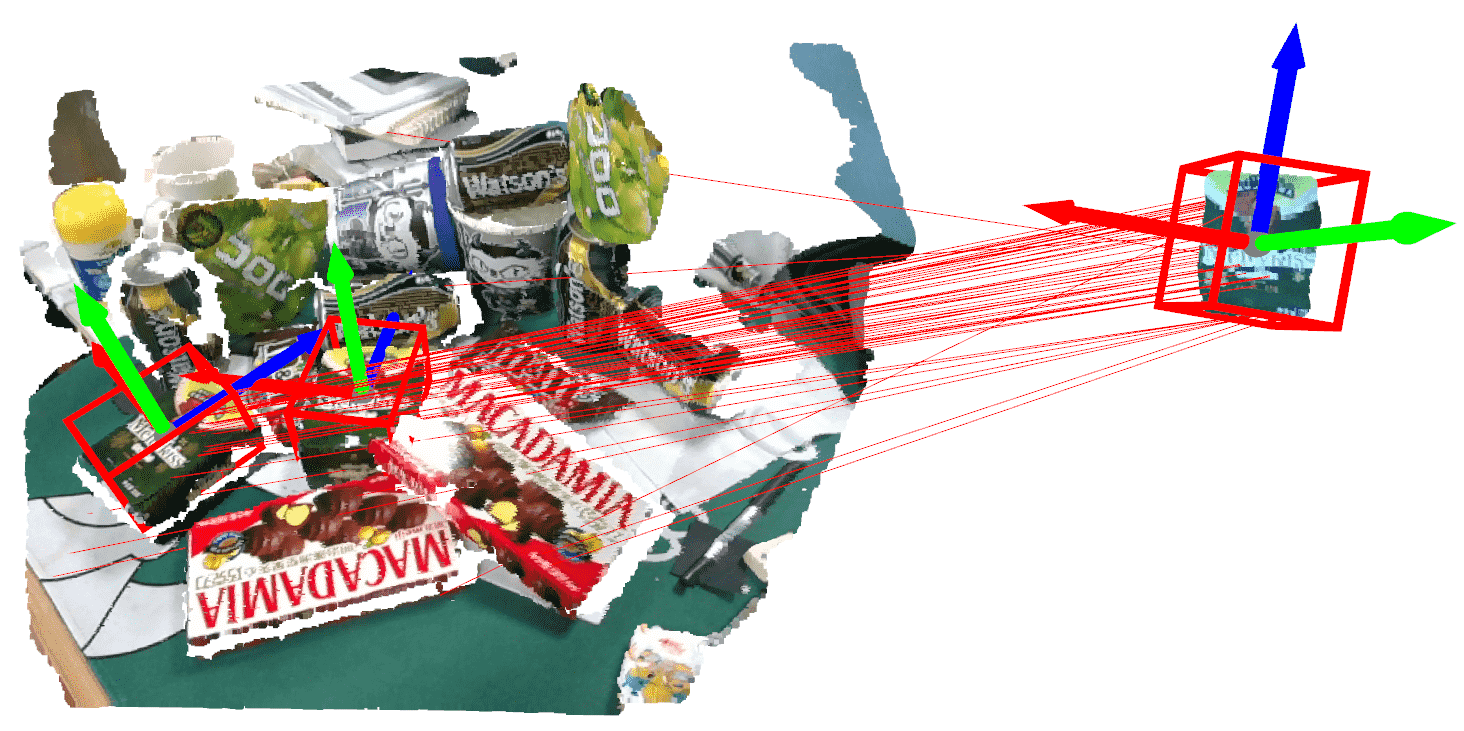}
    \caption{Ours}
    \label{fig:real-result15}
\end{subfigure}\hfill
\begin{subfigure}{0.3\textwidth}
  \centering
  \includegraphics[height=2.8cm]{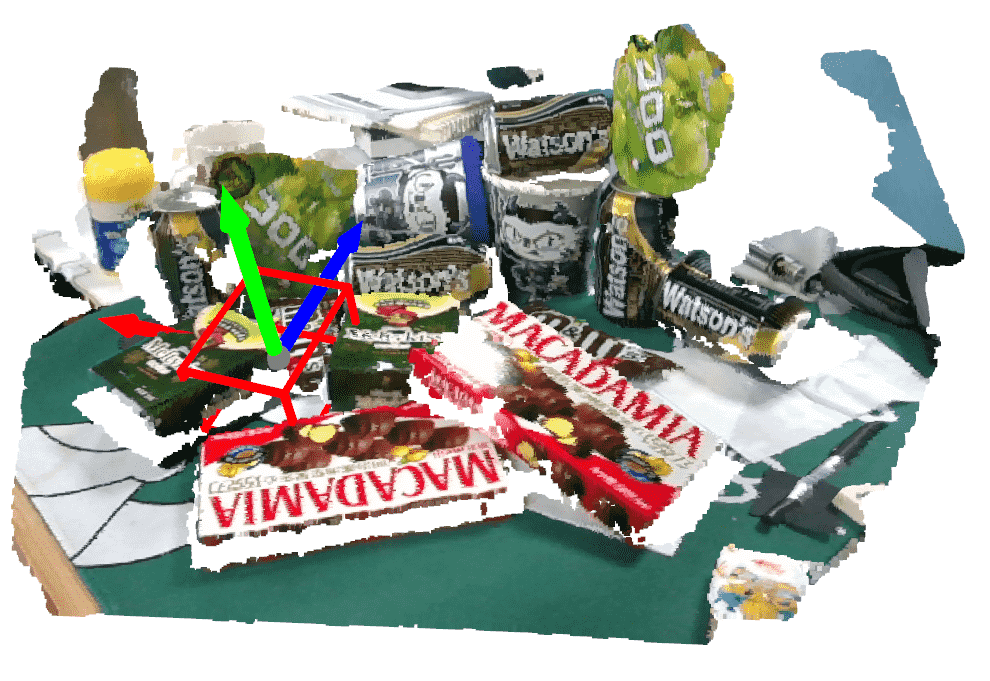}
    \caption{Progressive-X(2019) \cite{ProgressiveX}}
    \label{fig:real-prox15}
\end{subfigure}\hfill
\begin{subfigure}{0.3\textwidth}
  \centering
  \includegraphics[height=2.8cm]{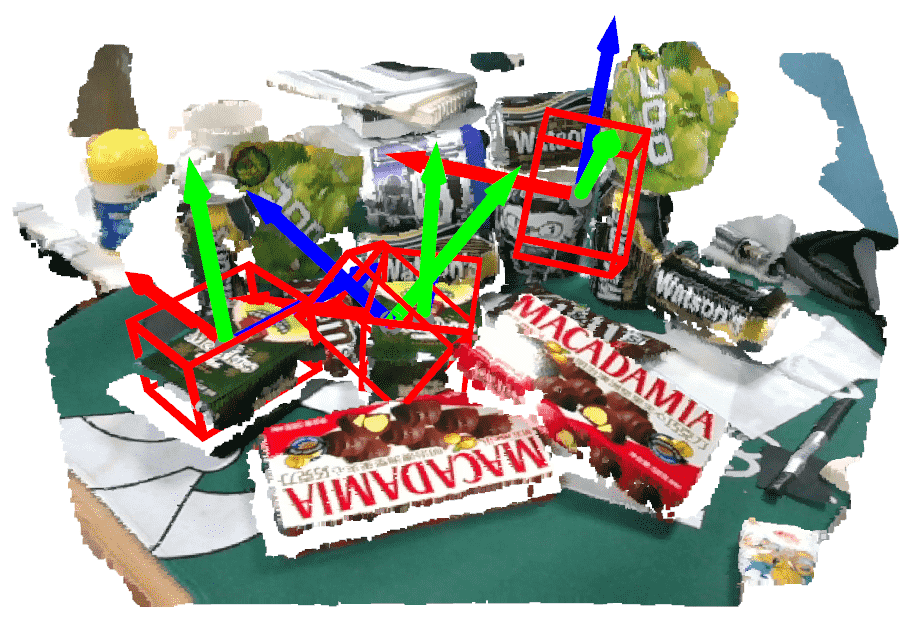}
    \caption{CONSAC(2020)\cite{CONSAC}}
    \label{fig:real-consac15}
\end{subfigure}\hfill
\caption{\textbf{Real-world tests on RGB-D scans.}}
\label{fig:Real15}
\end{figure*}

\begin{figure*}[ht]
  \centering
\begin{subfigure}{0.3\textwidth}
  \centering
  \includegraphics[height=2.8cm]{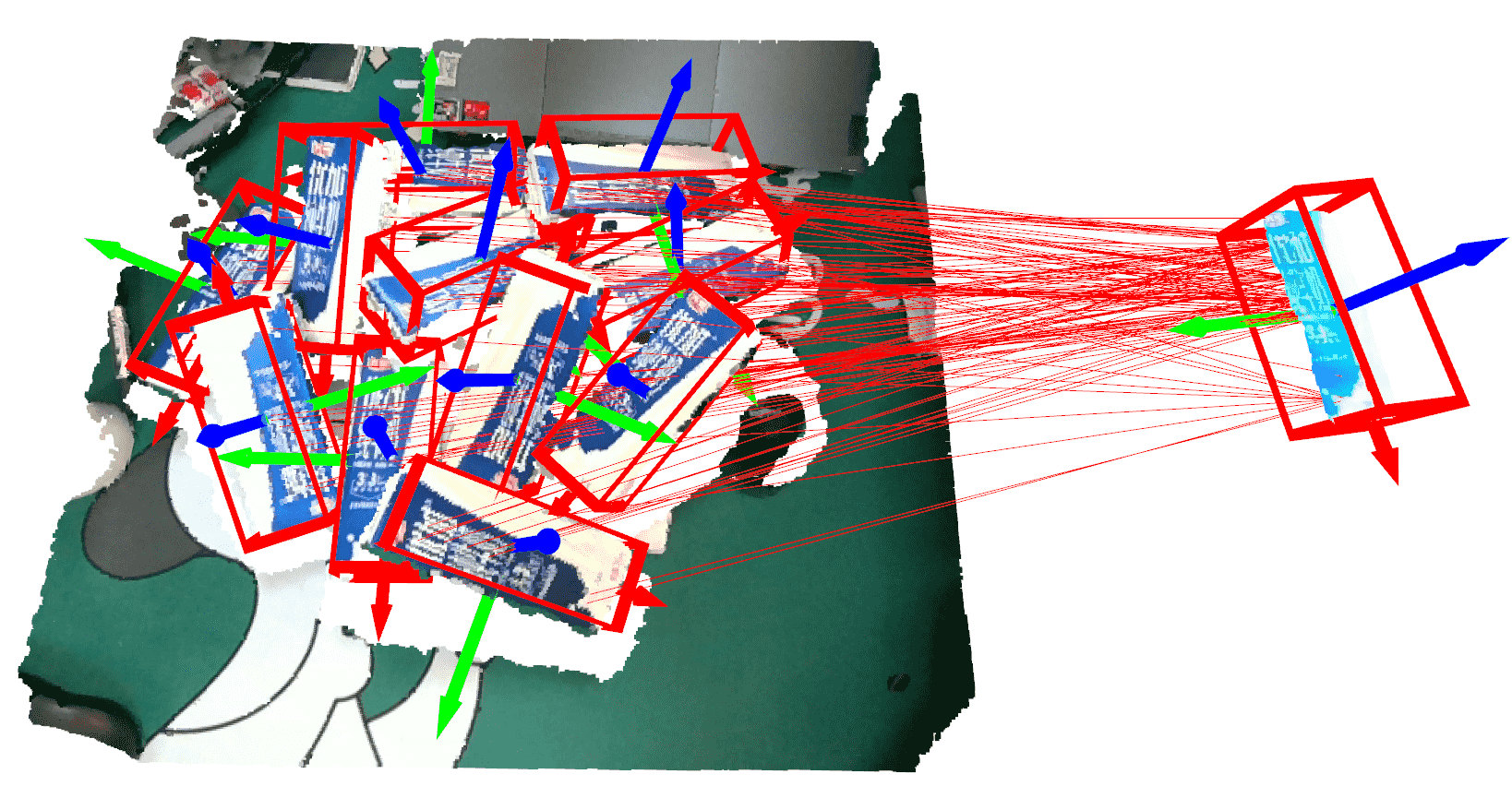}
    \caption{Ours}
    \label{fig:real-result16}
\end{subfigure}\hfill
\begin{subfigure}{0.3\textwidth}
  \centering
  \includegraphics[height=2.8cm]{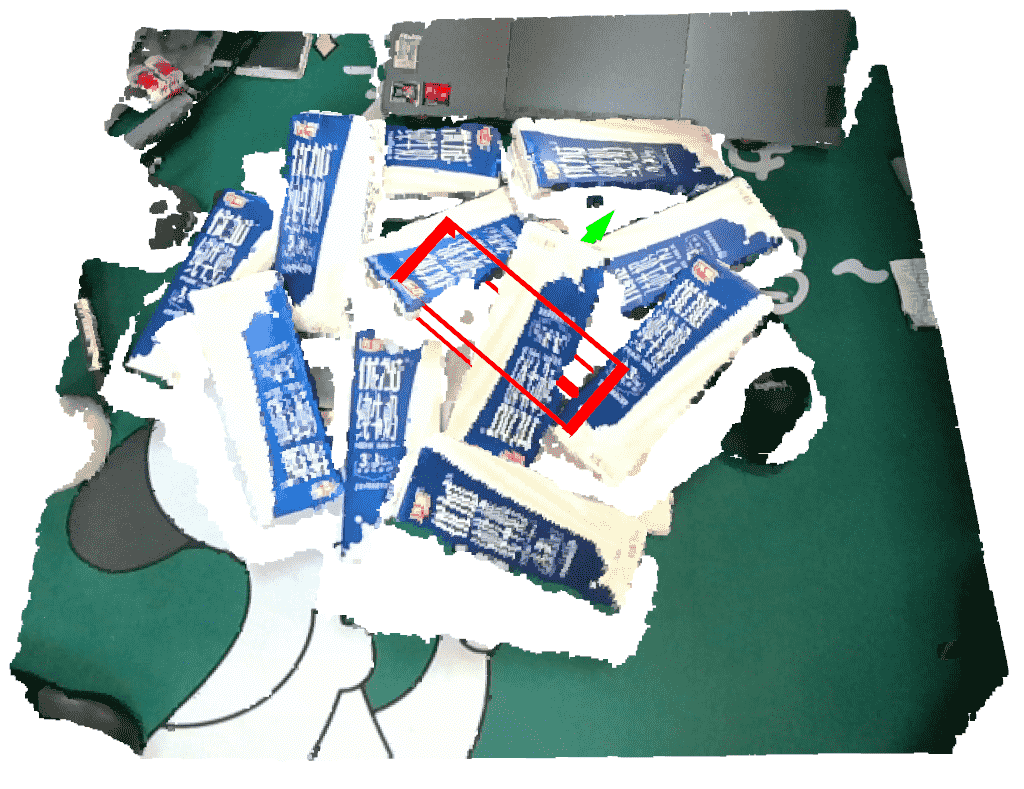}
    \caption{Progressive-X(2019) \cite{ProgressiveX}}
    \label{fig:real-prox16}
\end{subfigure}\hfill
\begin{subfigure}{0.3\textwidth}
  \centering
  \includegraphics[height=2.8cm]{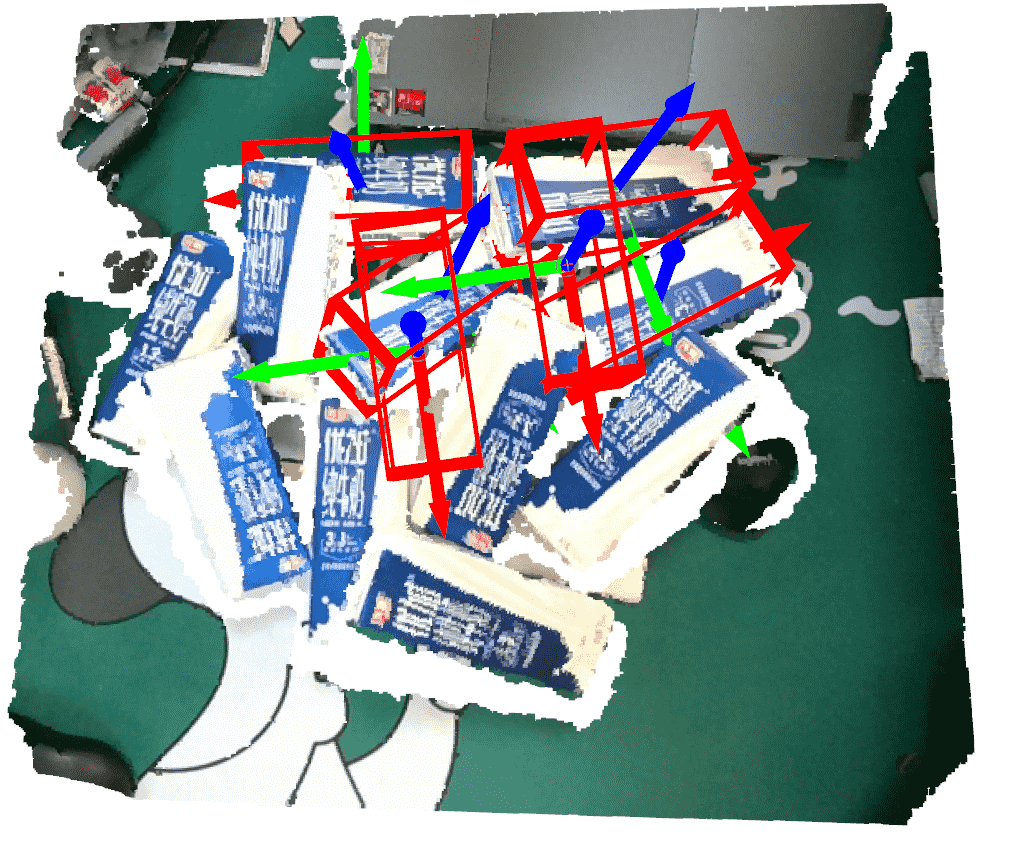}
    \caption{CONSAC(2020)\cite{CONSAC}}
    \label{fig:real-consac16}
\end{subfigure}\hfill
\caption{\textbf{Real-world tests on RGB-D scans.}}
\label{fig:Real16}
\end{figure*}

\begin{figure*}[ht]
  \centering
\begin{subfigure}{0.3\textwidth}
  \centering
  \includegraphics[height=2.8cm]{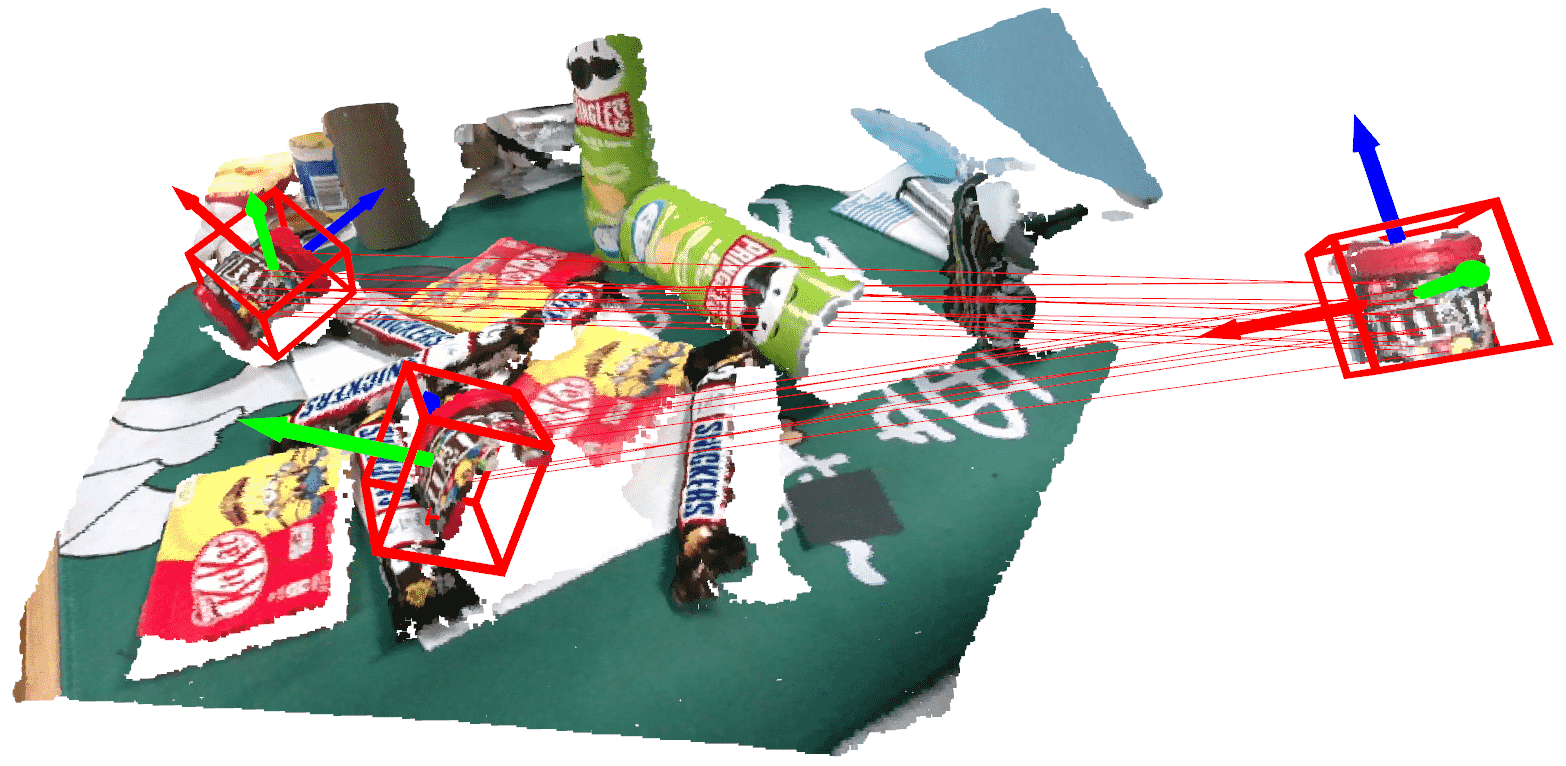}
    \caption{Ours}
    \label{fig:real-result17}
\end{subfigure}\hfill
\begin{subfigure}{0.3\textwidth}
  \centering
  \includegraphics[height=2.8cm]{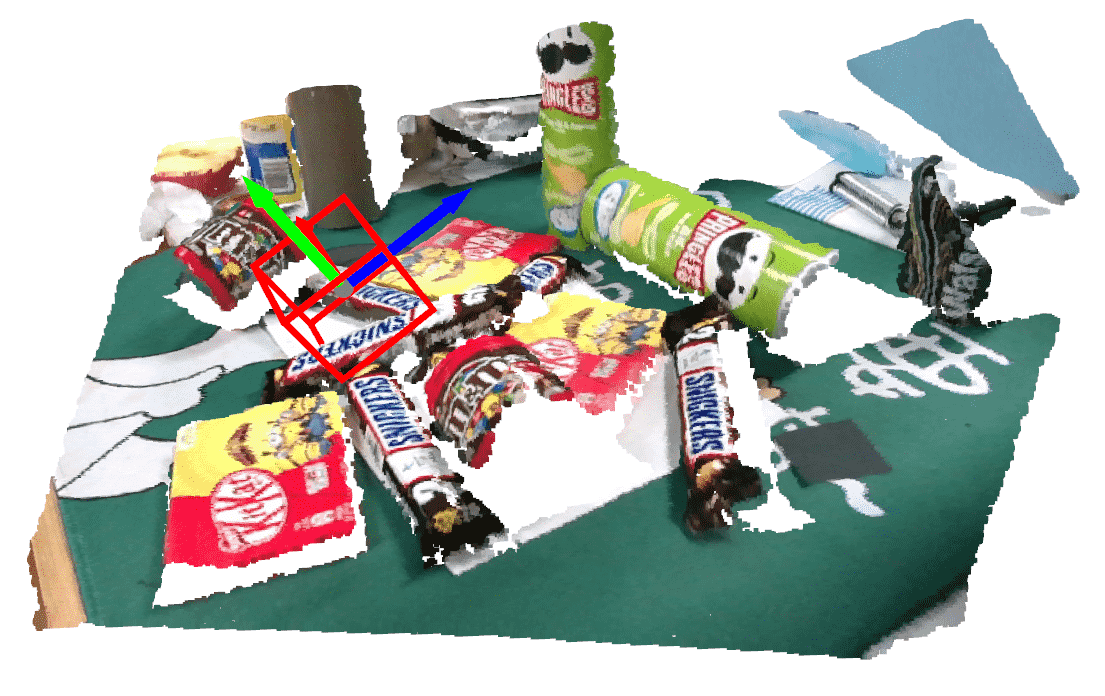}
    \caption{Progressive-X(2019) \cite{ProgressiveX}}
    \label{fig:real-prox17}
\end{subfigure}\hfill
\begin{subfigure}{0.3\textwidth}
  \centering
  \includegraphics[height=2.8cm]{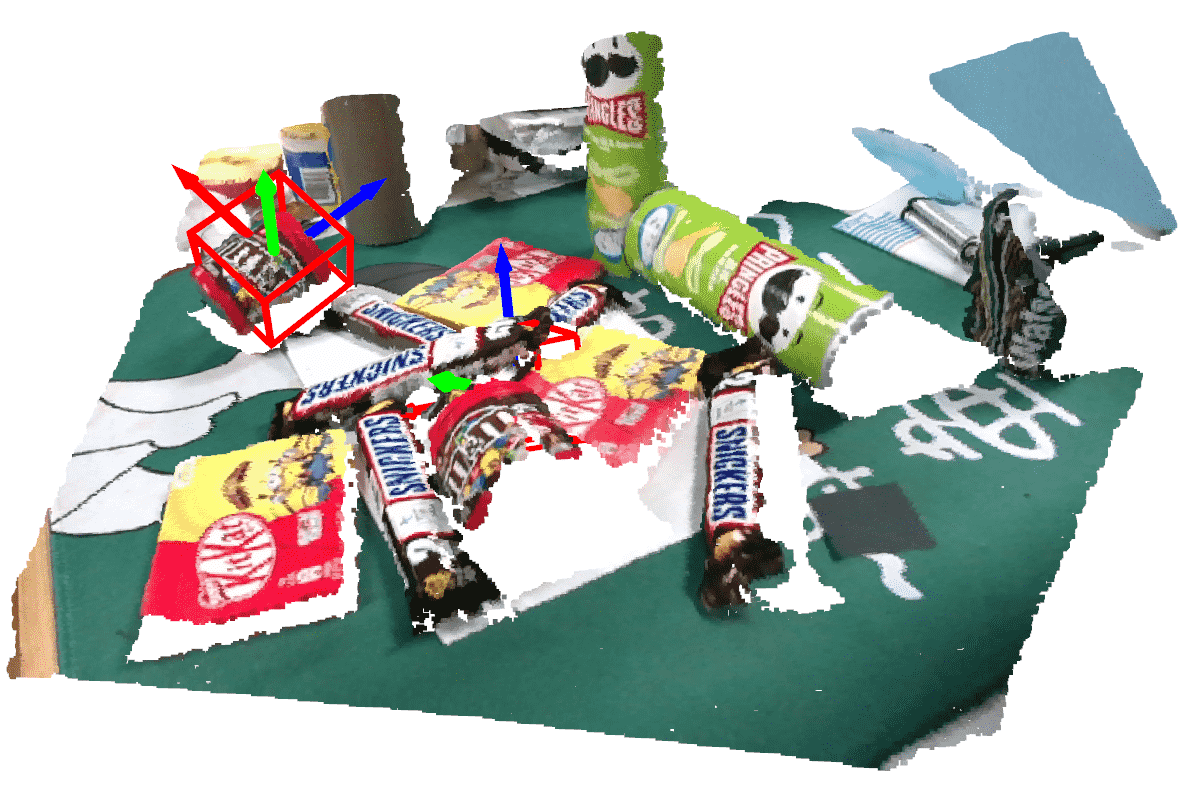}
    \caption{CONSAC(2020)\cite{CONSAC}}
    \label{fig:real-consac17}
\end{subfigure}\hfill
\caption{\textbf{Real-world tests on RGB-D scans.}}
\label{fig:Real17}
\end{figure*}

\begin{figure*}[ht]
  \centering
\begin{subfigure}{0.3\textwidth}
  \centering
  \includegraphics[height=2.8cm]{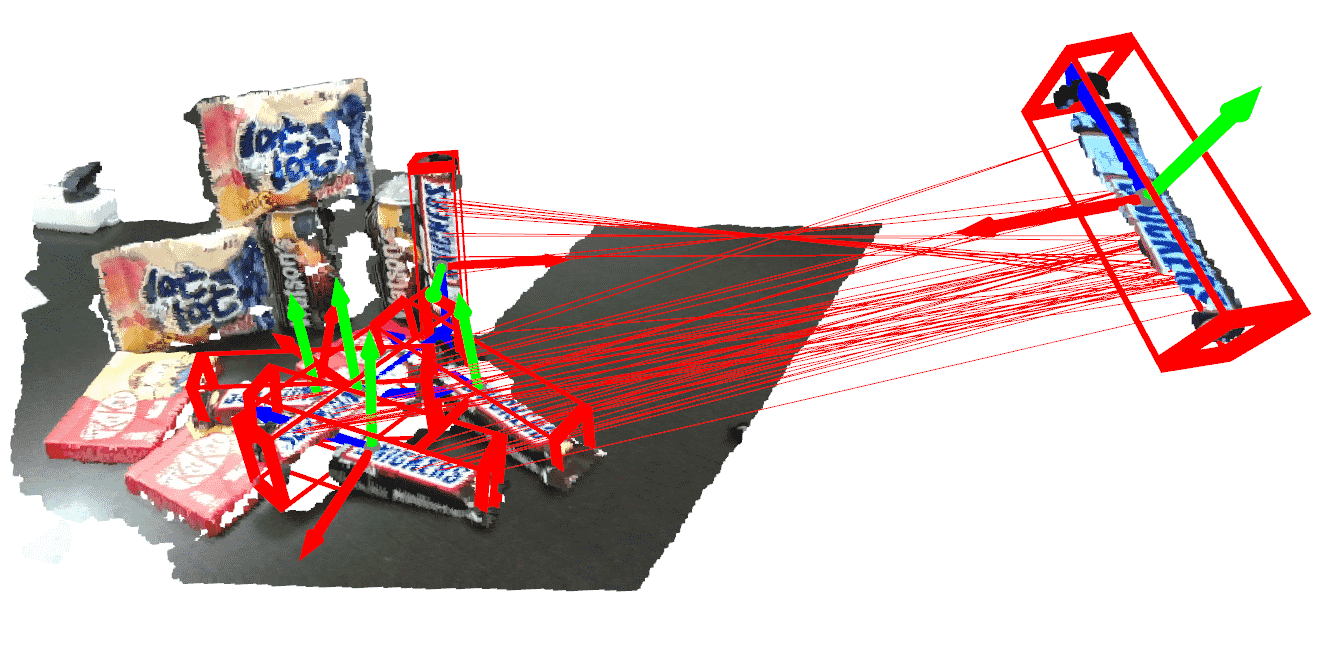}
    \caption{Ours}
    \label{fig:real-result18}
\end{subfigure}\hfill
\begin{subfigure}{0.3\textwidth}
  \centering
  \includegraphics[height=2.8cm]{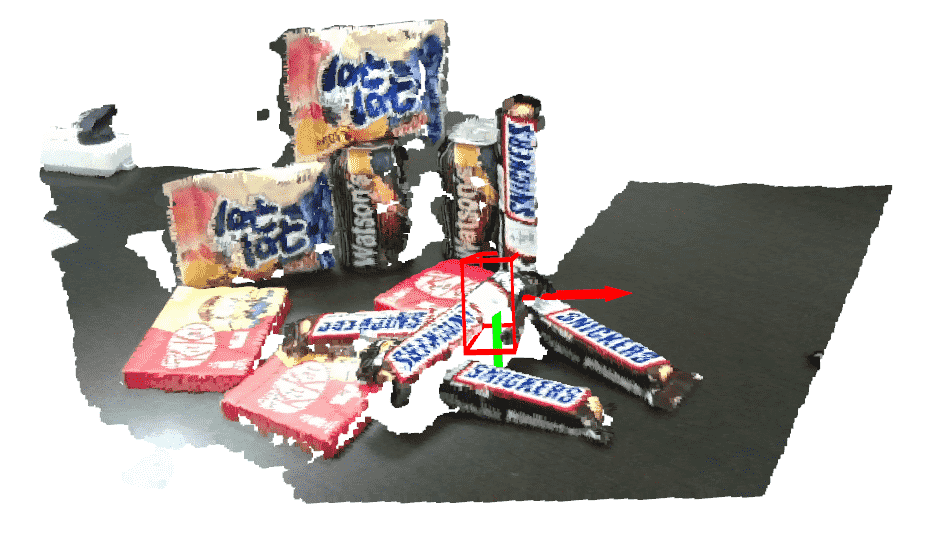}
    \caption{Progressive-X(2019) \cite{ProgressiveX}}
    \label{fig:real-prox18}
\end{subfigure}\hfill
\begin{subfigure}{0.3\textwidth}
  \centering
  \includegraphics[height=2.8cm]{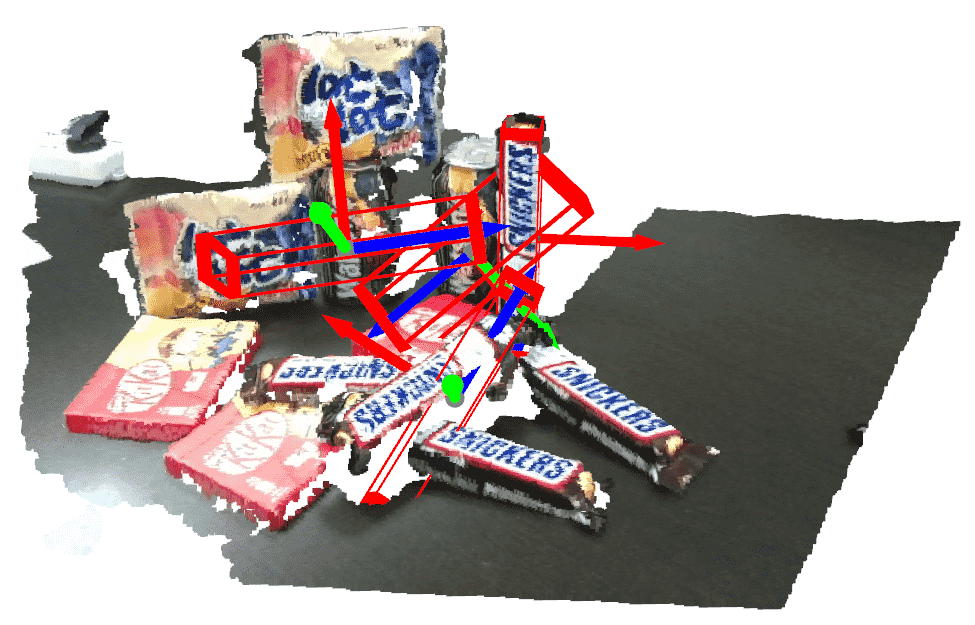}
    \caption{CONSAC(2020)\cite{CONSAC}}
    \label{fig:real-consac18}
\end{subfigure}\hfill
\caption{\textbf{Real-world tests on RGB-D scans.}}
\label{fig:Real18}
\end{figure*}

\begin{figure*}[ht]
  \centering
\begin{subfigure}{0.3\textwidth}
  \centering
  \includegraphics[height=2.8cm]{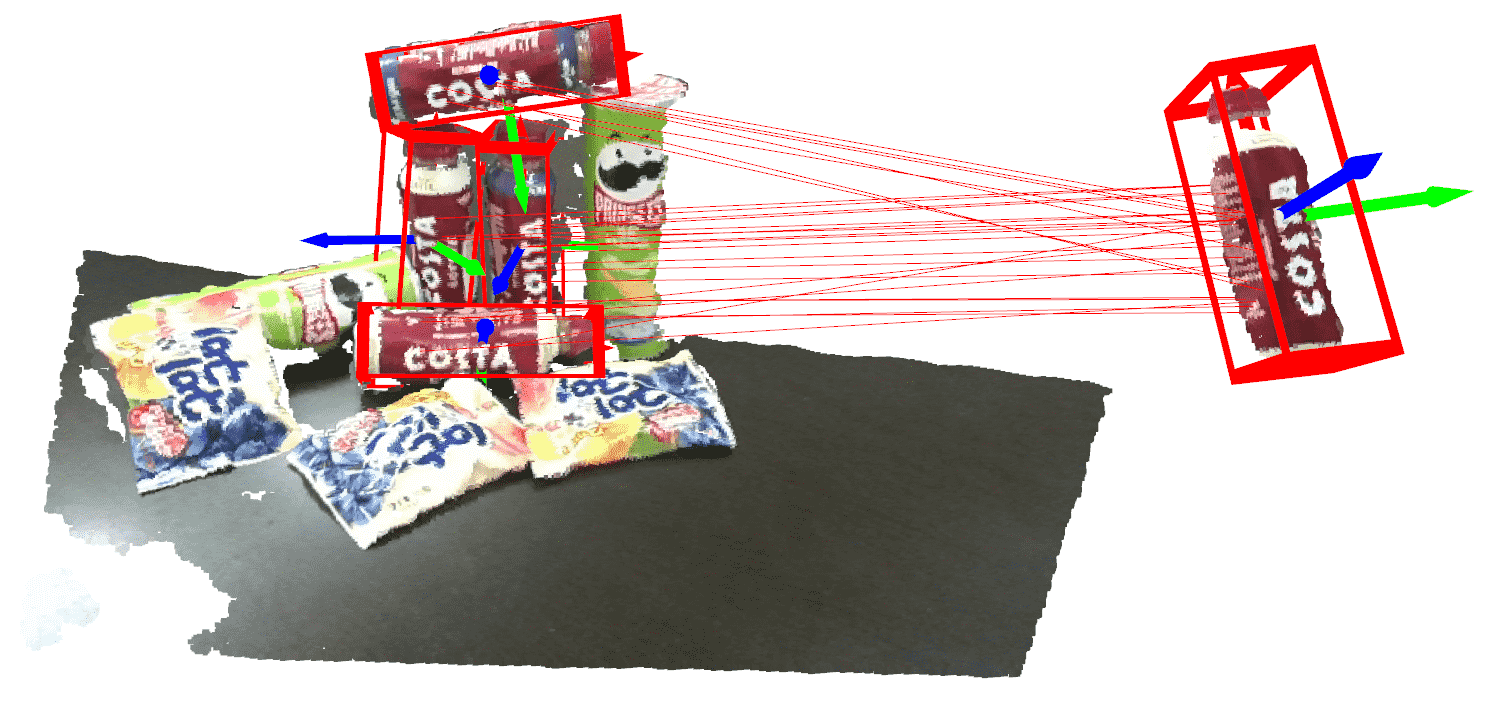}
    \caption{Ours}
    \label{fig:real-result20}
\end{subfigure}\hfill
\begin{subfigure}{0.3\textwidth}
  \centering
  \includegraphics[height=2.8cm]{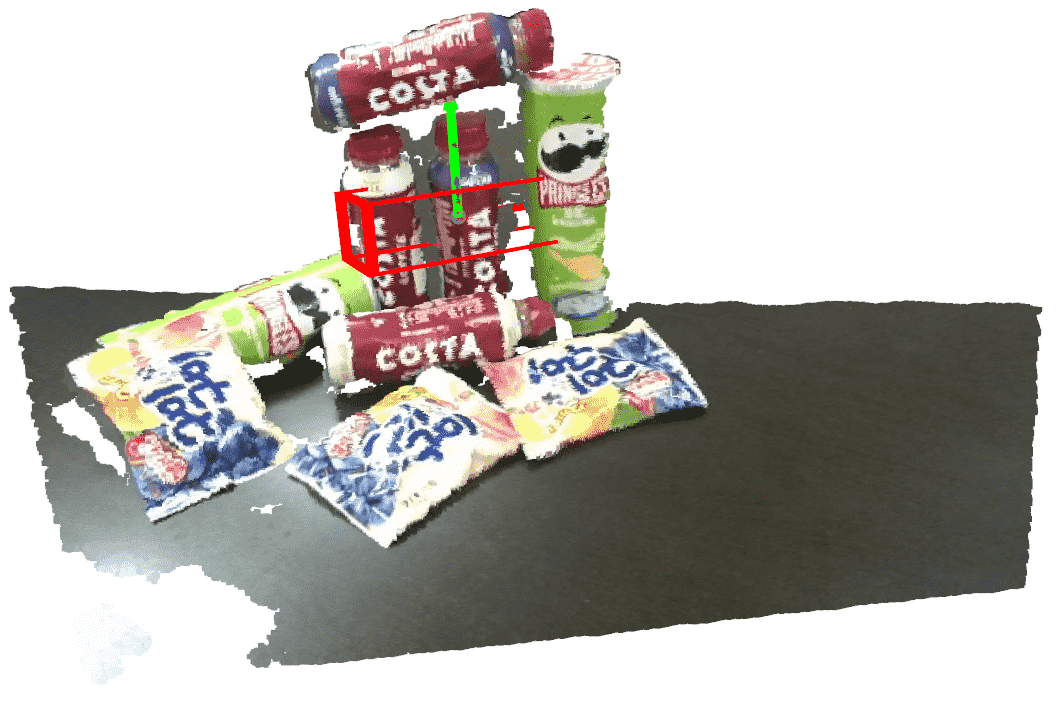}
    \caption{Progressive-X(2019) \cite{ProgressiveX}}
    \label{fig:real-prox20}
\end{subfigure}\hfill
\begin{subfigure}{0.3\textwidth}
  \centering
  \includegraphics[height=2.8cm]{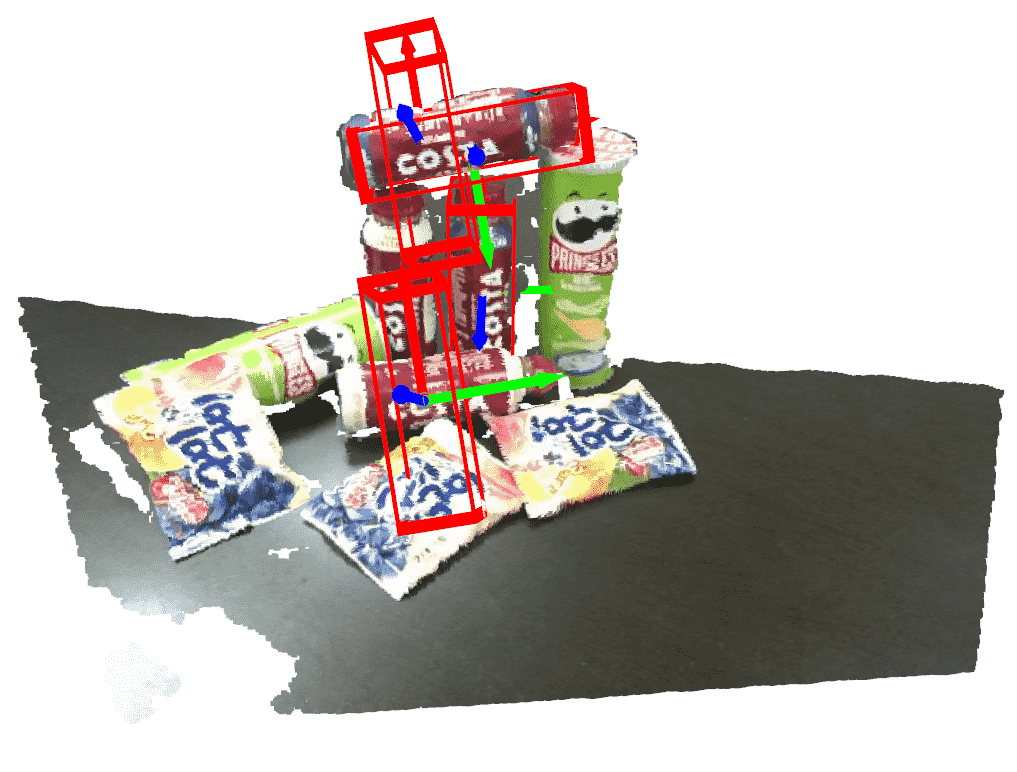}
    \caption{CONSAC(2020)\cite{CONSAC}}
    \label{fig:real-consac20}
\end{subfigure}\hfill
\caption{\textbf{Real-world tests on RGB-D scans.}}
\label{fig:Real20}
\end{figure*}

\begin{figure*}[ht]
  \centering
\begin{subfigure}{0.3\textwidth}
  \centering
  \includegraphics[height=2.8cm]{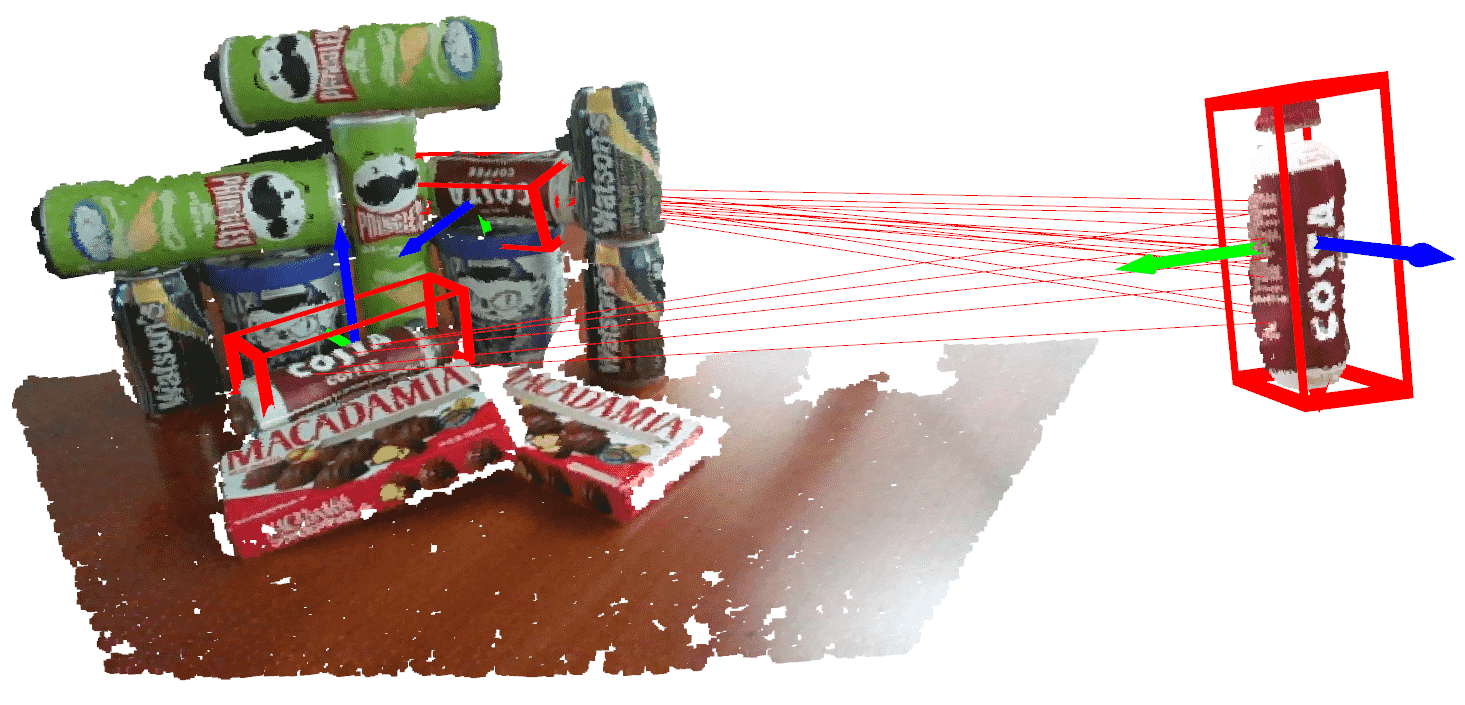}
    \caption{Ours}
    \label{fig:real-result21}
\end{subfigure}\hfill
\begin{subfigure}{0.3\textwidth}
  \centering
  \includegraphics[height=2.8cm]{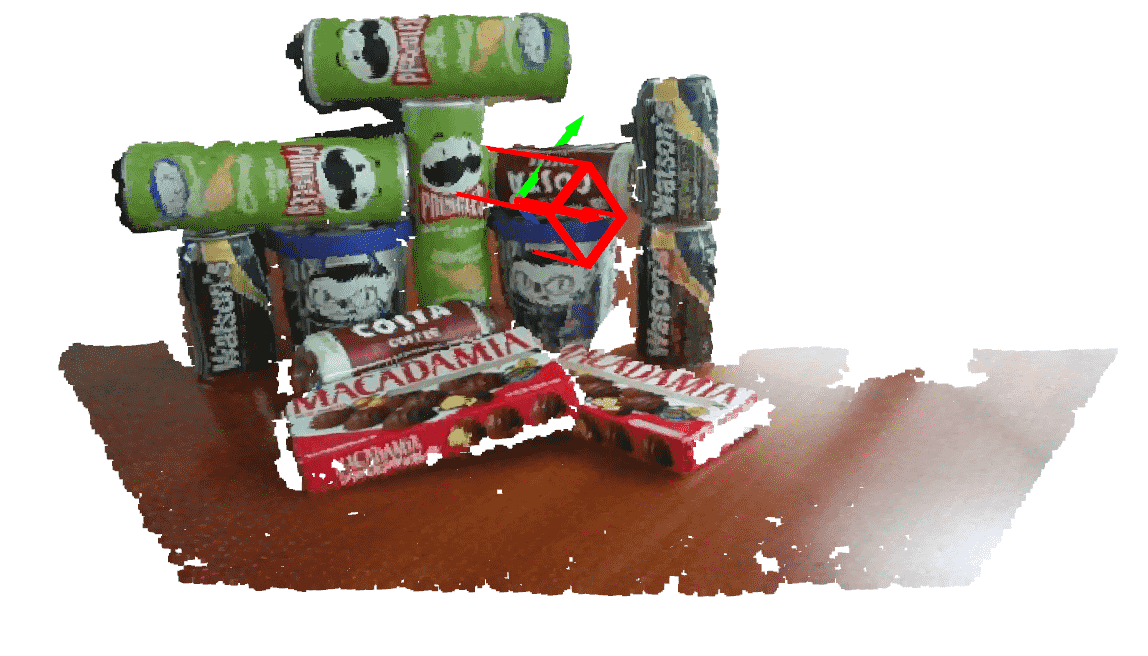}
    \caption{Progressive-X(2019) \cite{ProgressiveX}}
    \label{fig:real-prox21}
\end{subfigure}\hfill
\begin{subfigure}{0.3\textwidth}
  \centering
  \includegraphics[height=2.8cm]{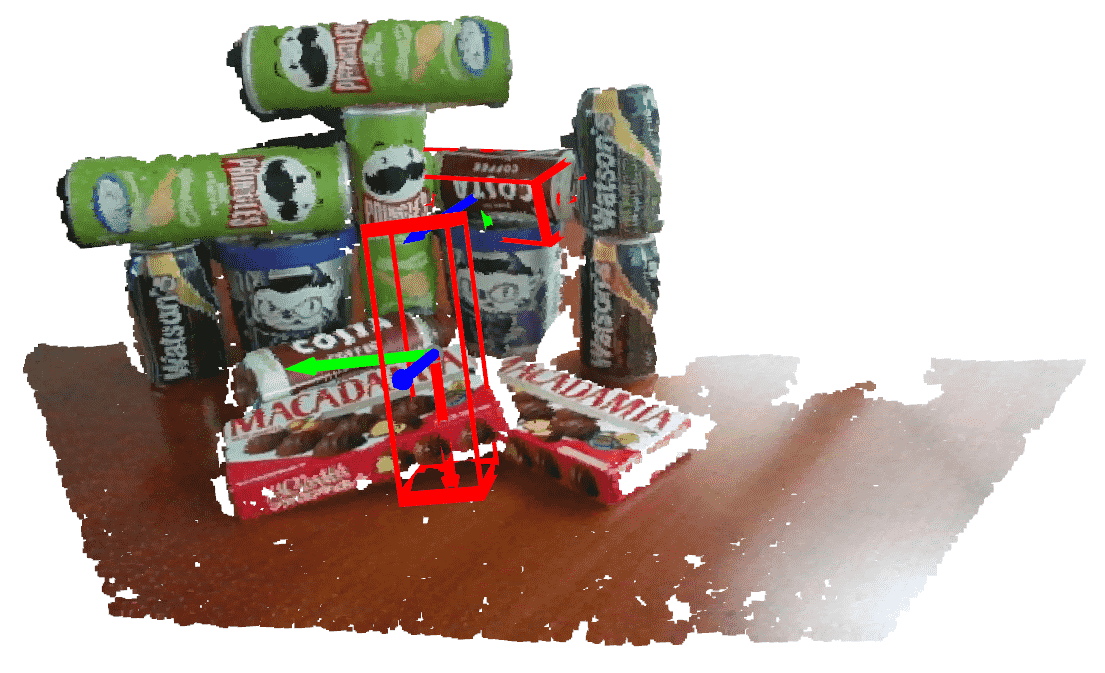}
    \caption{CONSAC(2020)\cite{CONSAC}}
    \label{fig:real-consac21}
\end{subfigure}\hfill
\caption{\textbf{Real-world tests on RGB-D scans.}}
\label{fig:Real21}
\end{figure*}

\begin{figure*}[ht]
  \centering
\begin{subfigure}{0.3\textwidth}
  \centering
  \includegraphics[height=2.8cm]{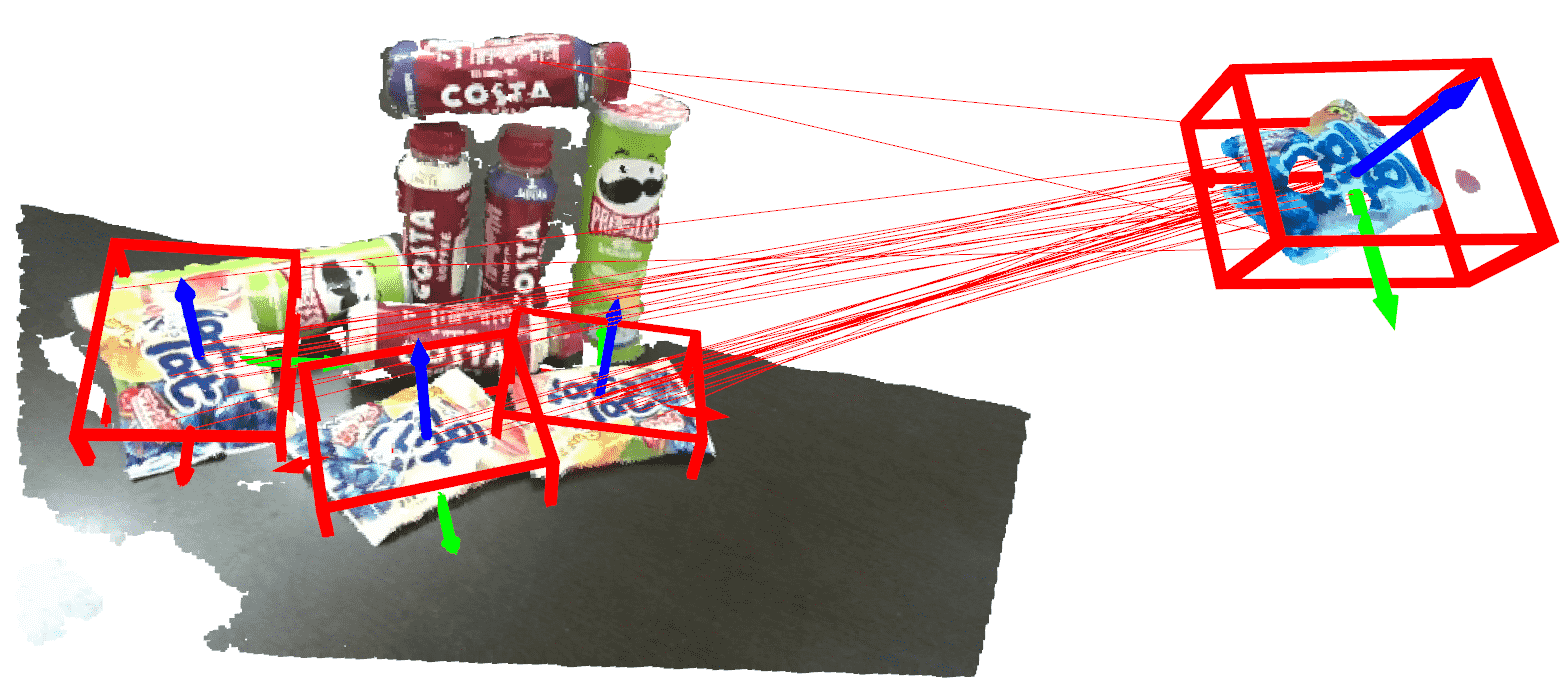}
    \caption{Ours}
    \label{fig:real-result22}
\end{subfigure}\hfill
\begin{subfigure}{0.3\textwidth}
  \centering
  \includegraphics[height=2.8cm]{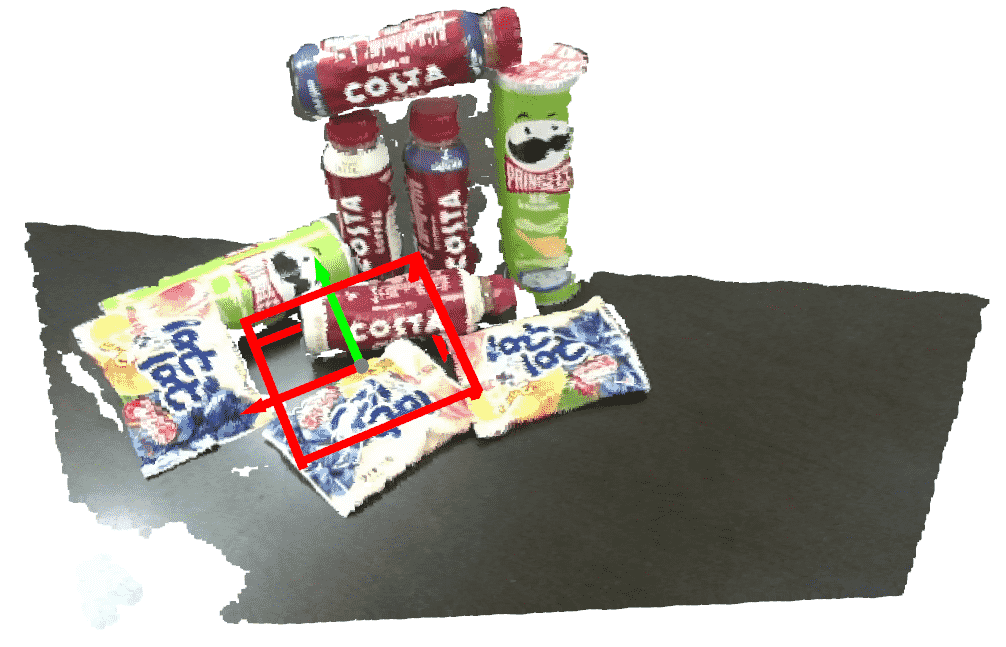}
    \caption{Progressive-X(2019) \cite{ProgressiveX}}
    \label{fig:real-prox22}
\end{subfigure}\hfill
\begin{subfigure}{0.3\textwidth}
  \centering
  \includegraphics[height=2.8cm]{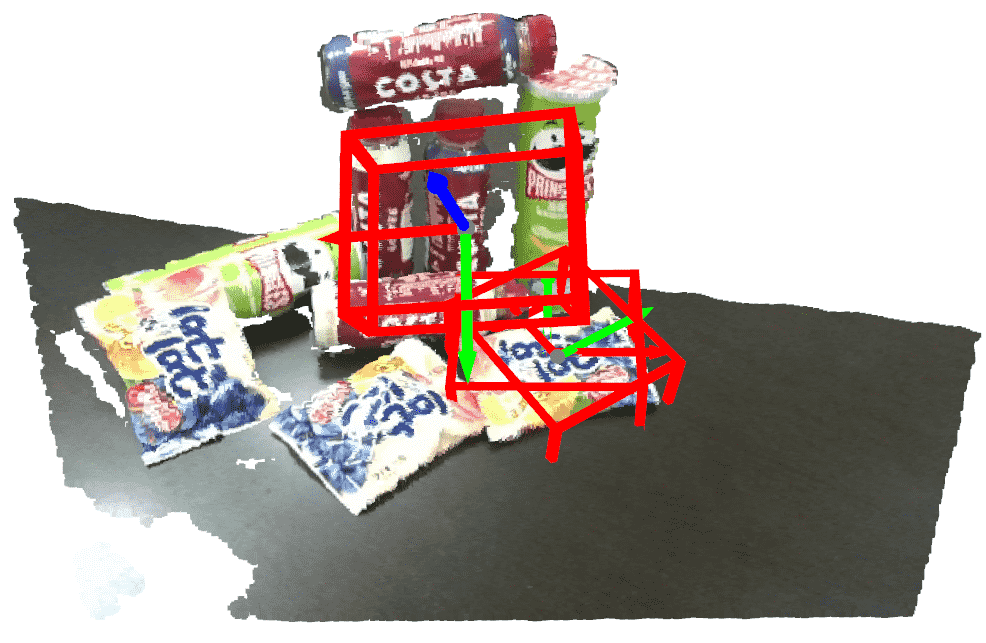}
    \caption{CONSAC(2020)\cite{CONSAC}}
    \label{fig:real-consac22}
\end{subfigure}\hfill
\caption{\textbf{Real-world tests on RGB-D scans.}}
\label{fig:Real22}
\end{figure*}

\begin{figure*}[ht]
  \centering
\begin{subfigure}{0.3\textwidth}
  \centering
  \includegraphics[height=2.8cm]{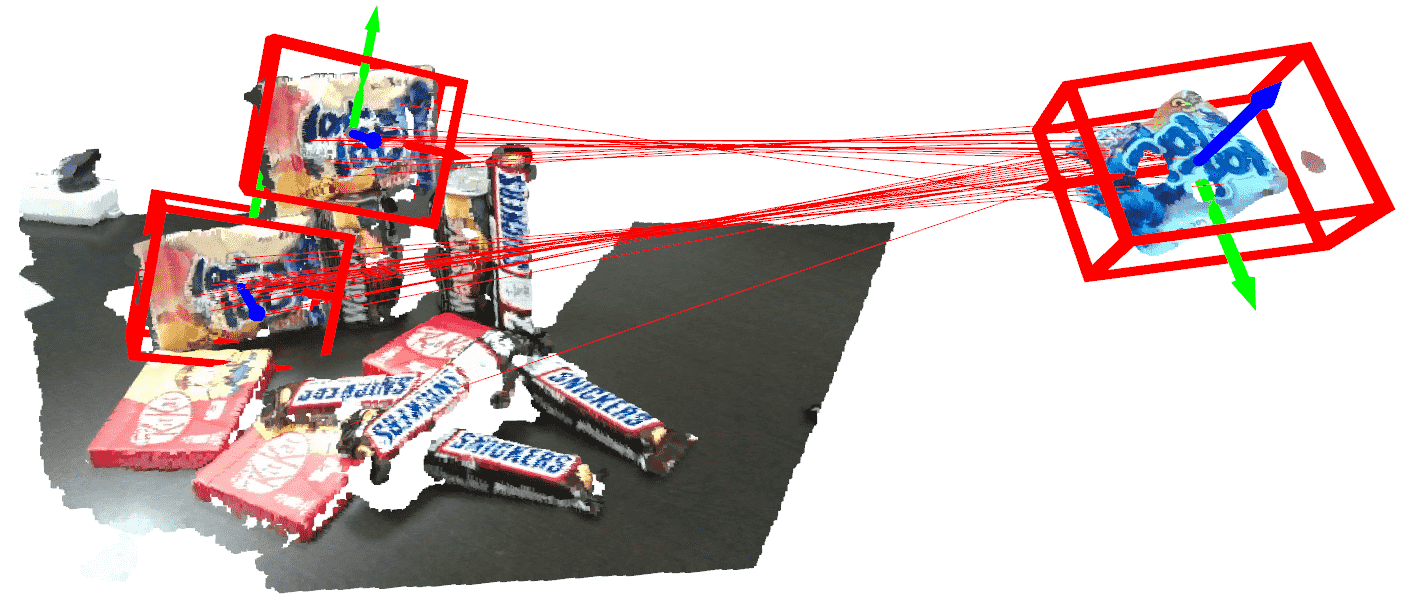}
    \caption{Ours}
    \label{fig:real-result23}
\end{subfigure}\hfill
\begin{subfigure}{0.3\textwidth}
  \centering
  \includegraphics[height=2.8cm]{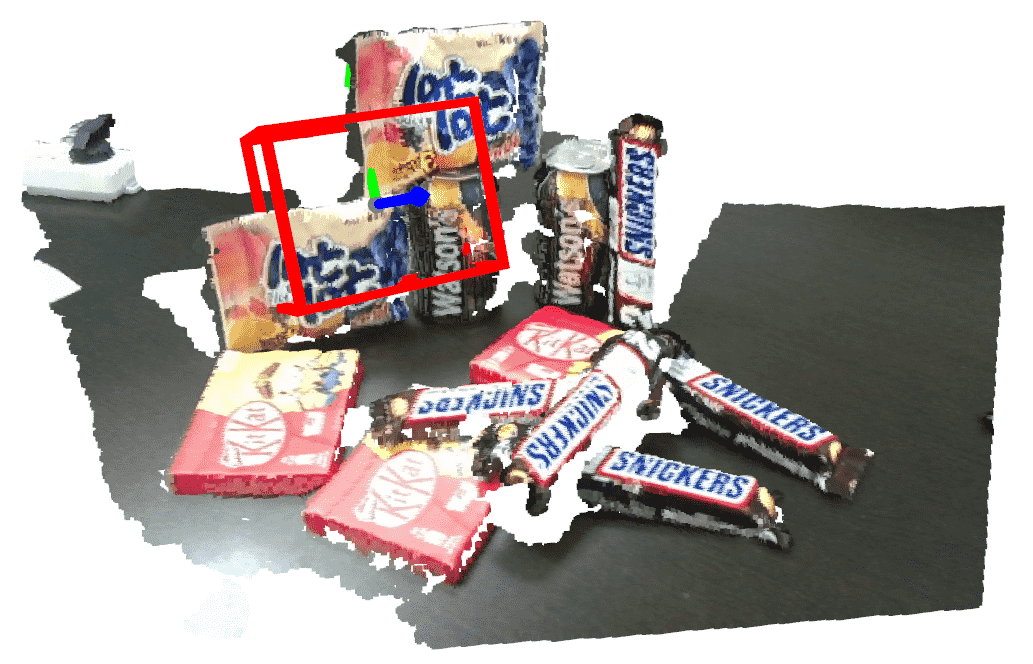}
    \caption{Progressive-X(2019) \cite{ProgressiveX}}
    \label{fig:real-prox23}
\end{subfigure}\hfill
\begin{subfigure}{0.3\textwidth}
  \centering
  \includegraphics[height=2.8cm]{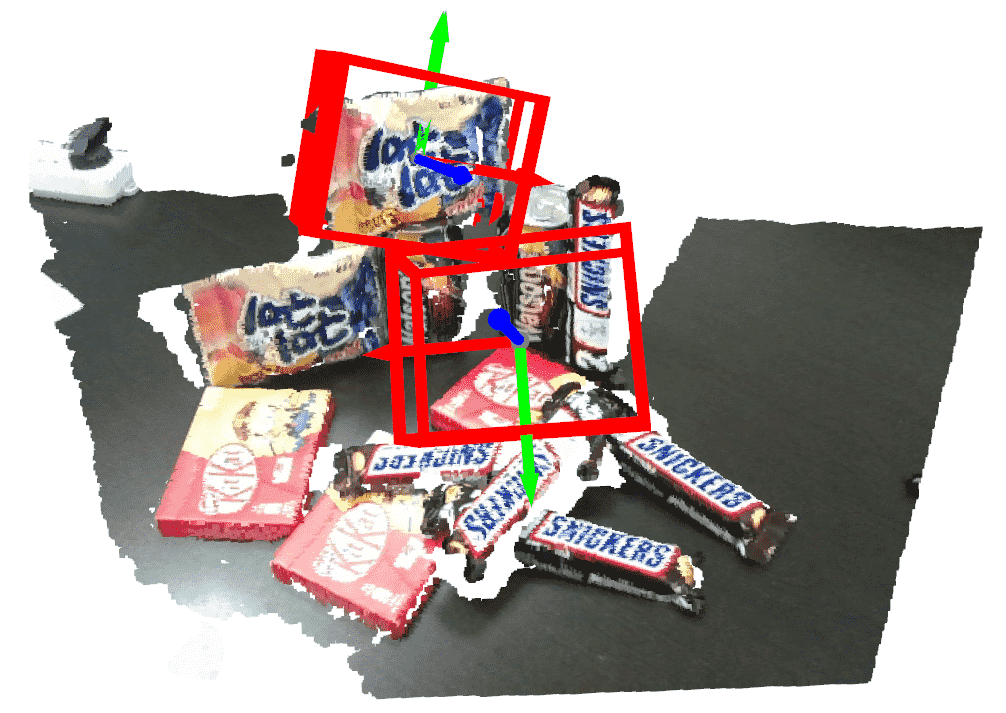}
    \caption{CONSAC(2020)\cite{CONSAC}}
    \label{fig:real-consac23}
\end{subfigure}\hfill
\caption{\textbf{Real-world tests on RGB-D scans.}}
\label{fig:Real23}
\end{figure*}

\begin{figure*}[ht]
  \centering
\begin{subfigure}{0.3\textwidth}
  \centering
  \includegraphics[height=2.8cm]{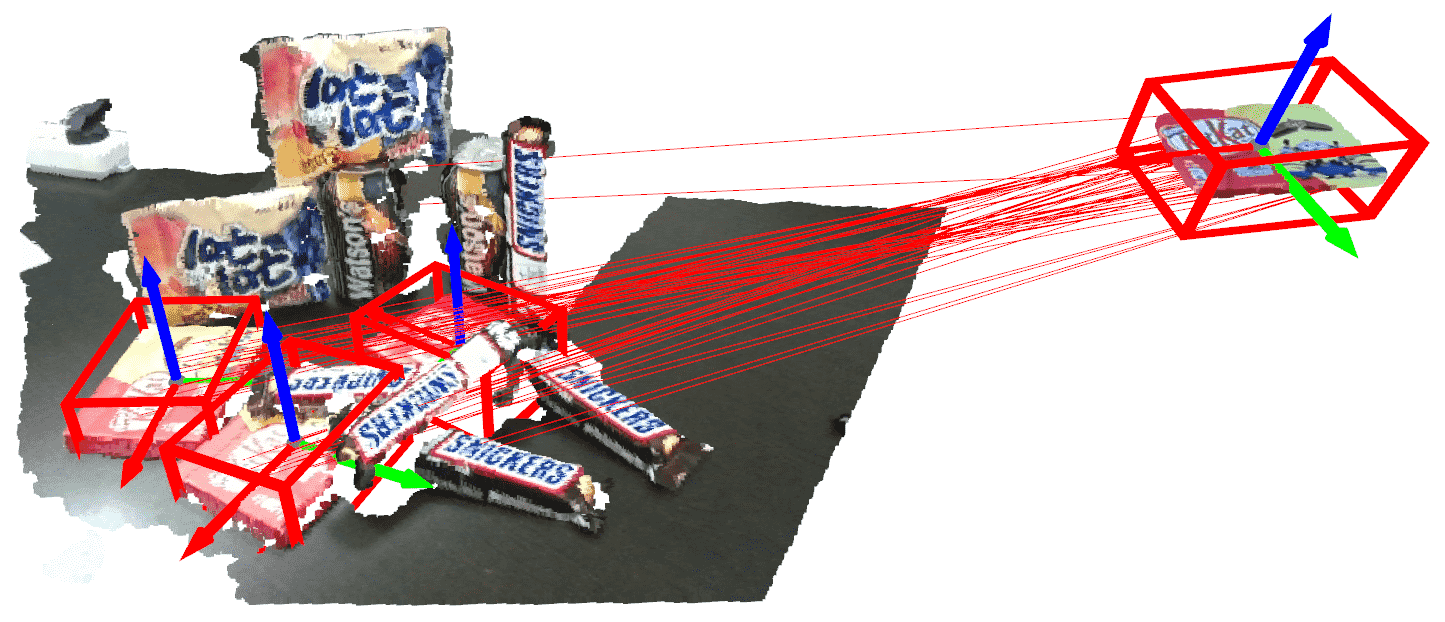}
    \caption{Ours}
    \label{fig:real-result24}
\end{subfigure}\hfill
\begin{subfigure}{0.3\textwidth}
  \centering
  \includegraphics[height=2.8cm]{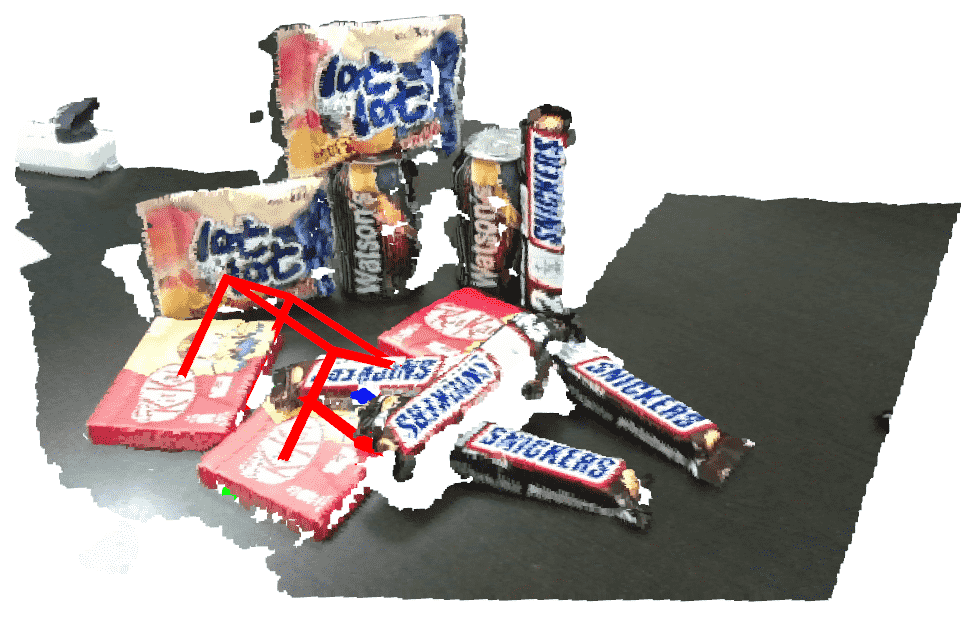}
    \caption{Progressive-X(2019) \cite{ProgressiveX}}
    \label{fig:real-prox24}
\end{subfigure}\hfill
\begin{subfigure}{0.3\textwidth}
  \centering
  \includegraphics[height=2.8cm]{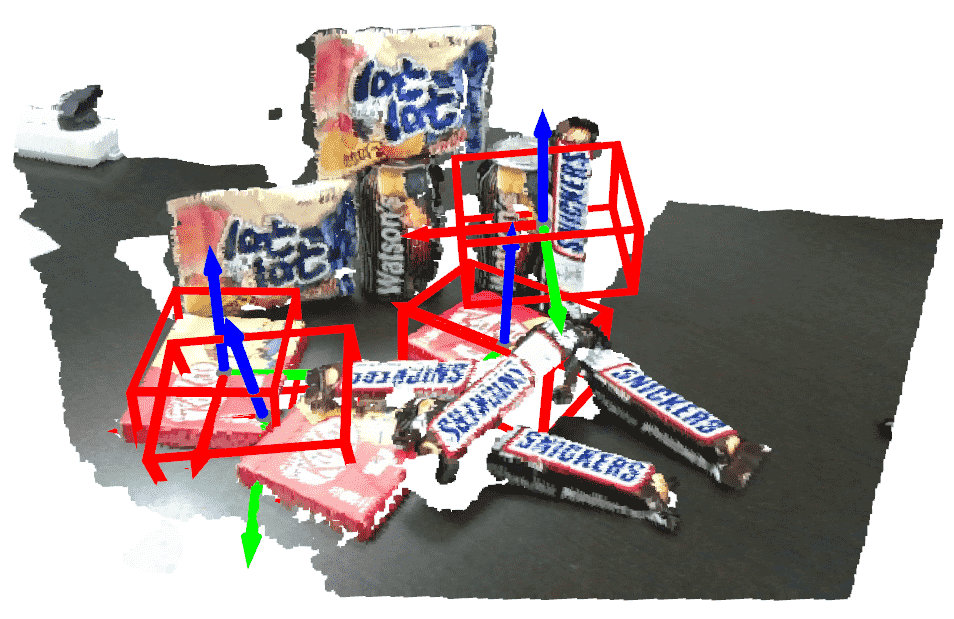}
    \caption{CONSAC(2020)\cite{CONSAC}}
    \label{fig:real-consac24}
\end{subfigure}\hfill
\caption{\textbf{Real-world tests on RGB-D scans.}}
\label{fig:Real24}
\end{figure*}

\begin{figure*}[ht]
  \centering
\begin{subfigure}{0.3\textwidth}
  \centering
  \includegraphics[height=2.8cm]{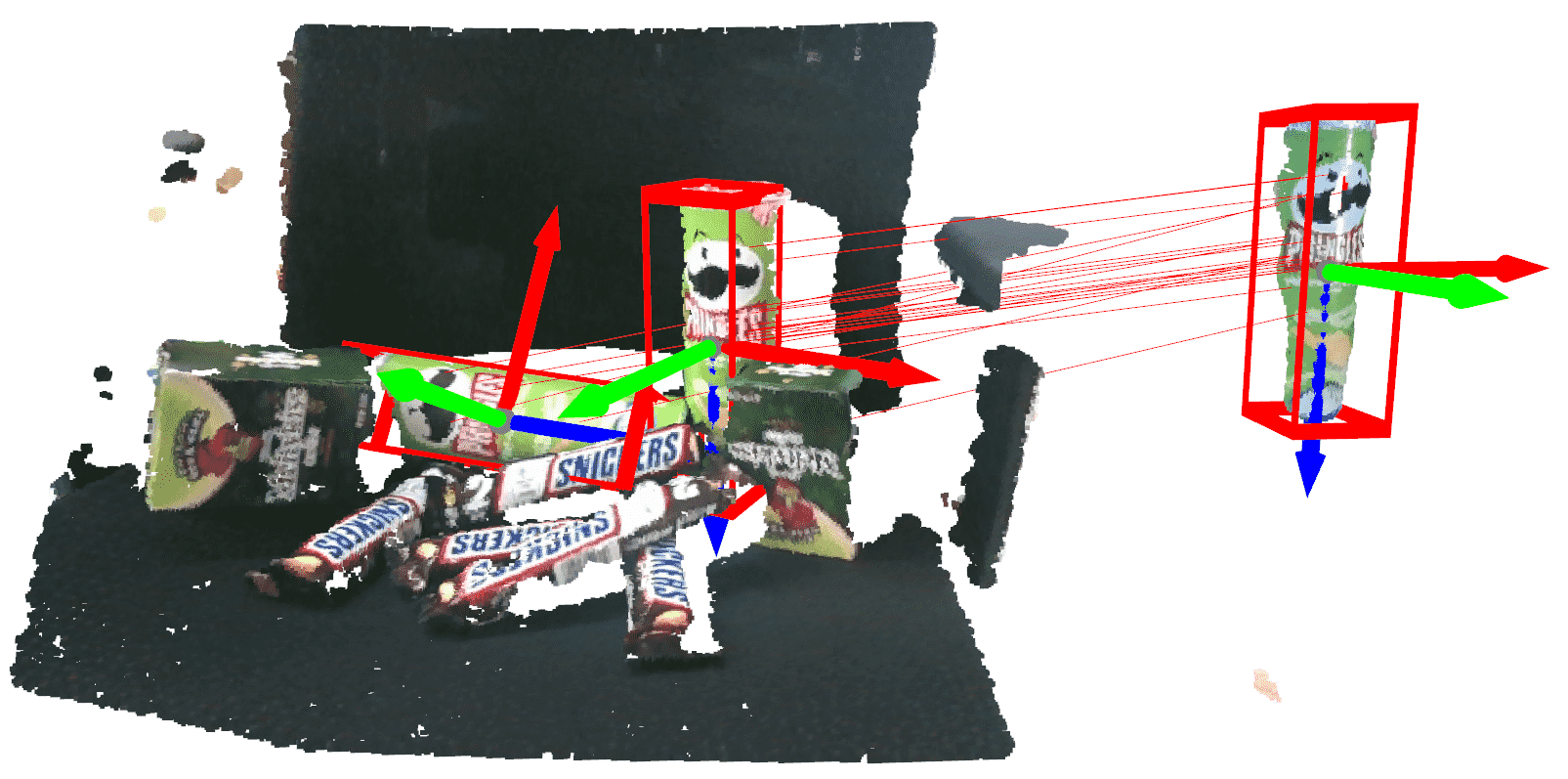}
    \caption{Ours}
    \label{fig:real-result25}
\end{subfigure}\hfill
\begin{subfigure}{0.3\textwidth}
  \centering
  \includegraphics[height=2.8cm]{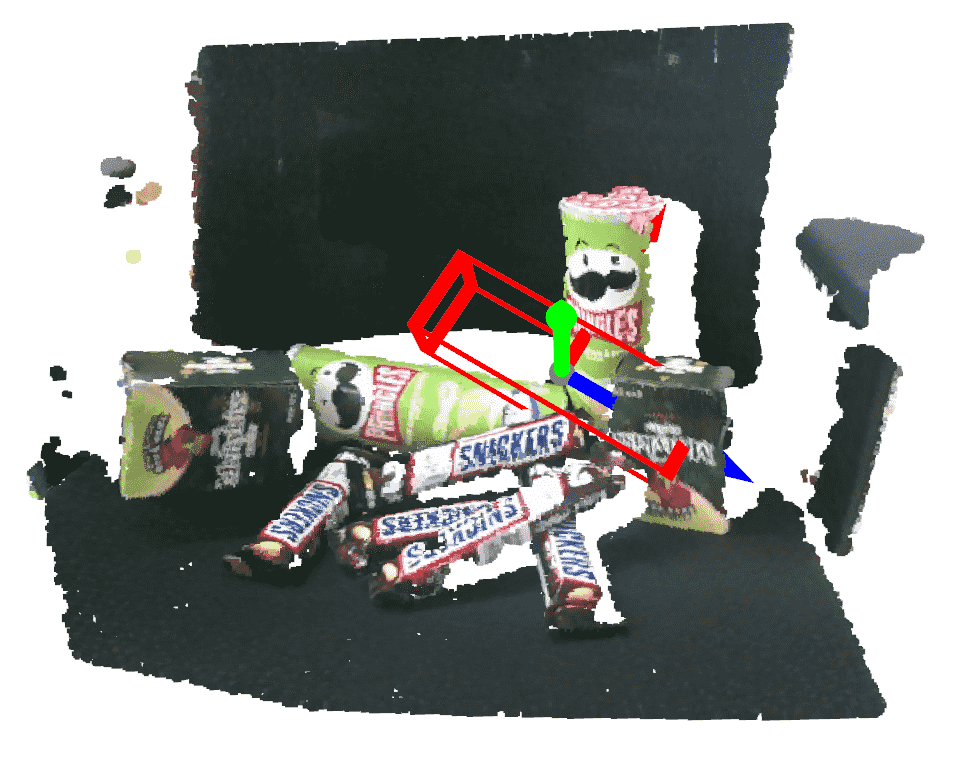}
    \caption{Progressive-X(2019) \cite{ProgressiveX}}
    \label{fig:real-prox25}
\end{subfigure}\hfill
\begin{subfigure}{0.3\textwidth}
  \centering
  \includegraphics[height=2.8cm]{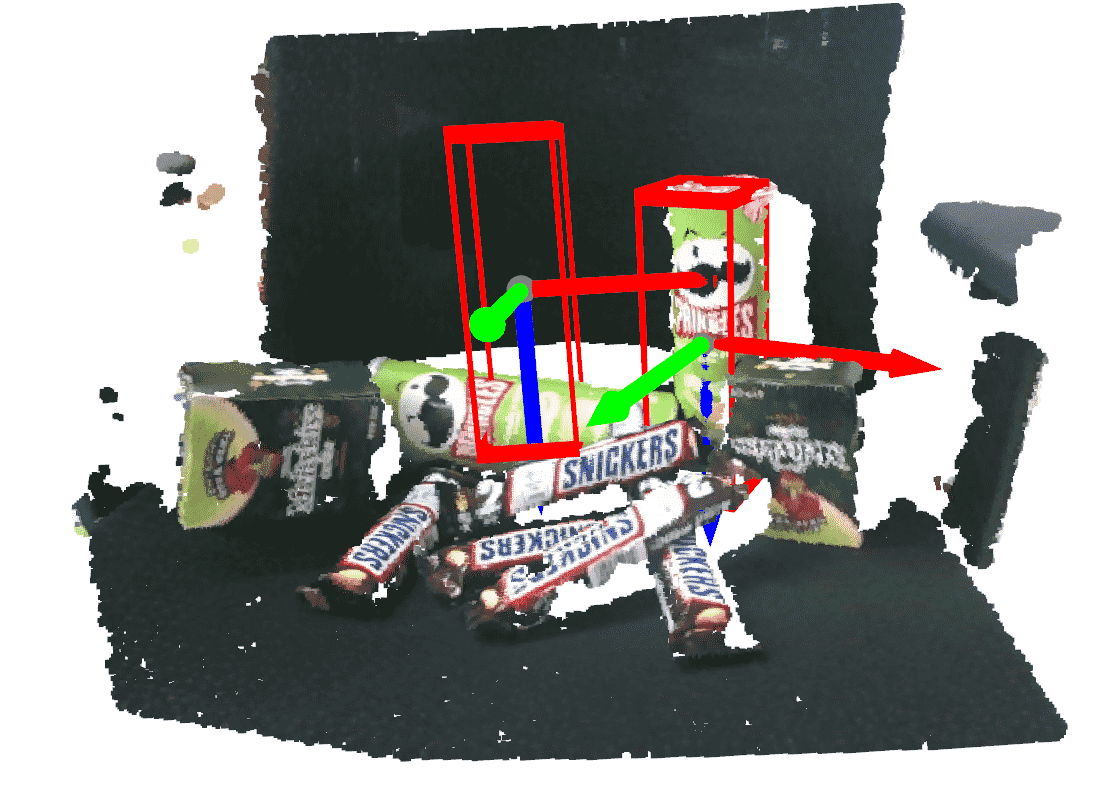}
    \caption{CONSAC(2020)\cite{CONSAC}}
    \label{fig:real-consac25}
\end{subfigure}\hfill
\caption{\textbf{Real-world tests on RGB-D scans.}}
\label{fig:Real25}
\end{figure*}

\begin{figure*}[ht]
  \centering
\begin{subfigure}{0.3\textwidth}
  \centering
  \includegraphics[height=2.8cm]{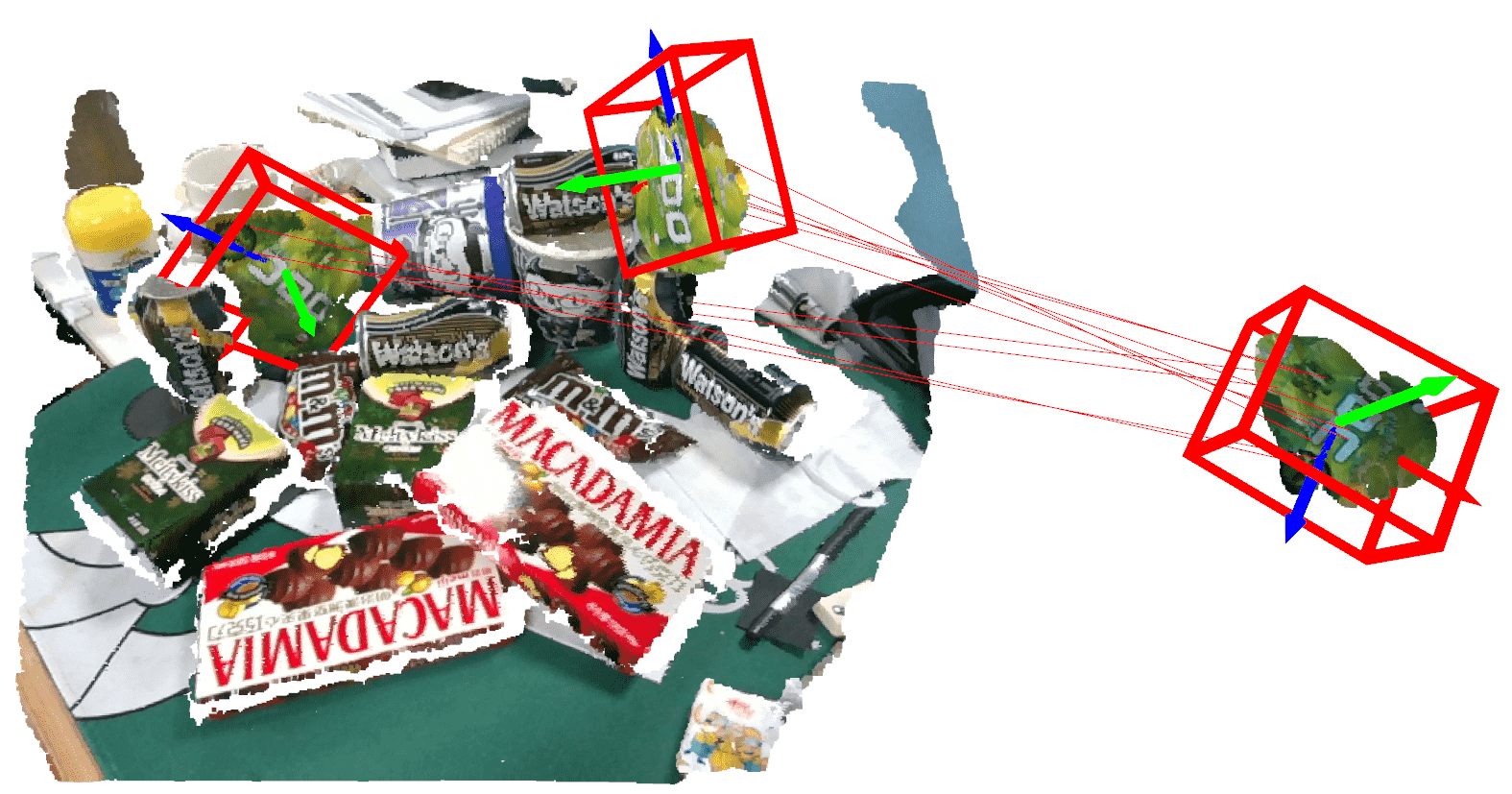}
    \caption{Ours}
    \label{fig:real-result28}
\end{subfigure}\hfill
\begin{subfigure}{0.3\textwidth}
  \centering
  \includegraphics[height=2.8cm]{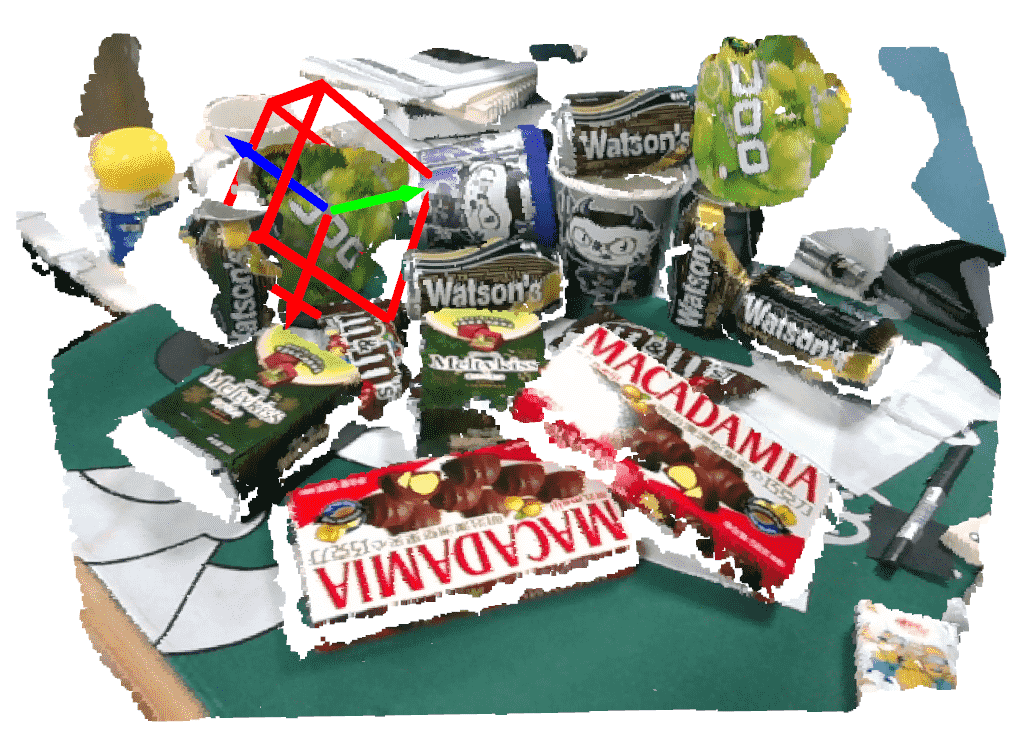}
    \caption{Progressive-X(2019) \cite{ProgressiveX}}
    \label{fig:real-prox28}
\end{subfigure}\hfill
\begin{subfigure}{0.3\textwidth}
  \centering
  \includegraphics[height=2.8cm]{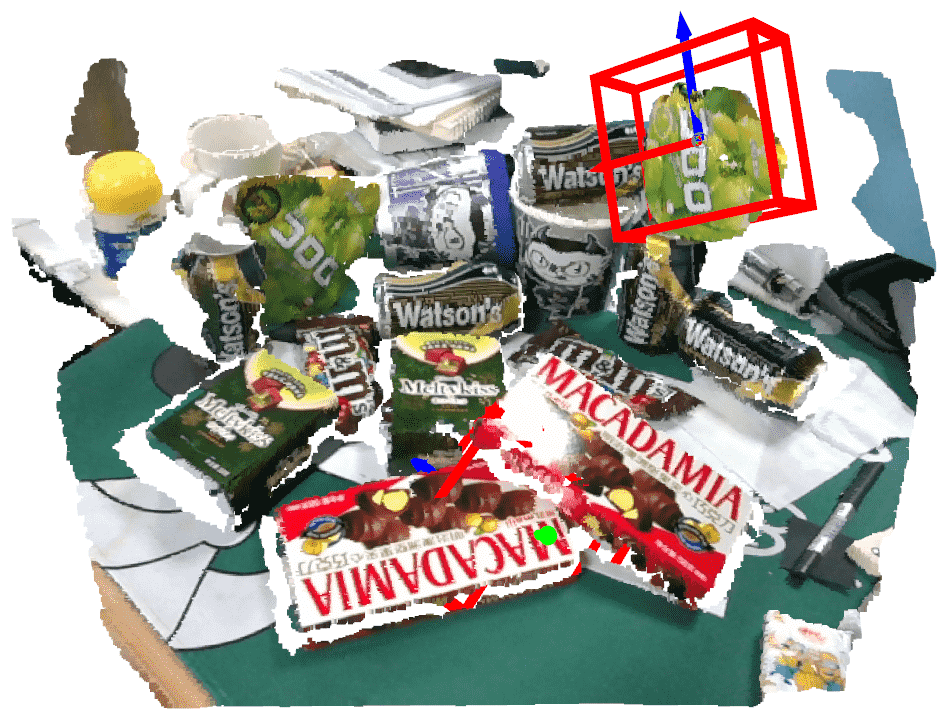}
    \caption{CONSAC(2020)\cite{CONSAC}}
    \label{fig:real-consac28}
\end{subfigure}\hfill
\caption{\textbf{Real-world tests on RGB-D scans.}}
\label{fig:Real28}
\end{figure*}

We show extra qualitative results of our experiments. 
The results of Scan2CAD are shown in Figure \ref{fig:aScan2CAD-cadresult1}-\ref{fig:Scan2CAD-cadresult3}, while the results of real-world tests are shown in Figure \ref{fig:Real1}-\ref{fig:Real28}. Video results are also available for real-world results.

%In this section, we give more qualitive result of our experiment.
%\subsection{Scan2cad datasets}
% Scan2cad 1

% Scan2cad 2

% Scan2cad 19

% Scan2cad 20

% Scan2cad 11

%\clearpage

% Scan2cad 12

% Scan2cad 17
% Scan2cad 22

% Scan2cad 24

% %\clearpage
% Scan2cad 4
% Scan2cad 27

% Scan2cad 3
%\clearpage

%%%%%%%%%%%%%%%%%%%%%%%%%%%%%%%%%%%%%%%%%%%%%%%%%%%%%%%%%%%%%%%%  
%%%%%%%%%%%%%%%%%%%%%%%%%%%%%%%%%%%%%%%%%%%%%%%%%%%%%%%%%%%%%%%%
%%%%%%%%%%%%%%%%%%%%%%%%%%%%%%%%%%%%%%%%%%%%%%%%%%%%%%%%%%%%%%%%
%%%%%%%%%%%%%%%%%%%%%%%%%%%%%%%%%%%%%%%%%%%%%%%%%%%%%%%%%%%%%%%%
%%%%%%%%%%%%%%%%%%%%%%%%%%%%%%%%%%%%%%%%%%%%%%%%%%%%%%%%%%%%%%%%
%%%%%%%%%%%%%%%%%%%%%%%%%%%%%%%%%%%%%%%%%%%%%%%%%%%%%%%%%%%%%%%%
%\subsubsection{Real}

% Real 1

% Real 2

% Real 3

% Real 4

% Real 6

% Real 10

% Real 13

% Real 15

% Real 16

% Real 17

% Real 18

% Real 20

% Real 21

% Real 22

% Real 23

% Real 24

% Real 25

% Real 28

\clearpage

% %%%%%%%%% REFERENCES
% {\small
% \bibliographystyle{ieee_fullname}
% \bibliography{egbib}
% }

\end{document}